\documentclass{article}
\usepackage[utf8]{inputenc}
\usepackage{graphicx}
\usepackage{multicol}
\usepackage{authblk}
\usepackage[margin=0.8in]{geometry}
\usepackage{array}
\usepackage{makecell}
\usepackage{hyperref}
\usepackage{tcolorbox}
\usepackage{float}
\usepackage{markdown}
\usepackage{enumitem}
\usepackage{placeins}
\usepackage{listings}
\usepackage[sorting = none, backend = bibtex, style=numeric-comp]{biblatex}
\title{Automated Multi-Language to English Machine Translation Using Generative Pre-Trained Transformers}

\author[1]{Elijah Pelofske\thanks{E-mail: elijah.pelofske@protonmail.com}}
\author[1]{Vincent Urias}
\author[2]{Lorie M. Liebrock}
\affil[1]{Sandia National Laboratories}
\affil[2]{New Mexico Cybersecurity Center of Excellence, New Mexico Tech}

\date{\vspace{-6ex}}

\begin{document}

\maketitle

\begin{abstract}
The task of accurate and efficient language translation is an extremely important information processing task. Machine learning enabled and automated translation that is accurate and fast is often a large topic of interest in the machine learning and data science communities. In this study, we examine using local Generative Pretrained Transformer (GPT) models to perform automated zero shot black-box, sentence wise, multi-natural-language translation into English text. We benchmark 16 different open-source GPT models, with no custom fine-tuning, from the Huggingface LLM repository for translating 50 different non-English languages into English using translated TED Talk transcripts as the reference dataset. These GPT model inference calls are performed strictly locally, on single A100 Nvidia GPUs. Benchmark metrics that are reported are language translation accuracy, using BLEU, GLEU, METEOR, and chrF text overlap measures, and wall-clock time for each sentence translation. The best overall performing GPT model for translating into English text for the BLEU metric is \texttt{ReMM-v2-L2-13B} with a mean score across all tested languages of $0.152$, for the GLEU metric is \texttt{ReMM-v2-L2-13B} with a mean score across all tested languages of $0.256$, for the chrF metric is \texttt{Llama2-chat-AYT-13B} with a mean score across all tested languages of $0.448$, and for the METEOR metric is \texttt{ReMM-v2-L2-13B} with a mean score across all tested languages of $0.438$. 
\end{abstract}


\section{Introduction}
\label{section:introduction}

Large Language Models (LLMs), specifically transformer based architecture~\cite{vaswani2023attention}, have been shown to be incredibly effective at learning tasks that require significant abstraction. Generative Pre-Trained Transformers (GPT)~\cite{yenduri2023generative} have been used to demonstrate numerous highly consequential learning and information processing tasks~\cite{openai2023gpt4}, including code generation~\cite{narasimhan2021cgems, zhong2023chatgpt, olausson2023selfrepair}, text summarization~\cite{goyal2023news, liu2023abstractive, bajaj2021long, pu2023summarization}, and chemistry experimental design~\cite{boiko2023autonomous}. 

In this study, we examine the capabilities of GPT models for the task of translating natural language text in an automated black-box fashion. Multi-language translation using GPT models has been investigated before using OpenAI's GPT models~\cite{hendy2023good}, and using deep learning~\cite{popel2020transforming}. In this study, we evaluate 16 open source GPT models, run locally and offline in order to assess the effectiveness of black-box translation using current local GPT models. We consider the language translation task of going from $50$ natural languages into English text, using the dataset of translated TED talk transcripts. 

This study is motivated by machine translation being of fundamental interest in computing and information sharing, and given the evident demonstrations of GPT model capabilities, it makes sense to evaluate how well current GPT models perform at this task. Many GPT chat models are available to users as cloud based resources. However, there are significant privacy and security concerns with this model of computation. Therefore, we are interested in using offline, entirely local, GPT inference calls. This also lets us quantify the scale of the computation required for a task such as automated multi-language machine translation - in this case we use single A100 GPU's to perform the inference for each model. Lastly, we aim to evaluate the \emph{automated} machine translation capabilities of the current GPT models - in particular we do not heavily optimize the inference hyperparameters, or the chat prompts. The goal is to measure a reasonably large and language agnostic (e.g., not prompt tuned for each language) benchmark of the translation capabilities of these models. The GPT translation quality is compared against the \texttt{Google translate} API, in Python~\cite{python_google_translate}.

\begin{table*}[ht!]
\centering
\begin{tabular}{ |c|c||c|c|c| }
 \hline
 \hline
 Model name & Reference(s) & Context Length & Architecture type & Model Size \\ 
 \hline
 \hline
 \hline
 \texttt{zephyr-7b-alpha} &~\cite{rafailov2023direct} & 32768 Tokens & mistral & 7.24B params \\ 
 \hline
 \texttt{zephyr-7b-beta} &~\cite{rafailov2023direct, tunstall2023zephyr} & 32768 Tokens & mistral & 7.24B params \\ 
 \hline
 \texttt{Mistral-7B-Instruct-v0.1} &~\cite{jiang2023mistral} & 32768 Tokens & mistral & 7.24B params \\ 
 \hline
 \texttt{Turdus} &~\cite{udk_dot_ai_turdus} & 32768 Tokens & mistral & 7.24B params \\ 
 \hline
 \texttt{vicuna-7b-v1.5} &~\cite{zheng2023judging, touvron2023llama} & 4096 Tokens & llama & 7B params \\ 
 \hline
 \texttt{phi-2} &~\cite{phi_2_huggingface} & 2048 Tokens & phi & 2.78B params \\ 
 \hline
 \texttt{phi-1} &~\cite{gunasekar2023textbooks} & 2048 Tokens & phi & 1.3B params \\ 
 \hline
 \texttt{phi-1\_5} & ~\cite{textbooks2} & 2048 Tokens & phi & 1.3B params \\ 
 \hline
 \texttt{ReMM-v2-L2-13B} &~\cite{ReMM-v2-L2-13B} & 4096 Tokens & llama & 13B params \\ 
 \hline
 \texttt{wizardLM-7B-HF} &~\cite{xu2023wizardlm} & 2048 Tokens & llama & 7B params \\ 
 \hline
 \texttt{wizardLM-13B-1.0-fp16} &~\cite{xu2023wizardlm} & 2048 Tokens & llama & 13B params \\ 
 \hline
 \texttt{Llama-2-13b-chat-hf} &~\cite{touvron2023llama} & 4096 Tokens & llama & 13B params \\ 
 \hline
 \texttt{Llama2-chat-AYT-13B} &~\cite{mukherjee2023orca, touvron2023llama} & 4096 Tokens & llama  & 13B params \\ 
 \hline
 \texttt{TinyLlama-1.1B-Chat-v1.0} &~\cite{zhang2024tinyllama, lit-gpt, dao2023flashattention2} & 2048 Tokens & llama & 1.1B params \\ 
 \hline
 \texttt{gpt4all-13b-snoozy} &~\cite{gpt4all} & 2048 Tokens & llama & 13B params \\ 
 \hline
 \texttt{falcon-7b-instruct} &~\cite{refinedweb, falcon40b} & 2048 Tokens & falcon & 7B params \\ 
 \hline
\end{tabular}
\caption{Summary of the $16$ Generative Pre-trained Transformers models used in this study}
\label{table:GPT_model_summary}
\end{table*}

\section{Methods}
\label{section:methods}

The GPT models used in this study are summarized in Table~\ref{table:GPT_model_summary} - the model weights were downloaded from huggingface~\cite{wolf2020huggingfaces}, where the trained model weights are open sourced. Each of these GPT models have been fine tuned, with varying levels of success, to be prompted in a chat-type mode. These models run using the PyTorch python library~\cite{Paszke_PyTorch_An_Imperative_2019}. The context window for the GPT models, summarized in Table~\ref{table:GPT_model_summary}, is not always clearly defined, but for several of the models the context window is given explicitly in the model weights repository. In other cases, the context window is in the metadata under the parameter \texttt{max\_position\_embeddings}, \texttt{n\_embd}, or is not explicitly stated. No fine-tuning of the model weights is performed, all $16$ of these GPT models are evaluated as-is in this black-box benchmarking comparison for language translation. Importantly, the underlying architectures of all of these GPT models rely on a large number of remarkable machine learning developments in recent years, many of which are described in refs.~\cite{su2023roformer, dao2022flashattention, dao2023flashattention2, brown2020language, shazeer2019fast, openai2023gpt4, vaswani2023attention}.

The translation dataset is a set of Ted Talk transcripts aggregated by the study in ref.~\cite{Ye2018WordEmbeddings}. Specifically, for $50$ of the foreign languages in the transcript dataset, $1,000$ of those sentences are translated into English. Due to the nature of the dataset, the same $1,000$ sentences are not necessarily translated across the $50$ foreign languages (many of the transcript translations are incomplete). Then, those translated sentences are compared against the corresponding reference English sentence. This GPT translation is performed on a per-sentence basis because, as detailed in Table~\ref{table:GPT_model_summary}, each of these models have a maximum token context window that is relatively small compared to the size of a complete document (which could be comprised of tens or hundreds of thousands of tokens). Therefore, we apply the translations for each individual sentence primarily to mitigate the problems that arise if we attempt to generate text that has a longer token length than what the GPT model was designed to process. In order to assess the quality of the translations, four metrics are used; METEOR~\cite{banarjee2005}, chrF (CHaRacter-level F-score)~\cite{Popovic2015chrFCN}, BLEU (Bilingual Evaluation Understudy)~\cite{Papineni02bleu, lin-och-2004-orange}, GLEU (General Language Understanding Evaluation)~\cite{mutton2007gleu}. The METEOR, GLEU, BLEU, chrF and metrics are computed using NLTK~\cite{bird2009natural}, using all default hyper-parameters. The metrics are computed after the reference sentence and the translated sentence have been tokenized, all punctuation is removed, and all text is made lower case in order to strictly evaluate the words used for the translation. All four of these metrics are defined to be in between (or equal to) $[0, 1]$, where $1$ indicates the translations completely agree and $0$ indicates the translated document shares no overlap with the original reference document. Note that even high-quality human translations do not guarantee a score of $1$ for all four metrics; generally, it is a difficult task to capture language translation quality~\cite{wu2016googles, math11041006}. Additionally, the multi-language dataset that is used in this study is a collection of TED Talk video transcripts, which themselves are not guaranteed to always be accurate. Therefore, when analyzing the translation quality metrics, we should not always expect to be able to reach scores of $1$, but rather we should be aiming to get closer to $1$ than $0$. 

Minimal GPT output postprocessing is applied in the form of removing language-agnostic key phrases from the beginning of the generated text, if it matches certain commonly used phrases that are not the actual content of the translation, such as \emph{This translated text is}. The full list of removed phrases is given in Appendix~\ref{section:appendix_phrase_post_processing}. 

For each sentence (regardless of the language of the text), the following text prompt is used in order to prompt the GPT model to translate the sentence into English text using a one-shot inference call.

\begin{tcolorbox}
Translate the following sentence into clearly written English text. Respond only with the translated text; do not write explanations or justifications in your reply. 

Text to be translated
\end{tcolorbox}

The text that we want translated is put where the phrase \texttt{Text to be translated} is in the above prompt example. This prompt is not changed to instruct the GPT model on what the input language is -- meaning that this automated translation method has the advantage of being entirely language agnostic, specifically meaning that language detection does not need to be applied so as to have the translation be performed correctly. Or more specifically, this is the prompting method that is applied to the GPT models with the aim of benchmarking how well they perform at the task of automated, and language agnostic, sentence-wise translation. All of the experimental results reported in this study use this fixed prompt so as to simplify the data analysis (and the total required compute time). This prompt was chosen based on minimal small experimentation with prompts that performed reasonably well - but better prompts could likely be found.

The GPT model inference is performed using the Python 3 module \emph{transformers}~\cite{wolf2020huggingfaces}, and each model inference call is performed on a single Nvidia A100 GPU~\cite{9361255} with 82 Gigabytes of memory, with CUDA Version 12.4. The text generation calls are performed using the \texttt{pipeline} method in \emph{transformers}~\cite{wolf2020huggingfaces} using all default parameters, except the inference temperature is set to $0.01$ which results in nearly deterministic output where the chosen token at each step of the model is very likely to be the highest probability token. The timing of the inference calls is reported using wall-clock time to generate the translation of each sentence. Importantly, multiple inference calls were performed on several GPUs concurrently - although the computations were independent, the timing statistics that are reported may be slightly greater than what could be achieved on a completed isolated computing platform with no concurrent GPU computations.

Finally, the translation quality from the GPT models is compared against automated translation (performed by supplying only the target language of English) using \texttt{Google translate}. This is performed using a python 3 library~\cite{python_google_translate} that calls the \texttt{Google translate} API. In cases where the output from the Google API is None, the ``translated'' text is set to an empty string (this happened for a couple of sentences, but was not very common).

\begin{figure}[h!]
    \centering
    \includegraphics[width=0.47\textwidth]{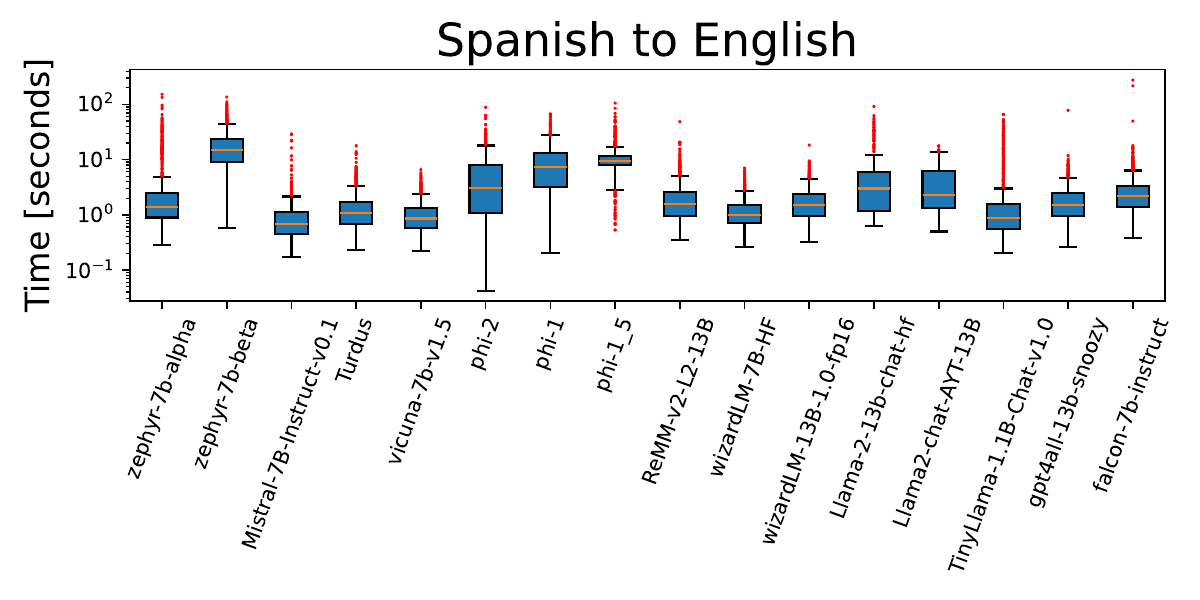}\\
    \includegraphics[width=0.47\textwidth]{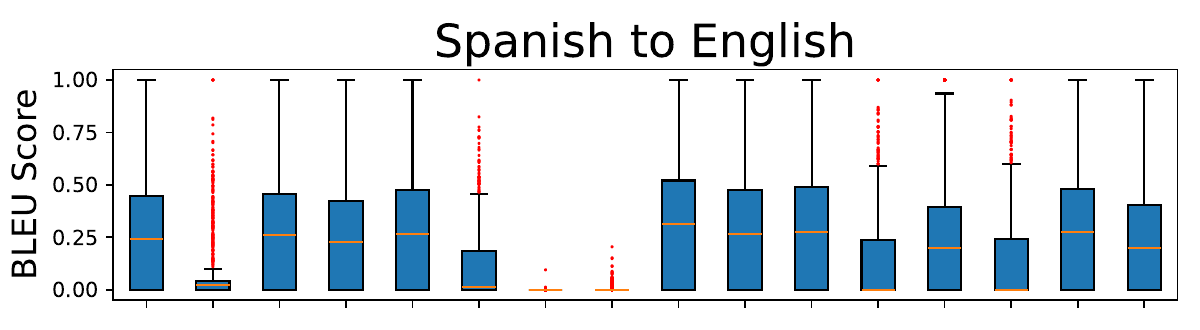}
    \includegraphics[width=0.47\textwidth]{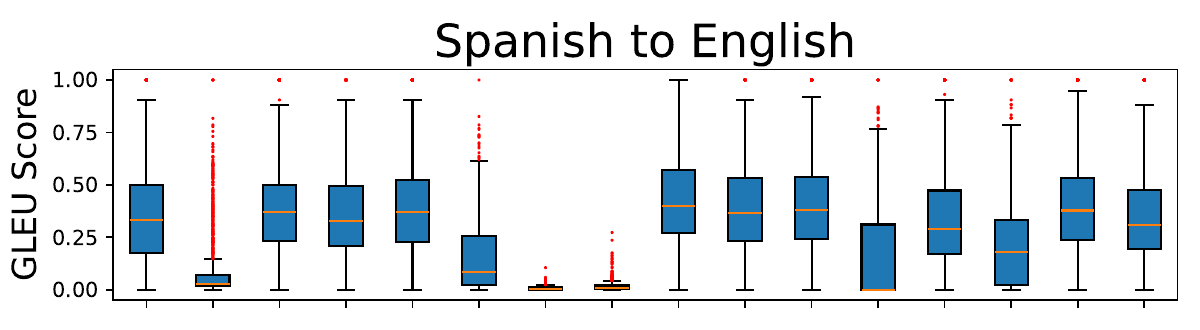}
    \includegraphics[width=0.47\textwidth]{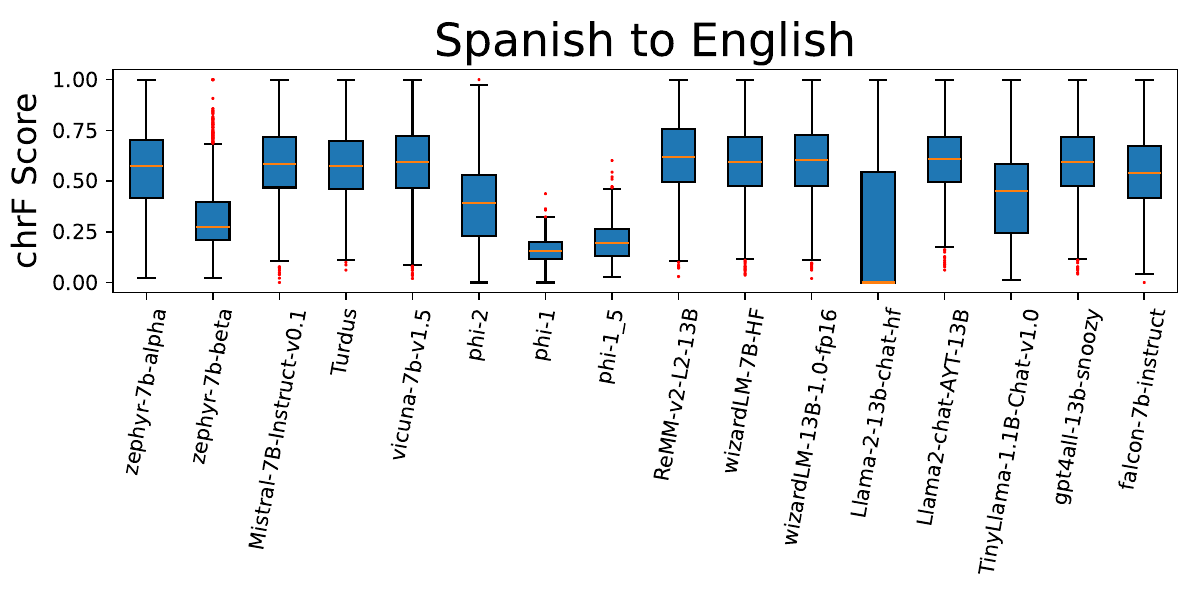}
    \includegraphics[width=0.47\textwidth]{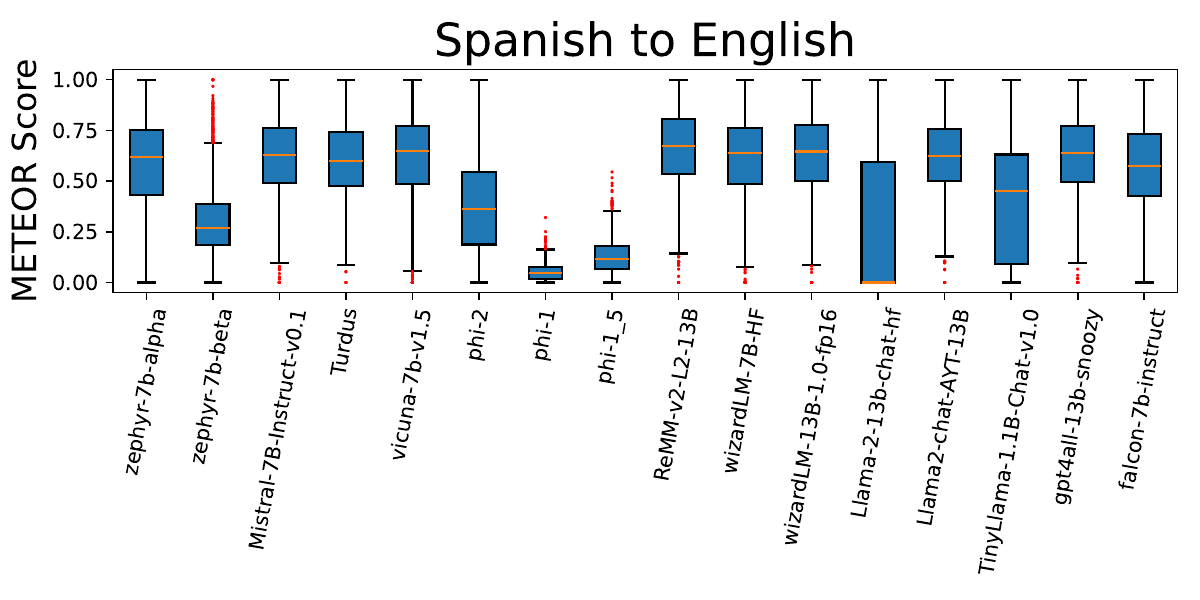}
    \caption{Spanish-to-English dataset per-sentence translation quality and timing statistics for each of the $16$ GPT models. Timing is reported in the top sub-plot, on a log scale y-axis, using direct wall-clock compute time to produce the generated text per sentence. Datapoints which are smaller in the time plot mean that the GTP model output took less wall-clock time to generate. The bottom four sub-plots report the distribution of language quality metrics (one datapoint for each sentence), using the four different language quality measures. For all four of the language quality translation plots, scores closer to $1$ indicate better translation quality, and scores near $0$ indicate bad translation quality. All distributions are shown as box-plot representations, where the red dots indicate outlier points and the blue rectangles indicate the region between the first and third quartile's, the orange line denotes the median. }
    \label{fig:Spanish_translate_stats}
\end{figure}

\begin{figure}[h!]
    \centering
    \includegraphics[width=0.47\textwidth]{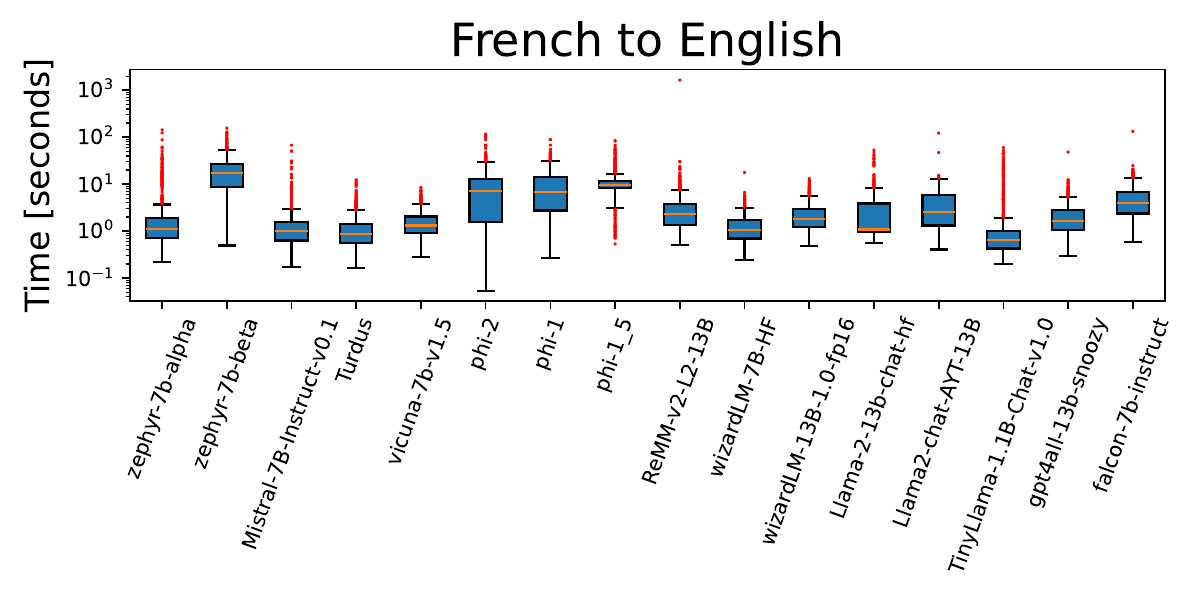}\\
    \includegraphics[width=0.47\textwidth]{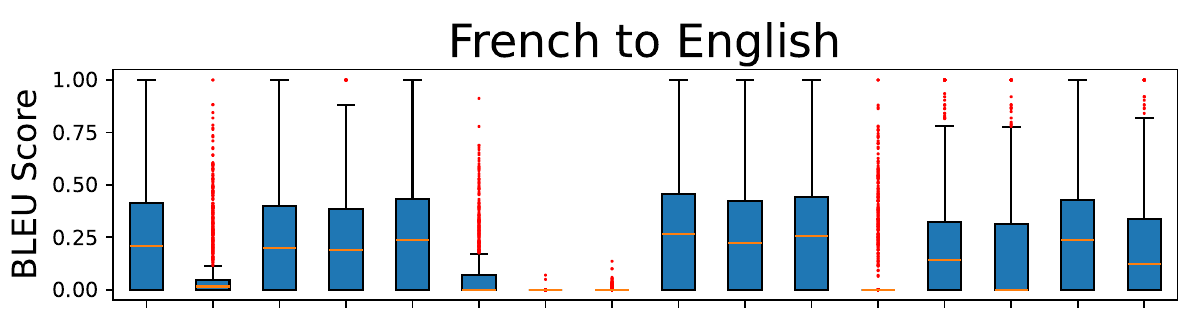}
    \includegraphics[width=0.47\textwidth]{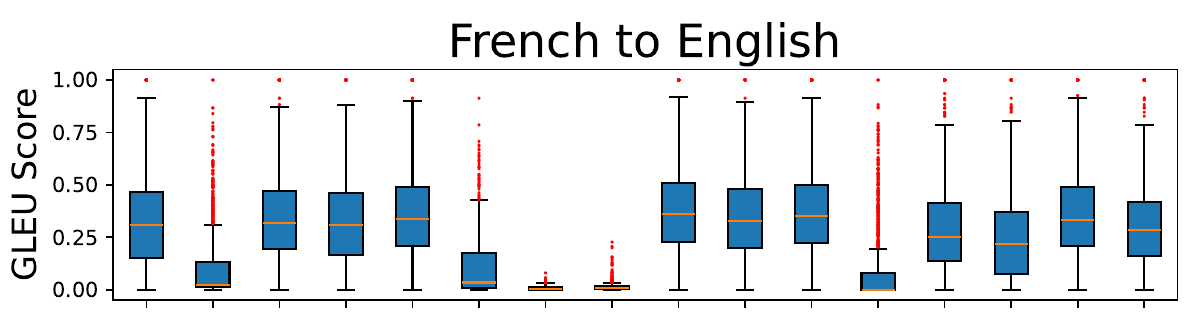}
    \includegraphics[width=0.47\textwidth]{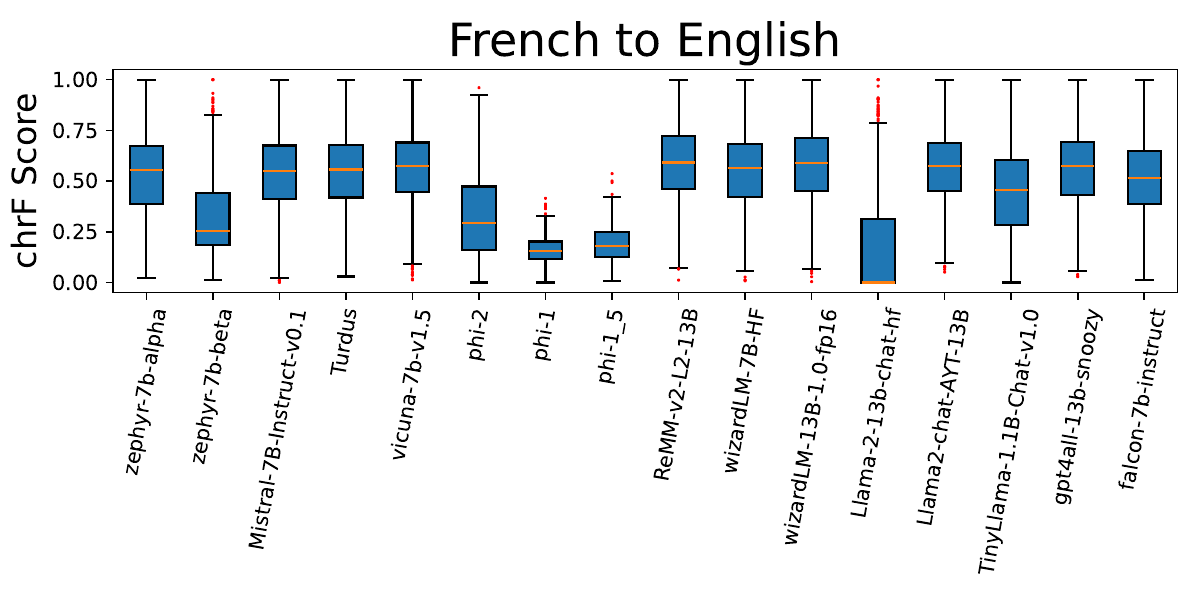}
    \includegraphics[width=0.47\textwidth]{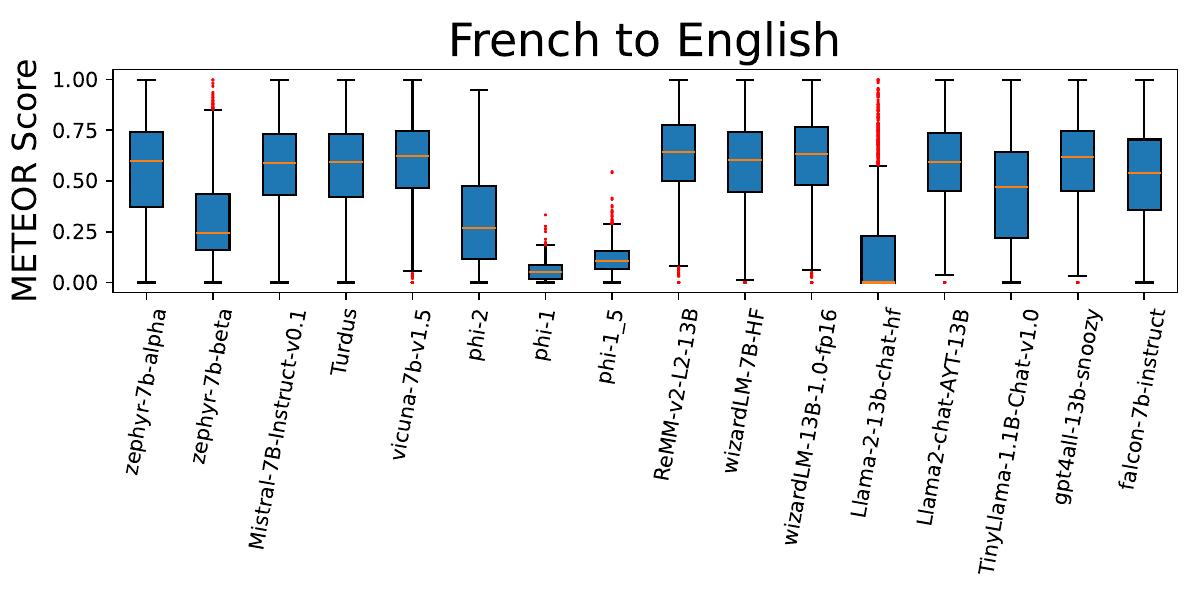}
    \caption{French-to-English dataset per-sentence translation quality and timing statistics for each of the $16$ GPT models. Timing is reported in the top sub-plot, on a log scale y-axis, using direct wall-clock compute time to produce the generated text per sentence. Datapoints which are smaller in the time plot mean that the GTP model output took less wall-clock time to generate. The bottom four sub-plots report the distribution of language quality metrics (one datapoint for each sentence), using the four different language quality measures. For all four of the language quality translation plots, scores closer to $1$ indicate better translation quality, and scores near $0$ indicate bad translation quality. All distributions are shown as box-plot representations, where the red dots indicate outlier points and the blue rectangles indicate the region between the first and third quartile's, the orange line denotes the median. }
    \label{fig:French_translate_stats}
\end{figure}

\section{Results}
\label{section:results}

Table~\ref{table:GPT_translation_results} summarizes the best performing GPT models for translating $50$ foreign languages, using the four different translation metrics. Table~\ref{table:GPT_translation_results} reports the best mean translation quality per sentence which are given by the rounded value to $3$ decimal places. The final aggregate metric of the best performing GPT model across all languages, for each of the language quality metrics, is computed as the mean of the vector of all $50$ language scores (this aggregate metric is not weighted by the different amounts of sentences that were translated in the language dataset). 

Notably, of the $16$ GPT models, only a small subset of these was the best performing for any tuple of language and translation quality metric. Specifically, the models that had the best mean scores (for any combination of language and translation quality measure) were; \texttt{ReMM-v2-L2-13B}, \texttt{Turdus}, \texttt{Llama2-chat-AYT-13B}, \texttt{wizardLM-13B-1.0-fp16}, and \texttt{zephyr-7b-alpha}. \texttt{ReMM-v2-L2-13B} was the best performing model overall. Importantly, for each language, the translation scores shown in Table~\ref{table:GPT_translation_results} were computed on the exact same translated sentences, but the best performing GPT model was not always the same across the $4$ translation quality metrics. 

On average, the best performing of the $16$ GPT models were not always able to generate good translations. The languages that the GPT models scored the lowest on were Mongolian, Burmese, Kazakh, Kurdish, Armenian, and Georgian.

\begin{figure}[th!]
    \centering
    \includegraphics[width=0.47\textwidth]{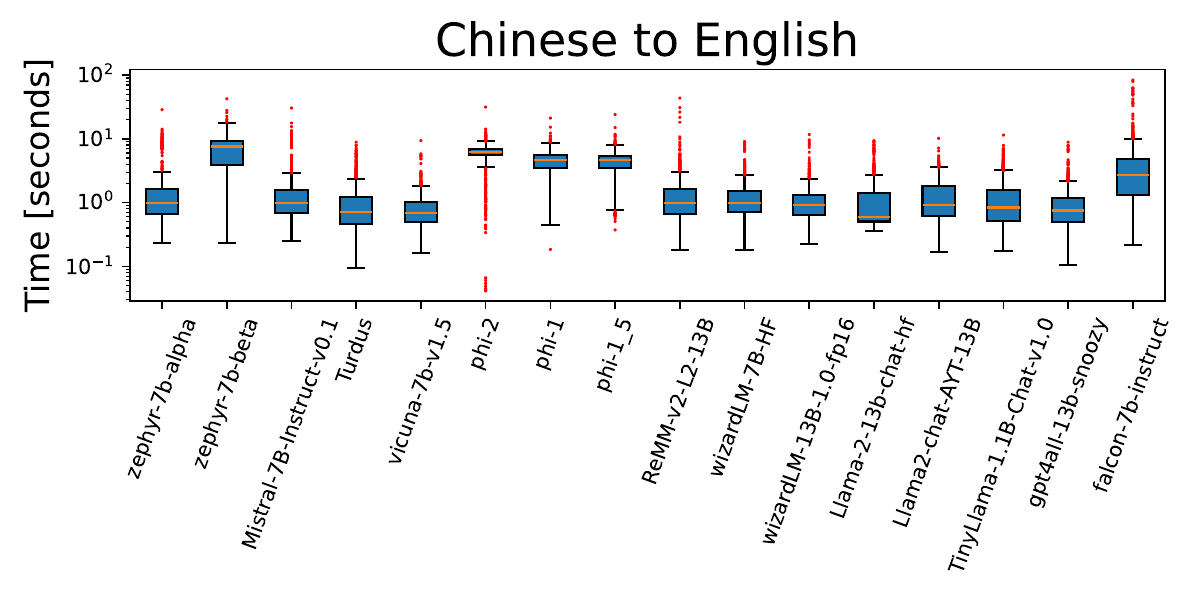}\\
    \includegraphics[width=0.47\textwidth]{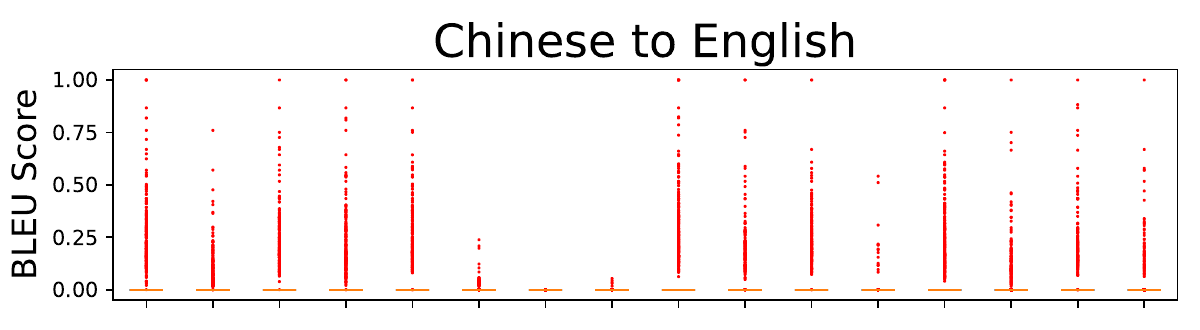}
    \includegraphics[width=0.47\textwidth]{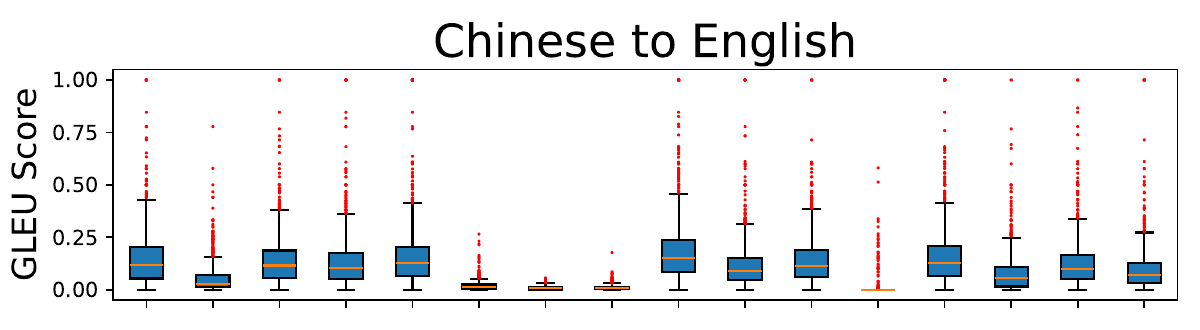}
    \includegraphics[width=0.47\textwidth]{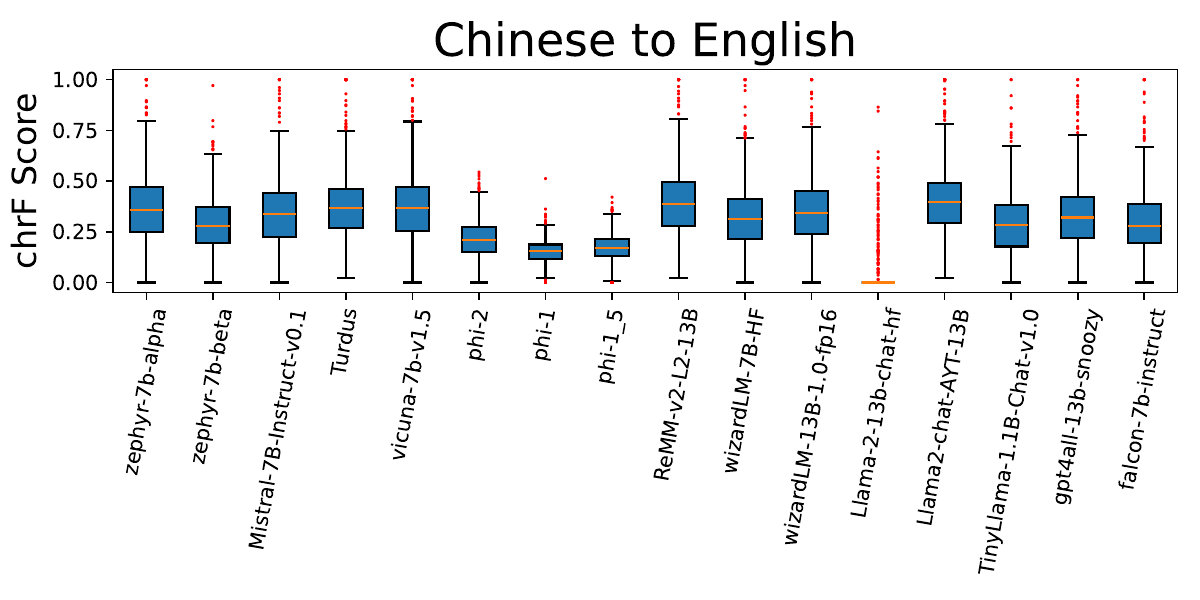}
    \includegraphics[width=0.47\textwidth]{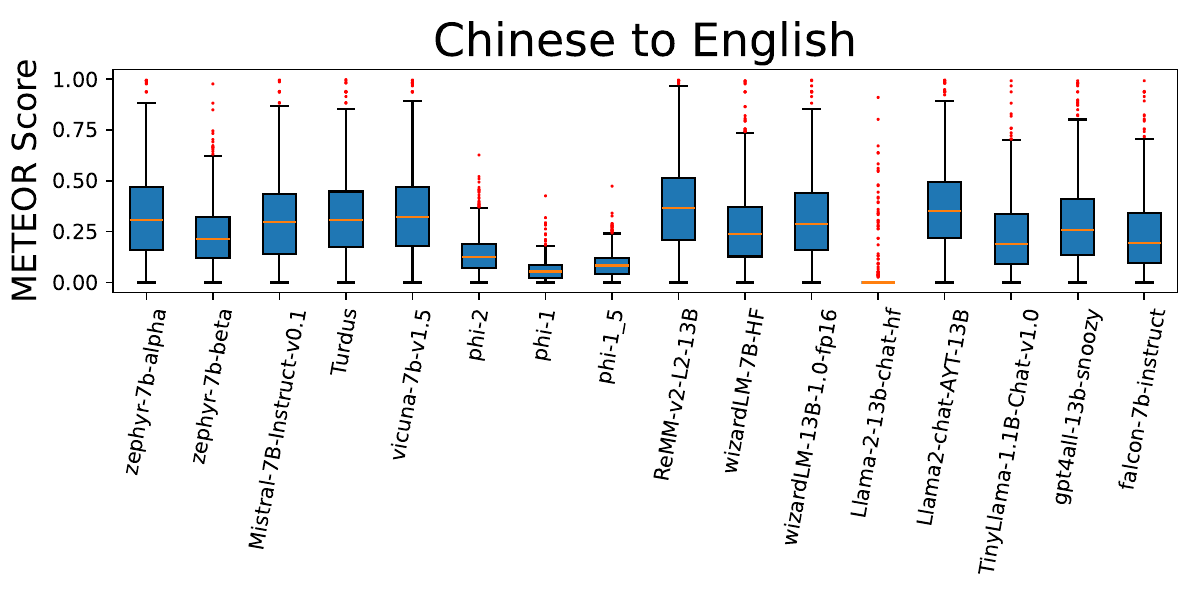}
    \caption{Chinese-to-English dataset per-sentence translation quality and timing statistics }
    \label{fig:Chinese_translate_stats}
\end{figure}

\begin{figure}[th!]
    \centering
    \includegraphics[width=0.47\textwidth]{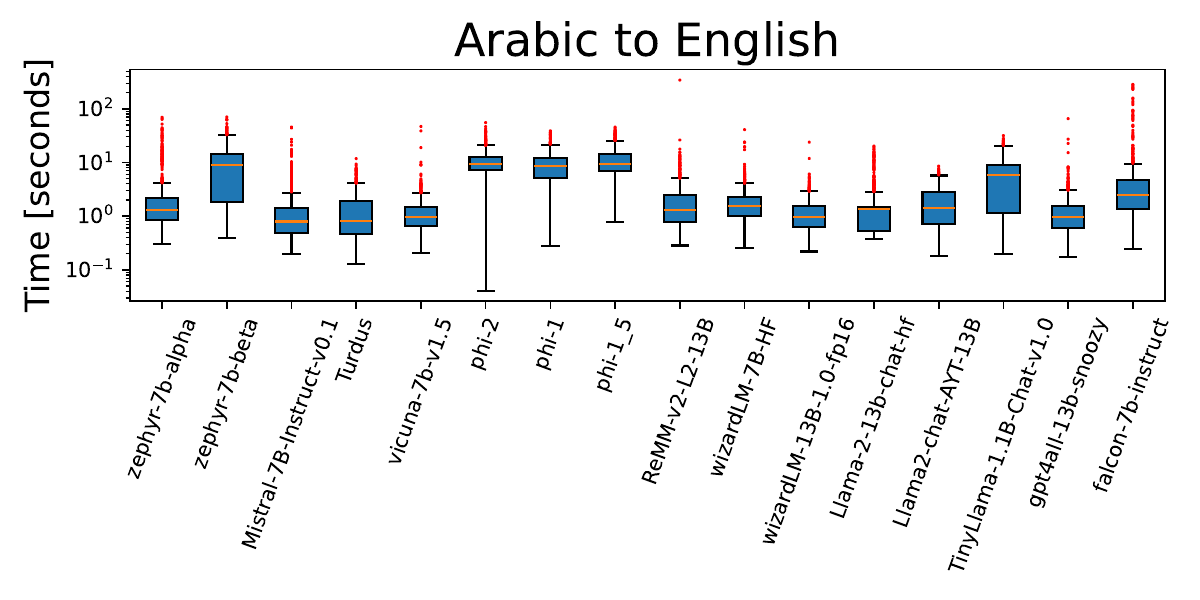}\\
    \includegraphics[width=0.47\textwidth]{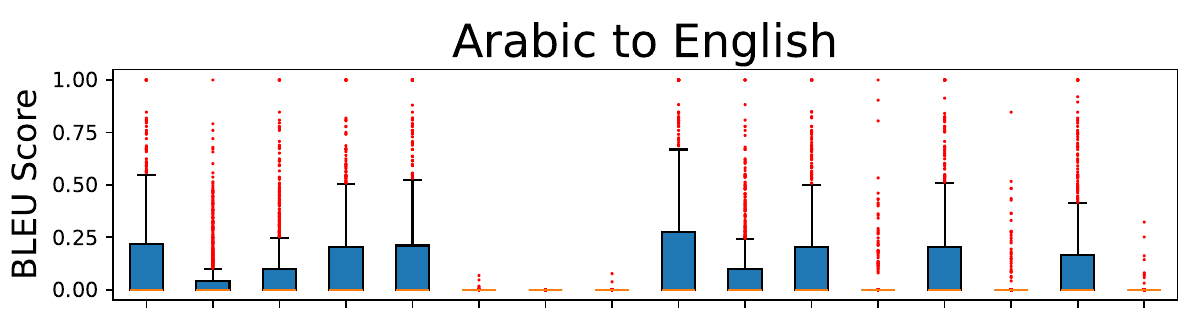}
    \includegraphics[width=0.47\textwidth]{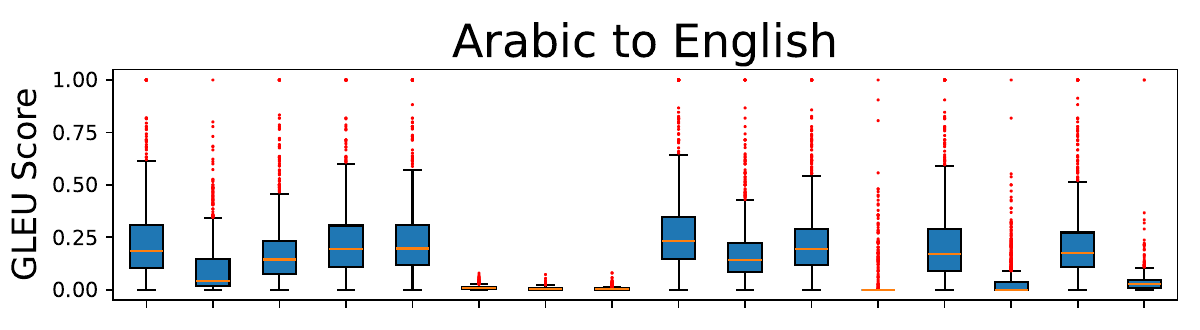}
    \includegraphics[width=0.47\textwidth]{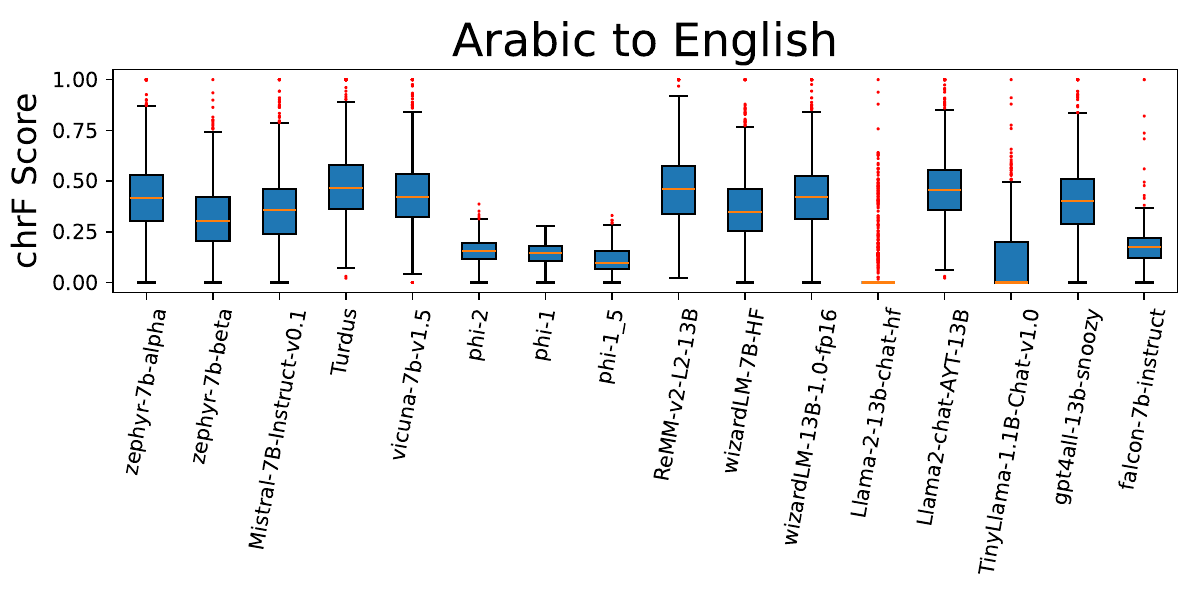}
    \includegraphics[width=0.47\textwidth]{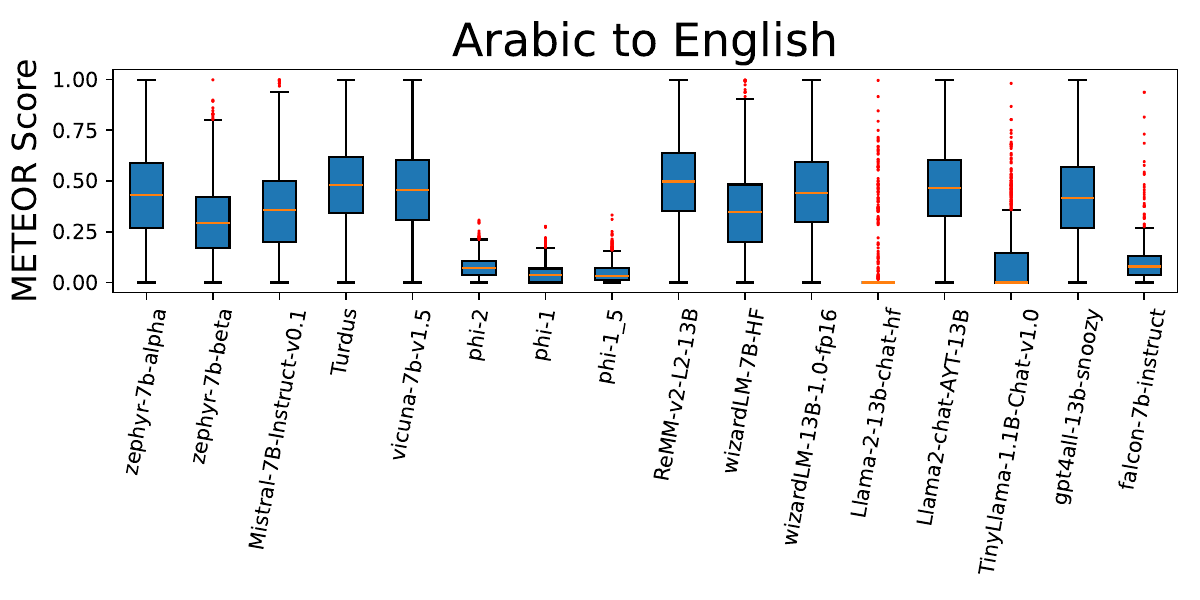}
    \caption{Arabic-to-English dataset per-sentence translation quality and timing statistics  }
    \label{fig:Arabic_translate_stats}
\end{figure}

\begin{figure}[th!]
    \centering
    \includegraphics[width=0.47\textwidth]{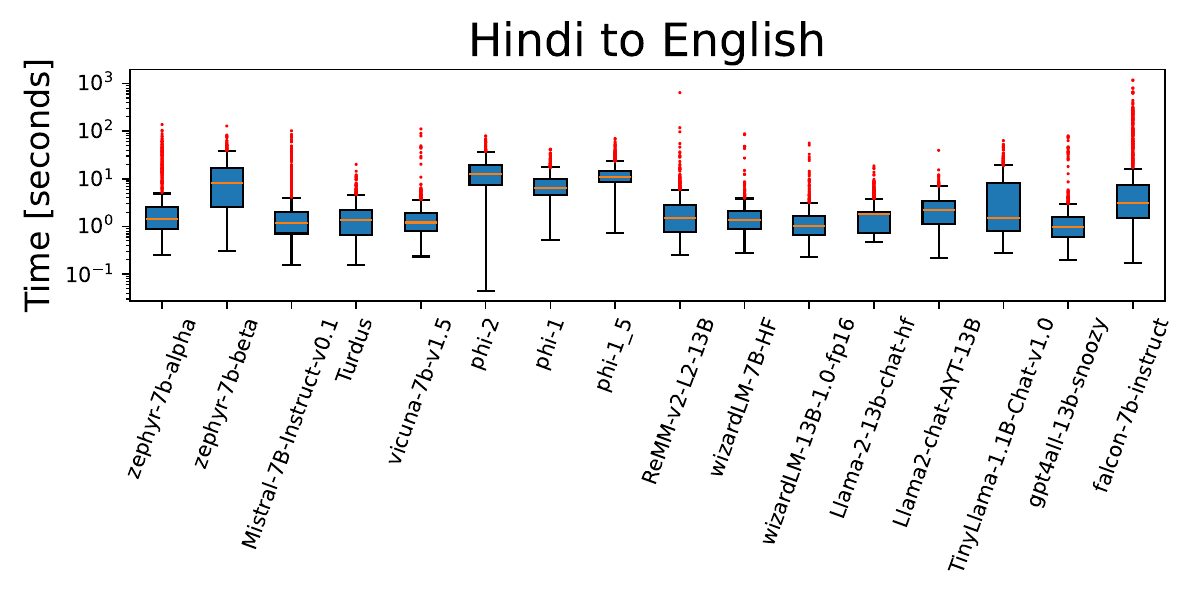}\\
    \includegraphics[width=0.47\textwidth]{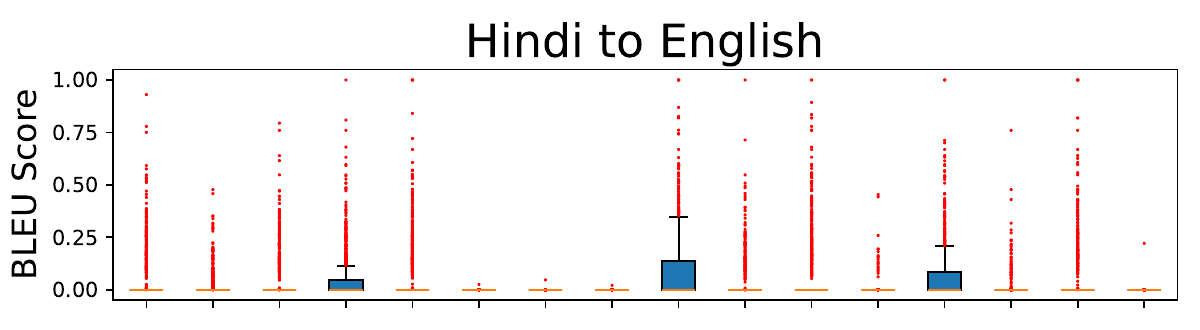}
    \includegraphics[width=0.47\textwidth]{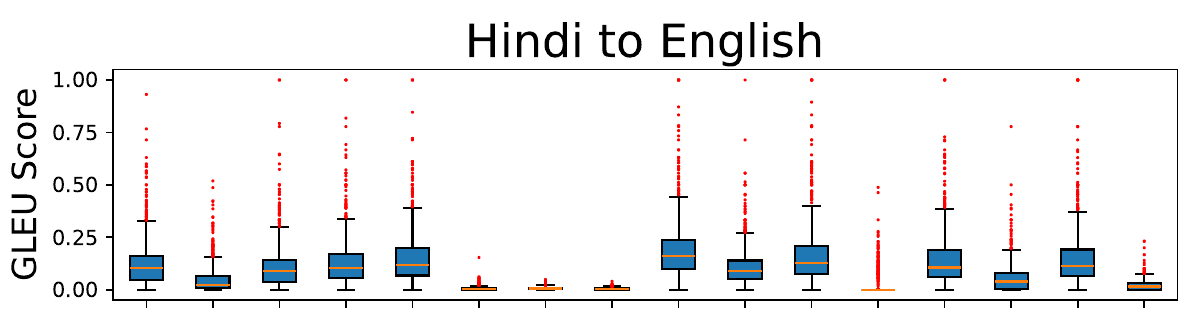}
    \includegraphics[width=0.47\textwidth]{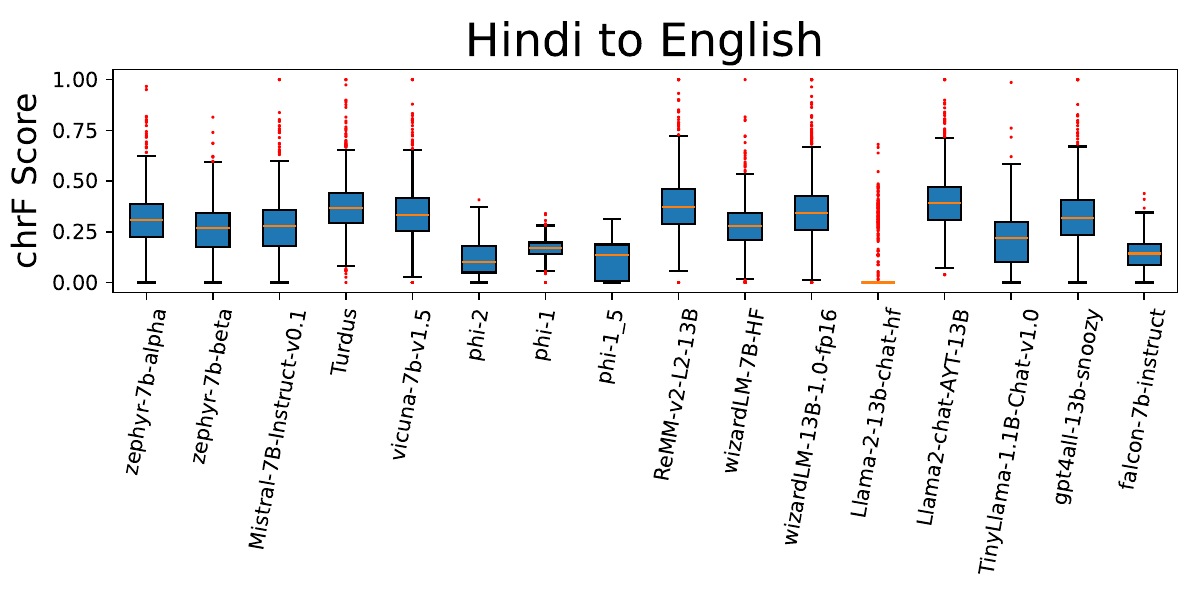}
    \includegraphics[width=0.47\textwidth]{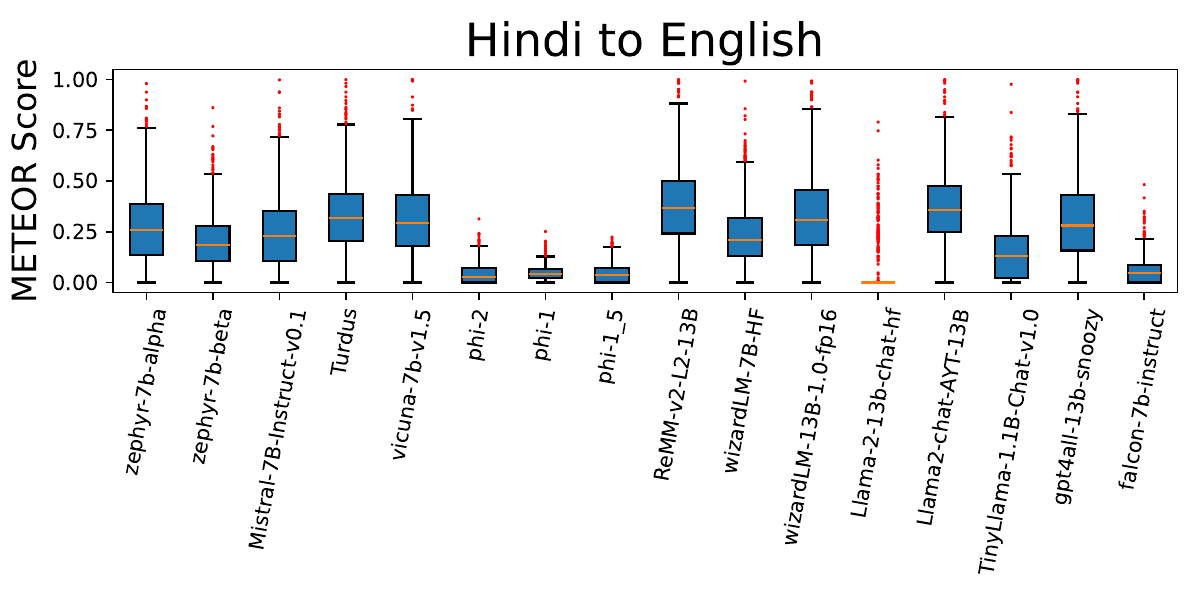}
    \caption{Hindi-to-English dataset per-sentence translation quality and timing statistics  }
    \label{fig:Hindi_translate_stats}
\end{figure}

\begin{figure}[th!]
    \centering
    \includegraphics[width=0.47\textwidth]{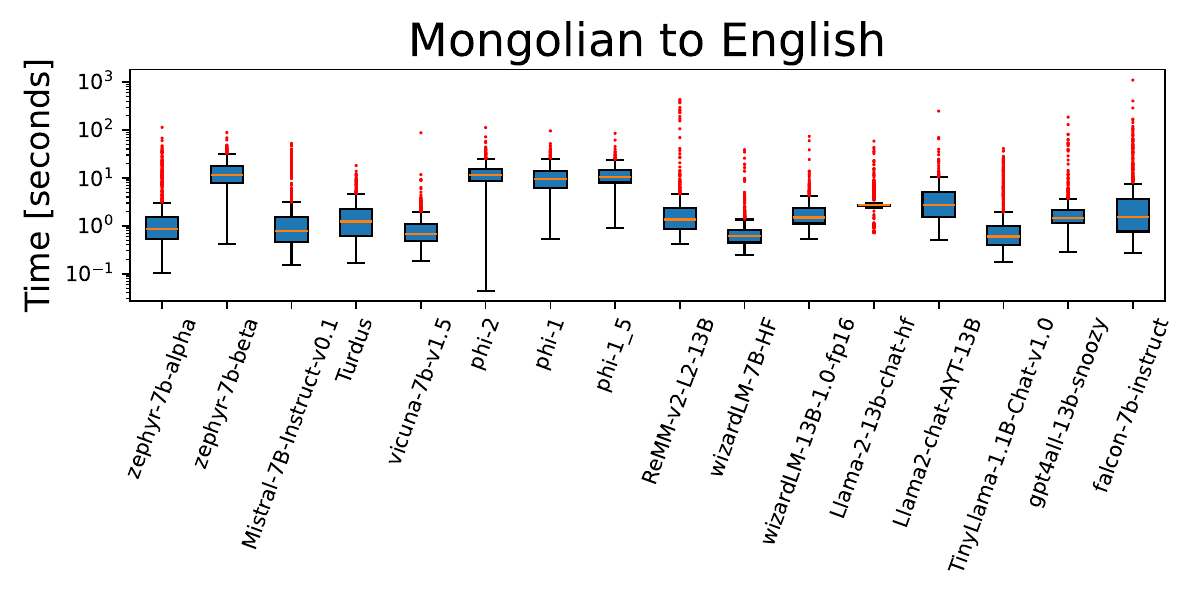}\\
    \includegraphics[width=0.47\textwidth]{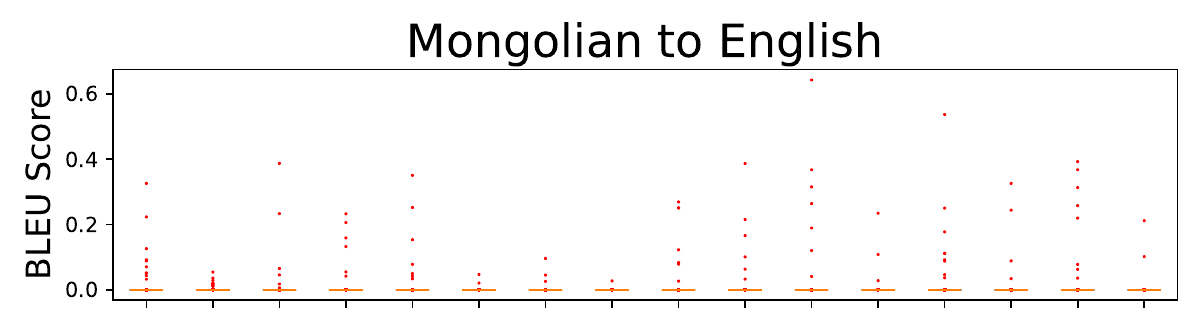}
    \includegraphics[width=0.47\textwidth]{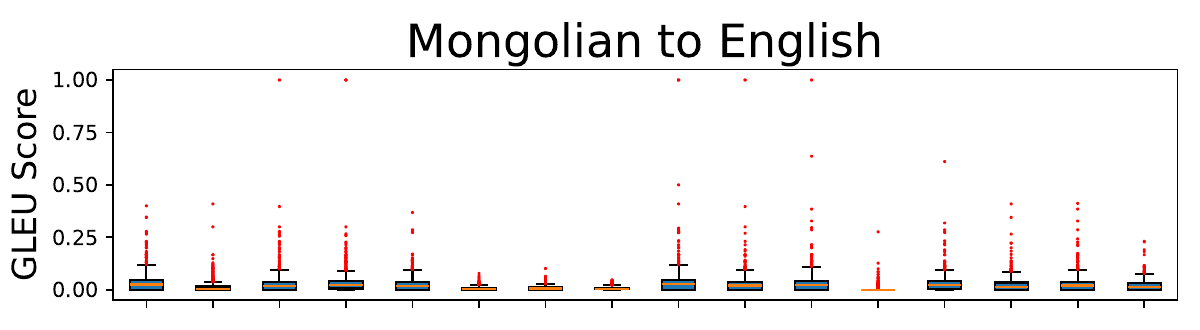}
    \includegraphics[width=0.47\textwidth]{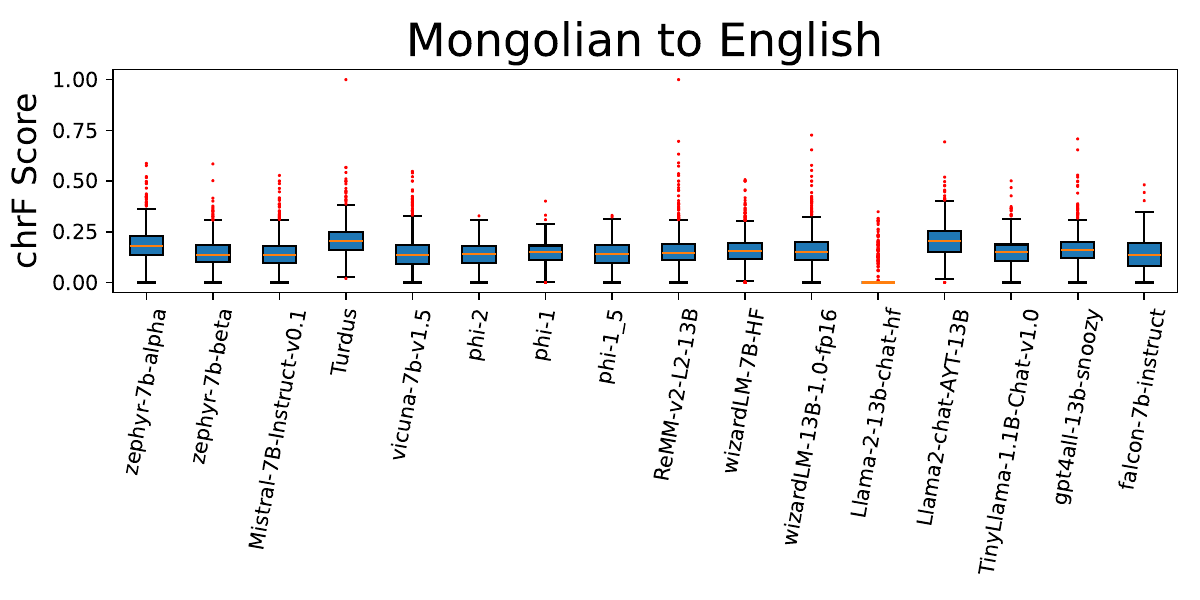}
    \includegraphics[width=0.47\textwidth]{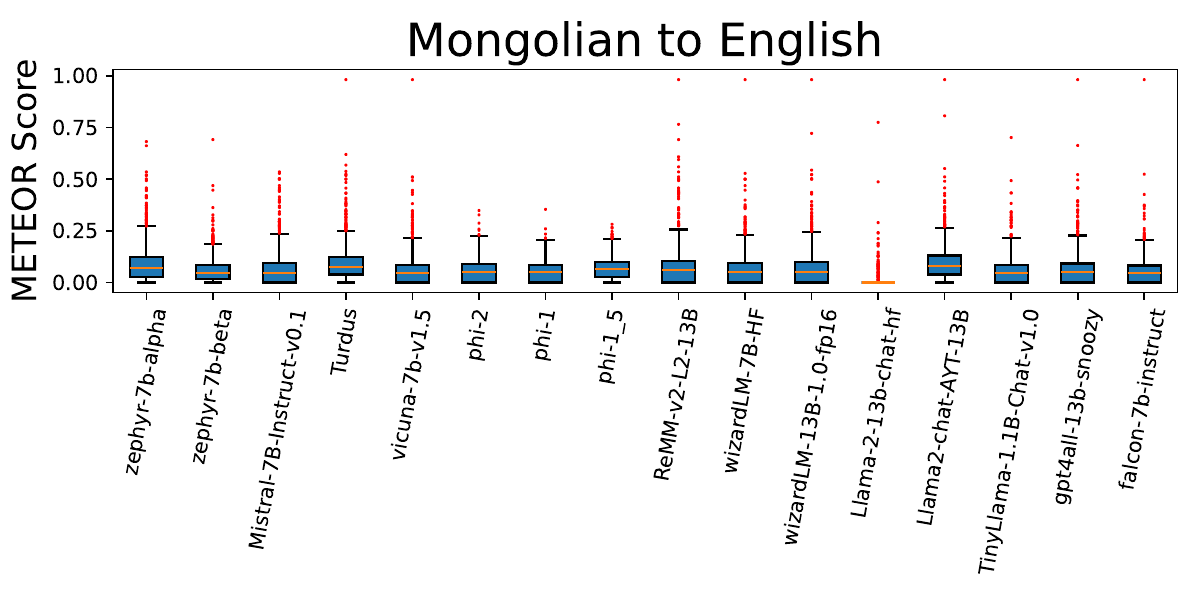}
    \caption{Mongolian-to-English dataset per-sentence translation quality and timing statistics  }
    \label{fig:Mongolian_translate_stats}
\end{figure}

\begin{figure}[th!]
    \centering
    \includegraphics[width=0.47\textwidth]{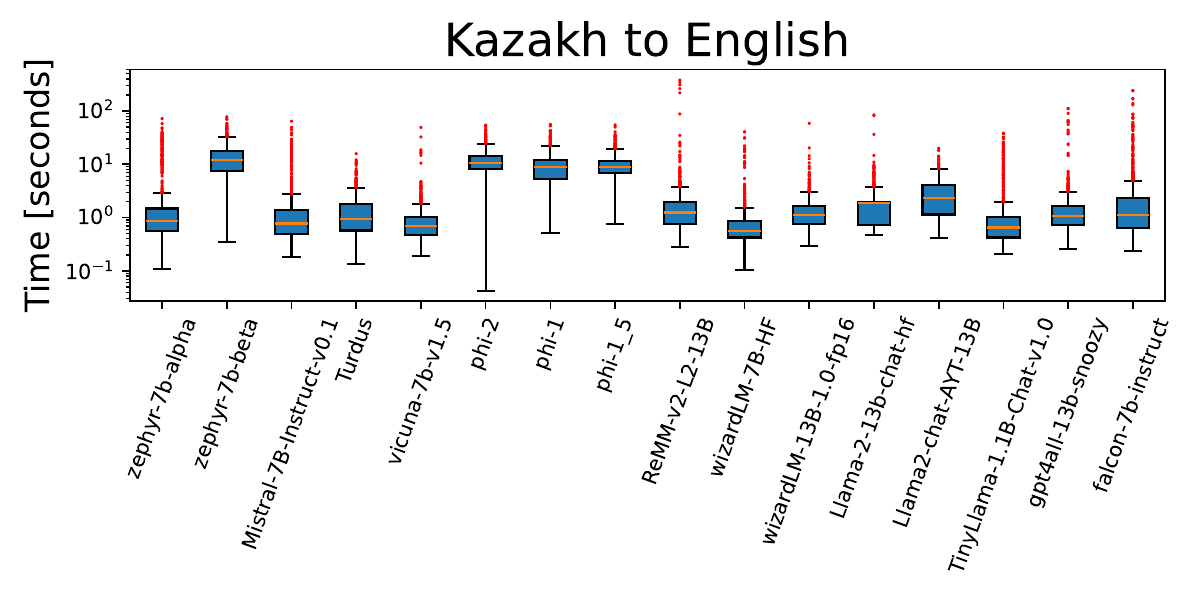}\\
    \includegraphics[width=0.47\textwidth]{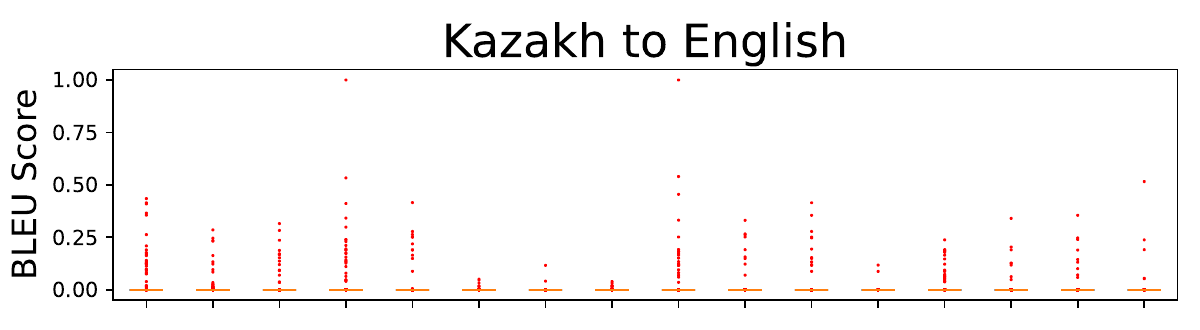}
    \includegraphics[width=0.47\textwidth]{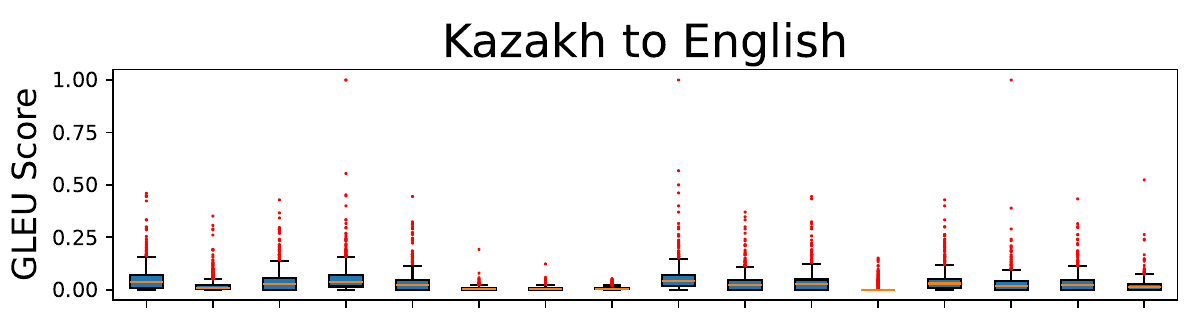}
    \includegraphics[width=0.47\textwidth]{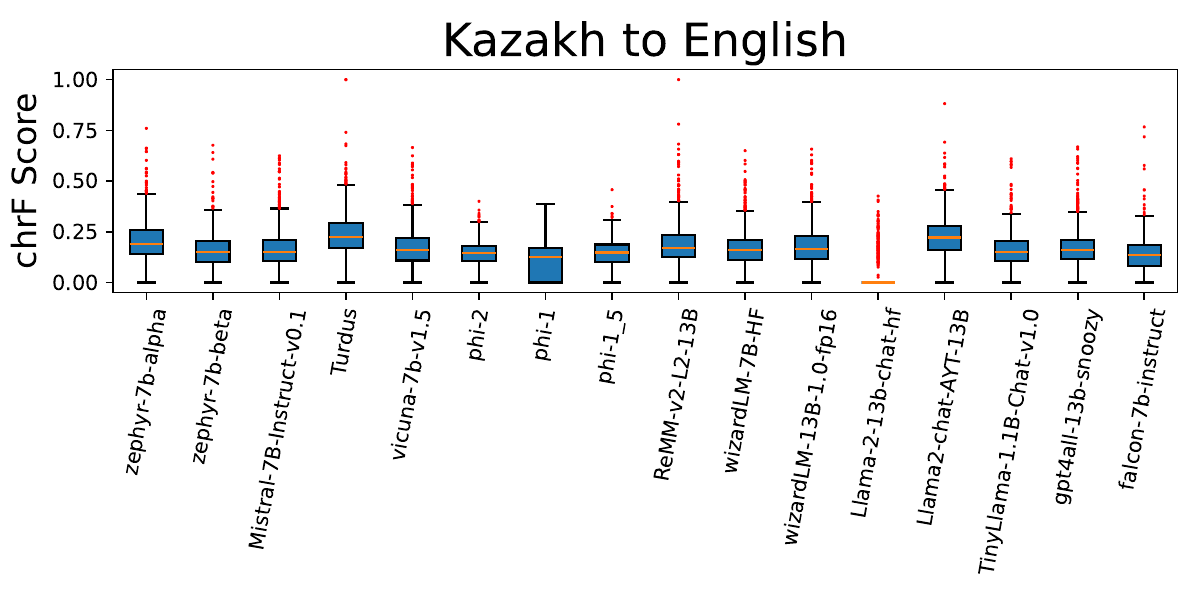}
    \includegraphics[width=0.47\textwidth]{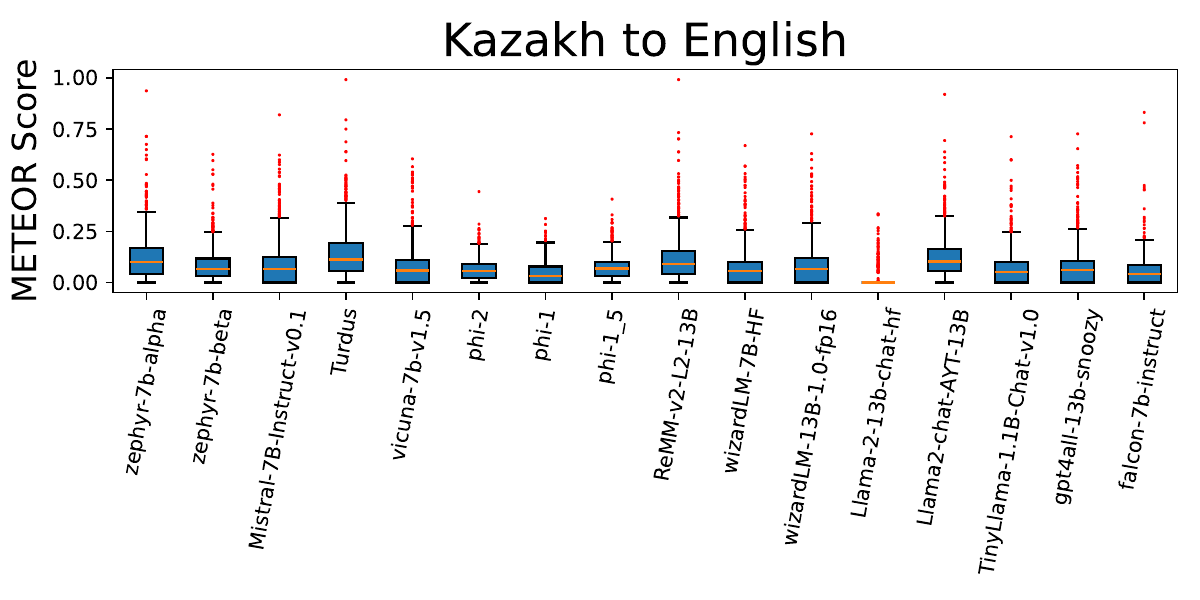}
    \caption{Kazakh-to-English dataset per-sentence translation quality and timing statistics  }
    \label{fig:Kazakh_translate_stats}
\end{figure}

\begin{figure}[th!]
    \centering
    \includegraphics[width=0.47\textwidth]{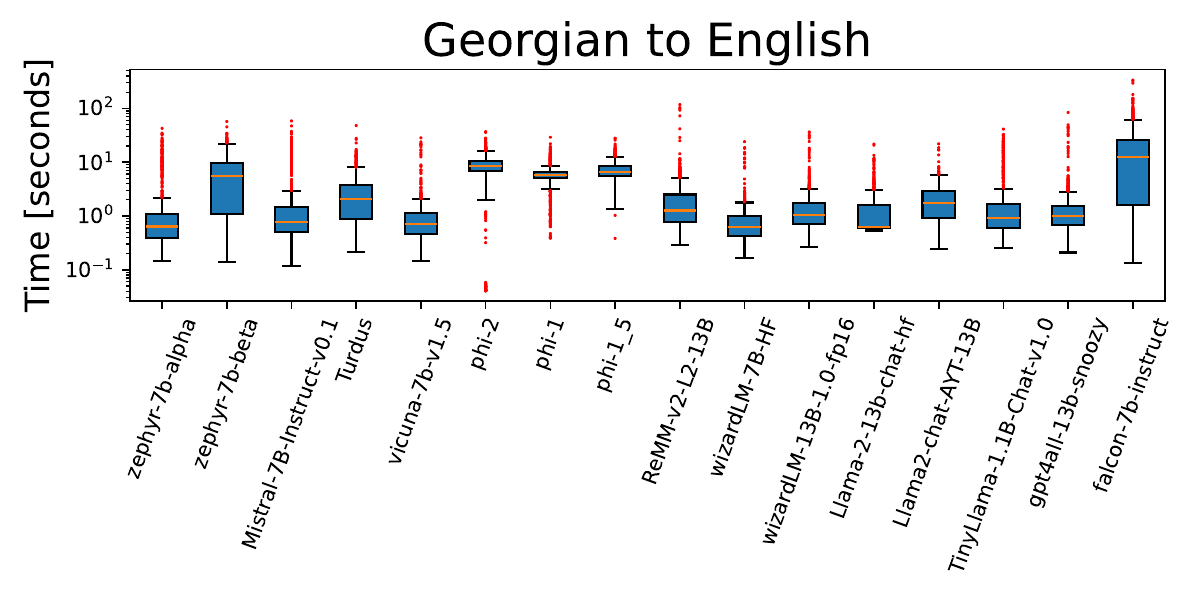}\\
    \includegraphics[width=0.47\textwidth]{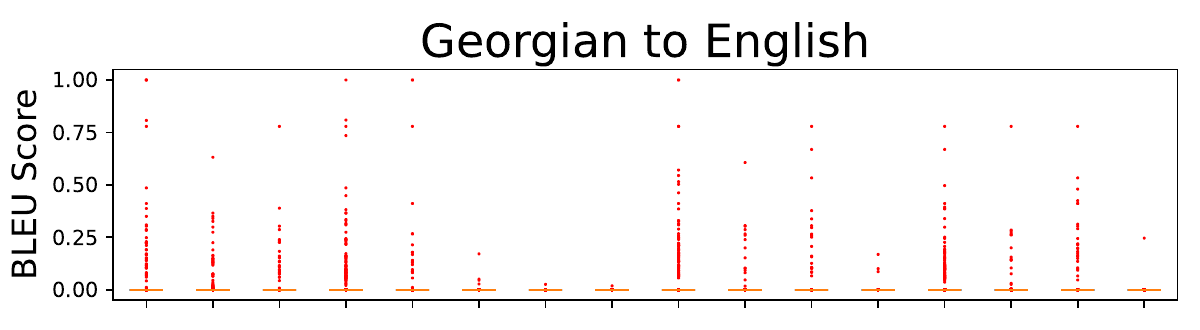}
    \includegraphics[width=0.47\textwidth]{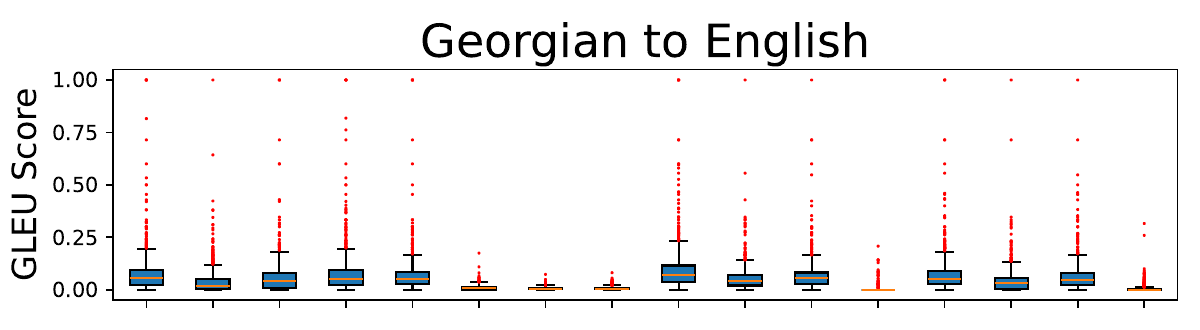}
    \includegraphics[width=0.47\textwidth]{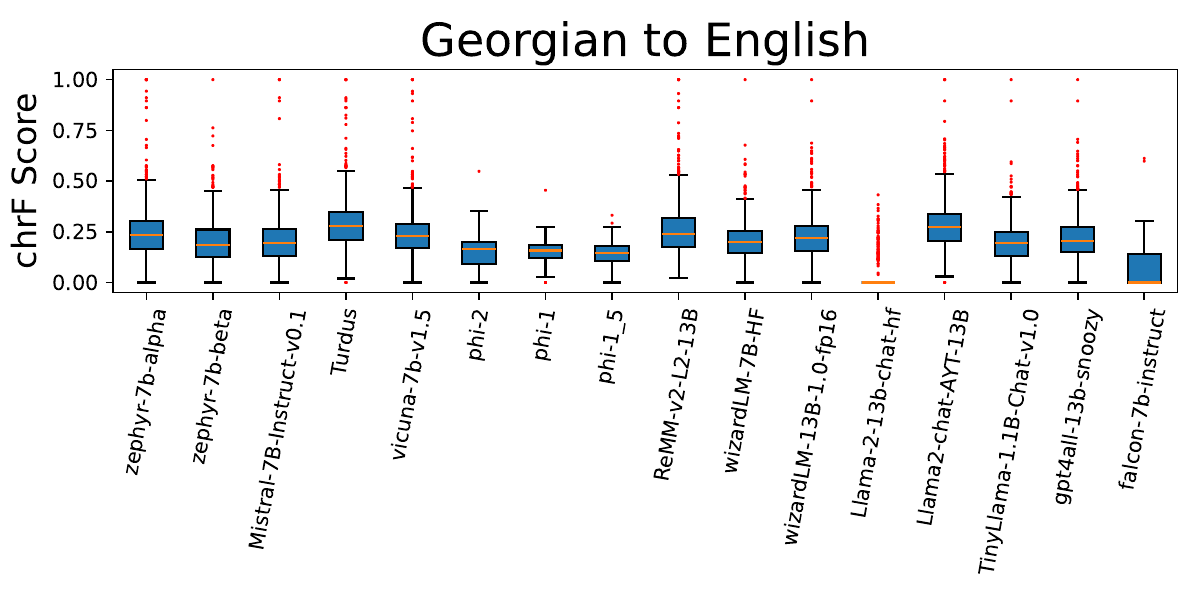}
    \includegraphics[width=0.47\textwidth]{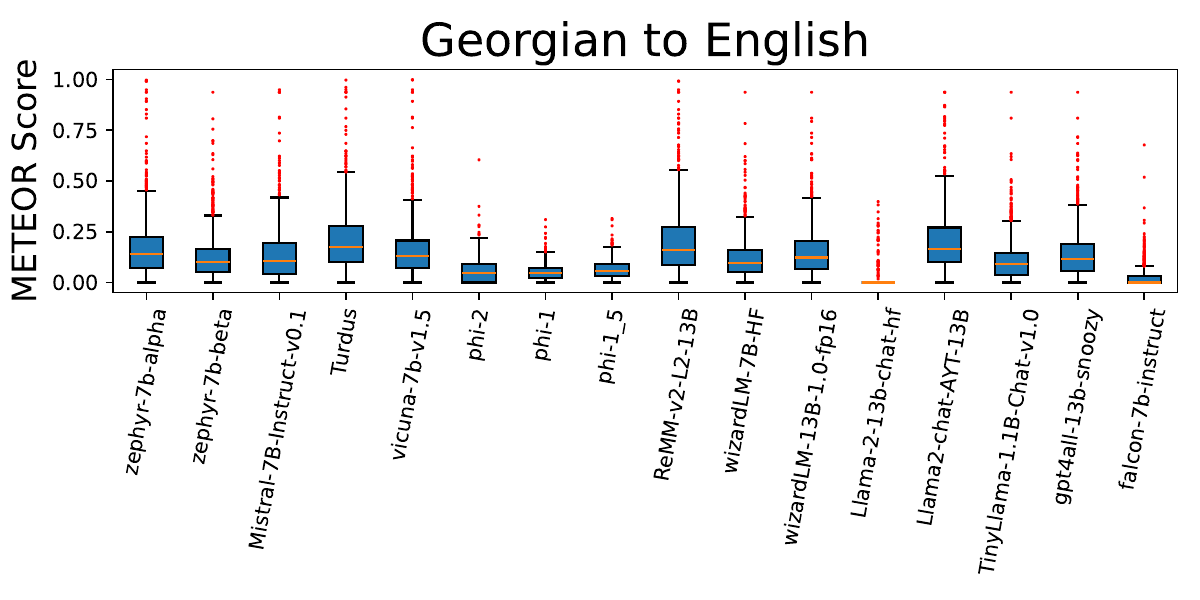}
    \caption{Georgian-to-English dataset per-sentence translation quality and timing statistics  }
    \label{fig:Georgian_translate_stats}
\end{figure}

Figures \ref{fig:Spanish_translate_stats}, \ref{fig:French_translate_stats}, \ref{fig:Chinese_translate_stats}, \ref{fig:Arabic_translate_stats}, and \ref{fig:Hindi_translate_stats} shows detailed performance and wall-clock timing statistics for Spanish, French, Chinese, Arabic, and Hindi -- which are the five most commonly spoken natural languages (besides English). These plots are representative of the expected language translation quality for the most commonly used languages. Figures \ref{fig:Mongolian_translate_stats}, \ref{fig:Kazakh_translate_stats}, \ref{fig:Georgian_translate_stats} show detailed per-GPT model performance for Mongolian, Kazakh, and Georgian, which were languages for which the GPT models were unable to produce good translations for, on average. The detailed per-GPT translation metrics and timing statistics for translating all of the other $50$ languages into English are enumerated in Appendix \ref{section:appendix_translation_stats}.

There are a number of consistent trends seen in the translation quality box-plot figures - namely that the three \texttt{phi} models generally have very low accuracy. \texttt{Llama-2-13b-chat-hf}, notably, also consistently has very low translation accuracy, which is surprising because nearly all of the best performing models were fine-tuned from Llama-2 models. The mechanism that caused this low accuracy is not clear, but this behavior could be due to the particular prompt that was used and testing other prompts could improve the translation accuracy for future study. 
In terms of translation speed, the slowest GPT models were \texttt{phi-1}, \texttt{phi-2}, \texttt{phi-1\_5}, \texttt{zephyr-7b-beta}, and \texttt{falcon-7b-instruct}.

Table~\ref{table:google_translate_results} shows the mean translation quality metrics, for the four language metrics, across all $50$ languages being translated into English, using \texttt{Google translate}. The same test sentences translated by the GPT models, for each language, were also translated using \texttt{Google translate} - therefore the entries in Table~\ref{table:google_translate_results} should be compared against the best performing GPT models in Table~\ref{table:GPT_translation_results}. These results show the performance of \texttt{Google translate}, using it as a reasonable performance benchmark for automated machine translation of languages. Interestingly, there were exactly two languages where, for at least one of the language metrics (although, in these cases it was for all four language quality metrics), the best performing GPT model had better mean sentence translation quality than google translate. These two languages were French and Chinese. For all other languages, either the best performing GPT model was definitively worse at translating, or was comparable to within a small margin. The languages for which the best performing GPT model and google translate performed marginally the same were German, Spanish, Italian, Russian, Korean, Serbian, Japanese, Ukrainian, Vietnamese, and Bosnian.

\begin{table*}[h!]
\centering
\vspace{-1.5cm}
\scalebox{0.77}{
\begin{tabular}{ |c||c|c|c|c| }
 \hline
 \hline
 Language & Best Mean GLEU & Best Mean BLEU & Best Mean chrF  & Best Mean METEOR \\ 
 \hline
 \hline
Arabic & ReMM-v2-L2-13B (0.271) & ReMM-v2-L2-13B (0.157) & Turdus (0.478) & ReMM-v2-L2-13B (0.489)\\ 
\hline
Azerbaijani & ReMM-v2-L2-13B (0.121) & ReMM-v2-L2-13B (0.035) & Turdus (0.316) & Turdus (0.257)\\ 
\hline
Belarusian & ReMM-v2-L2-13B (0.2) & ReMM-v2-L2-13B (0.093) & Llama2-chat-AYT-13B (0.392) & ReMM-v2-L2-13B (0.369)\\ 
\hline
Bulgarian & ReMM-v2-L2-13B (0.363) & ReMM-v2-L2-13B (0.246) & ReMM-v2-L2-13B (0.554) & ReMM-v2-L2-13B (0.582)\\ 
\hline
Bengali & ReMM-v2-L2-13B (0.121) & ReMM-v2-L2-13B (0.028) & Turdus (0.335) & Turdus (0.282)\\ 
\hline
Bosnian & ReMM-v2-L2-13B (0.342) & ReMM-v2-L2-13B (0.229) & Llama2-chat-AYT-13B (0.546) & ReMM-v2-L2-13B (0.559)\\ 
\hline
Czech & ReMM-v2-L2-13B (0.326) & ReMM-v2-L2-13B (0.207) & Llama2-chat-AYT-13B (0.518) & ReMM-v2-L2-13B (0.546)\\ 
\hline
Danish & ReMM-v2-L2-13B (0.454) & ReMM-v2-L2-13B (0.368) & ReMM-v2-L2-13B (0.632) & ReMM-v2-L2-13B (0.673)\\ 
\hline
German & ReMM-v2-L2-13B (0.361) & ReMM-v2-L2-13B (0.238) & ReMM-v2-L2-13B (0.555) & ReMM-v2-L2-13B (0.579)\\ 
\hline
Greek & ReMM-v2-L2-13B (0.294) & ReMM-v2-L2-13B (0.176) & ReMM-v2-L2-13B (0.467) & ReMM-v2-L2-13B (0.499)\\ 
\hline
Spanish & ReMM-v2-L2-13B (0.443) & ReMM-v2-L2-13B (0.331) & ReMM-v2-L2-13B (0.624) & ReMM-v2-L2-13B (0.658)\\ 
\hline
Estonian & ReMM-v2-L2-13B (0.134) & Turdus (0.044) & Turdus (0.349) & Turdus (0.287)\\ 
\hline
Persian & ReMM-v2-L2-13B (0.226) & ReMM-v2-L2-13B (0.112) & Llama2-chat-AYT-13B (0.434) & Llama2-chat-AYT-13B (0.428)\\ 
\hline
Finnish & ReMM-v2-L2-13B (0.304) & ReMM-v2-L2-13B (0.192) & ReMM-v2-L2-13B (0.523) & ReMM-v2-L2-13B (0.526)\\ 
\hline
French & ReMM-v2-L2-13B (0.394) & ReMM-v2-L2-13B (0.278) & ReMM-v2-L2-13B (0.585) & ReMM-v2-L2-13B (0.614)\\ 
\hline
Galician & ReMM-v2-L2-13B (0.317) & ReMM-v2-L2-13B (0.192) & Llama2-chat-AYT-13B (0.53) & Llama2-chat-AYT-13B (0.528)\\ 
\hline
Hebrew & ReMM-v2-L2-13B (0.267) & ReMM-v2-L2-13B (0.149) & Turdus (0.463) & ReMM-v2-L2-13B (0.477)\\ 
\hline
Hindi & ReMM-v2-L2-13B (0.191) & ReMM-v2-L2-13B (0.084) & Llama2-chat-AYT-13B (0.396) & ReMM-v2-L2-13B (0.379)\\ 
\hline
Croatian & ReMM-v2-L2-13B (0.345) & ReMM-v2-L2-13B (0.237) & ReMM-v2-L2-13B (0.541) & ReMM-v2-L2-13B (0.561)\\ 
\hline
Hungarian & ReMM-v2-L2-13B (0.285) & ReMM-v2-L2-13B (0.171) & ReMM-v2-L2-13B (0.487) & ReMM-v2-L2-13B (0.495)\\ 
\hline
Armenian & Turdus (0.096) & Turdus (0.019) & Turdus (0.294) & Turdus (0.228)\\ 
\hline
Indonesian & ReMM-v2-L2-13B (0.29) & ReMM-v2-L2-13B (0.166) & Llama2-chat-AYT-13B (0.495) & ReMM-v2-L2-13B (0.514)\\ 
\hline
Italian & ReMM-v2-L2-13B (0.375) & ReMM-v2-L2-13B (0.262) & ReMM-v2-L2-13B (0.561) & ReMM-v2-L2-13B (0.59)\\ 
\hline
Japanese & ReMM-v2-L2-13B (0.186) & ReMM-v2-L2-13B (0.078) & Llama2-chat-AYT-13B (0.403) & Llama2-chat-AYT-13B (0.372)\\ 
\hline
Georgian & ReMM-v2-L2-13B (0.094) & ReMM-v2-L2-13B (0.017) & Turdus (0.292) & Turdus (0.206)\\ 
\hline
Kazakh & Turdus (0.053) & Turdus (0.006) & Turdus (0.237) & Turdus (0.138)\\ 
\hline
Korean & ReMM-v2-L2-13B (0.238) & ReMM-v2-L2-13B (0.116) & ReMM-v2-L2-13B (0.452) & ReMM-v2-L2-13B (0.446)\\ 
\hline
Kurdish & ReMM-v2-L2-13B (0.044) & Turdus (0.002) & Turdus (0.225) & Llama2-chat-AYT-13B (0.12)\\ 
\hline
Lithuanian & Turdus (0.12) & zephyr-7b-alpha (0.037) & Turdus (0.328) & Turdus (0.269)\\ 
\hline
Macedonian & ReMM-v2-L2-13B (0.276) & ReMM-v2-L2-13B (0.152) & Llama2-chat-AYT-13B (0.486) & ReMM-v2-L2-13B (0.479)\\ 
\hline
Mongolian & ReMM-v2-L2-13B (0.038) & wizardLM-13B-1.0-fp16 (0.002) & Turdus (0.21) & Turdus (0.095)\\ 
\hline
Malay & ReMM-v2-L2-13B (0.271) & ReMM-v2-L2-13B (0.157) & ReMM-v2-L2-13B (0.472) & ReMM-v2-L2-13B (0.486)\\ 
\hline
Burmese & ReMM-v2-L2-13B (0.028) & ReMM-v2-L2-13B (0.002) & Turdus (0.228) & Turdus (0.096)\\ 
\hline
Norwegian & ReMM-v2-L2-13B (0.413) & ReMM-v2-L2-13B (0.299) & ReMM-v2-L2-13B (0.597) & ReMM-v2-L2-13B (0.629)\\ 
\hline
Dutch & ReMM-v2-L2-13B (0.387) & ReMM-v2-L2-13B (0.277) & Llama2-chat-AYT-13B (0.561) & ReMM-v2-L2-13B (0.59)\\ 
\hline
Polish & ReMM-v2-L2-13B (0.292) & ReMM-v2-L2-13B (0.177) & Llama2-chat-AYT-13B (0.492) & ReMM-v2-L2-13B (0.494)\\ 
\hline
Portuguese & ReMM-v2-L2-13B (0.441) & ReMM-v2-L2-13B (0.32) & ReMM-v2-L2-13B (0.618) & ReMM-v2-L2-13B (0.643)\\ 
\hline
Romanian & ReMM-v2-L2-13B (0.367) & ReMM-v2-L2-13B (0.255) & ReMM-v2-L2-13B (0.562) & ReMM-v2-L2-13B (0.58)\\ 
\hline
Russian & ReMM-v2-L2-13B (0.305) & ReMM-v2-L2-13B (0.182) & ReMM-v2-L2-13B (0.501) & ReMM-v2-L2-13B (0.516)\\ 
\hline
Slovak & ReMM-v2-L2-13B (0.287) & ReMM-v2-L2-13B (0.166) & Llama2-chat-AYT-13B (0.493) & ReMM-v2-L2-13B (0.506)\\ 
\hline
Slovenian & ReMM-v2-L2-13B (0.254) & ReMM-v2-L2-13B (0.14) & ReMM-v2-L2-13B (0.46) & ReMM-v2-L2-13B (0.459)\\ 
\hline
Albanian & Turdus (0.122) & Turdus (0.041) & Turdus (0.329) & Turdus (0.266)\\ 
\hline
Serbian & ReMM-v2-L2-13B (0.347) & ReMM-v2-L2-13B (0.233) & ReMM-v2-L2-13B (0.535) & ReMM-v2-L2-13B (0.559)\\ 
\hline
Swedish & ReMM-v2-L2-13B (0.408) & ReMM-v2-L2-13B (0.294) & ReMM-v2-L2-13B (0.583) & ReMM-v2-L2-13B (0.619)\\ 
\hline
Thai & ReMM-v2-L2-13B (0.151) & ReMM-v2-L2-13B (0.051) & Turdus (0.338) & ReMM-v2-L2-13B (0.309)\\ 
\hline
Turkish & ReMM-v2-L2-13B (0.236) & ReMM-v2-L2-13B (0.118) & Llama2-chat-AYT-13B (0.437) & ReMM-v2-L2-13B (0.432)\\ 
\hline
Ukrainian & ReMM-v2-L2-13B (0.303) & ReMM-v2-L2-13B (0.178) & ReMM-v2-L2-13B (0.501) & ReMM-v2-L2-13B (0.507)\\ 
\hline
Urdu & ReMM-v2-L2-13B (0.197) & ReMM-v2-L2-13B (0.085) & Llama2-chat-AYT-13B (0.413) & Llama2-chat-AYT-13B (0.385)\\ 
\hline
Vietnamese & ReMM-v2-L2-13B (0.274) & ReMM-v2-L2-13B (0.155) & ReMM-v2-L2-13B (0.464) & ReMM-v2-L2-13B (0.497)\\ 
\hline
Chinese & ReMM-v2-L2-13B (0.186) & ReMM-v2-L2-13B (0.068) & Llama2-chat-AYT-13B (0.405) & ReMM-v2-L2-13B (0.372)\\ 
 \hline
 \hline
 All languages & ReMM-v2-L2-13B (0.256) & ReMM-v2-L2-13B (0.152) & Llama2-chat-AYT-13B (0.448) & ReMM-v2-L2-13B (0.438) \\ 
 \hline
\end{tabular}}
\caption{Best GPT translation metrics for each language, computed by the best mean translation quality over all tested sentence}
\label{table:GPT_translation_results}
\end{table*}

\begin{table*}[h!]
\centering
\vspace{-1.5cm}
\begin{tabular}{ |c||c|c|c|c| }
 \hline
 \hline
 Language & Mean GLEU & Mean BLEU & Mean chrF  & Mean METEOR \\ 
 \hline
 \hline
Arabic &  0.354 &  0.225 &  0.559 &  0.592\\ 
\hline
Azerbaijani &  0.231 &  0.099 &  0.431 &  0.417\\ 
\hline
Belarusian &  0.289 &  0.169 &  0.484 &  0.486\\ 
\hline
Bulgarian &  0.396 &  0.273 &  0.584 &  0.619\\ 
\hline
Bengali &  0.184 &  0.078 &  0.365 &  0.369\\ 
\hline
Bosnian &  0.381 &  0.26 &  0.58 &  0.611\\ 
\hline
Czech &  0.372 &  0.238 &  0.559 &  0.607\\ 
\hline
Danish &  0.527 &  0.438 &  0.689 &  0.744\\ 
\hline
German &  0.367 &  0.235 &  0.557 &  0.588\\ 
\hline
Greek &  0.379 &  0.25 &  0.559 &  0.602\\ 
\hline
Spanish &  0.438 &  0.324 &  0.623 &  0.66\\ 
\hline
Estonian &  0.32 &  0.198 &  0.526 &  0.545\\ 
\hline
Persian &  0.301 &  0.188 &  0.494 &  0.52\\ 
\hline
Finnish &  0.349 &  0.234 &  0.555 &  0.576\\ 
\hline
French &  0.326 &  0.202 &  0.552 &  0.579\\ 
\hline
Galician &  0.367 &  0.25 &  0.564 &  0.586\\ 
\hline
Hebrew &  0.39 &  0.262 &  0.569 &  0.615\\ 
\hline
Hindi &  0.234 &  0.113 &  0.437 &  0.445\\ 
\hline
Croatian &  0.413 &  0.282 &  0.599 &  0.637\\ 
\hline
Hungarian &  0.329 &  0.21 &  0.522 &  0.551\\ 
\hline
Armenian &  0.28 &  0.158 &  0.478 &  0.477\\ 
\hline
Indonesian &  0.327 &  0.2 &  0.527 &  0.567\\ 
\hline
Italian &  0.372 &  0.255 &  0.568 &  0.601\\ 
\hline
Japanese &  0.204 &  0.078 &  0.395 &  0.387\\ 
\hline
Georgian &  0.254 &  0.119 &  0.458 &  0.452\\ 
\hline
Kazakh &  0.206 &  0.076 &  0.383 &  0.379\\ 
\hline
Korean &  0.25 &  0.126 &  0.455 &  0.461\\ 
\hline
Kurdish &  0.21 &  0.095 &  0.394 &  0.407\\ 
\hline
Lithuanian &  0.308 &  0.176 &  0.507 &  0.52\\ 
\hline
Macedonian &  0.369 &  0.245 &  0.571 &  0.597\\ 
\hline
Mongolian &  0.165 &  0.037 &  0.338 &  0.312\\ 
\hline
Malay &  0.309 &  0.184 &  0.509 &  0.537\\ 
\hline
Burmese &  0.078 &  0.012 &  0.195 &  0.152\\ 
\hline
Norwegian &  0.462 &  0.351 &  0.635 &  0.685\\ 
\hline
Dutch &  0.409 &  0.292 &  0.589 &  0.621\\ 
\hline
Polish &  0.313 &  0.187 &  0.507 &  0.521\\ 
\hline
Portuguese &  0.459 &  0.335 &  0.64 &  0.672\\ 
\hline
Romanian &  0.389 &  0.268 &  0.581 &  0.612\\ 
\hline
Russian &  0.298 &  0.174 &  0.497 &  0.507\\ 
\hline
Slovak &  0.361 &  0.229 &  0.561 &  0.594\\ 
\hline
Slovenian &  0.321 &  0.197 &  0.525 &  0.544\\ 
\hline
Albanian &  0.383 &  0.256 &  0.575 &  0.608\\ 
\hline
Serbian &  0.388 &  0.261 &  0.57 &  0.617\\ 
\hline
Swedish &  0.446 &  0.329 &  0.618 &  0.666\\ 
\hline
Thai &  0.189 &  0.067 &  0.36 &  0.361\\ 
\hline
Turkish &  0.323 &  0.193 &  0.514 &  0.546\\ 
\hline
Ukrainian &  0.305 &  0.17 &  0.502 &  0.513\\ 
\hline
Urdu &  0.302 &  0.18 &  0.509 &  0.529\\ 
\hline
Vietnamese &  0.292 &  0.173 &  0.483 &  0.526\\ 
\hline
Chinese &  0.155 &  0.039 &  0.358 &  0.318\\ 
\hline
\end{tabular}
\caption{Mean translation quality metrics from using the Google translate service, taken across all test sentences }
\label{table:google_translate_results}
\end{table*}

\subsection{Translation Quality Metrics and Example Translations}
\label{section:results_example_sentences}
The following are some examples where the translations produced by the GPT models are reasonable, but the language quality scores are not very close to $1$. These examples are shown with the aim of conveying that the overall translation quality for many of the GPT models is quite good even if the mean language quality scores are on average not incredibly close to $1$. Importantly, most of the reason for this is that the translation quality metrics are computed for individual sentences, not the entirety of the translated document - and this can lead to unstable measurements of translation quality. However, the mean of the sentence translation quality is a good representation of the overall translation quality -- in particular the language quality metrics over the entire translated corpus are very similar (but not necessarily equal) to the mean of the translation metrics across all of the component sentences.

This is an example sentence translation from Spanish into English from the TED talk dataset where the translated sentence has a GLEU score of $0.435$, a BLEU score of $0.320$, a chrF score of $0.780$, and a METEOR score of $0.864$. Note that both of these sentences have been tokenized before the score was computed and are shown in their tokenized form. 

\begin{tcolorbox}
\textbf{Reference English sentence:} this is a viking lander photograph of the surface of mars

\textbf{GPT translated sentence from Spanish into English:} this is a photograph from the viking lander on the surface of mars
\end{tcolorbox}

This is another Spanish to English sentence translation where the translated sentence has a GLEU score of $0.427$, a BLEU score of $0.379$, a chrF score of $0.645$, and a METEOR score of $0.725$:
\begin{tcolorbox}
\textbf{Reference English sentence:} but there is intriguing evidence that suggests that the early history of mars there may have been rivers and fast flowing water

\textbf{GPT translated sentence from Spanish into English:} there is intriguing evidence suggesting that the early history of mars may have had rivers and streams of water
\end{tcolorbox}

This is an example sentence translation from Spanish into English which had a GLEU score of $0.481$, a BLEU score of $0.429$, a chrf score of $0.629$, and a METEOR score of $0.735$:
\begin{tcolorbox}
\textbf{Reference English sentence:} the answer is no there is no liquid water on the surface of mars today

\textbf{GPT translated sentence from Spanish into English:} there is no water liquid on the surface of mars today
\end{tcolorbox}

This is an example sentence translation from French into English which had a GLEU score of $0.587$, a BLEU score of $0.556$, a chrF score of $0.718$, and a METEOR score of $0.825$:
\begin{tcolorbox}
\textbf{Reference English sentence:} i want to talk to you about one of the biggest myths in medicine and that is the idea that all we need are more medical breakthroughs and then all of our problems will be solved

\textbf{GPT translated sentence from French into English:} i want to talk about one of the greatest myths of medicine and that is the idea that all we need are additional medical procedures and then all our problems will be solved
\end{tcolorbox}

\section{Discussion and Conclusion}
\label{section:conclusion}

The translation quality provided by sentence-wise GPT translations showed a clear stratification of the capabilities of the evaluated $16$ GPT models. The best performing GPT models, across all $50$ foreign languages, for translating into English is \texttt{ReMM-v2-L2-13B} and \texttt{Llama2-chat-AYT-13B}. This shows that language translation could serve as a clear, and very application-relevant, benchmark for GPT capabilities for processing natural language.

The best performing GPT model translations compare very well against automated machine translation using the Google translate API, although typically Google translate has marginally better scores. Importantly, the GPT model computations offer the security advantage of performing the computations locally, meaning that locally run GPT model automated language translation may be a good alternative depending on the importance of the security and the privacy of handling the information.

The GPT models do not uniformly perform well though -- several of the tested models performed noticeably worse than other GPT models, and these trends are consistent across all of the $50$ tested languages. Interestingly, there were also some languages that were not able to be translated well by any of the GPT models, for example Mongolian, Kazakh, Burmese, Kurdish, Armenian, and Georgian. This could be due to these languages being relatively low-resource in the training data used when training these GPT models. Notably, there were several GPT models that were consistently the slowest across the different languages; \texttt{phi-1}, \texttt{phi-2}, \texttt{phi-1\_5}, \texttt{zephyr-7b-beta}, and \texttt{falcon-7b-instruct}.

\section{Acknowledgments}
\label{section:acknowledgments}

Sandia National Laboratories is a multi-mission laboratory managed and operated by National Technology \& Engineering Solutions of Sandia, LLC (NTESS), a wholly owned subsidiary of Honeywell International Inc., for the U.S. Department of Energy’s National Nuclear Security Administration (DOE/NNSA) under contract DE-NA0003525. This written work is authored by an employee of NTESS. The employee, not NTESS, owns the right, title and interest in and to the written work and is responsible for its contents. Any subjective views or opinions that might be expressed in the written work do not necessarily represent the views of the U.S. Government. The publisher acknowledges that the U.S. Government retains a non-exclusive, paid-up, irrevocable, world-wide license to publish or reproduce the published form of this written work or allow others to do so, for U.S. Government purposes. The DOE will provide public access to results of federally sponsored research in accordance with the DOE Public Access Plan.

\clearpage

\setlength\bibitemsep{0pt}
\printbibliography

\appendix

\section{Key Phrases Removed From GPT Output}
\label{section:appendix_phrase_post_processing}

Any generated text that starts with any of the following strings has that text removed before the language quality metrics are computed. This list is not complete for the sake of space, but these serve as representative ancillary text that were commonly seen in the GPT model output. 

\begin{itemize}[noitemsep]
    \item Translated text:
    \item Translation:
    \item The translated text is:
    \item The sentence translates to:
    \item The translation is:
    \item The sentence should be translated to:
    \item The following sentence is translated into English text:
    \item Answer: The translation of the sentence is:
    \item This is the translated text:
    \item Clear English translation:
    \item You can translate this sentence into English as:
    \item This sentence translates to
    \item Solution: The sentence is translated into clearly written English text as follows:
    \item The following sentence is translated into clearly written English text:
    \item This is a clear and accurate translation:
    \item The sentence has been translated into English as:
    \item The given sentence translates to:
    \item A clear English translation of the given sentence is:
    \item In English, this translates to:
    \item The following is the translated text:
    \item The correct translation is:
    \item This should be translated as:
    \item Here is the translation of the sentence:
    \item The text translates to:
    \item Here is my translation:
    \item You can translate the sentence as:
    \item The English translation of the given sentence is:
    \item This sentence can be translated as:
    \item This sentence can be translated into English as:
    \item The following is a clearly written English translation of the provided text:
    \item The following sentence is translated into English:
    \item Translate the following sentence into clearly written English text. Respond only with the translated text; do not write explanations or justifications in your reply. 
\end{itemize}

Note that the last entry in the above list is actually just the input prompt; we found some of the GPT models would on occasion re-output the prompt before generating any further text, so this was included in the simple text removal post-processing.

\section{Complete Translation Quality Measure and Timing Plots}
\label{section:appendix_translation_stats}

\begin{figure}[th!]
    \centering
    \includegraphics[width=0.47\textwidth]{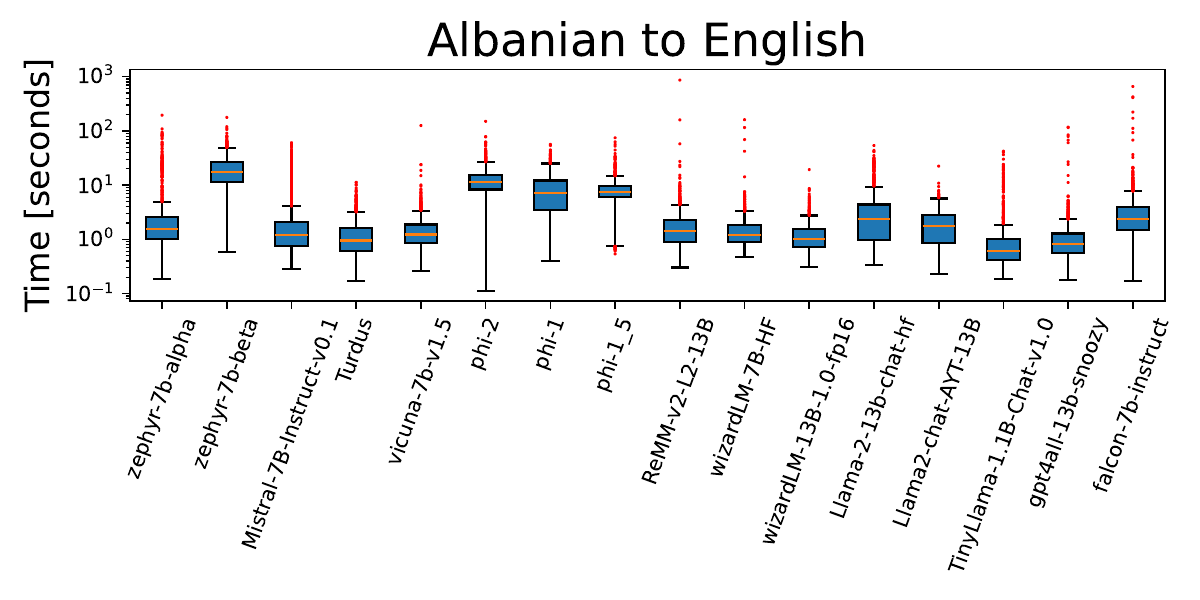}\\
    \includegraphics[width=0.47\textwidth]{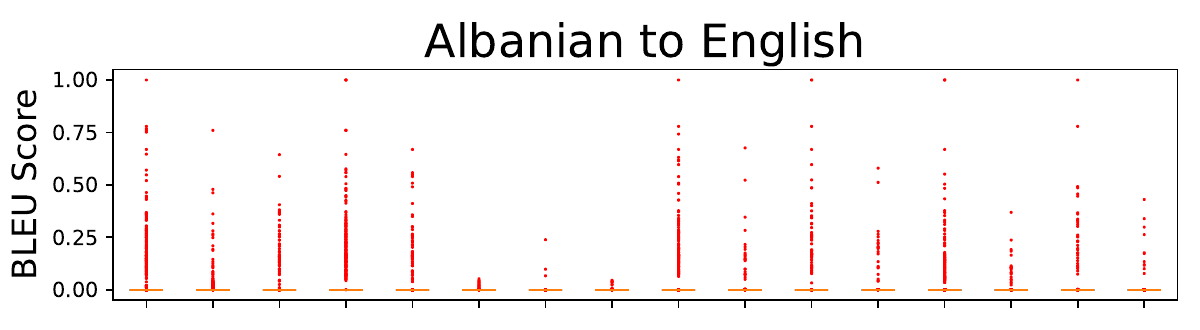}
    \includegraphics[width=0.47\textwidth]{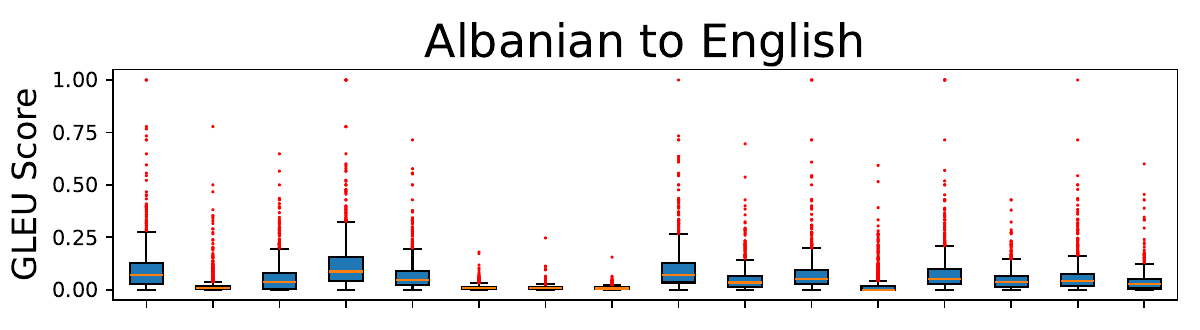}
    \includegraphics[width=0.47\textwidth]{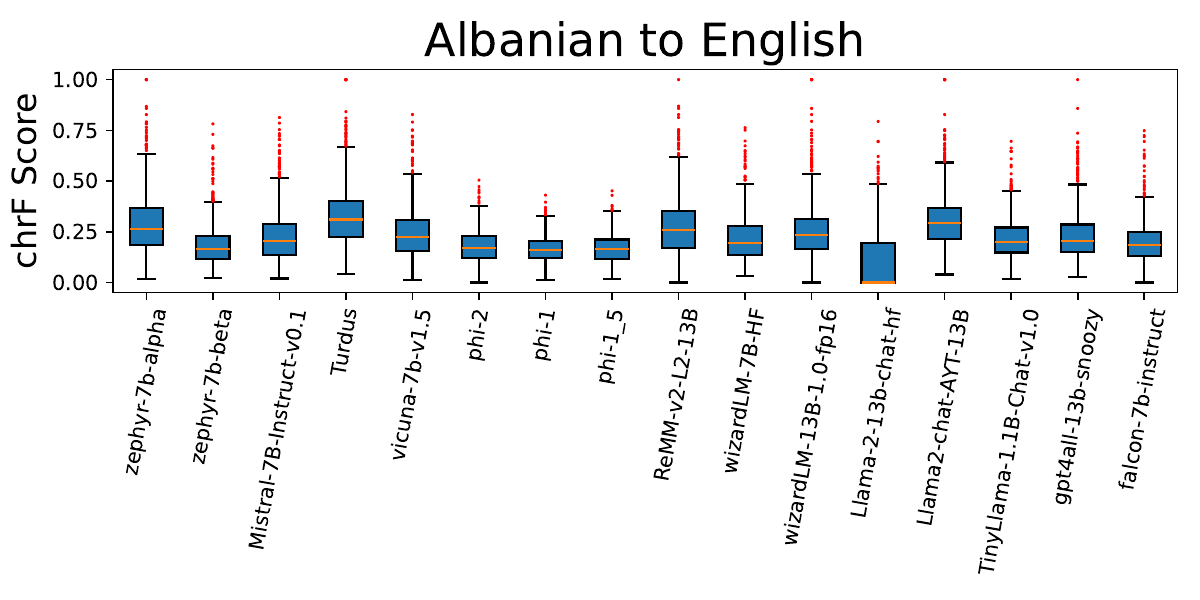}
    \includegraphics[width=0.47\textwidth]{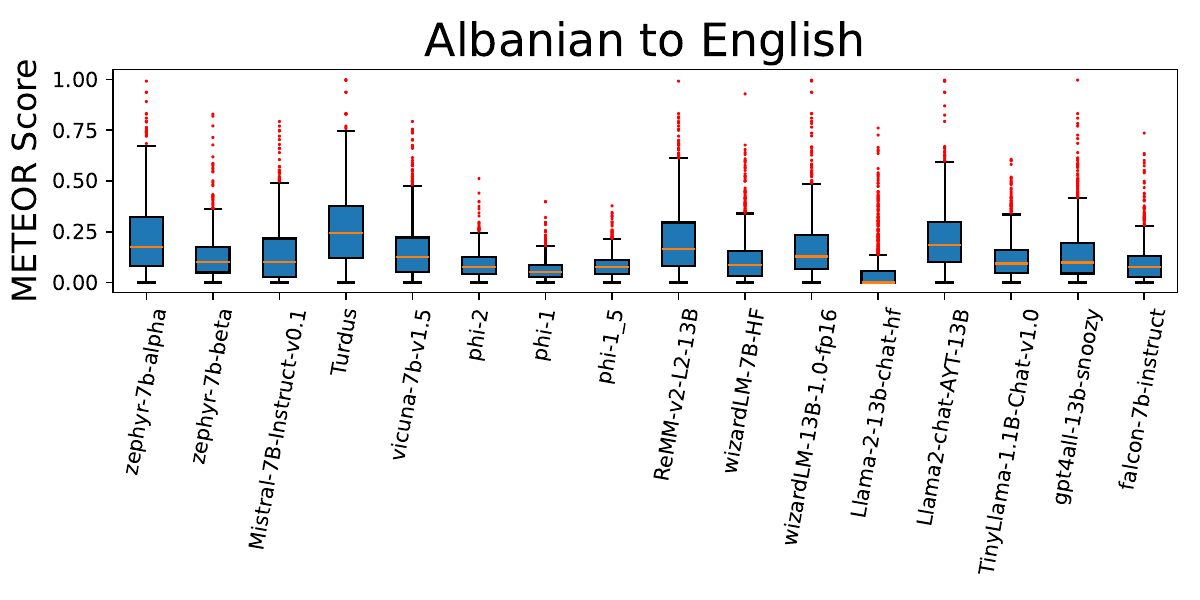}
    \caption{Albanian-to-English dataset per-sentence translation quality and timing statistics  }
    \label{fig:Albanian_translate_stats}
\end{figure}

\begin{figure}[th!]
    \centering
    \includegraphics[width=0.47\textwidth]{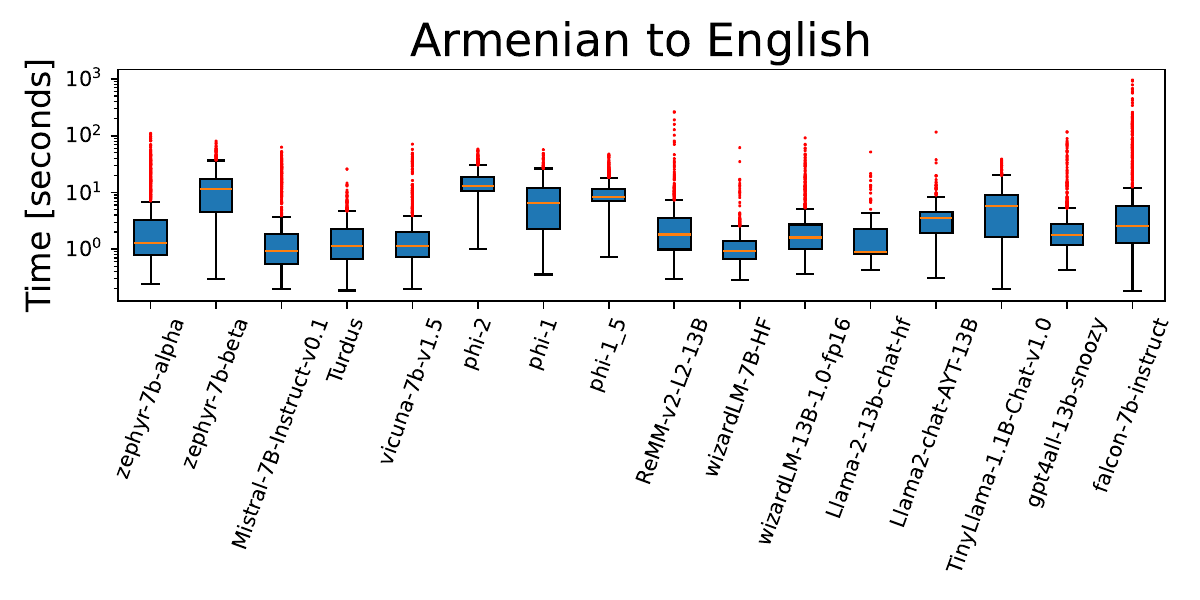}\\
    \includegraphics[width=0.47\textwidth]{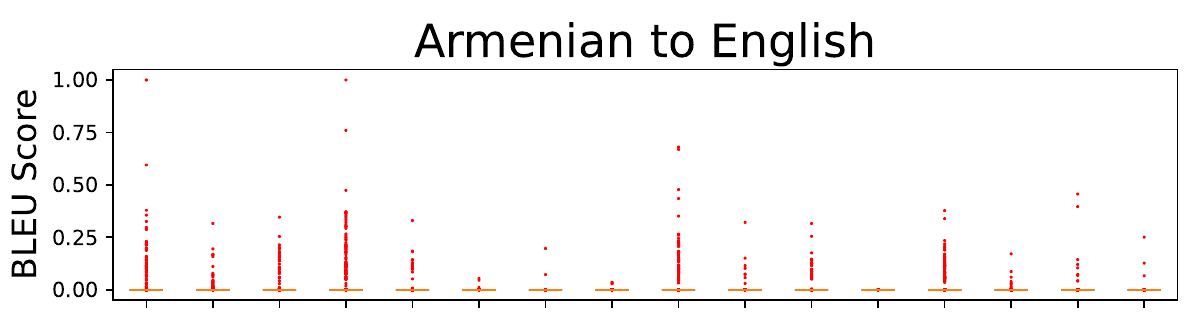}
    \includegraphics[width=0.47\textwidth]{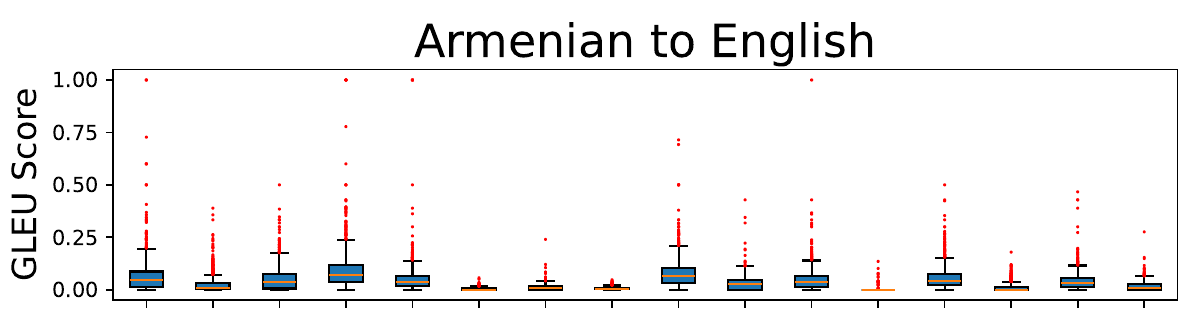}
    \includegraphics[width=0.47\textwidth]{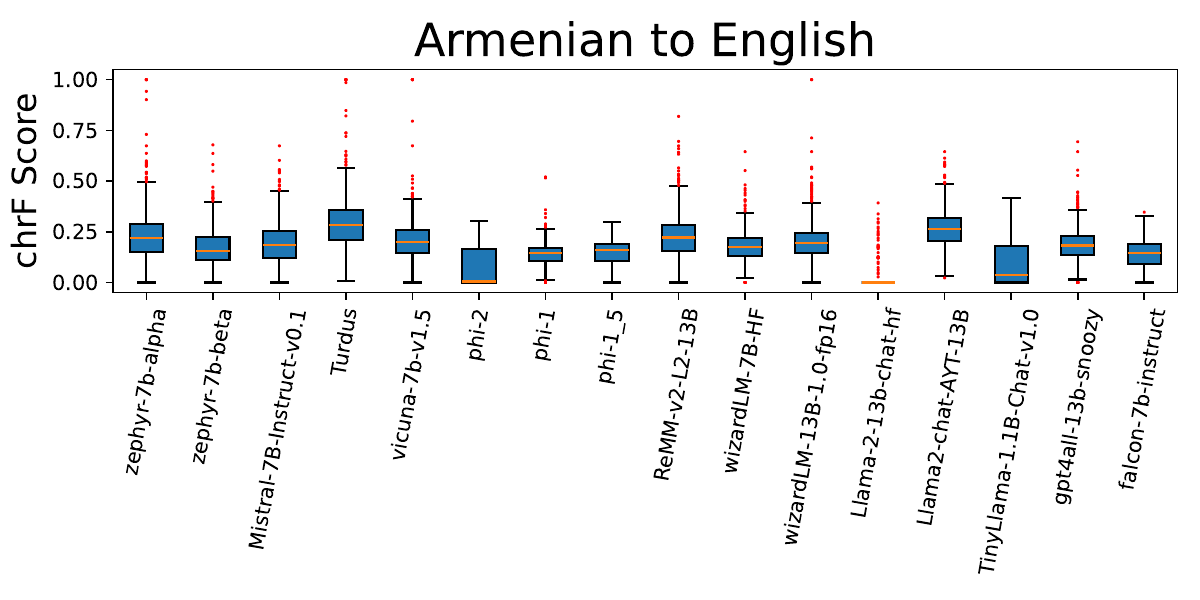}
    \includegraphics[width=0.47\textwidth]{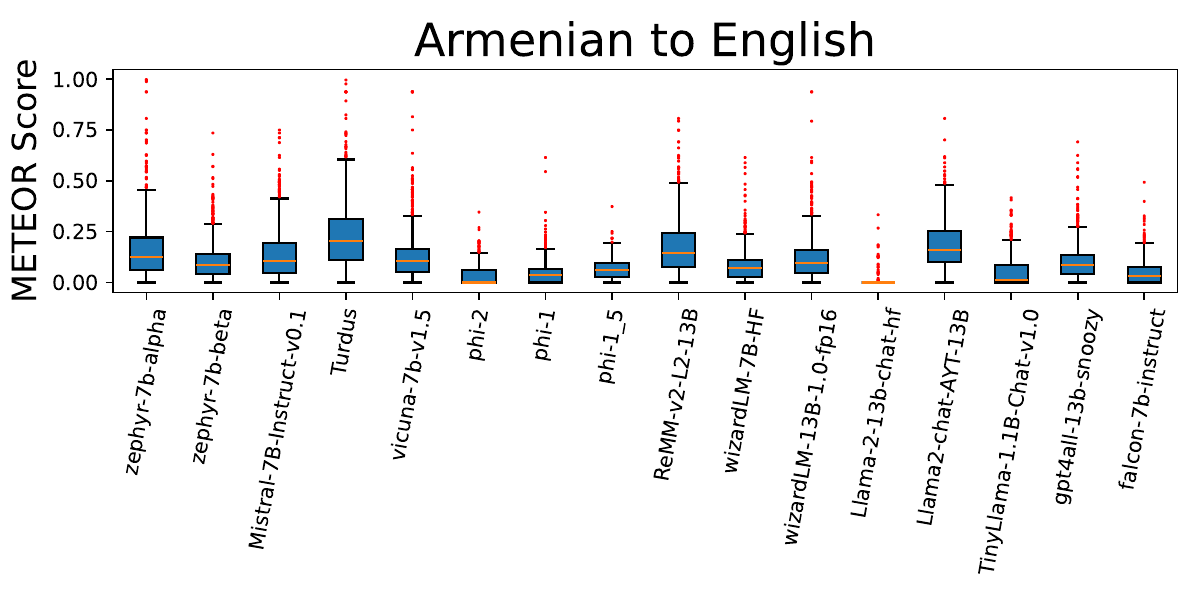}
    \caption{Armenian-to-English dataset per-sentence translation quality and timing statistics  }
    \label{fig:Armenian_translate_stats}
\end{figure}

\begin{figure}[th!]
    \centering
    \includegraphics[width=0.47\textwidth]{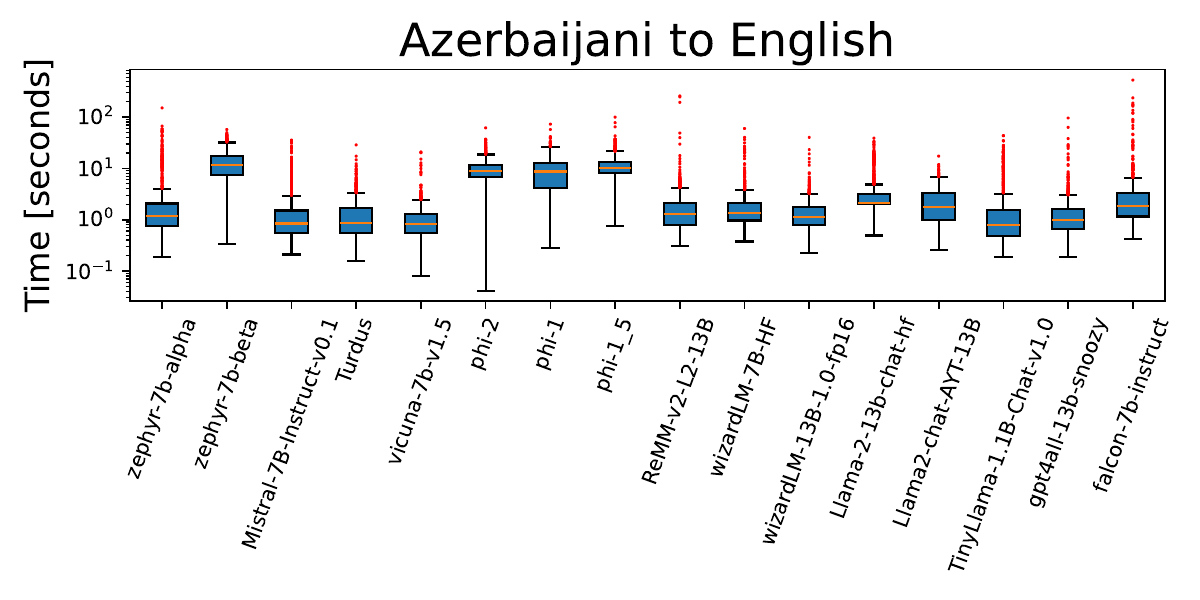}\\
    \includegraphics[width=0.47\textwidth]{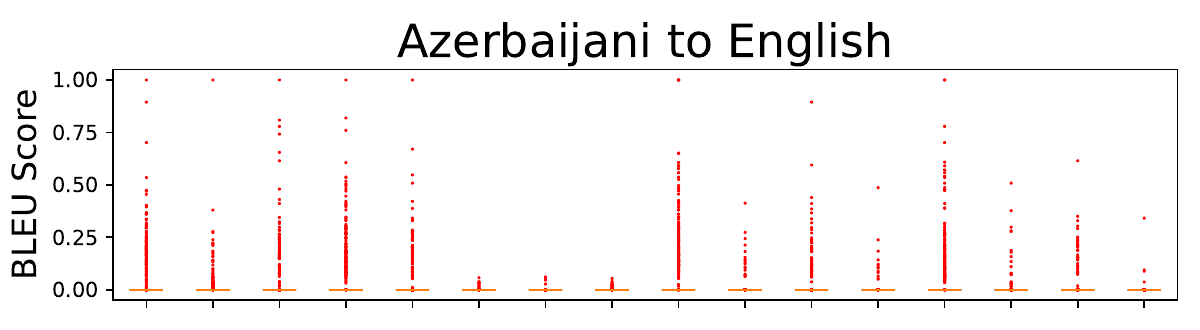}
    \includegraphics[width=0.47\textwidth]{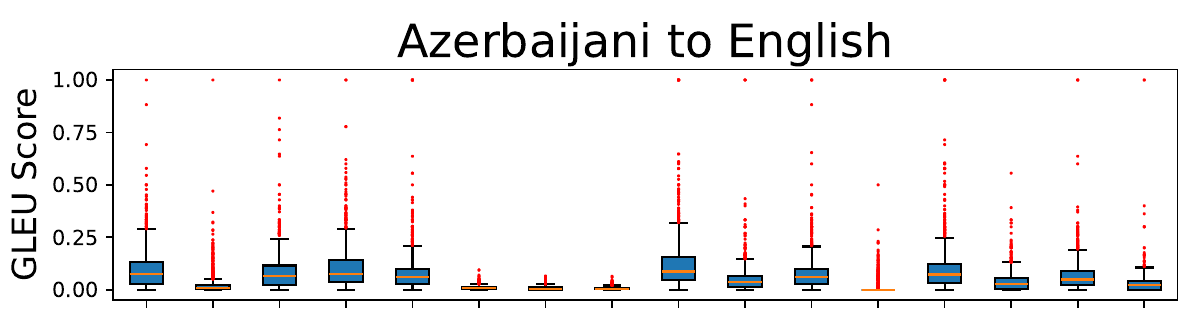}
    \includegraphics[width=0.47\textwidth]{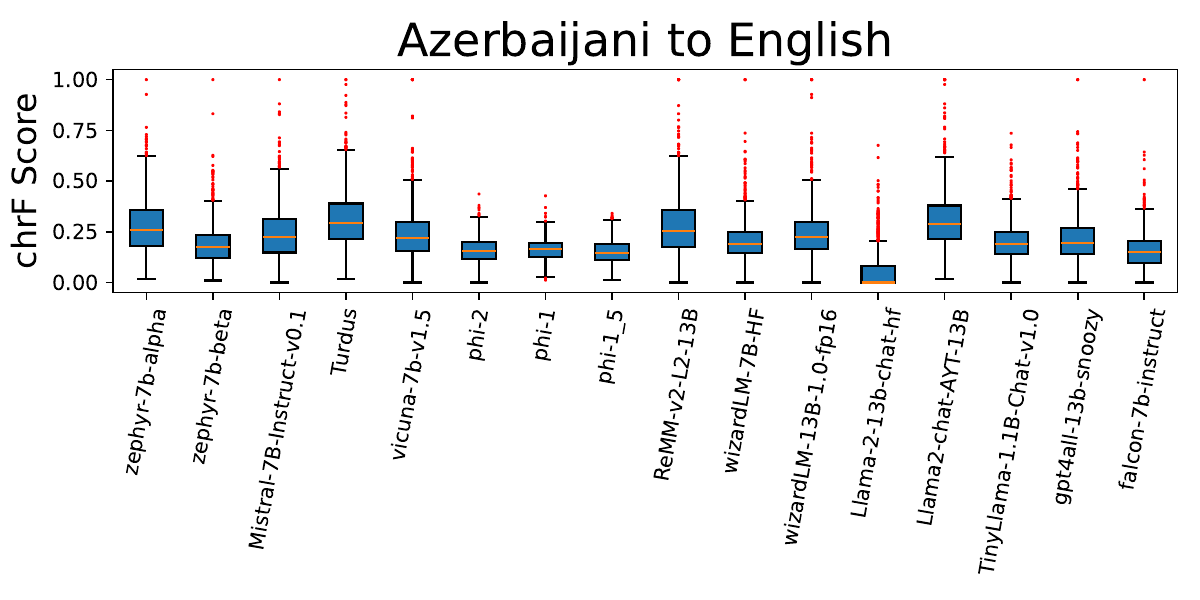}
    \includegraphics[width=0.47\textwidth]{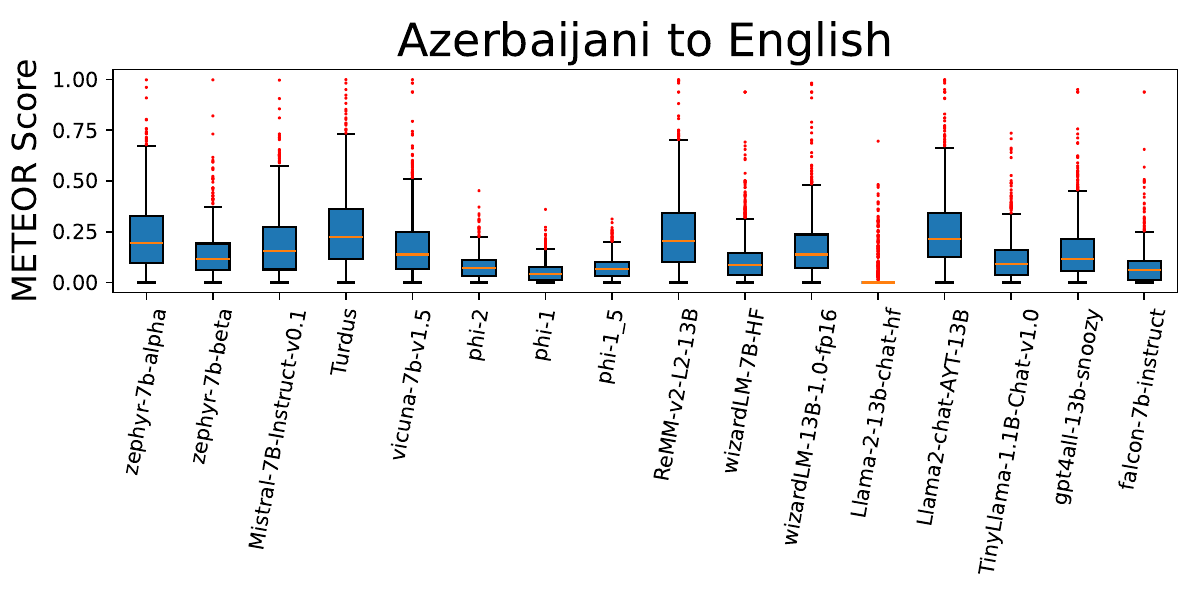}
    \caption{Azerbaijani-to-English dataset per-sentence translation quality and timing statistics }
    \label{fig:Azerbaijani_translate_stats}
\end{figure}

\begin{figure}[th!]
    \centering
    \includegraphics[width=0.47\textwidth]{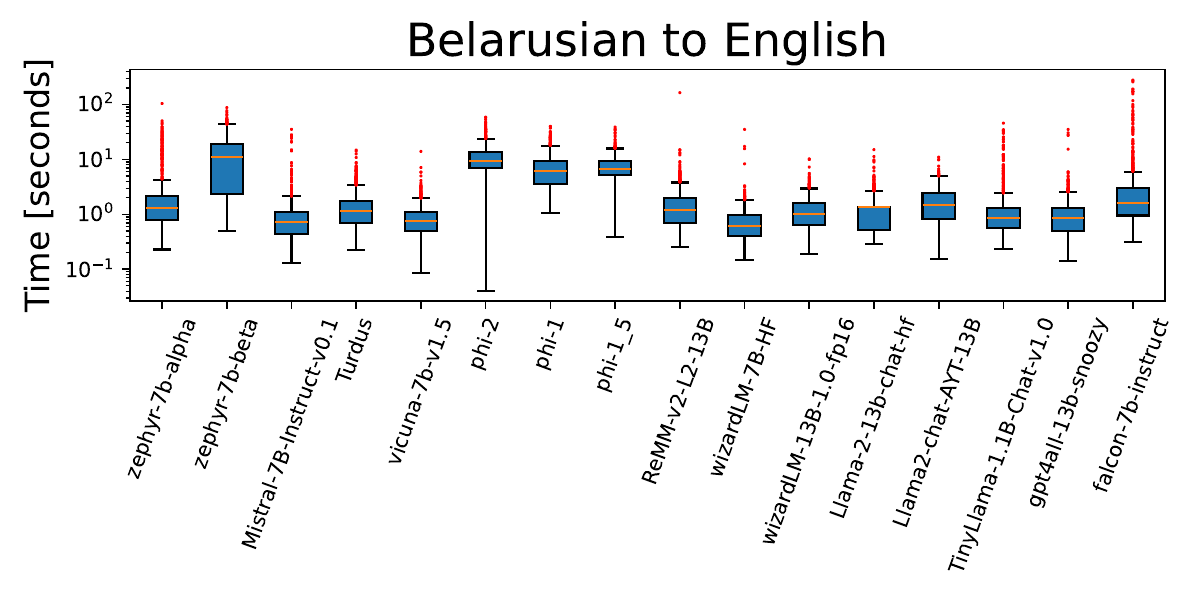}\\
    \includegraphics[width=0.47\textwidth]{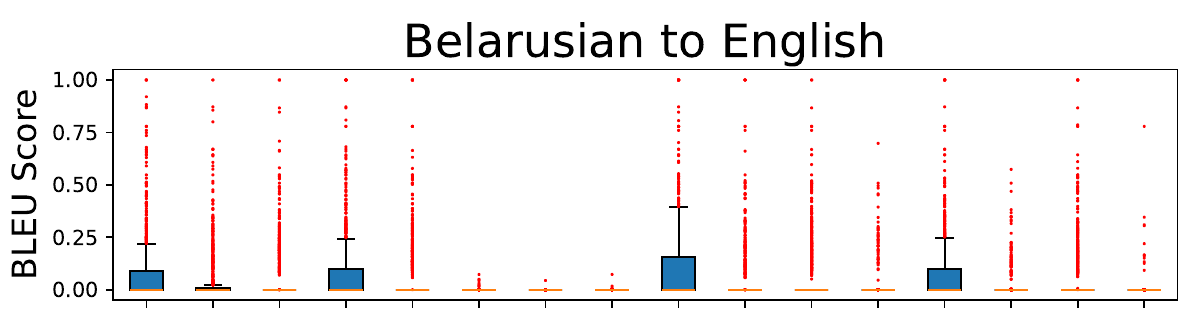}
    \includegraphics[width=0.47\textwidth]{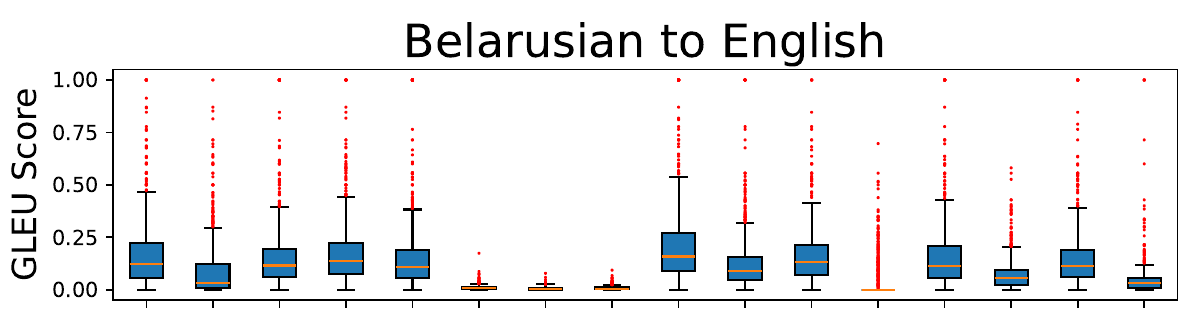}
    \includegraphics[width=0.47\textwidth]{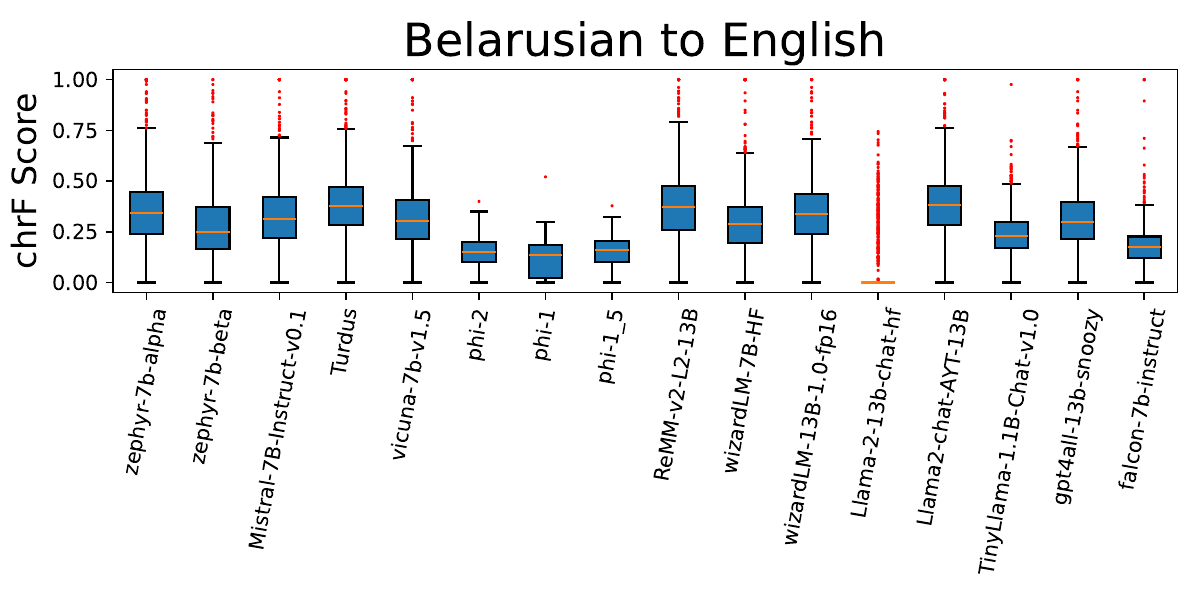}
    \includegraphics[width=0.47\textwidth]{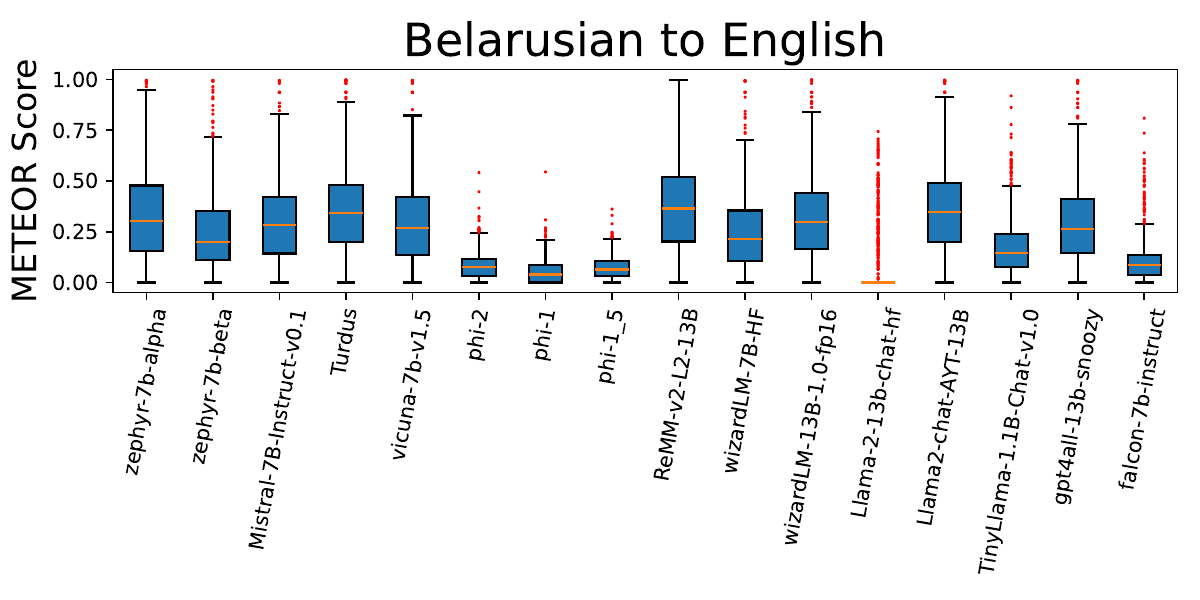}
    \caption{Belarusian-to-English dataset per-sentence translation quality and timing statistics  }
    \label{fig:Belarusian_translate_stats}
\end{figure}

\begin{figure}[th!]
    \centering
    \includegraphics[width=0.47\textwidth]{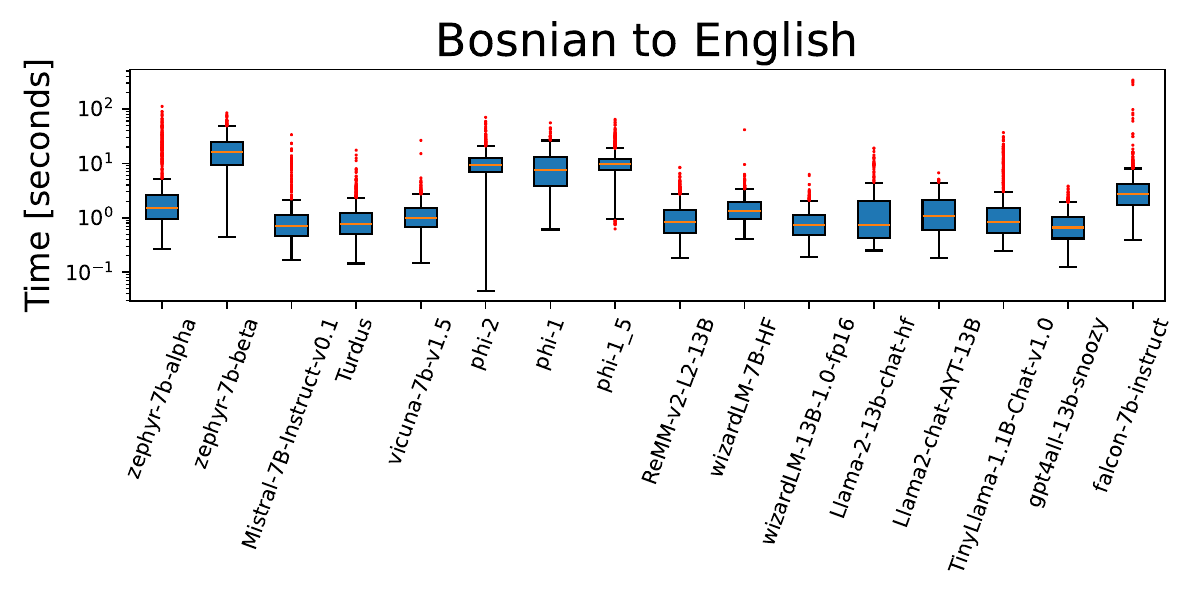}\\
    \includegraphics[width=0.47\textwidth]{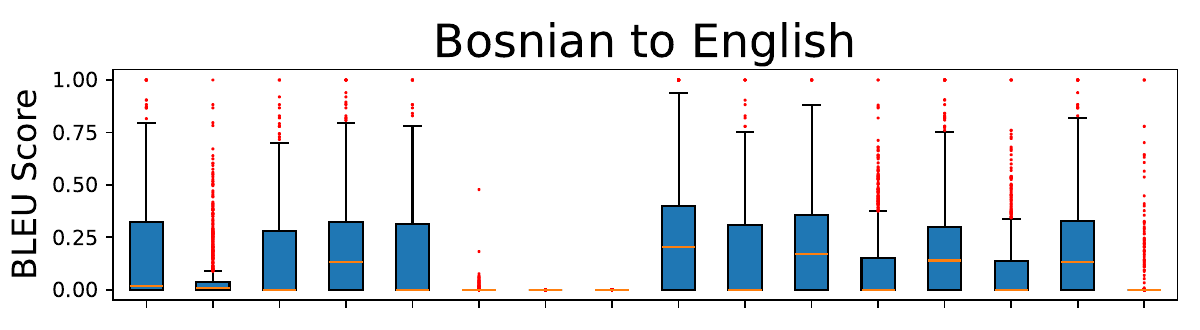}
    \includegraphics[width=0.47\textwidth]{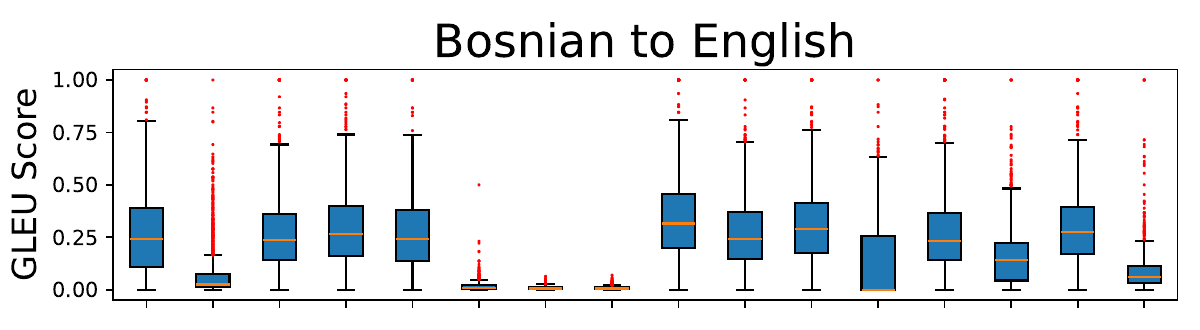}
    \includegraphics[width=0.47\textwidth]{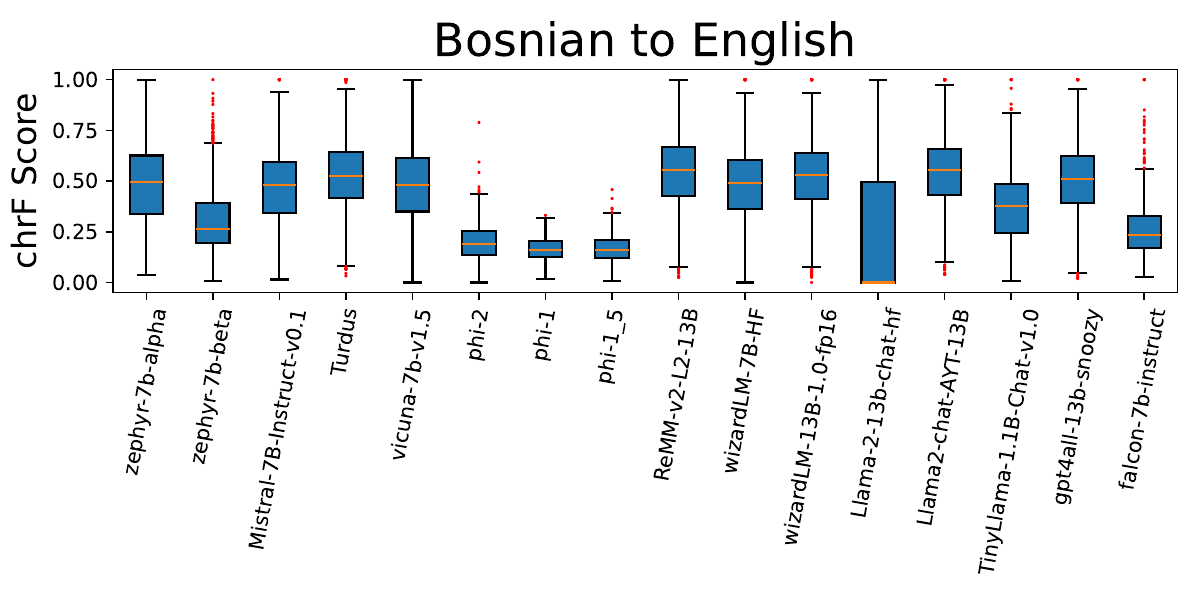}
    \includegraphics[width=0.47\textwidth]{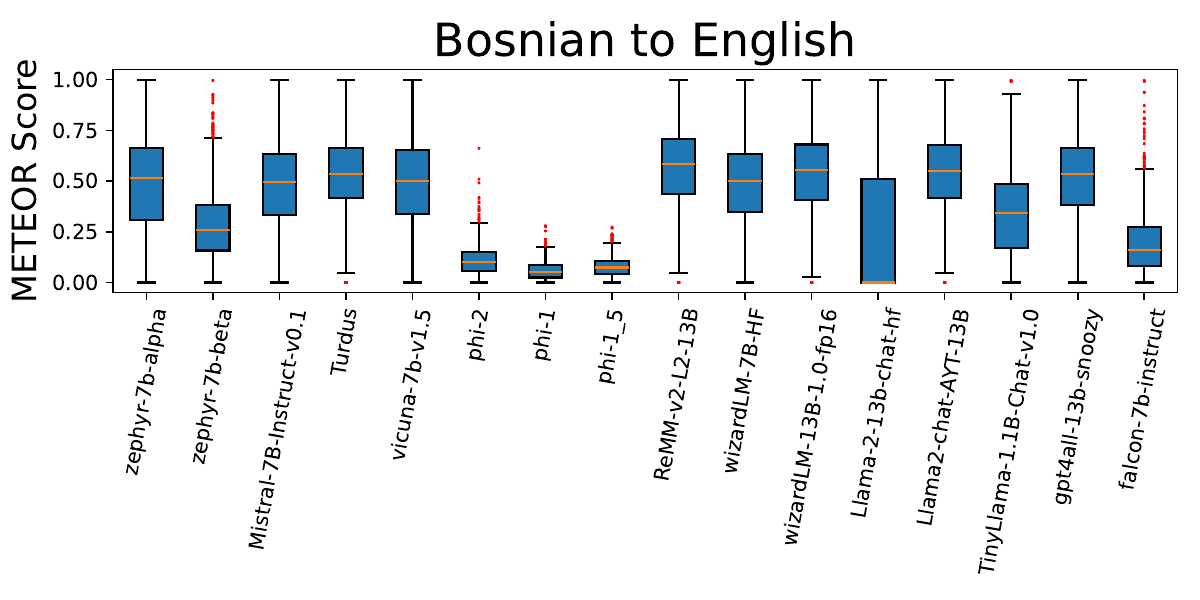}
    \caption{Bosnian-to-English dataset per-sentence translation quality and timing statistics }
    \label{fig:Bosnian_translate_stats}
\end{figure}

\begin{figure}[th!]
    \centering
    \includegraphics[width=0.47\textwidth]{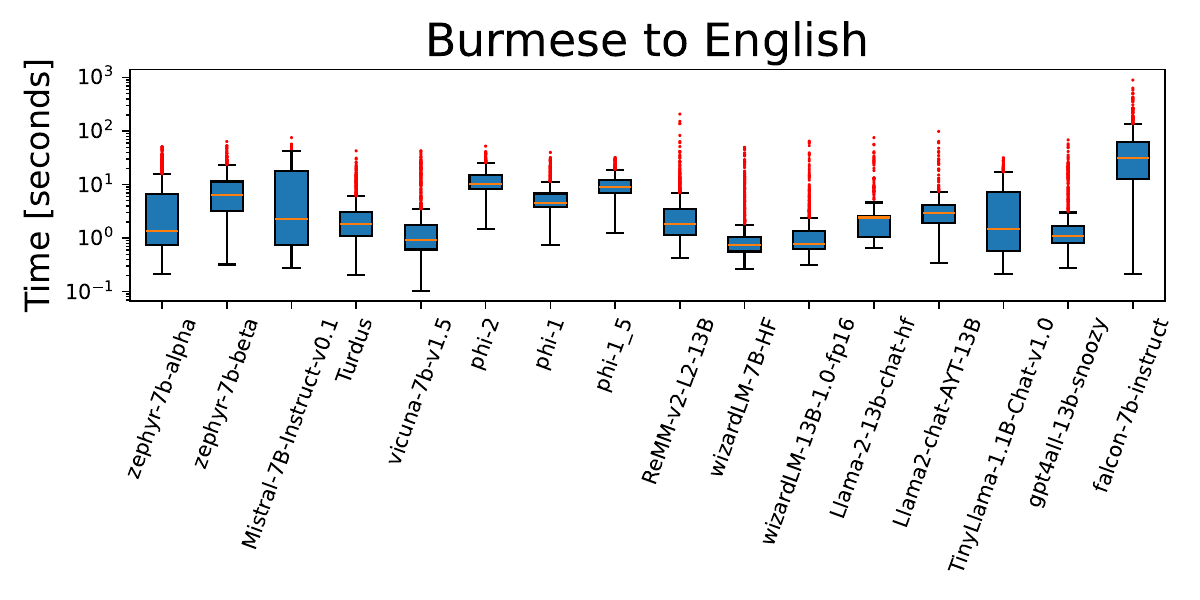}\\
    \includegraphics[width=0.47\textwidth]{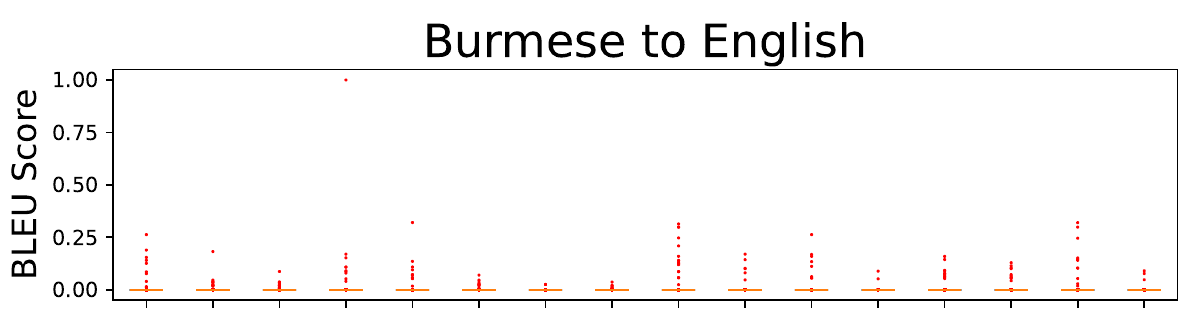}
    \includegraphics[width=0.47\textwidth]{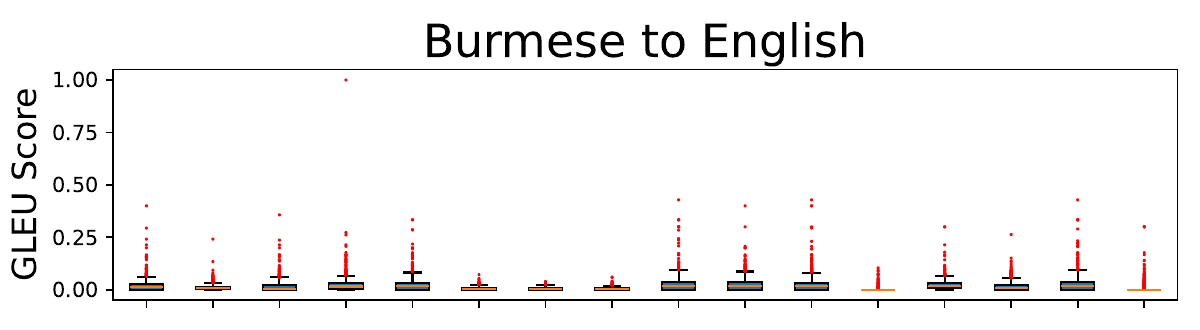}
    \includegraphics[width=0.47\textwidth]{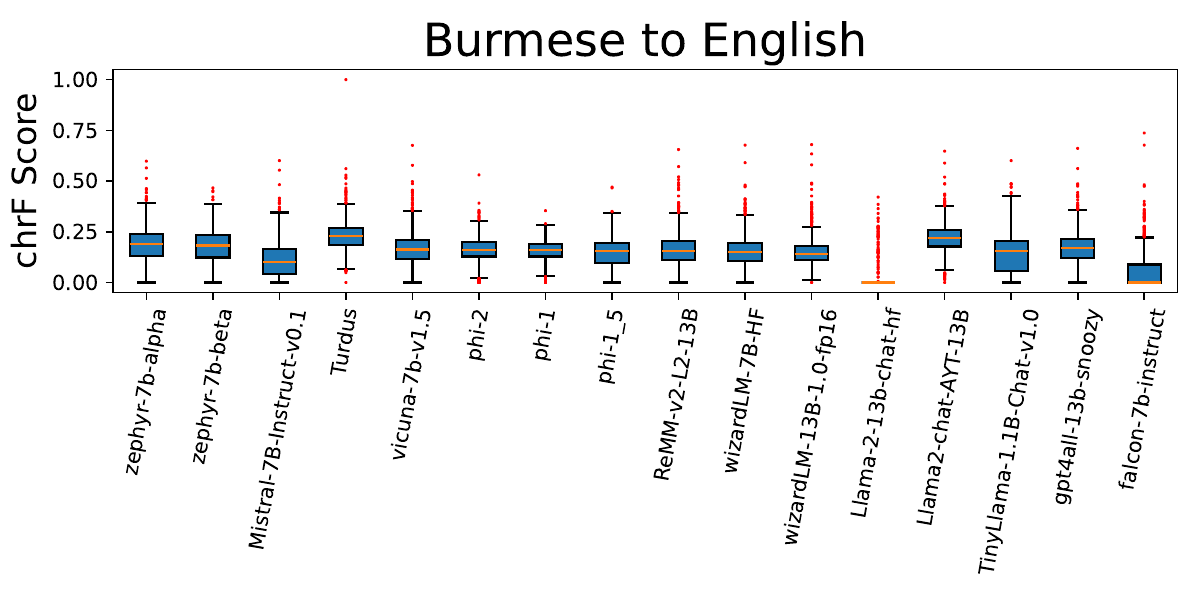}
    \includegraphics[width=0.47\textwidth]{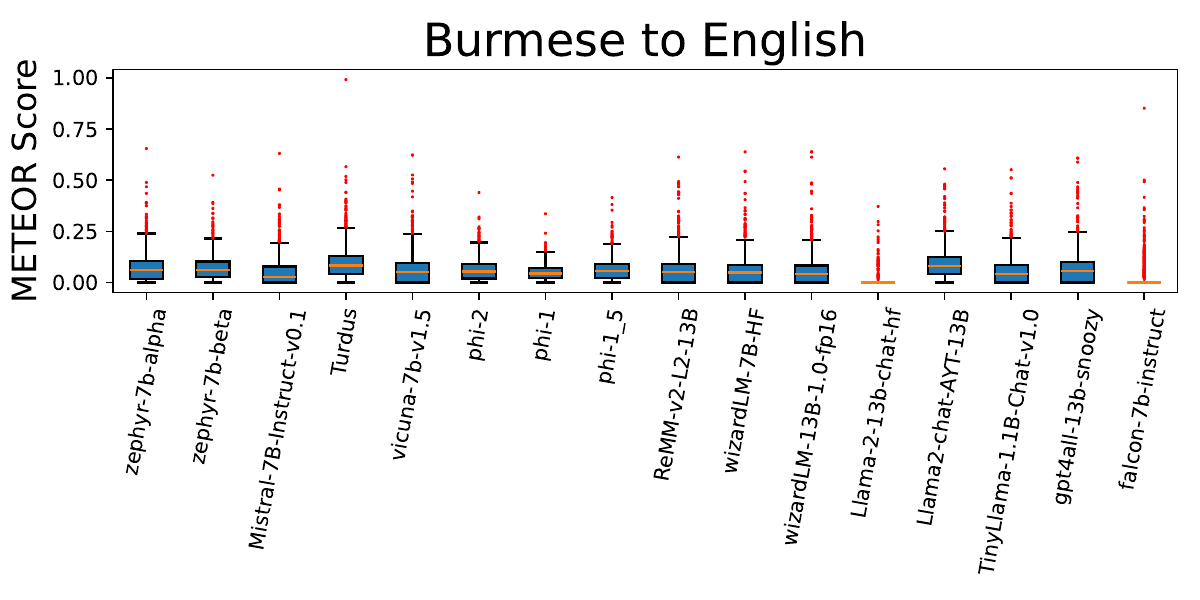}
    \caption{Burmese-to-English dataset per-sentence translation quality and timing statistics  }
    \label{fig:Burmese_translate_stats}
\end{figure}

\begin{figure}[th!]
    \centering
    \includegraphics[width=0.47\textwidth]{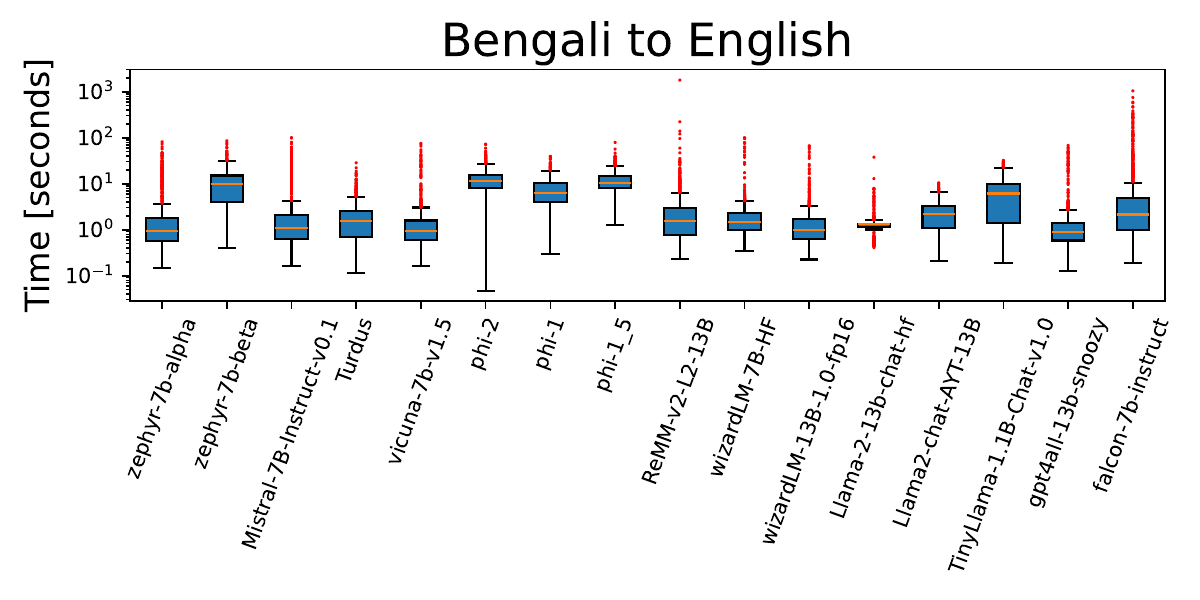}\\
    \includegraphics[width=0.47\textwidth]{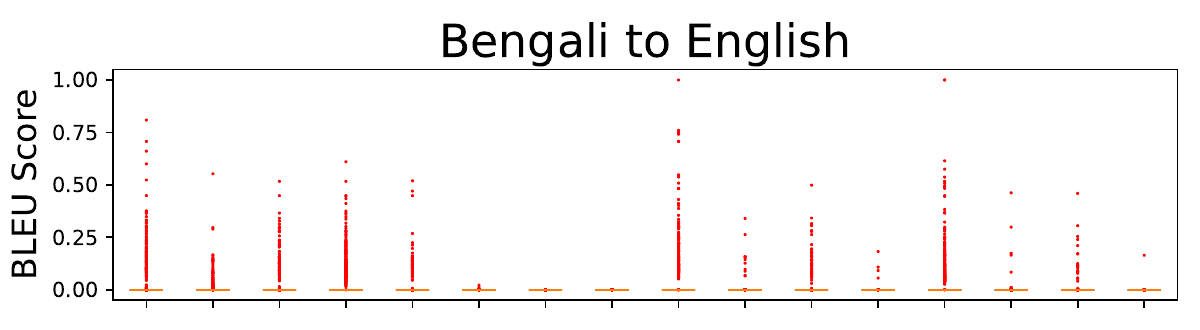}
    \includegraphics[width=0.47\textwidth]{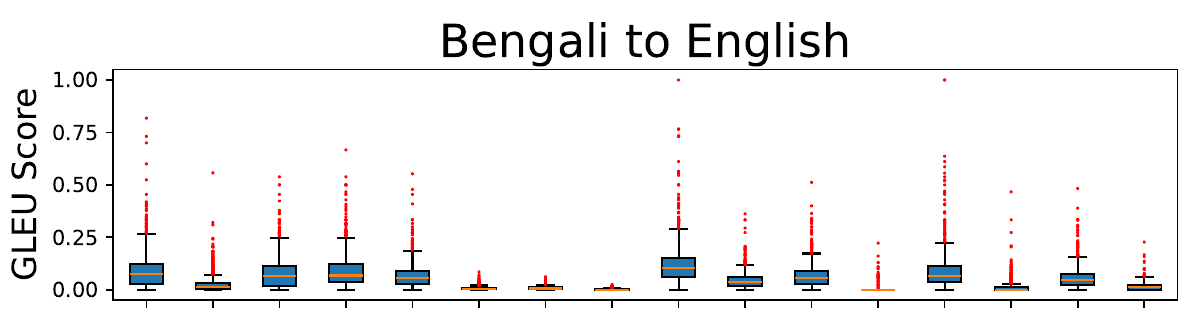}
    \includegraphics[width=0.47\textwidth]{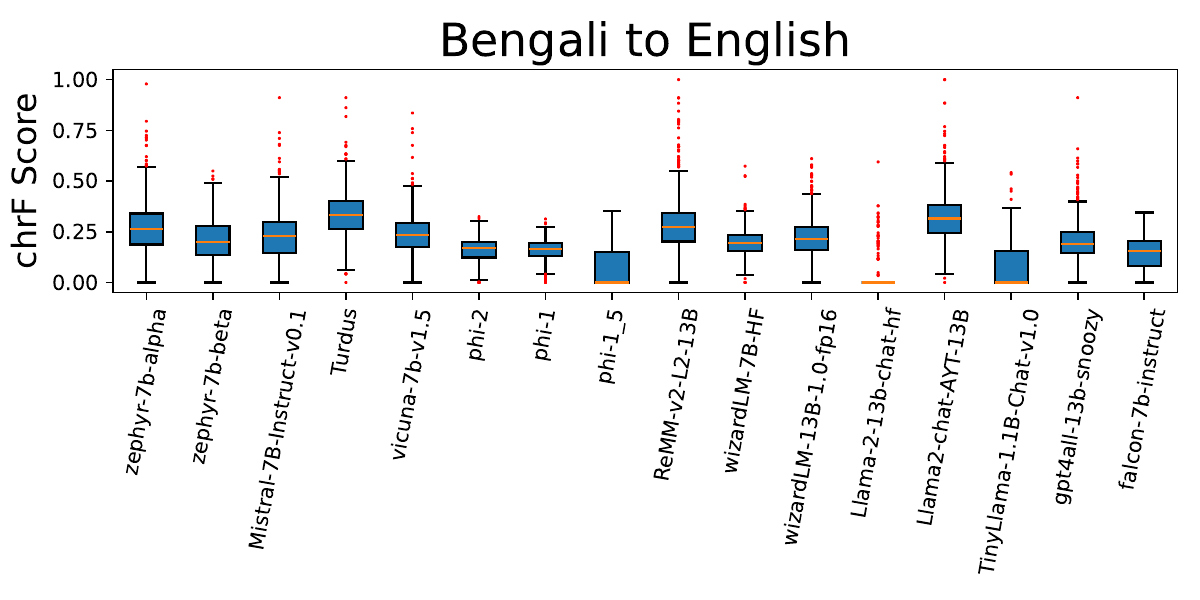}
    \includegraphics[width=0.47\textwidth]{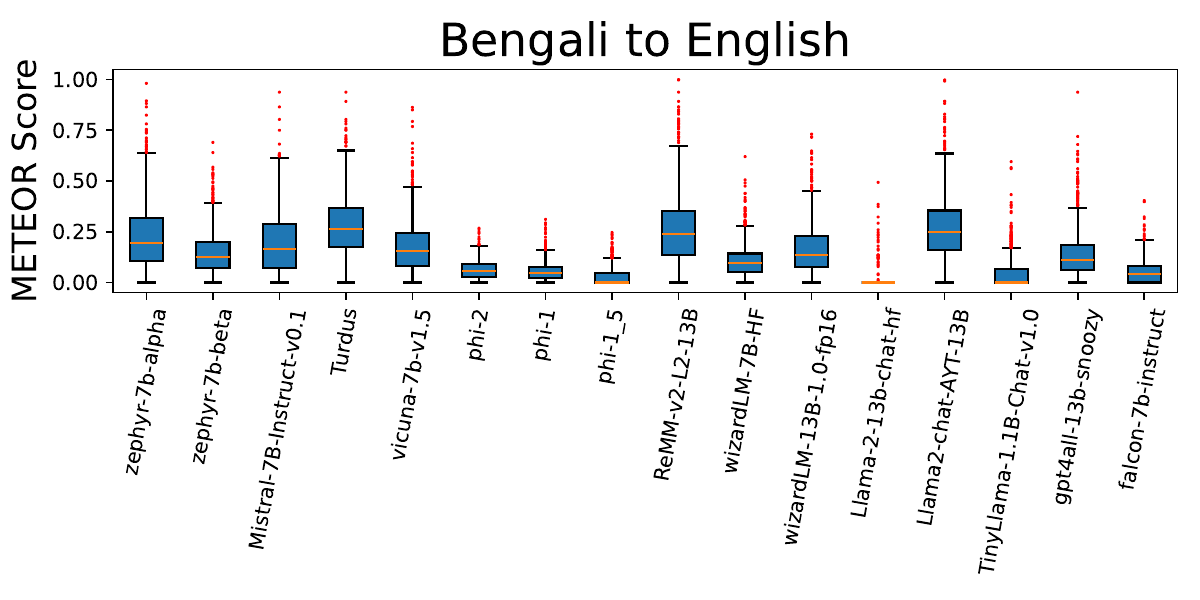}
    \caption{Bengali-to-English dataset per-sentence translation quality and timing statistics  }
    \label{fig:Bengali_translate_stats}
\end{figure}

\begin{figure}[th!]
    \centering
    \includegraphics[width=0.47\textwidth]{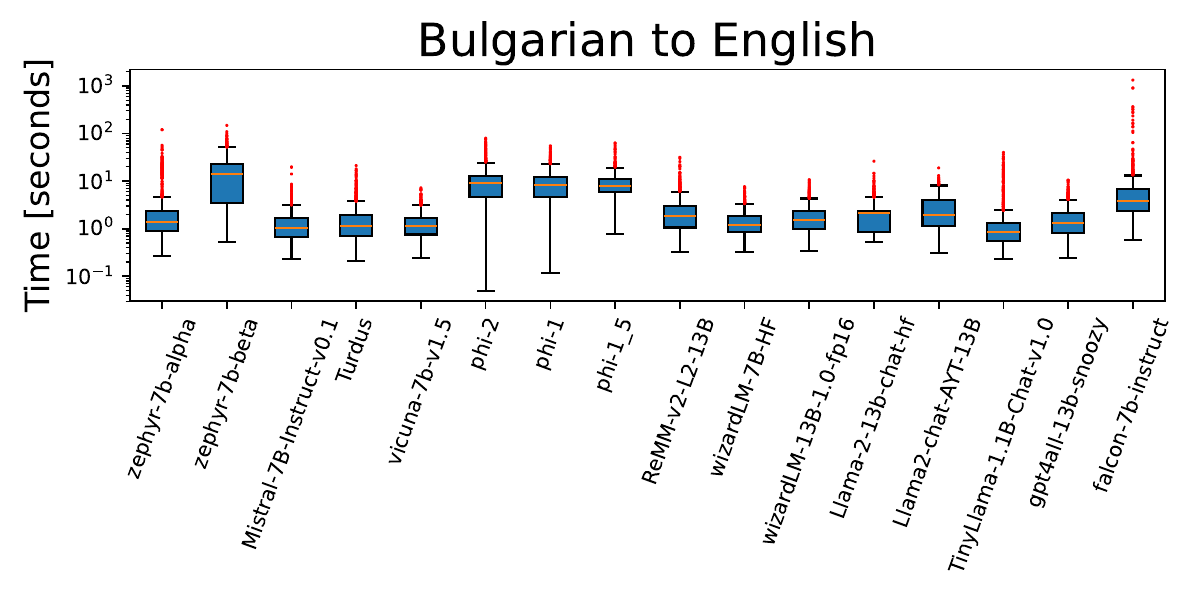}\\
    \includegraphics[width=0.47\textwidth]{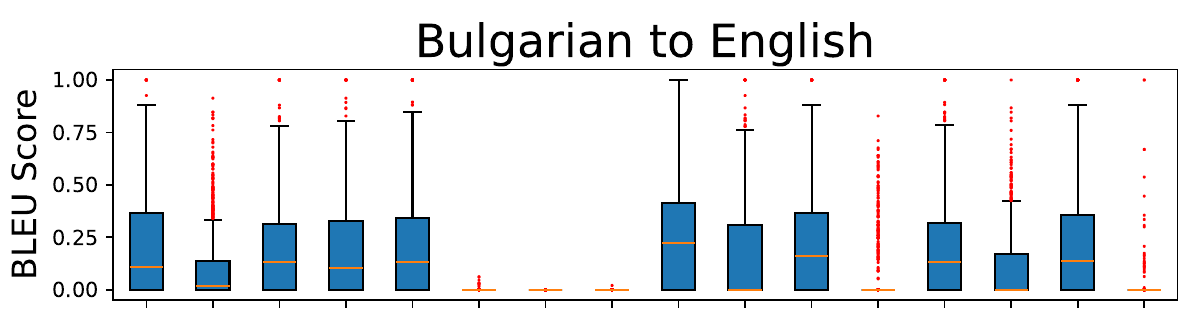}
    \includegraphics[width=0.47\textwidth]{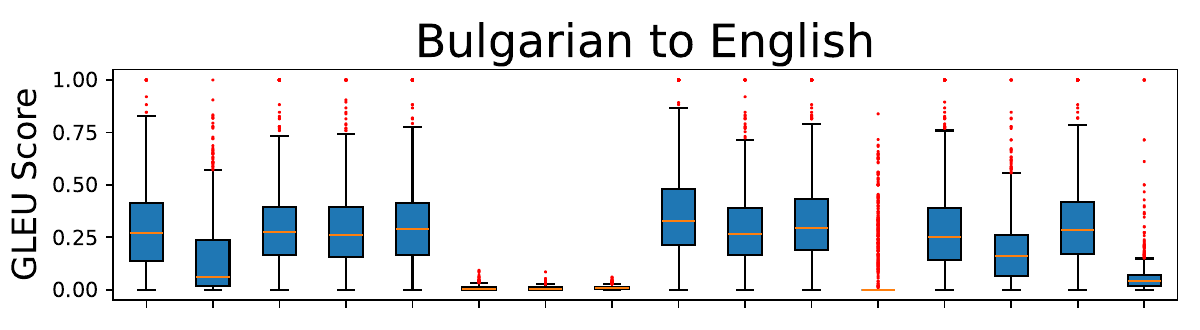}
    \includegraphics[width=0.47\textwidth]{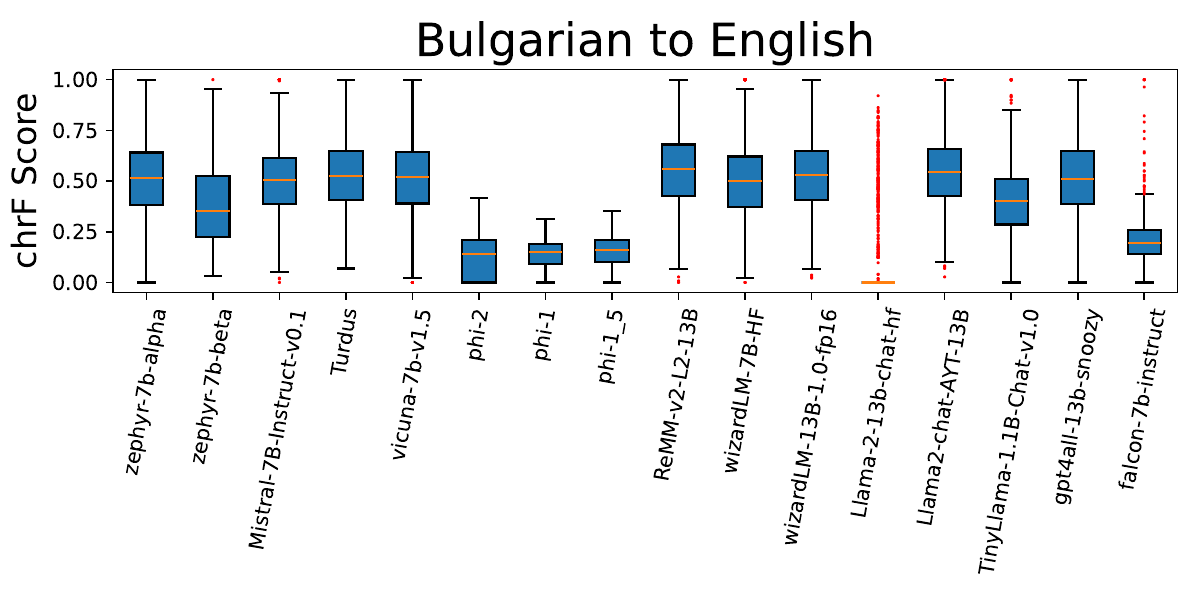}
    \includegraphics[width=0.47\textwidth]{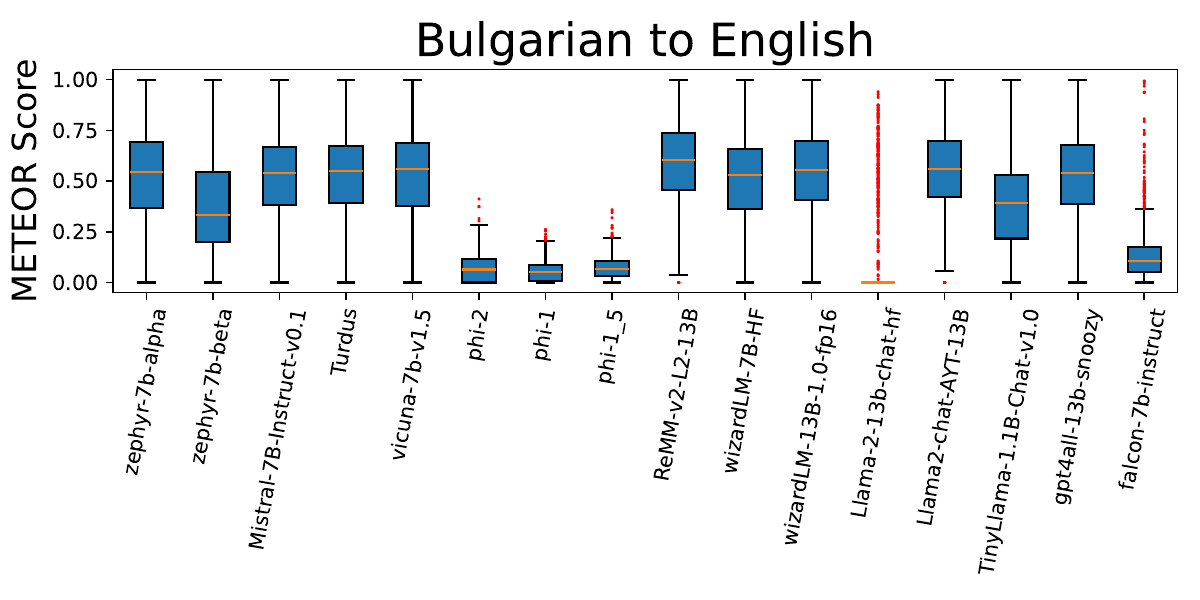}
    \caption{Bulgarian-to-English dataset per-sentence translation quality and timing statistics  }
    \label{fig:Bulgarian_translate_stats}
\end{figure}

\begin{figure}[th!]
    \centering
    \includegraphics[width=0.47\textwidth]{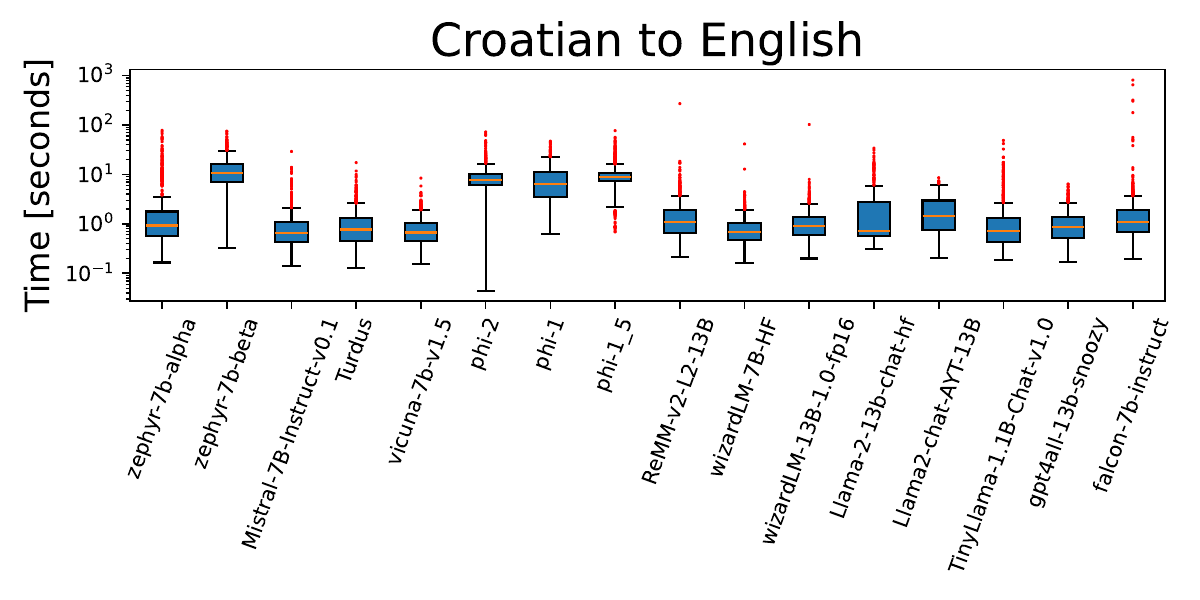}\\
    \includegraphics[width=0.47\textwidth]{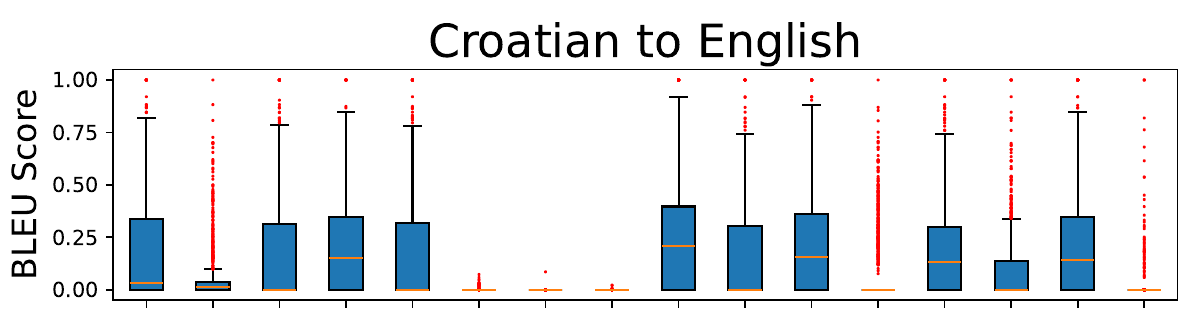}
    \includegraphics[width=0.47\textwidth]{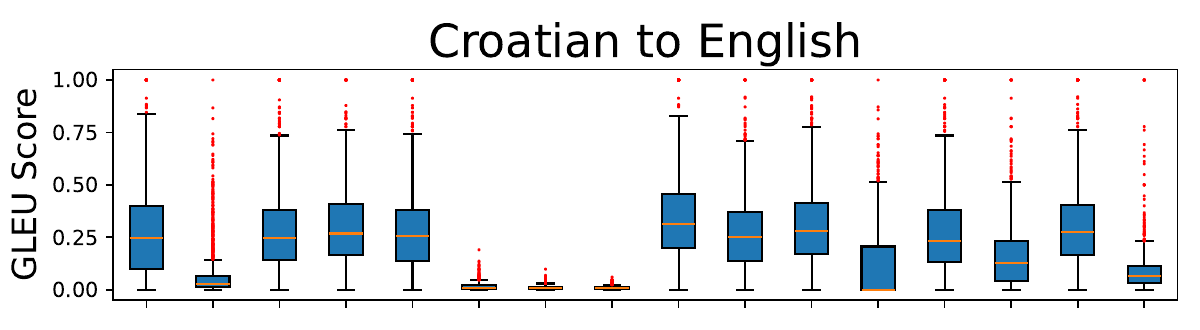}
    \includegraphics[width=0.47\textwidth]{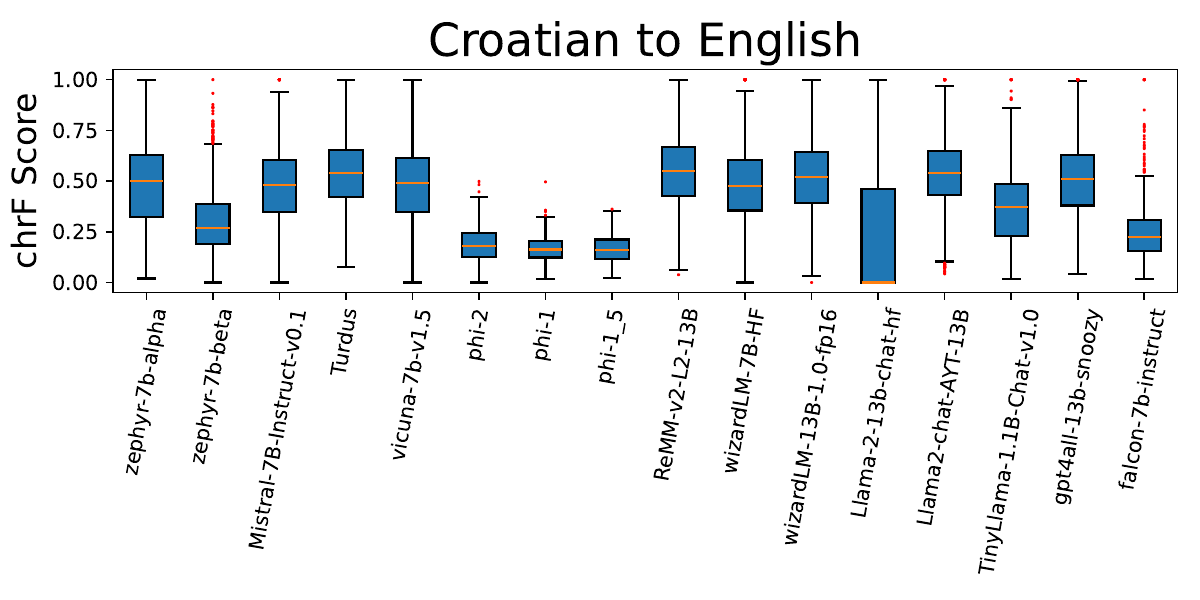}
    \includegraphics[width=0.47\textwidth]{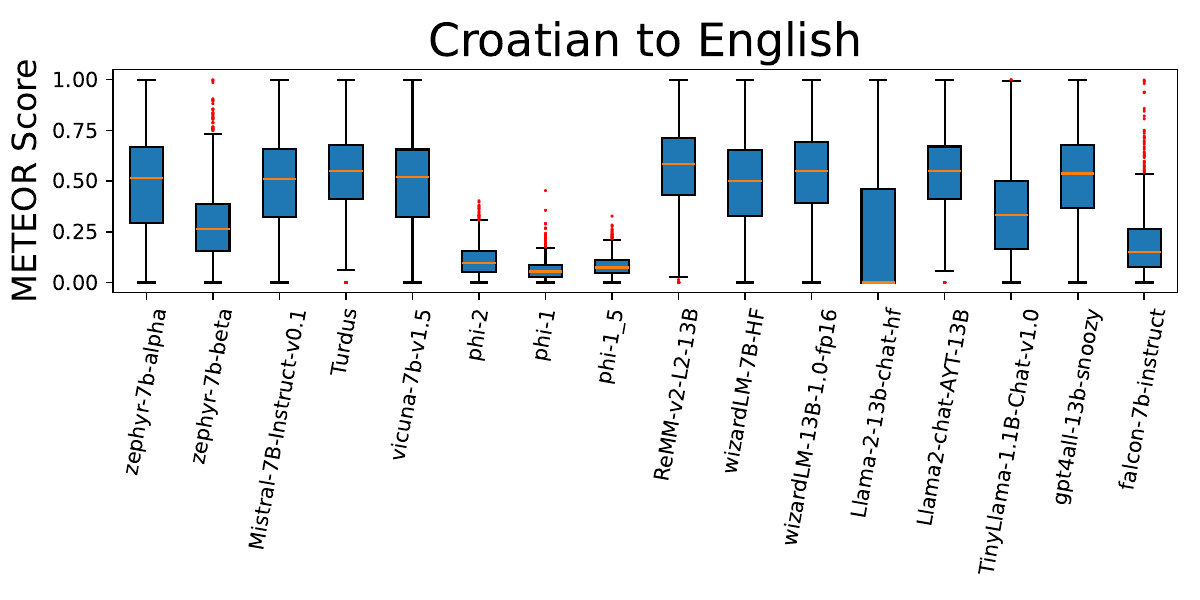}
    \caption{Croatian-to-English dataset per-sentence translation quality and timing statistics  }
    \label{fig:Croatian_translate_stats}
\end{figure}

\begin{figure}[h!]
    \centering
    \includegraphics[width=0.47\textwidth]{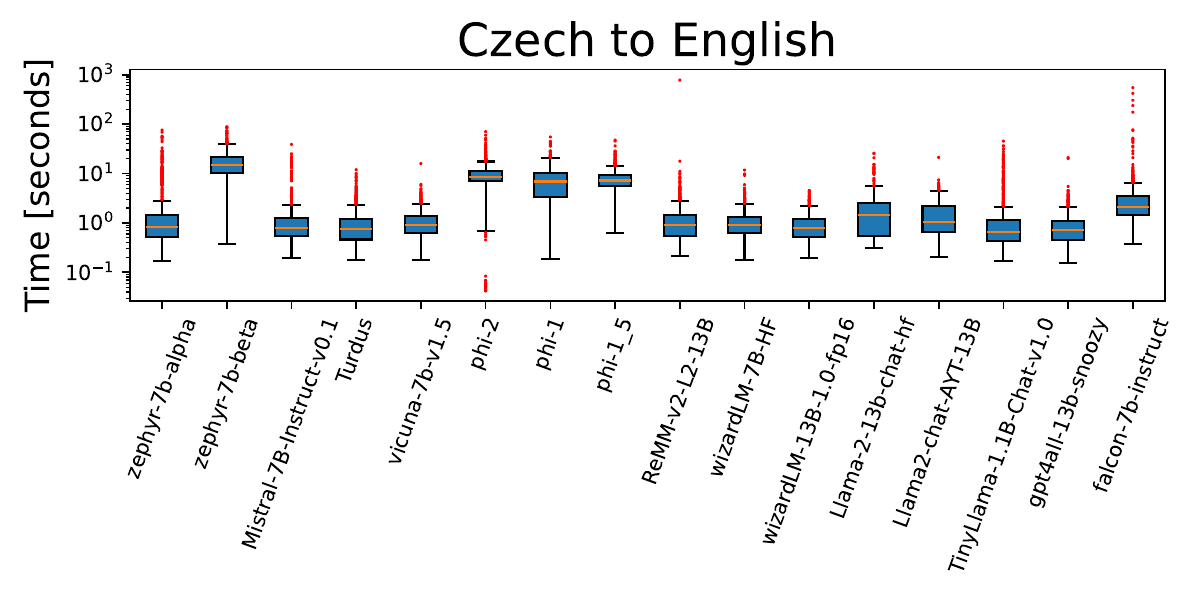}\\
    \includegraphics[width=0.47\textwidth]{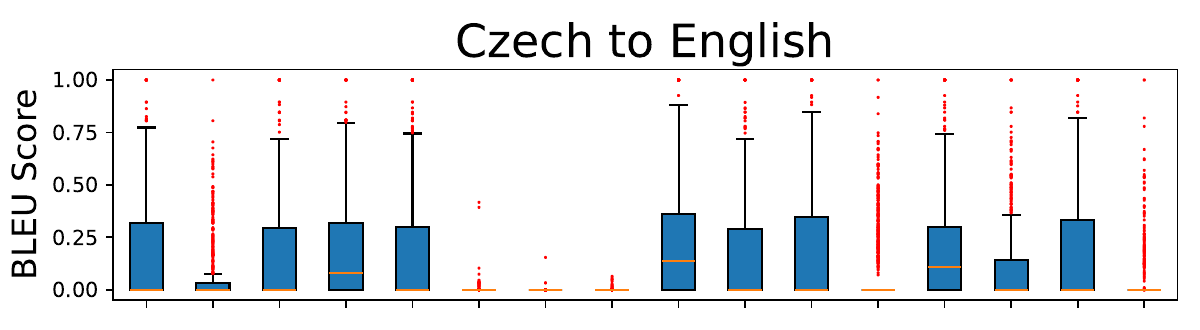}
    \includegraphics[width=0.47\textwidth]{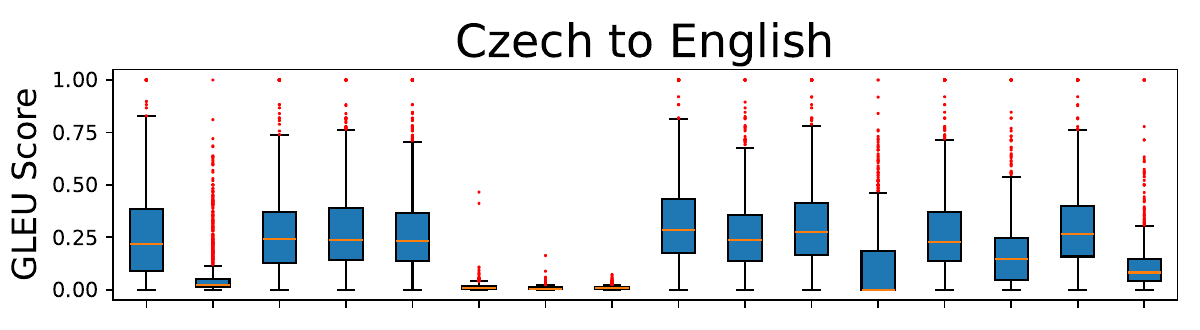}
    \includegraphics[width=0.47\textwidth]{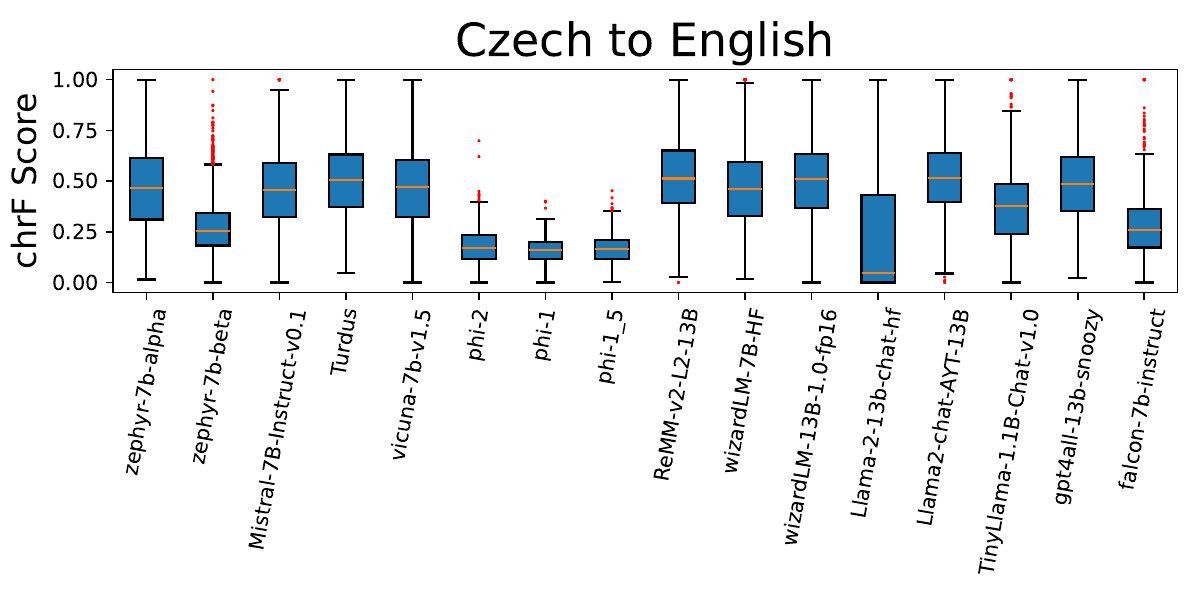}
    \includegraphics[width=0.47\textwidth]{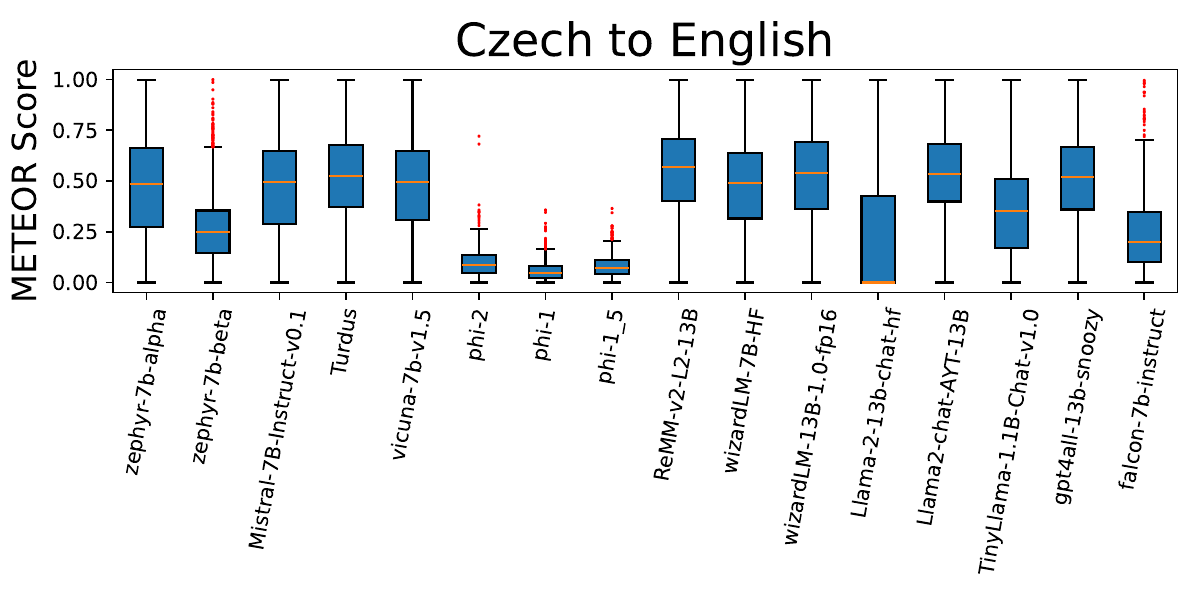}
    \caption{Czech-to-English dataset per-sentence translation quality and timing statistics }
    \label{fig:Czech_translate_stats}
\end{figure}

\begin{figure}[th!]
    \centering
    \includegraphics[width=0.47\textwidth]{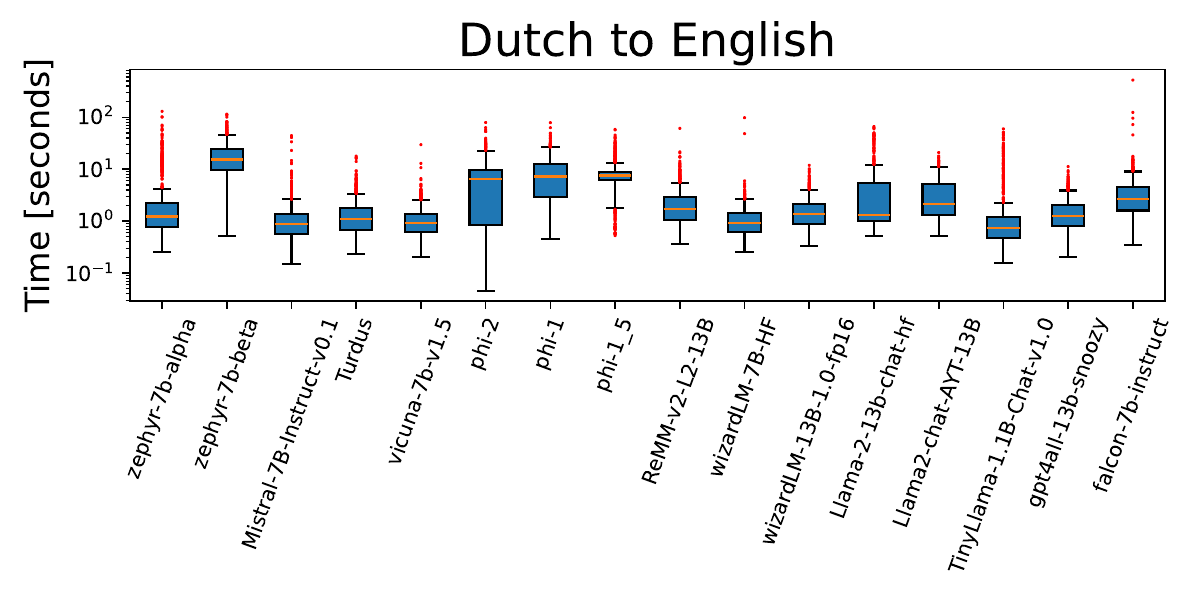}\\
    \includegraphics[width=0.47\textwidth]{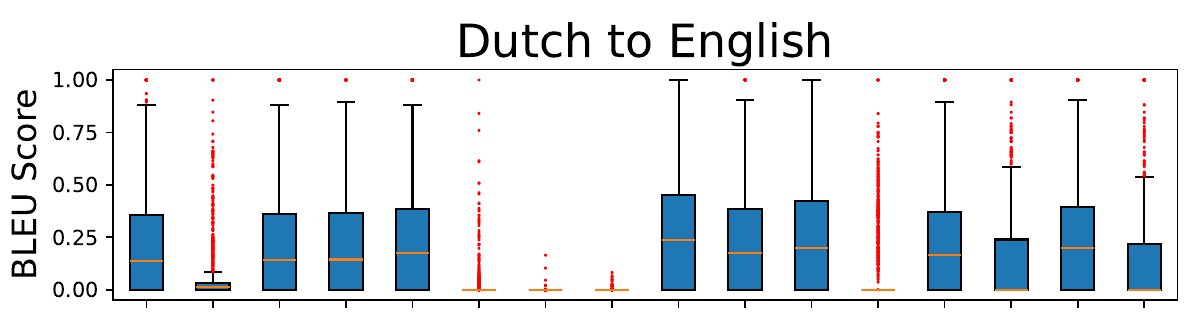}
    \includegraphics[width=0.47\textwidth]{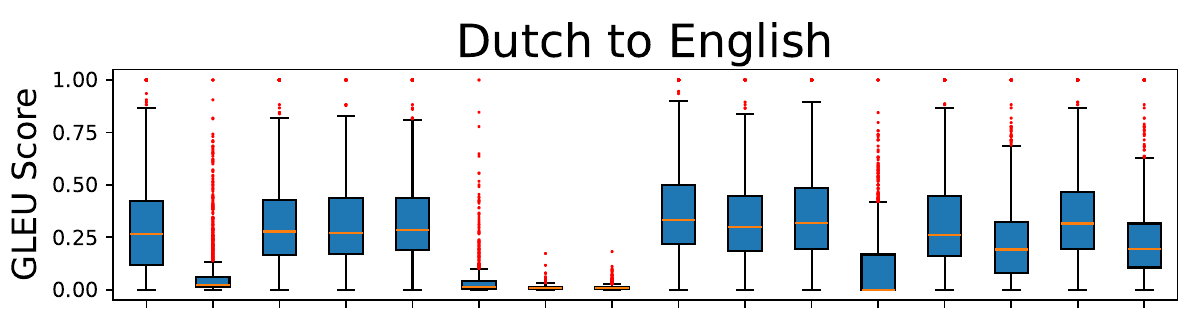}
    \includegraphics[width=0.47\textwidth]{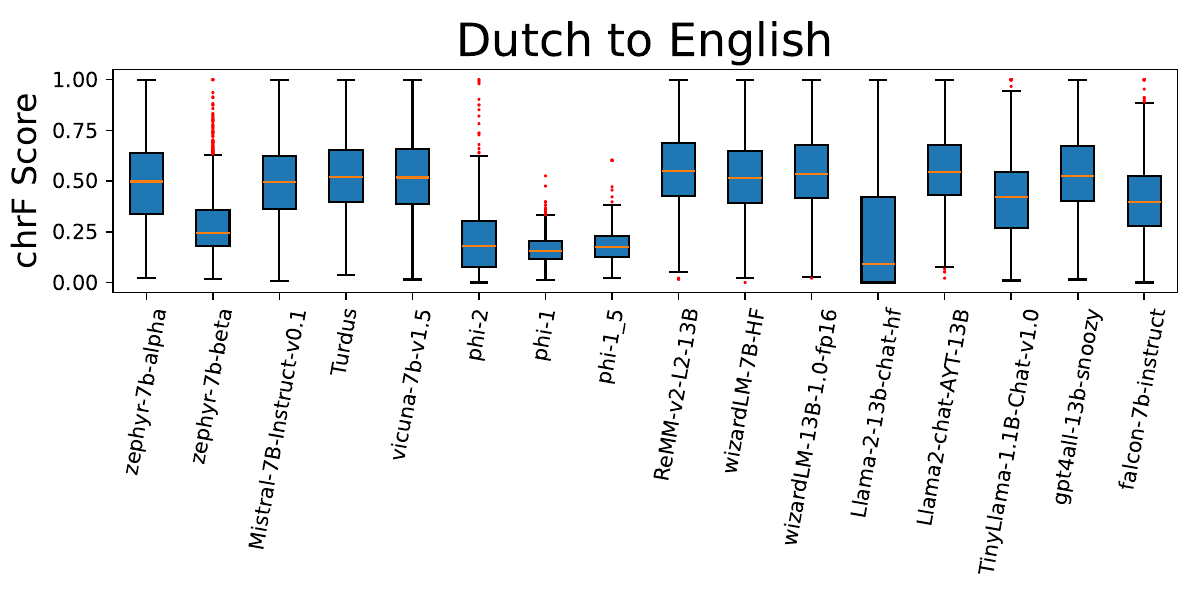}
    \includegraphics[width=0.47\textwidth]{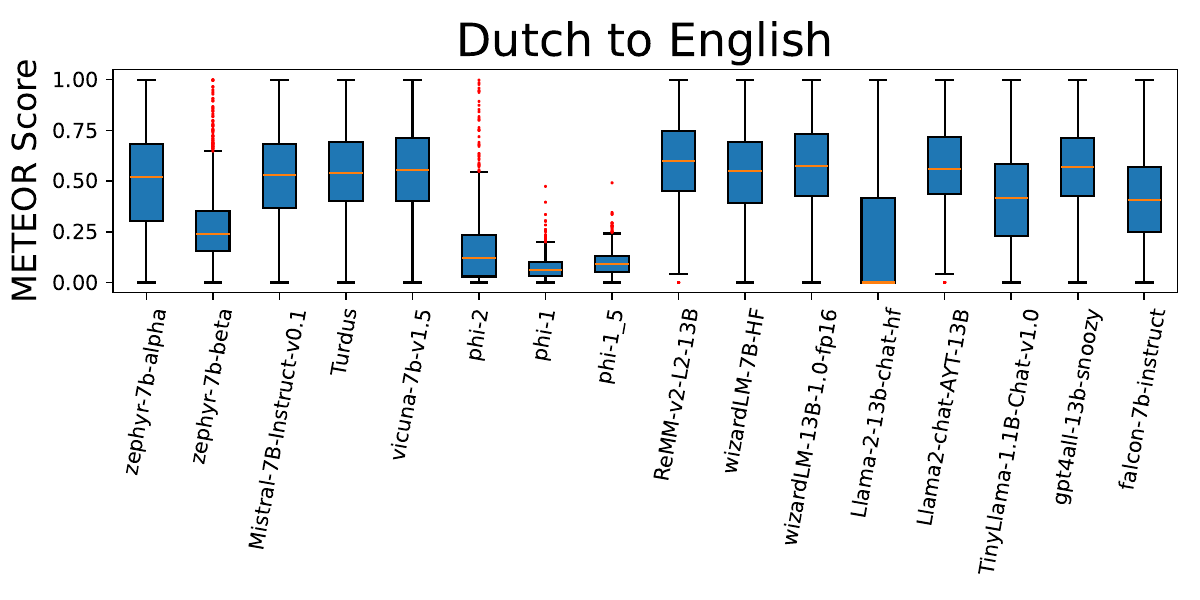}
    \caption{Dutch-to-English dataset per-sentence translation quality and timing statistics  }
    \label{fig:Dutch_translate_stats}
\end{figure}

\begin{figure}[th!]
    \centering
    \includegraphics[width=0.47\textwidth]{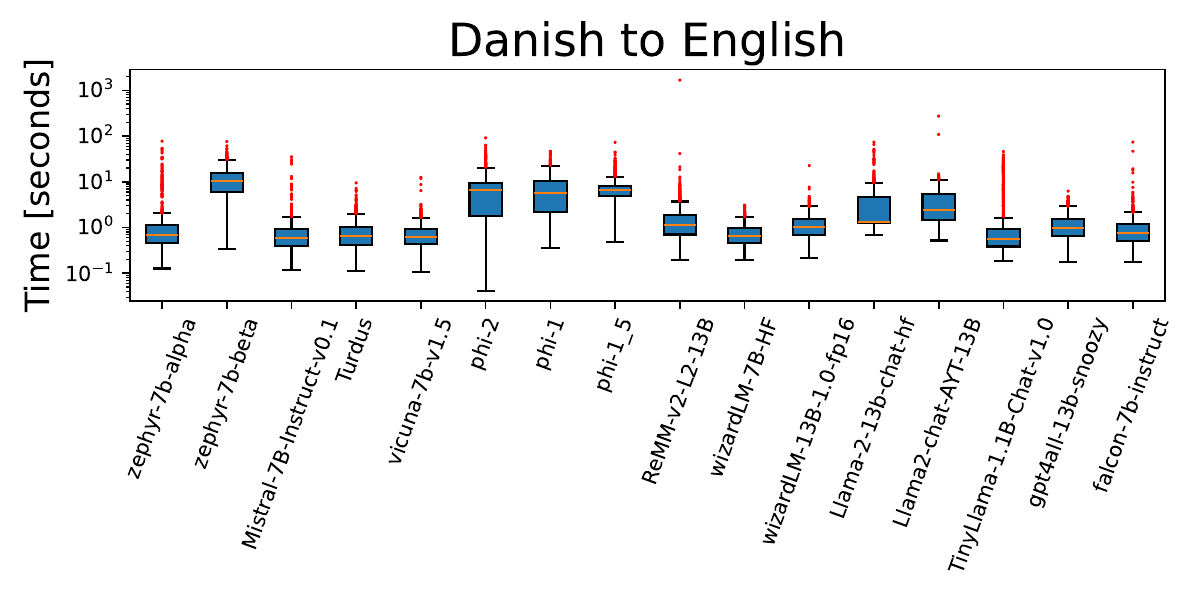}\\
    \includegraphics[width=0.47\textwidth]{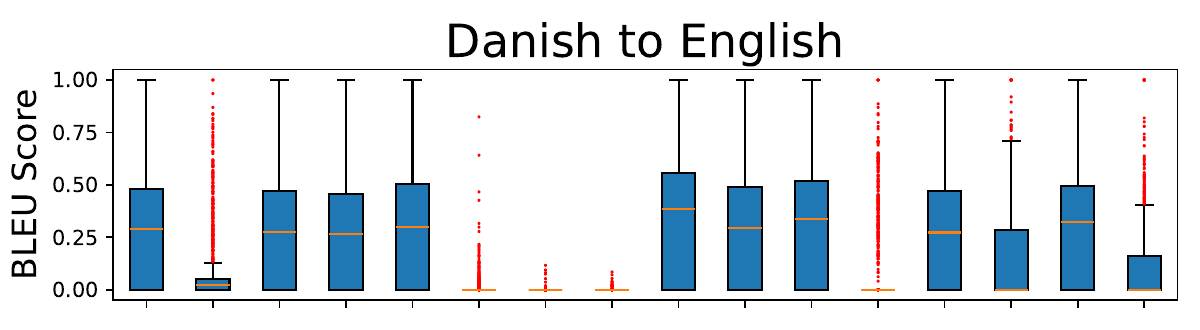}
    \includegraphics[width=0.47\textwidth]{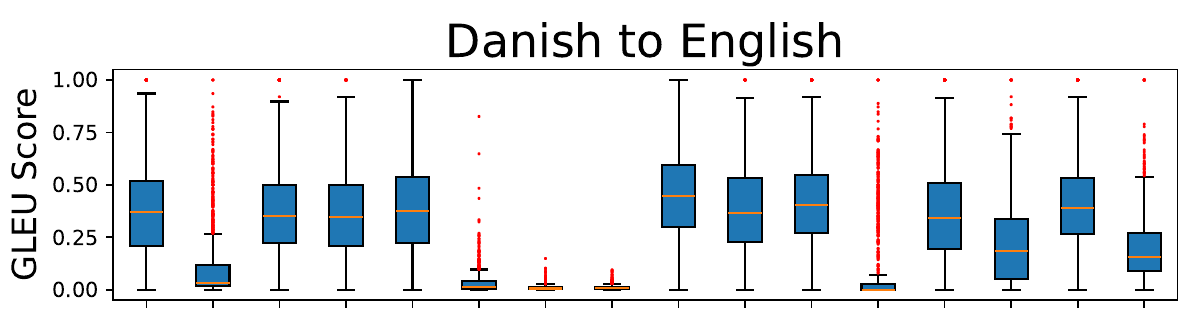}
    \includegraphics[width=0.47\textwidth]{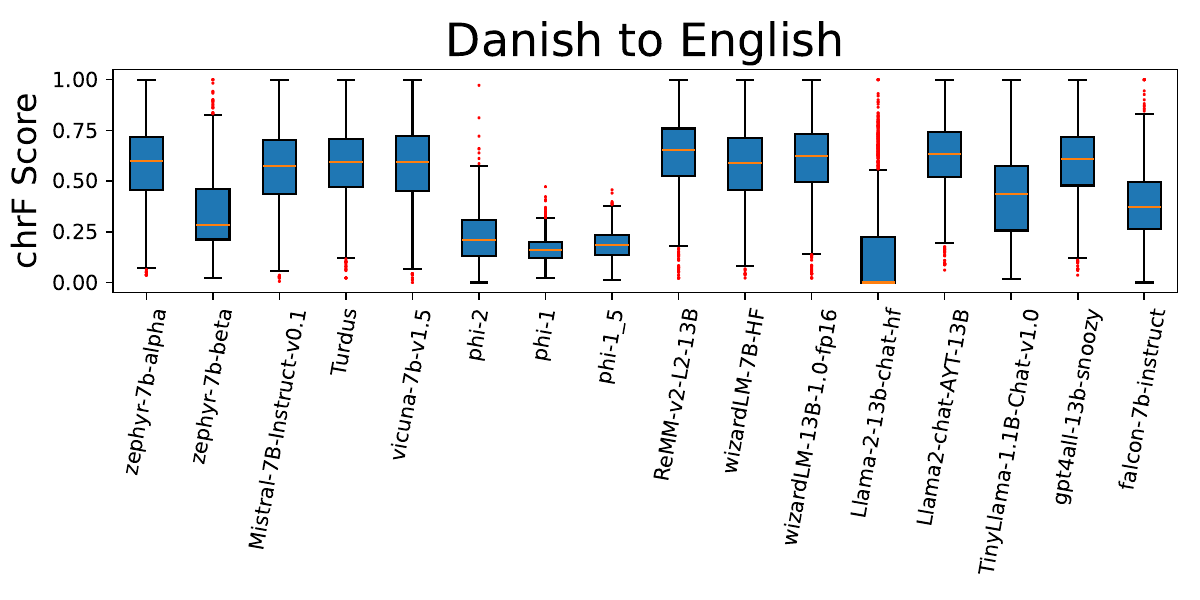}
    \includegraphics[width=0.47\textwidth]{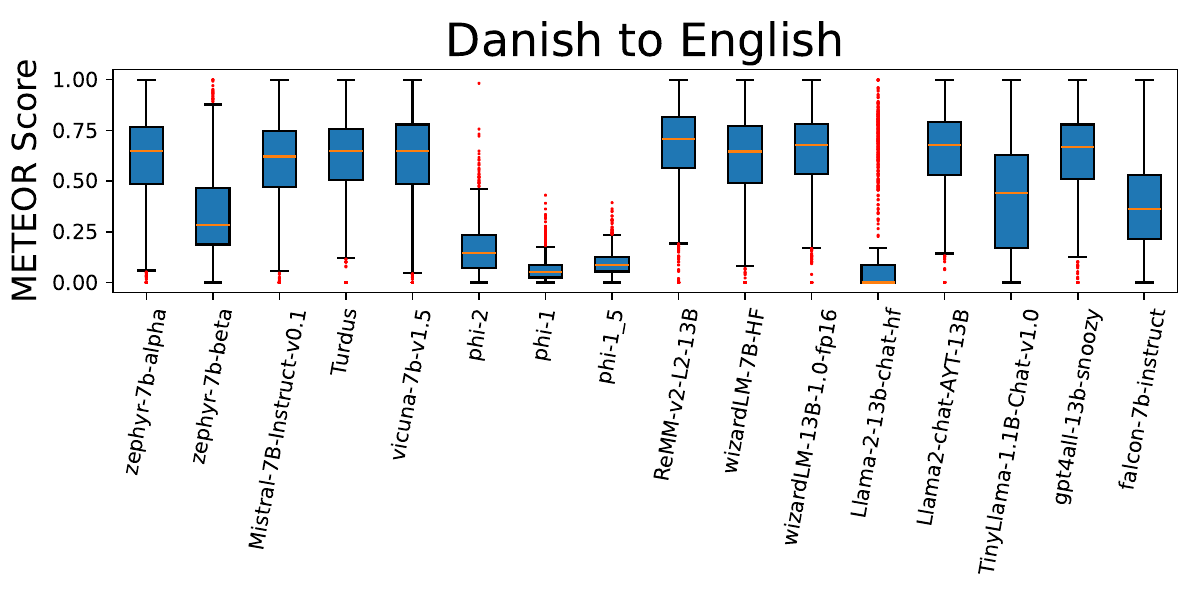}
    \caption{Danish-to-English dataset per-sentence translation quality and timing statistics  }
    \label{fig:Danish_translate_stats}
\end{figure}

\begin{figure}[th!]
    \centering
    \includegraphics[width=0.47\textwidth]{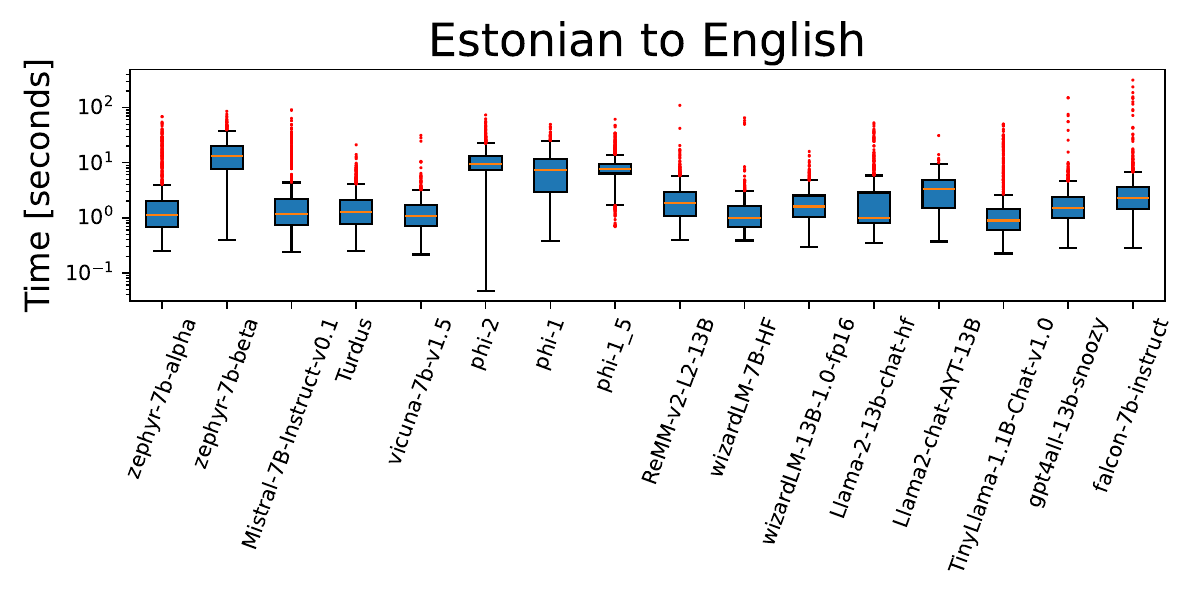}\\
    \includegraphics[width=0.47\textwidth]{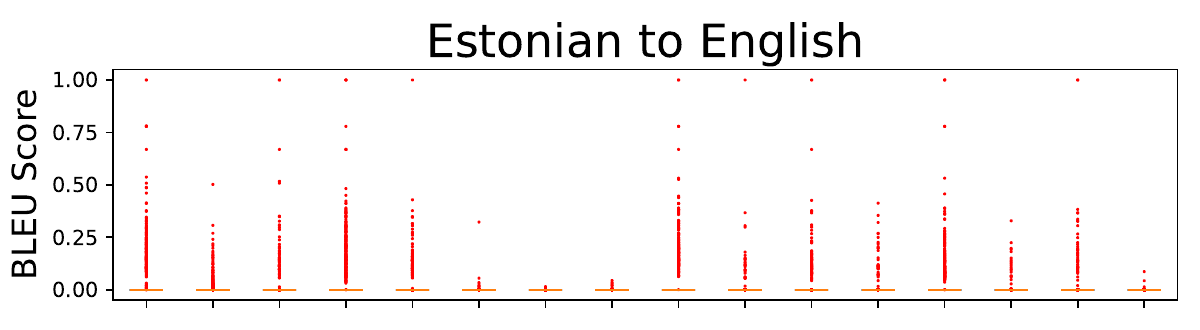}
    \includegraphics[width=0.47\textwidth]{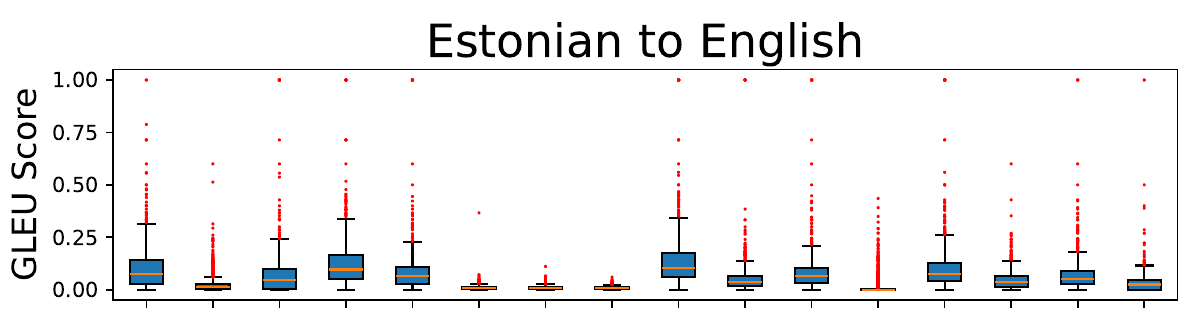}
    \includegraphics[width=0.47\textwidth]{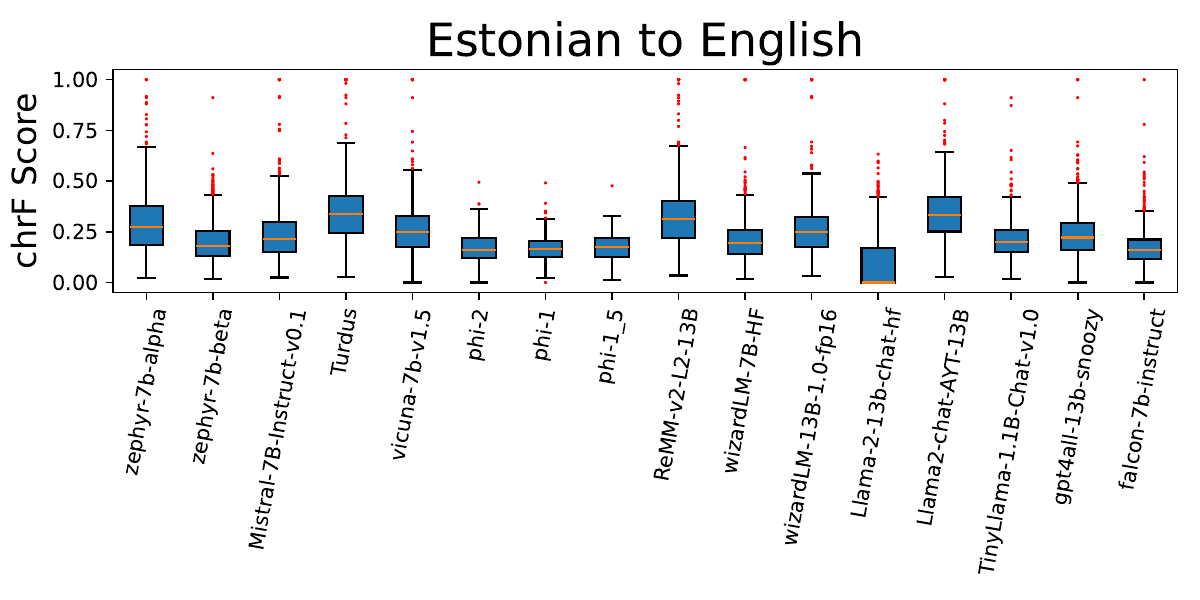}
    \includegraphics[width=0.47\textwidth]{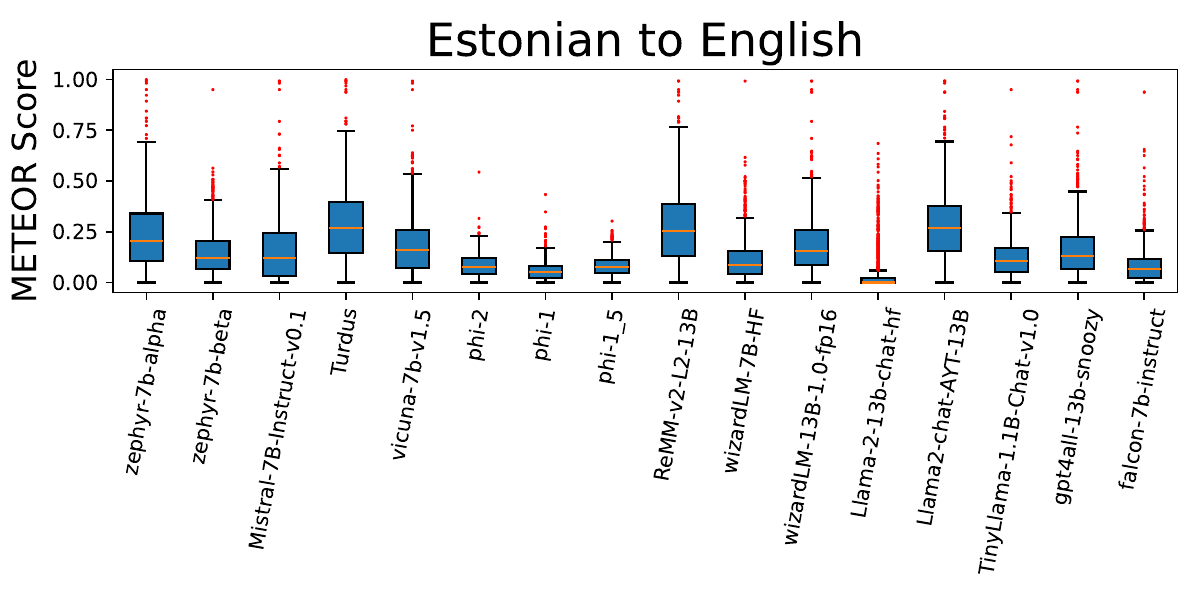}
    \caption{Estonian-to-English dataset per-sentence translation quality and timing statistics  }
    \label{fig:Estonian_translate_stats}
\end{figure}

\begin{figure}[th!]
    \centering
    \includegraphics[width=0.47\textwidth]{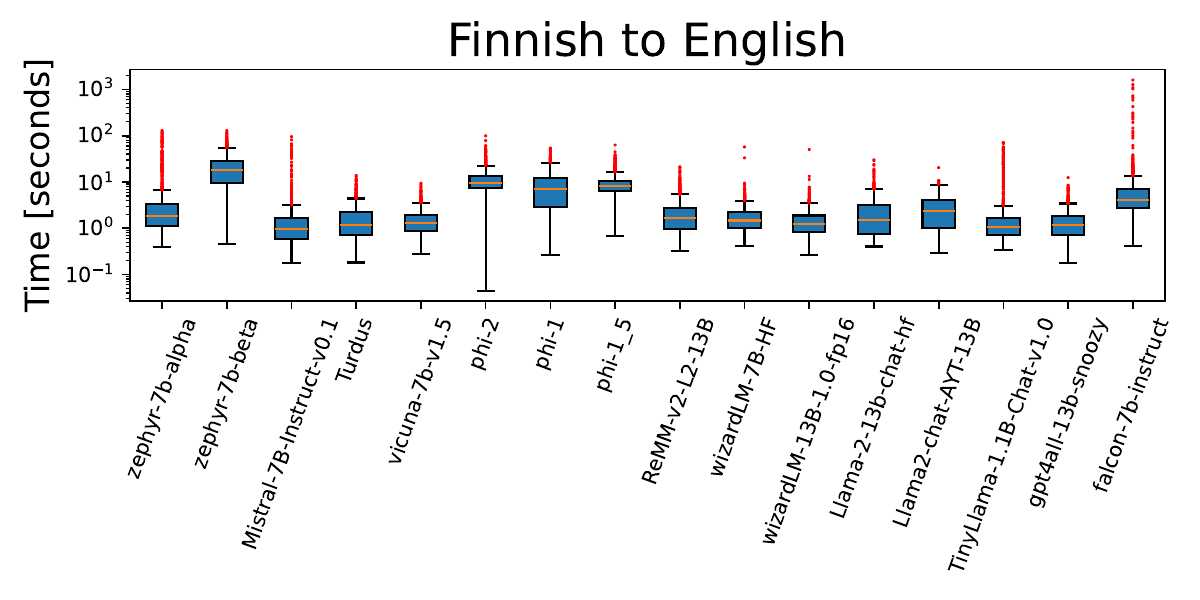}\\
    \includegraphics[width=0.47\textwidth]{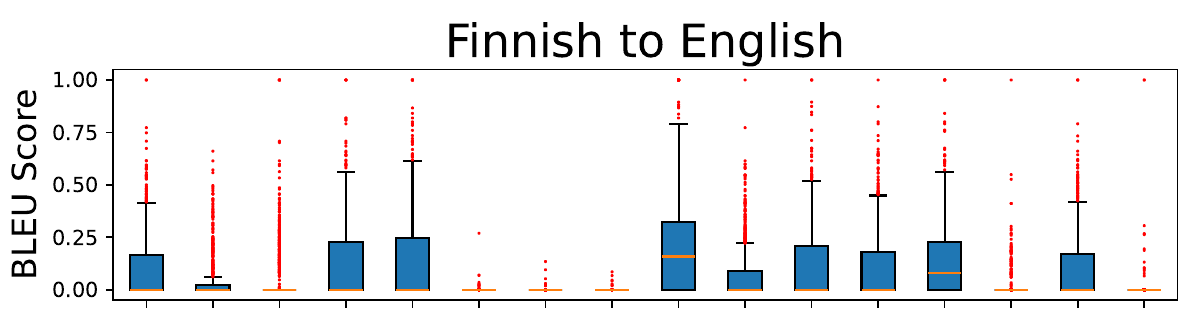}
    \includegraphics[width=0.47\textwidth]{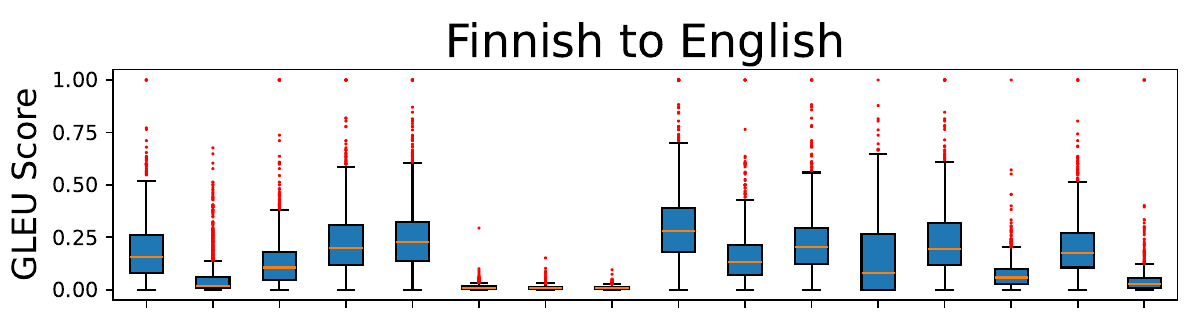}
    \includegraphics[width=0.47\textwidth]{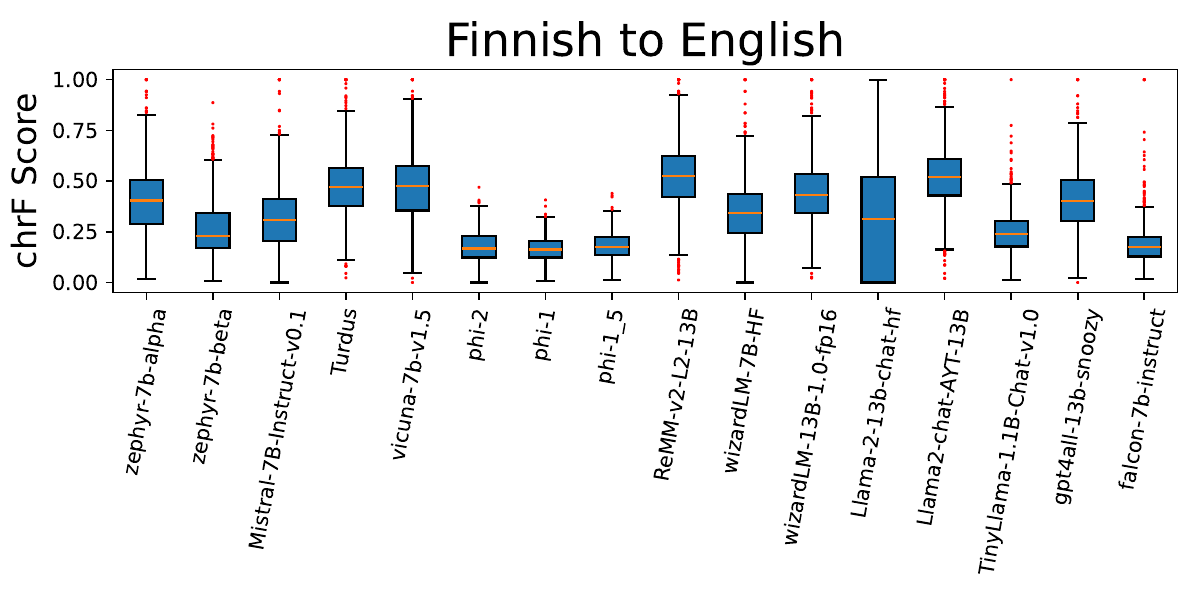}
    \includegraphics[width=0.47\textwidth]{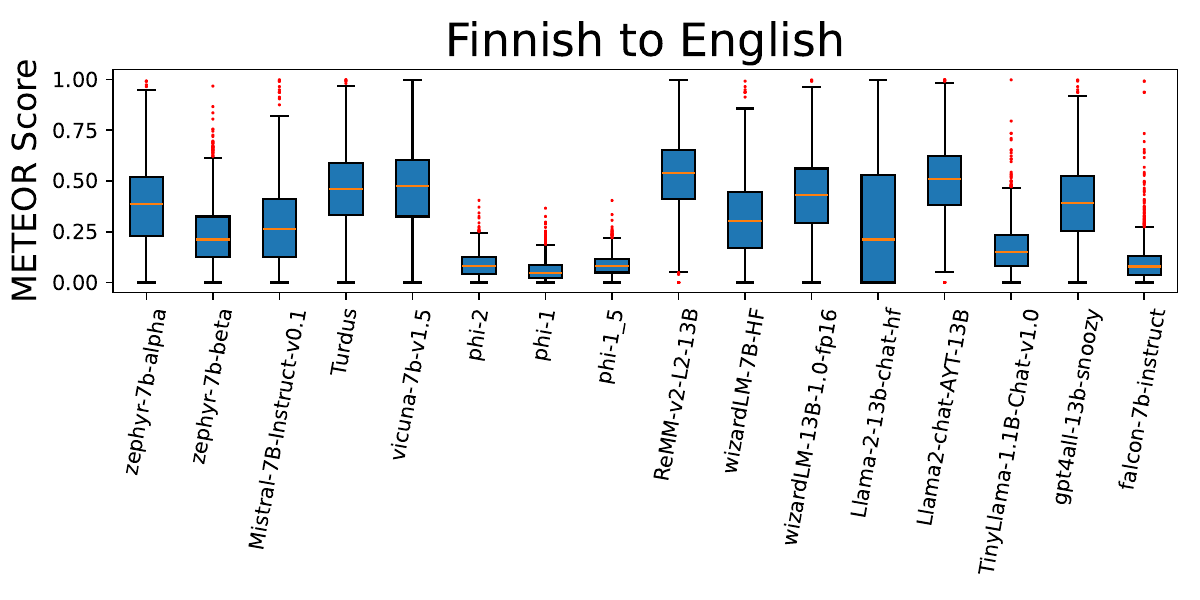}
    \caption{Finnish-to-English dataset per-sentence translation quality and timing statistics  }
    \label{fig:Finnish_translate_stats}
\end{figure}

\begin{figure}[h!]
    \centering
    \includegraphics[width=0.47\textwidth]{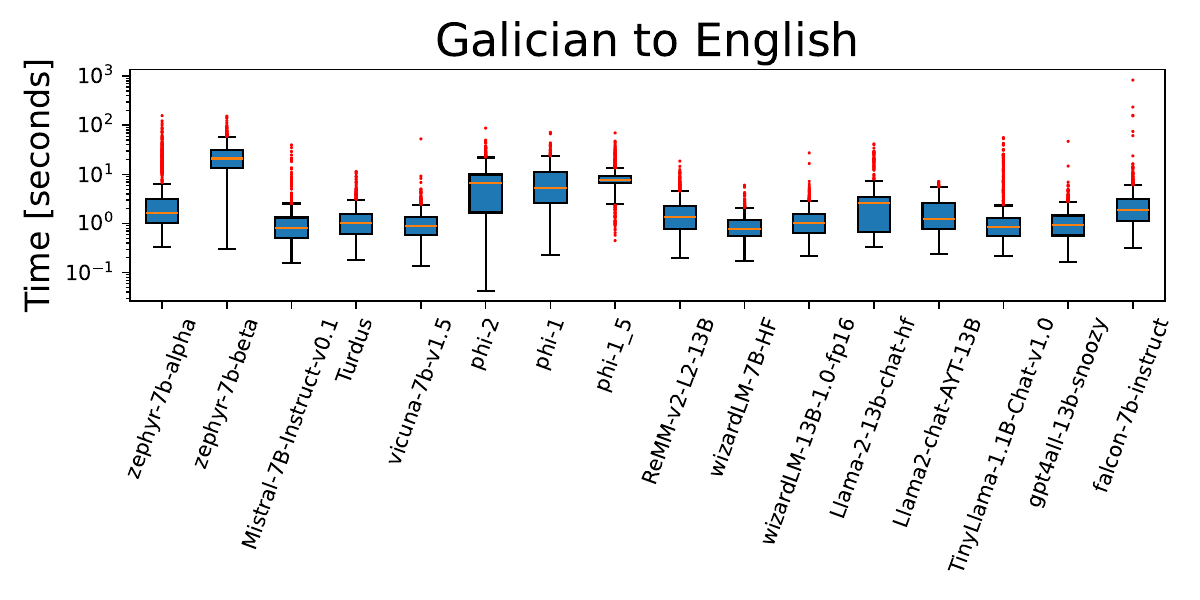}\\
    \includegraphics[width=0.47\textwidth]{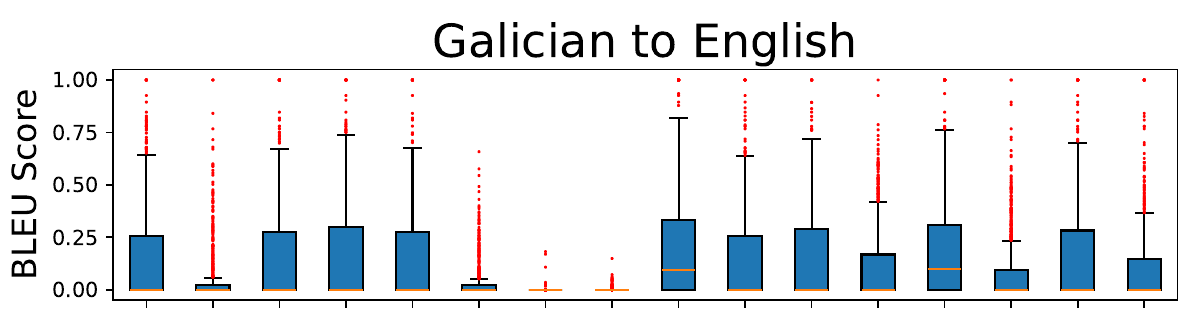}
    \includegraphics[width=0.47\textwidth]{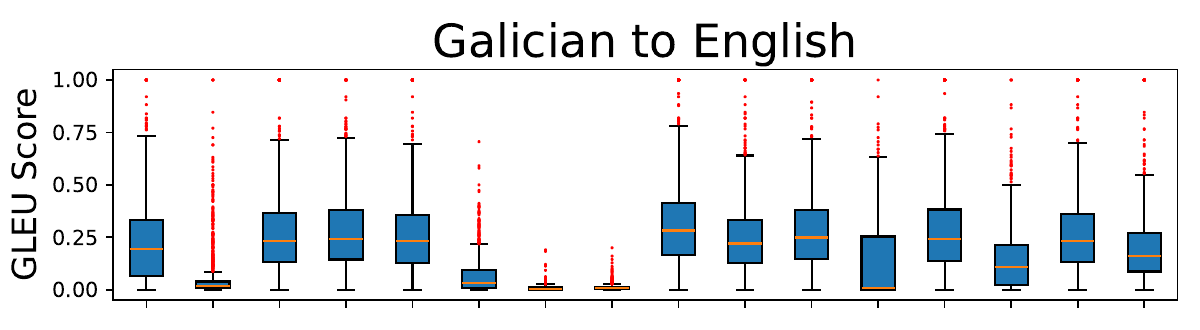}
    \includegraphics[width=0.47\textwidth]{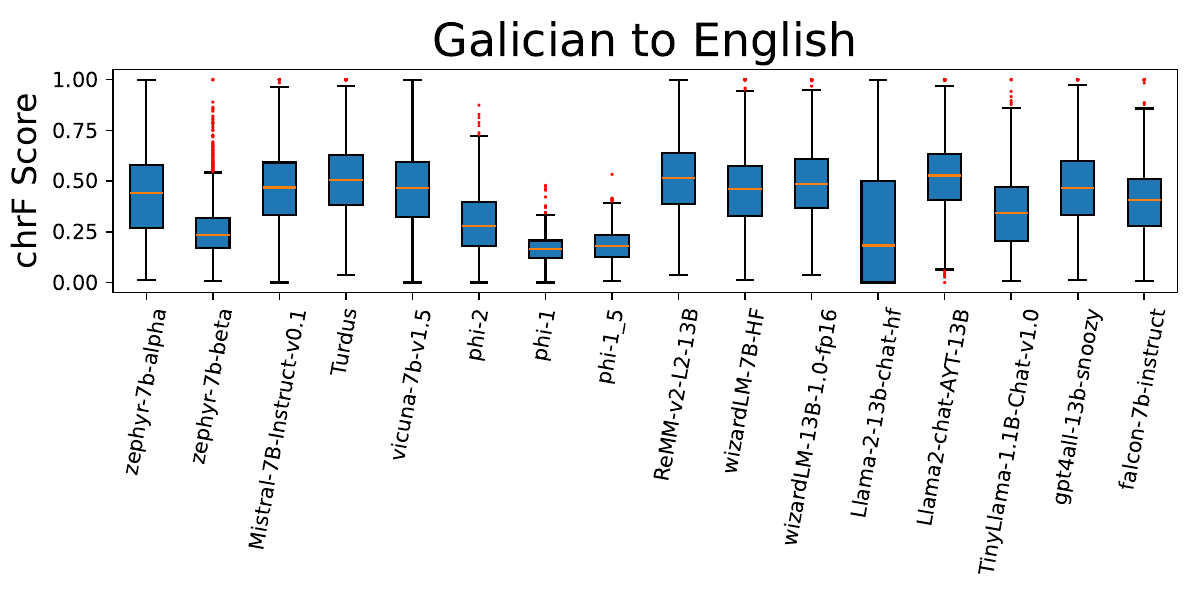}
    \includegraphics[width=0.47\textwidth]{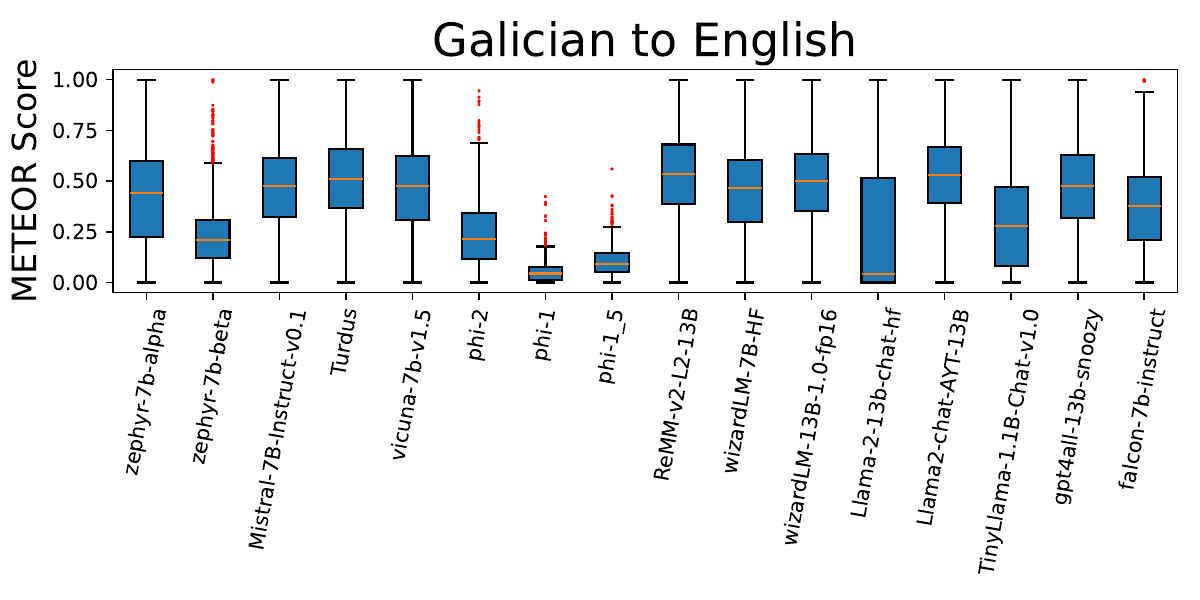}
    \caption{Galician-to-English dataset per-sentence translation quality and timing statistics }
    \label{fig:Galician_translate_stats}
\end{figure}

\begin{figure}[th!]
    \centering
    \includegraphics[width=0.47\textwidth]{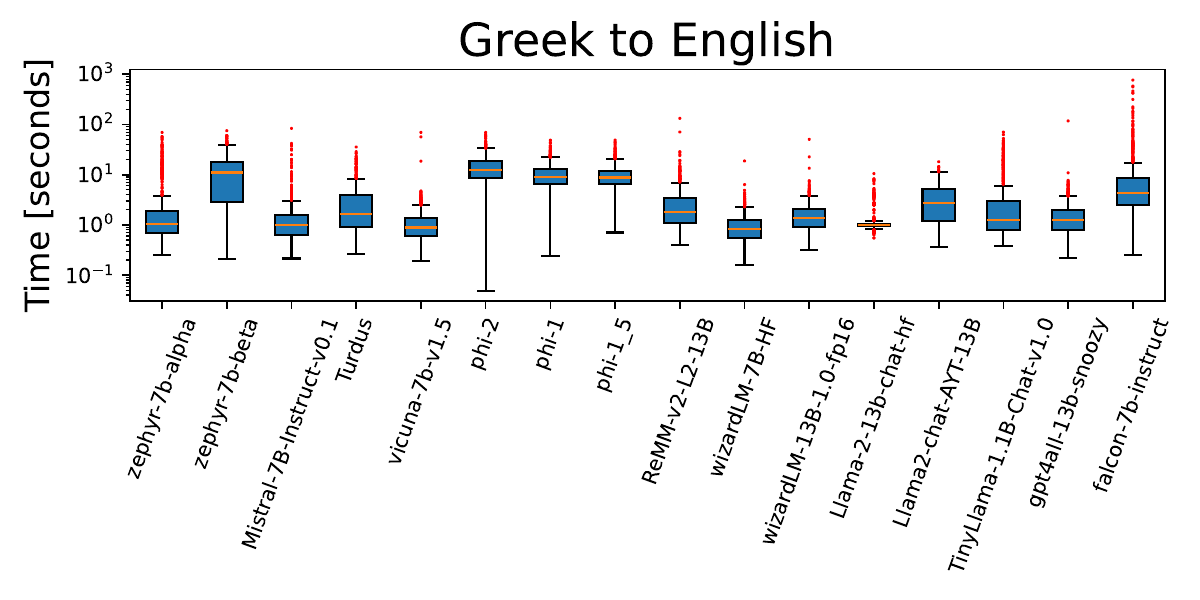}\\
    \includegraphics[width=0.47\textwidth]{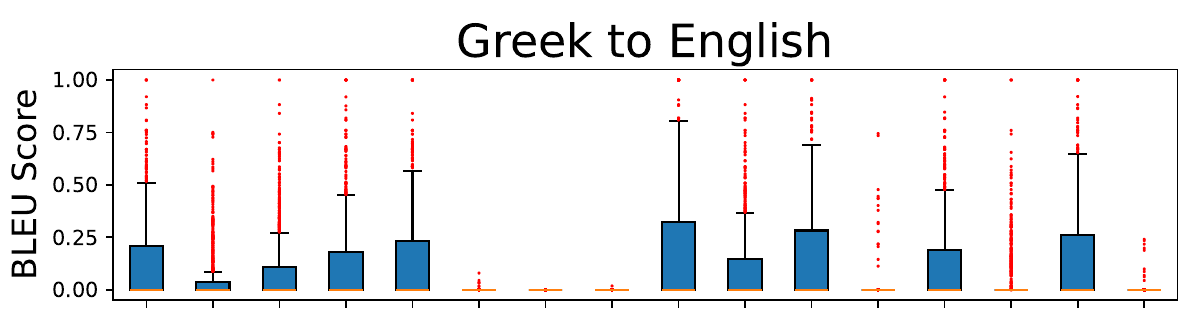}
    \includegraphics[width=0.47\textwidth]{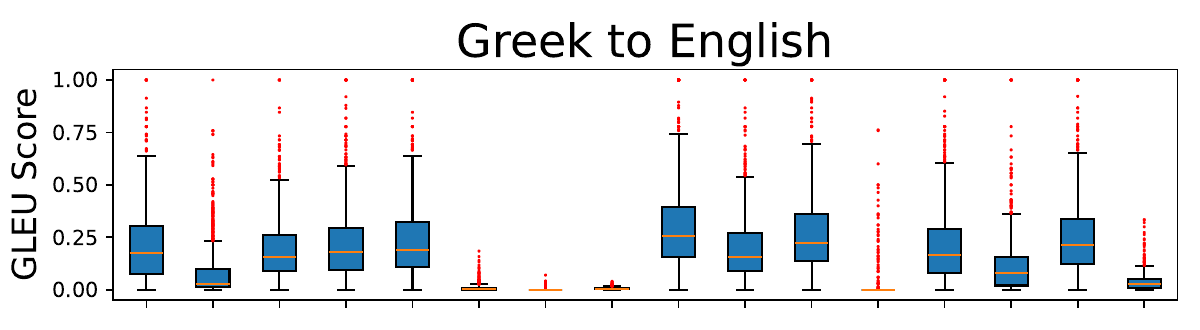}
    \includegraphics[width=0.47\textwidth]{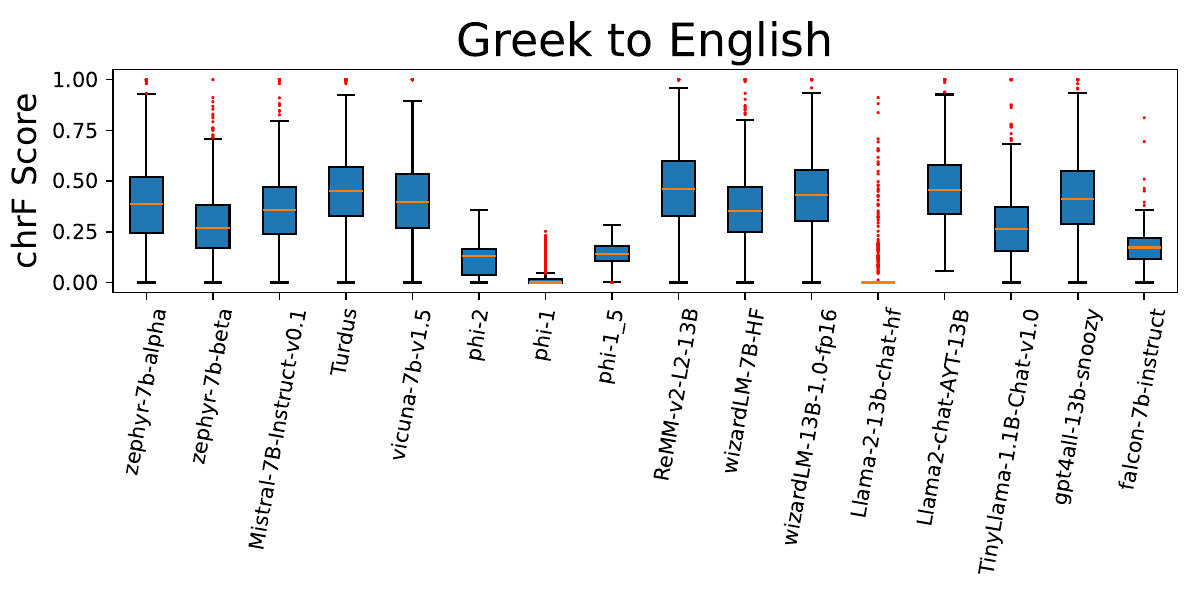}
    \includegraphics[width=0.47\textwidth]{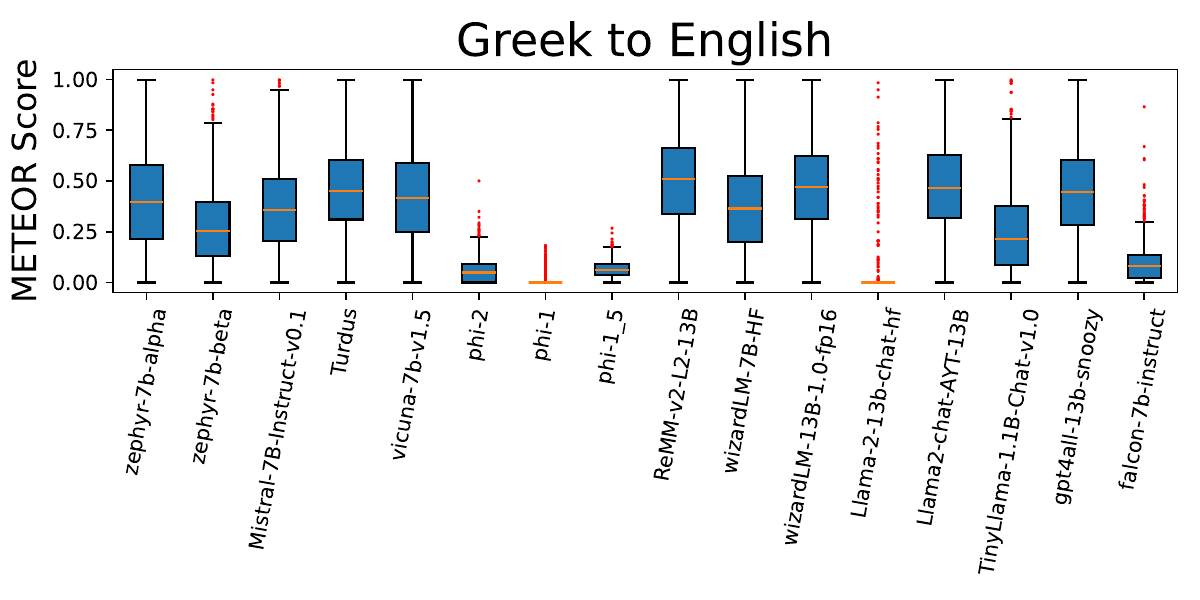}
    \caption{Greek-to-English dataset per-sentence translation quality and timing statistics  }
    \label{fig:Greek_translate_stats}
\end{figure}

\begin{figure}[th!]
    \centering
    \includegraphics[width=0.47\textwidth]{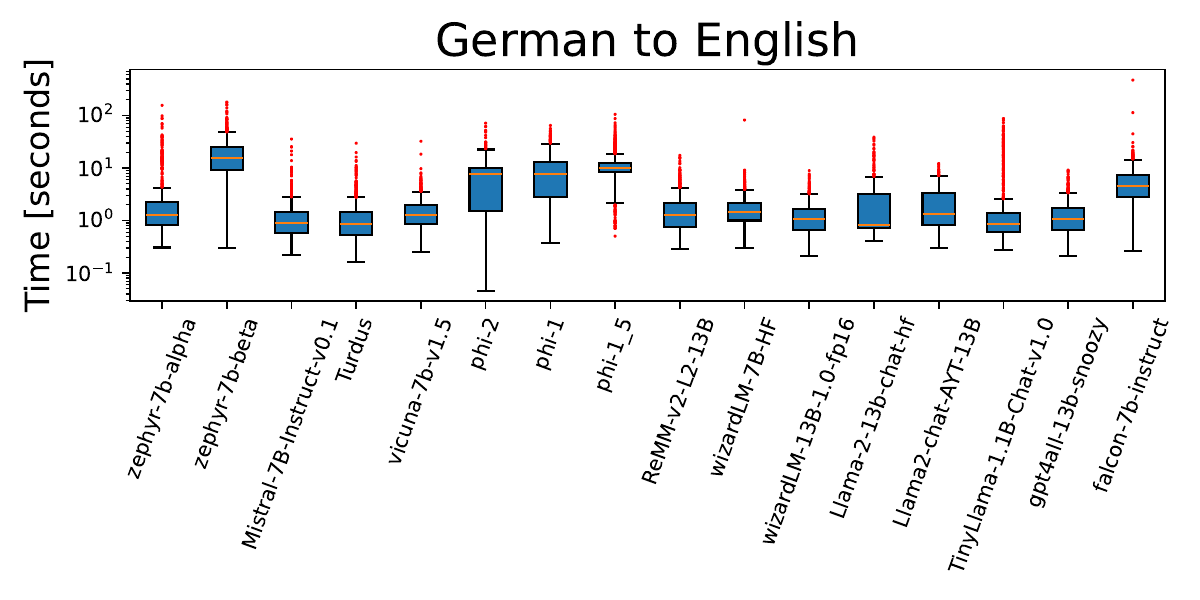}\\
    \includegraphics[width=0.47\textwidth]{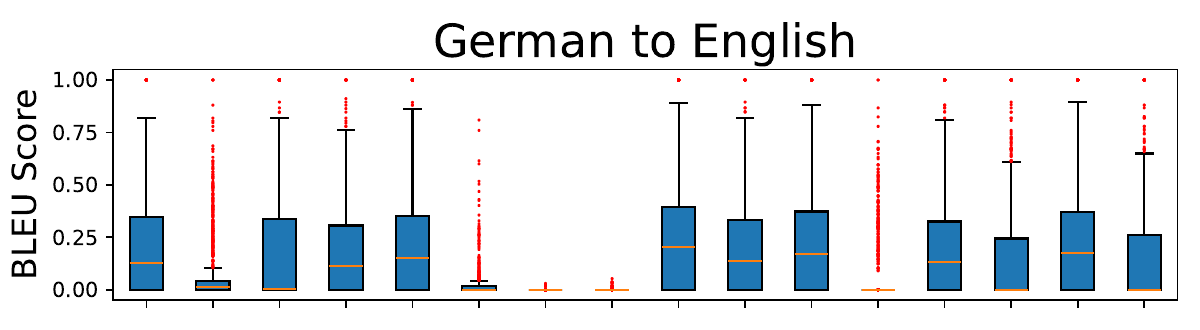}
    \includegraphics[width=0.47\textwidth]{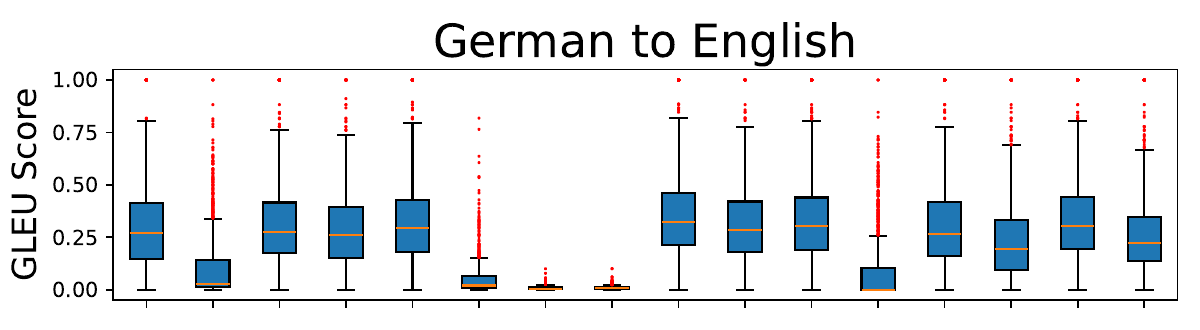}
    \includegraphics[width=0.47\textwidth]{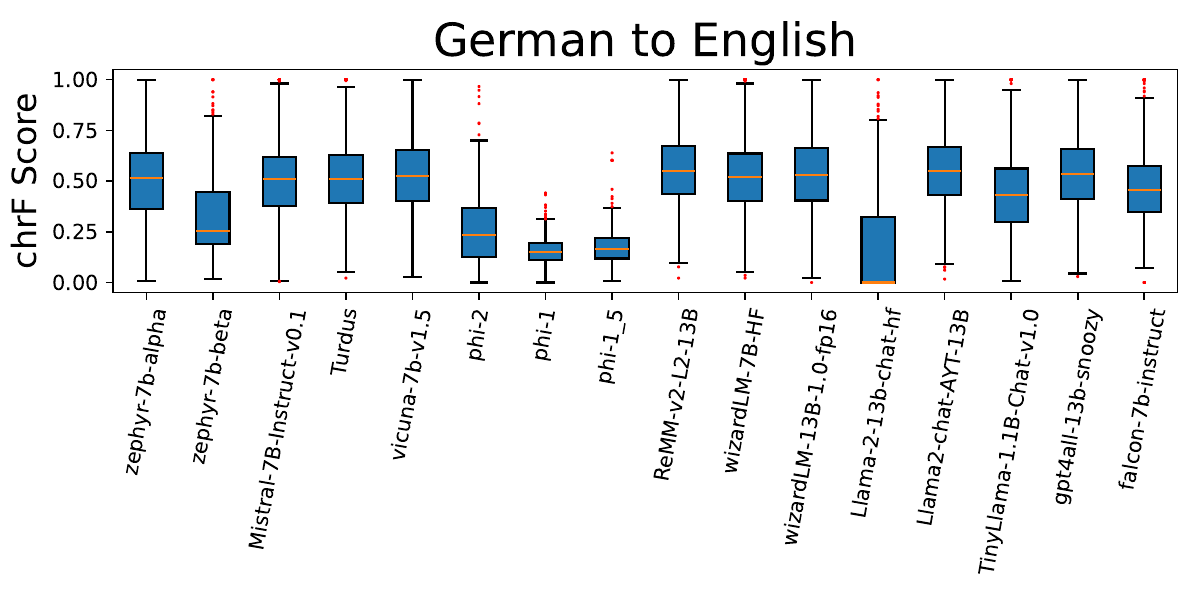}
    \includegraphics[width=0.47\textwidth]{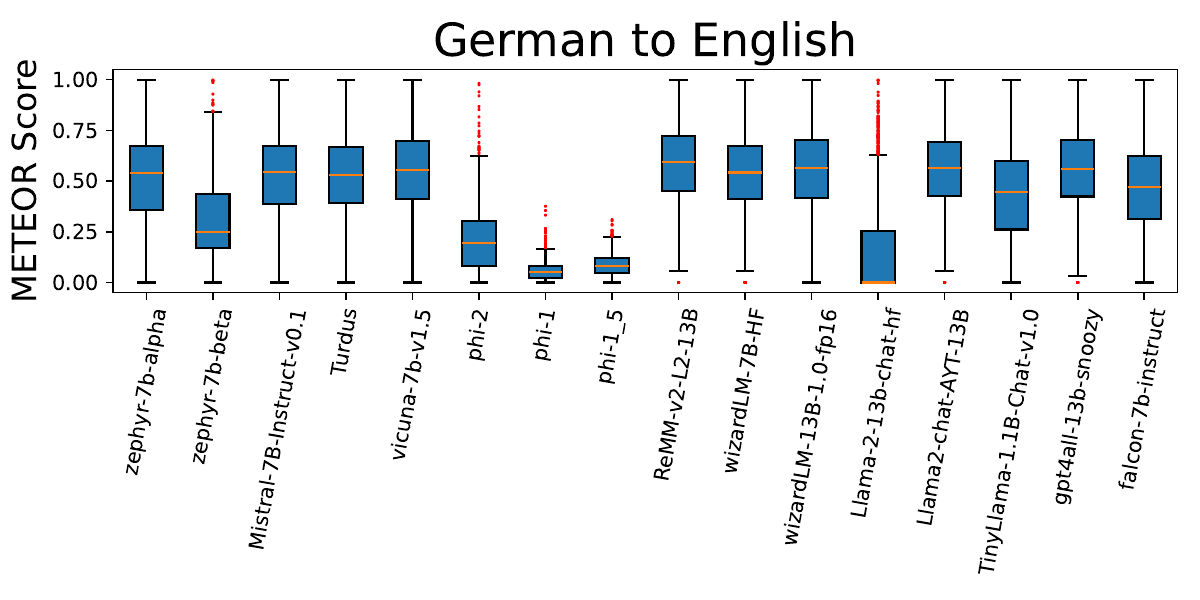}
    \caption{German-to-English dataset per-sentence translation quality and timing statistics  }
    \label{fig:German_translate_stats}
\end{figure}

\begin{figure}[th!]
    \centering
    \includegraphics[width=0.47\textwidth]{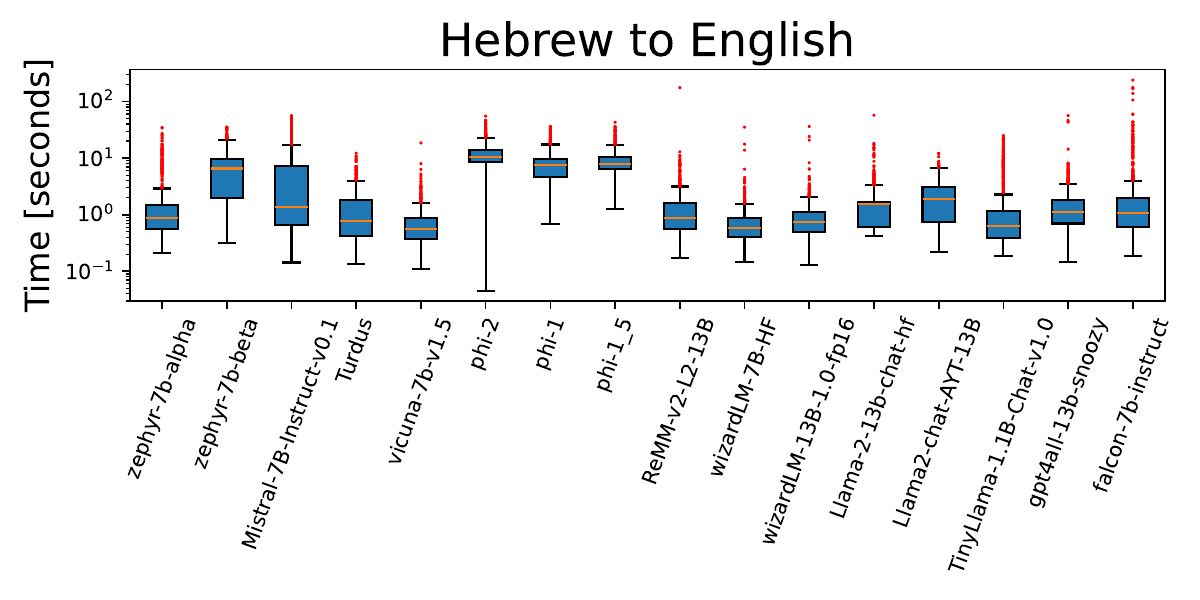}\\
    \includegraphics[width=0.47\textwidth]{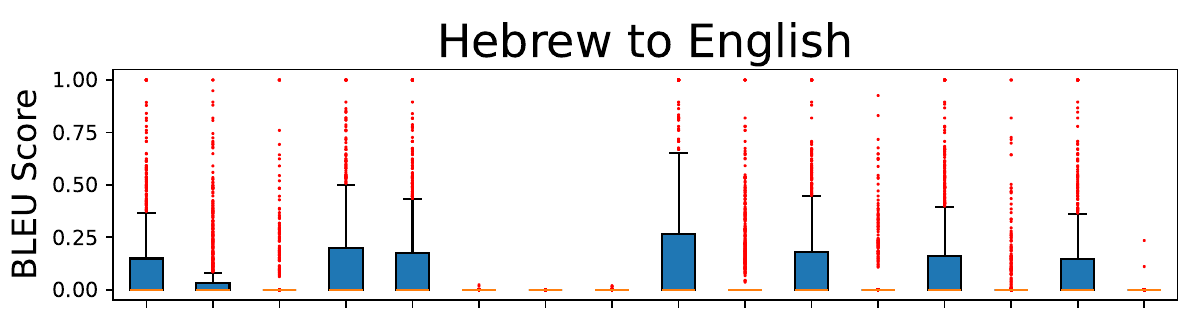}
    \includegraphics[width=0.47\textwidth]{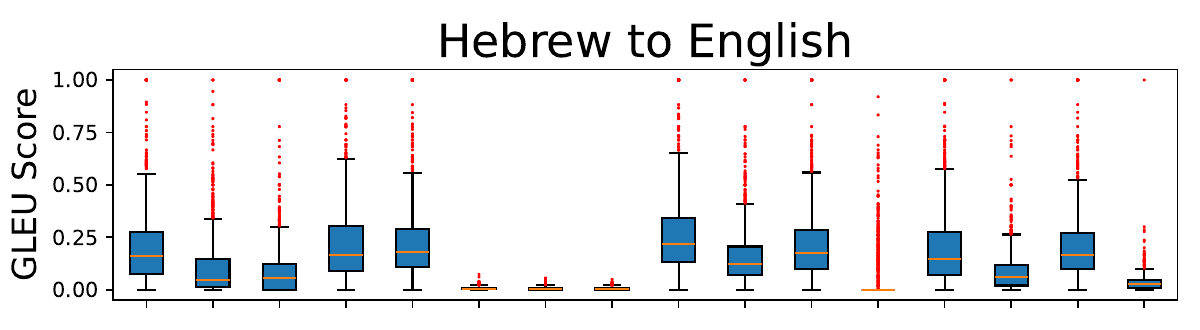}
    \includegraphics[width=0.47\textwidth]{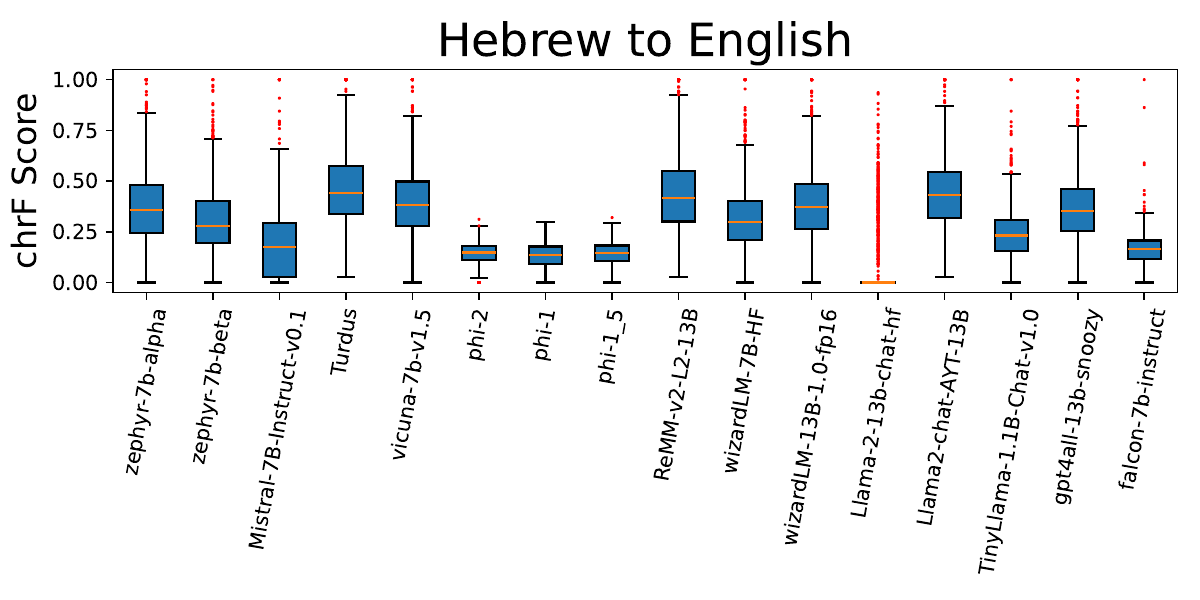}
    \includegraphics[width=0.47\textwidth]{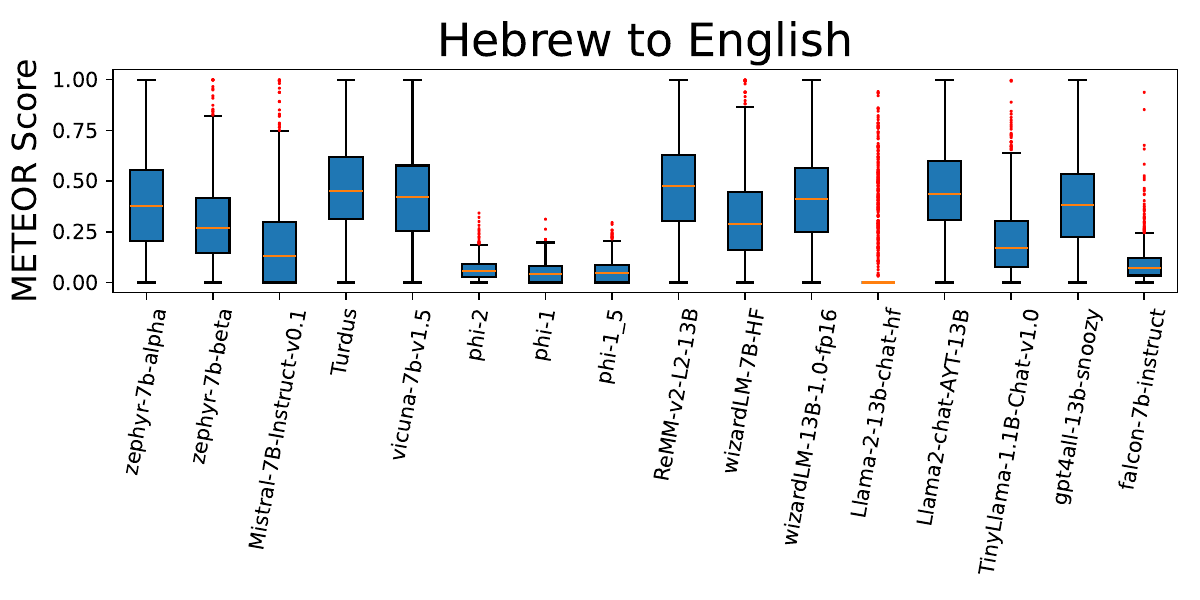}
    \caption{Hebrew-to-English dataset per-sentence translation quality and timing statistics  }
    \label{fig:Hebrew_translate_stats}
\end{figure}

\begin{figure}[th!]
    \centering
    \includegraphics[width=0.47\textwidth]{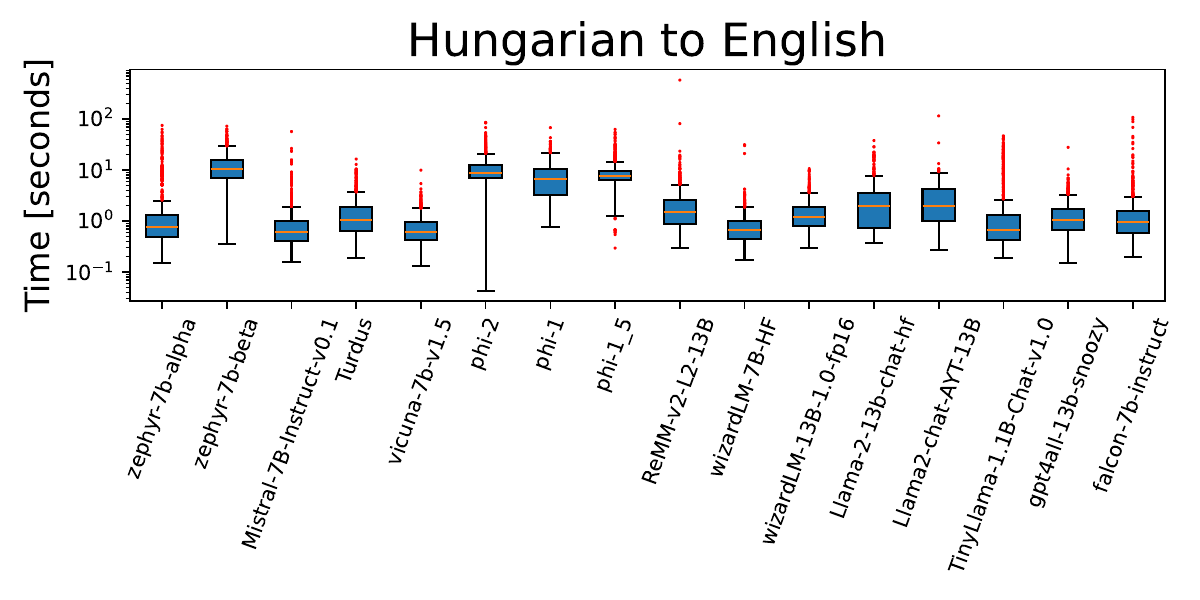}\\
    \includegraphics[width=0.47\textwidth]{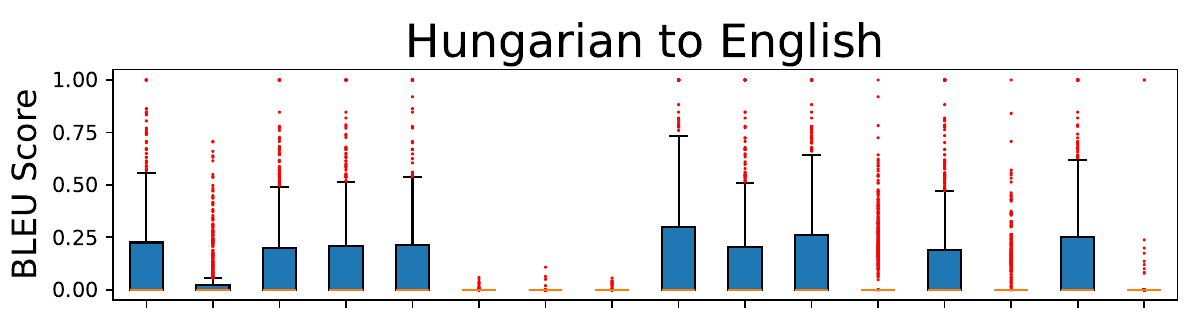}
    \includegraphics[width=0.47\textwidth]{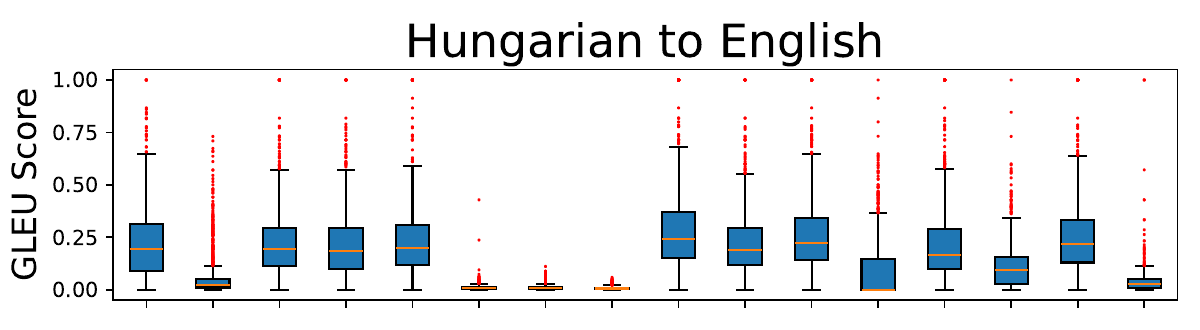}
    \includegraphics[width=0.47\textwidth]{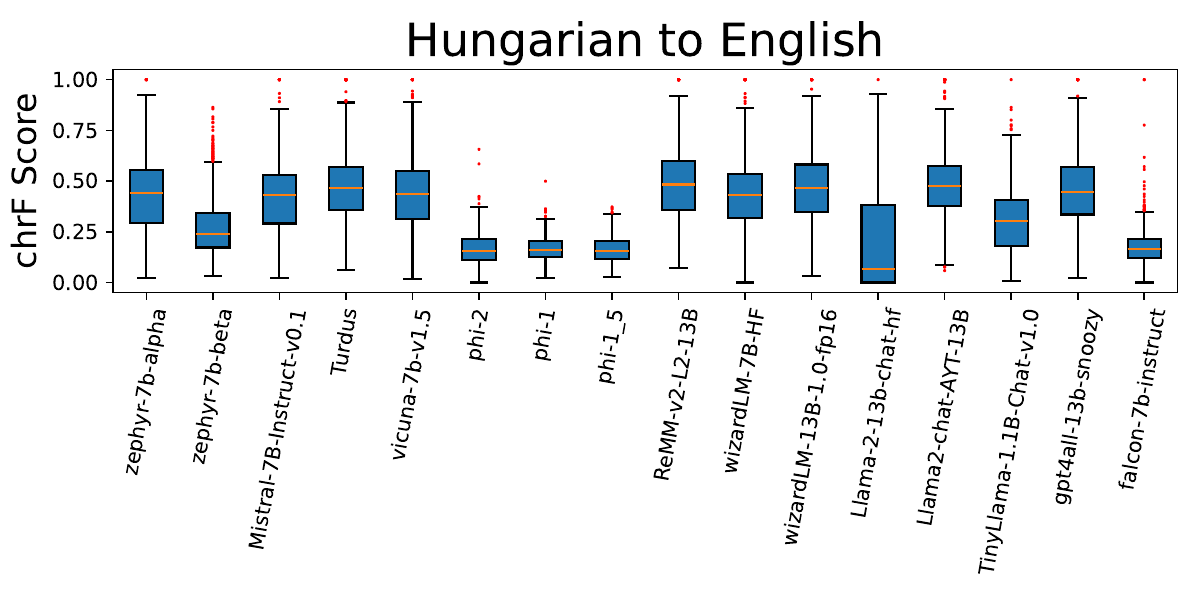}
    \includegraphics[width=0.47\textwidth]{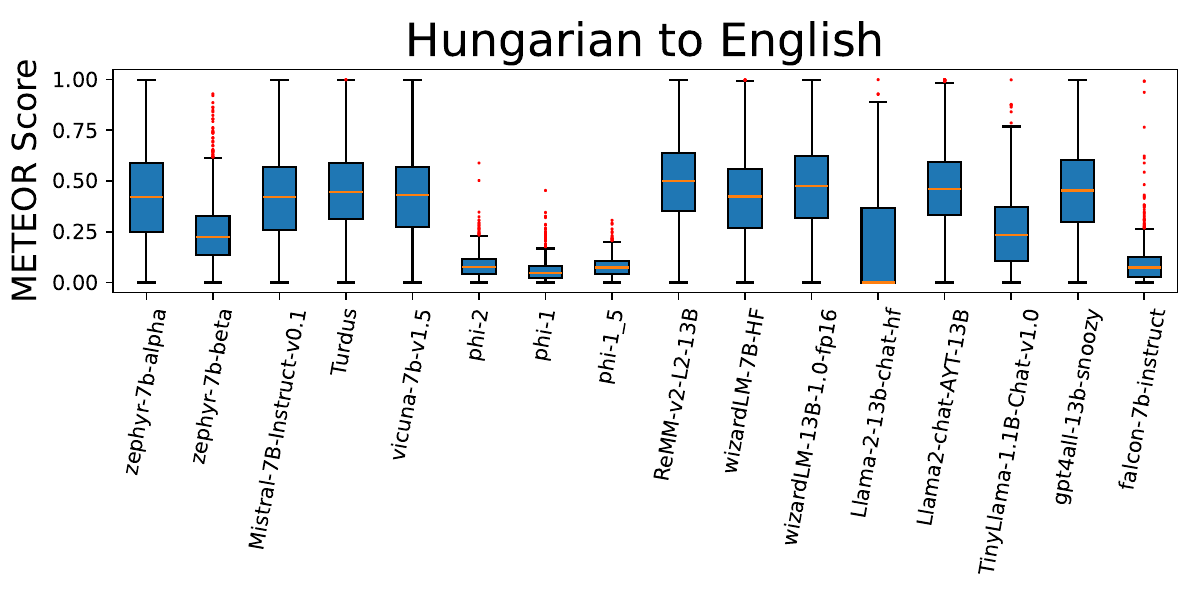}
    \caption{Hungarian-to-English dataset per-sentence translation quality and timing statistics  }
    \label{fig:Hungarian_translate_stats}
\end{figure}

\begin{figure}[th!]
    \centering
    \includegraphics[width=0.47\textwidth]{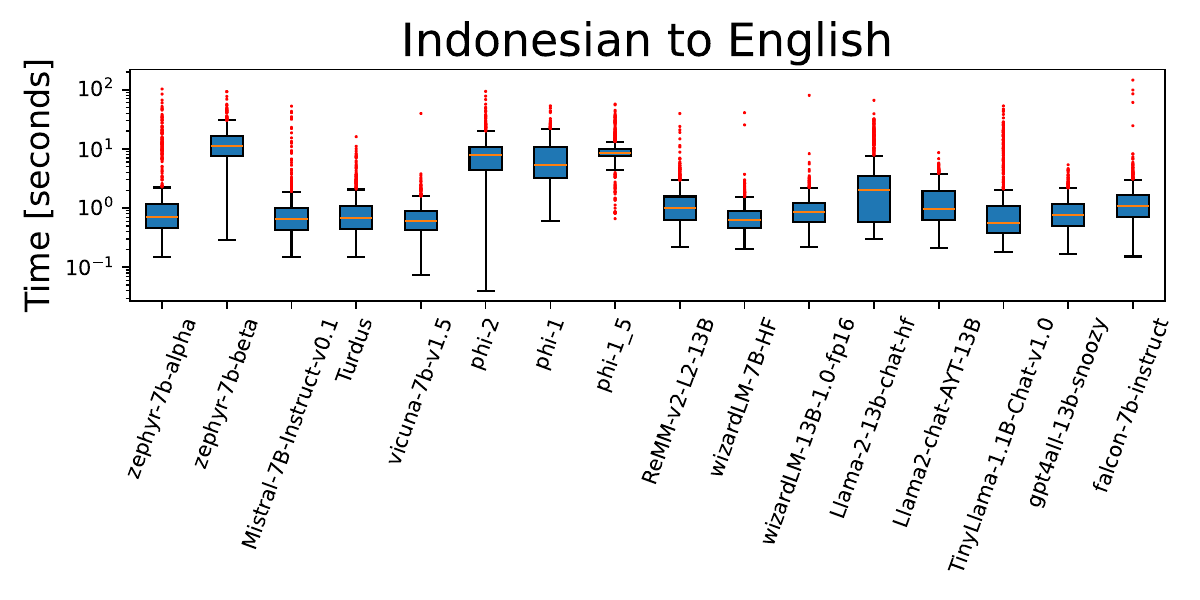}\\
    \includegraphics[width=0.47\textwidth]{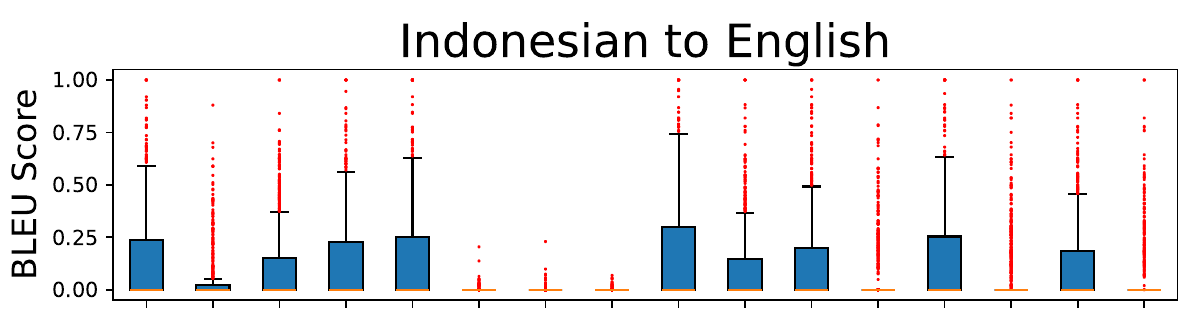}
    \includegraphics[width=0.47\textwidth]{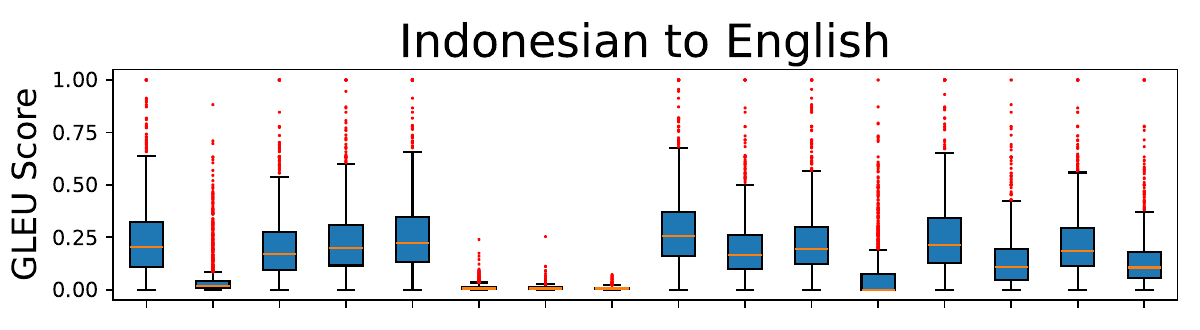}
    \includegraphics[width=0.47\textwidth]{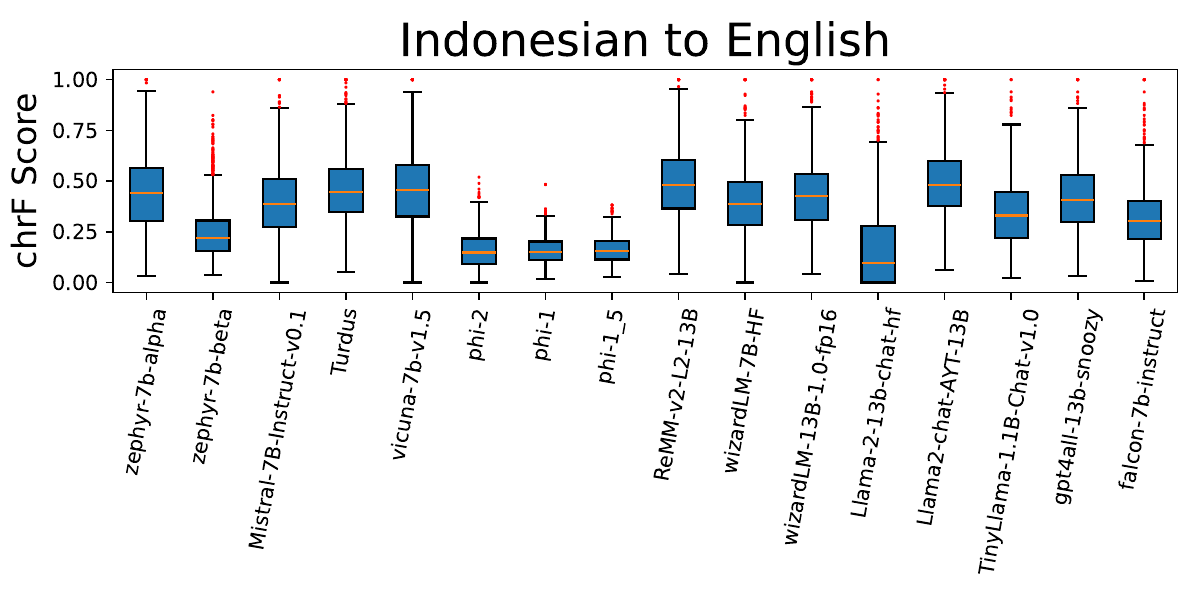}
    \includegraphics[width=0.47\textwidth]{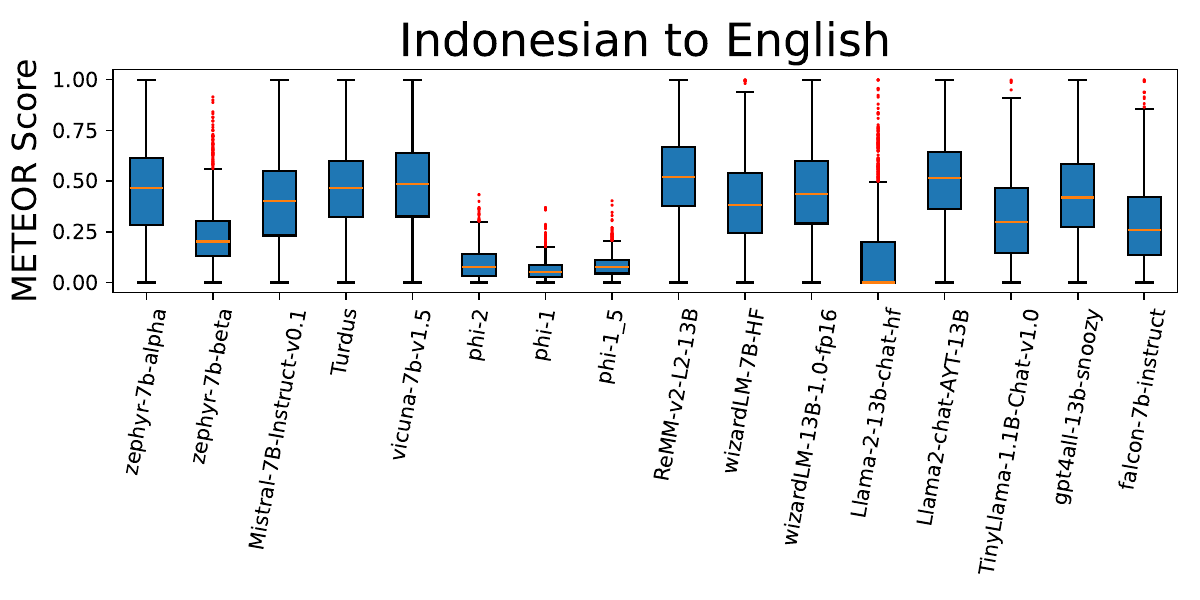}
    \caption{Indonesian-to-English dataset per-sentence translation quality and timing statistics  }
    \label{fig:Indonesian_translate_stats}
\end{figure}

\begin{figure}[h!]
    \centering
    \includegraphics[width=0.47\textwidth]{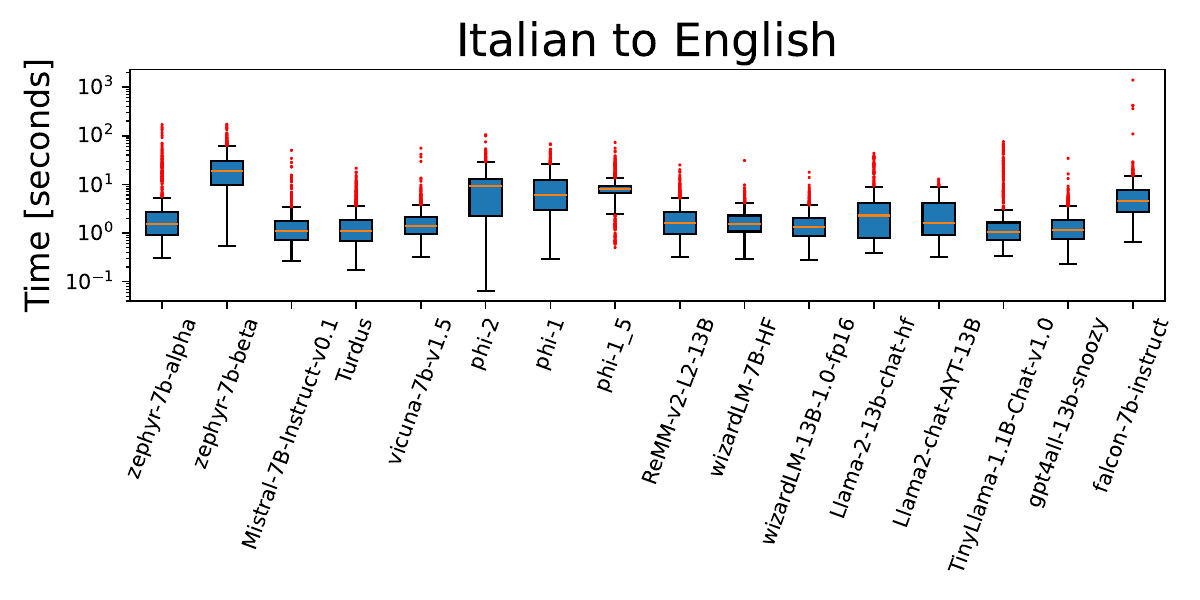}\\
    \includegraphics[width=0.47\textwidth]{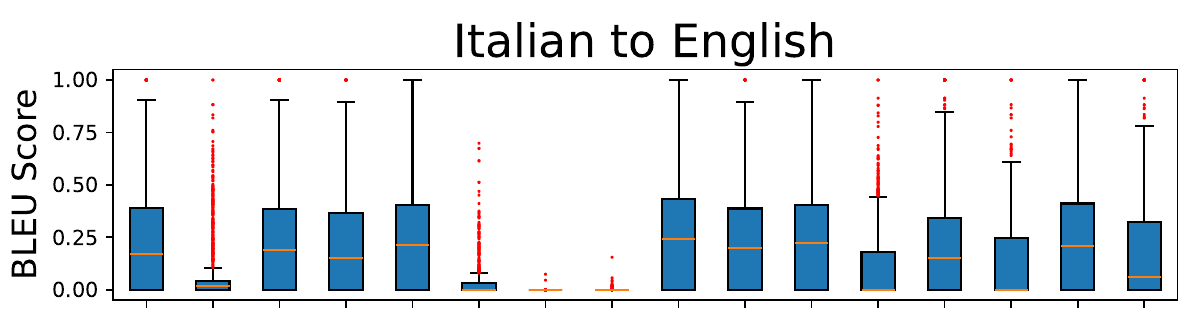}
    \includegraphics[width=0.47\textwidth]{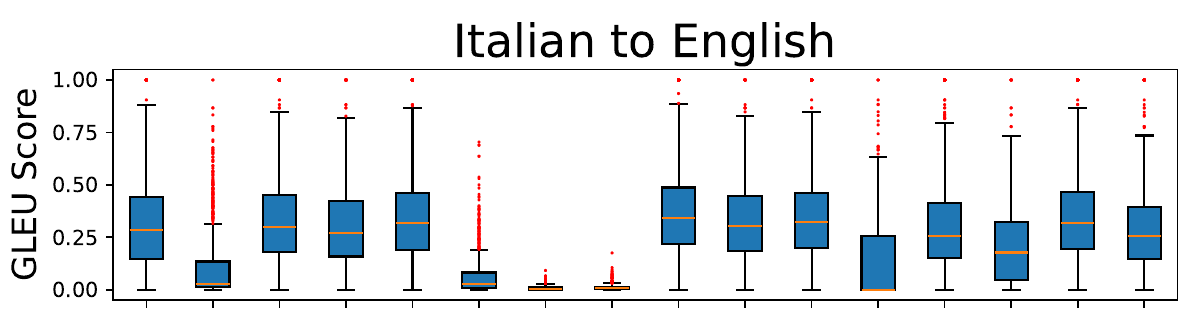}
    \includegraphics[width=0.47\textwidth]{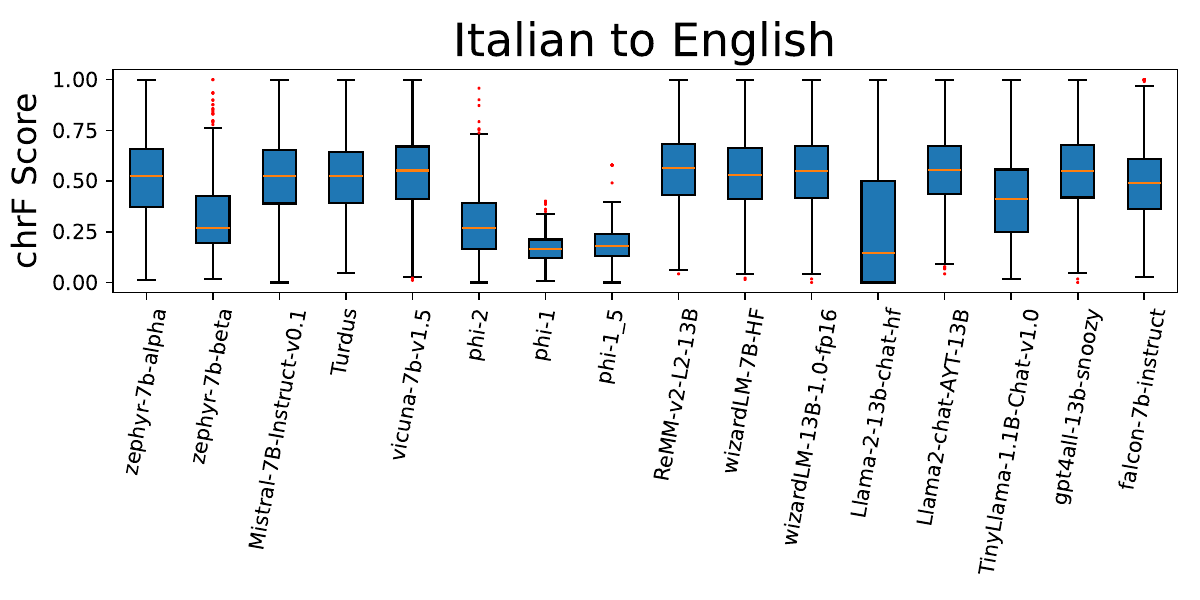}
    \includegraphics[width=0.47\textwidth]{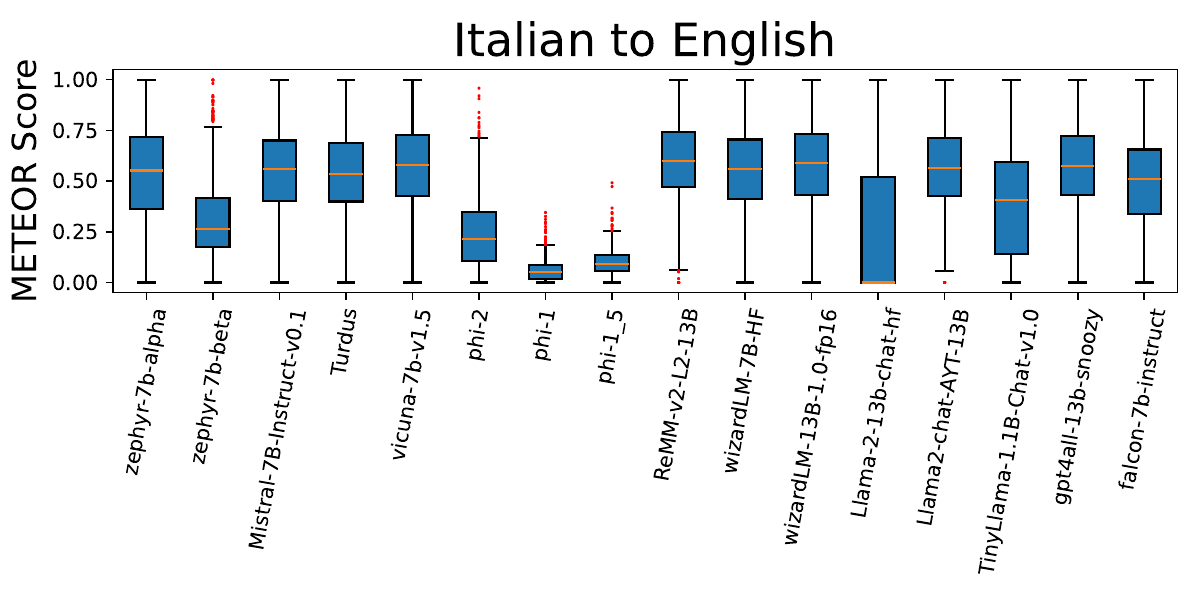}
    \caption{Italian-to-English dataset per-sentence translation quality and timing statistics }
    \label{fig:Italian_translate_stats}
\end{figure}

\begin{figure}[th!]
    \centering
    \includegraphics[width=0.47\textwidth]{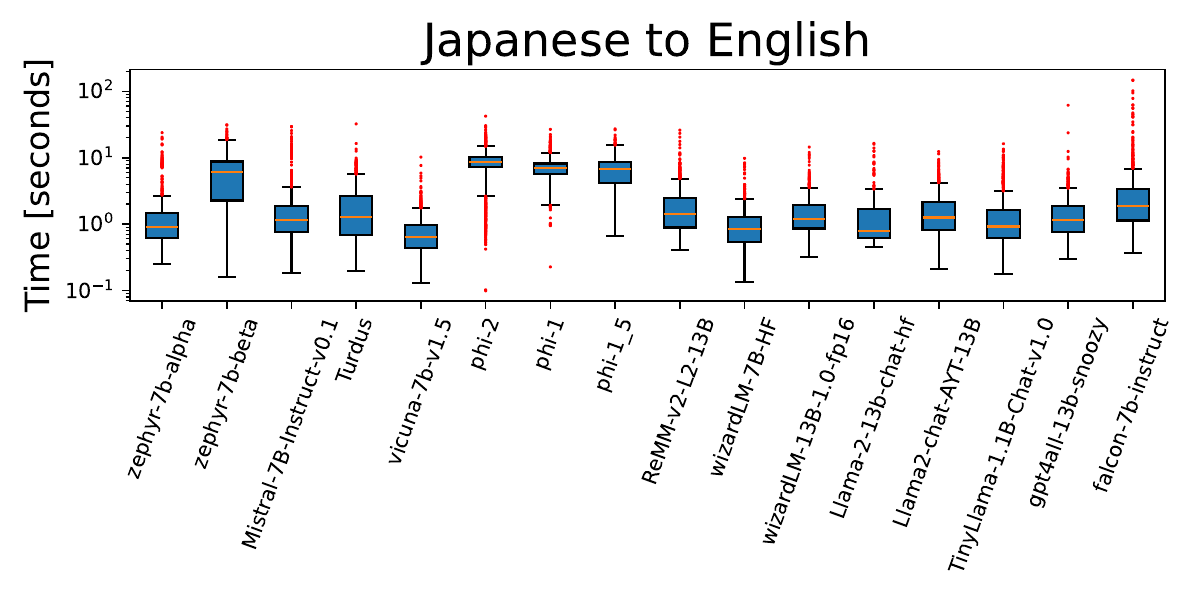}\\
    \includegraphics[width=0.47\textwidth]{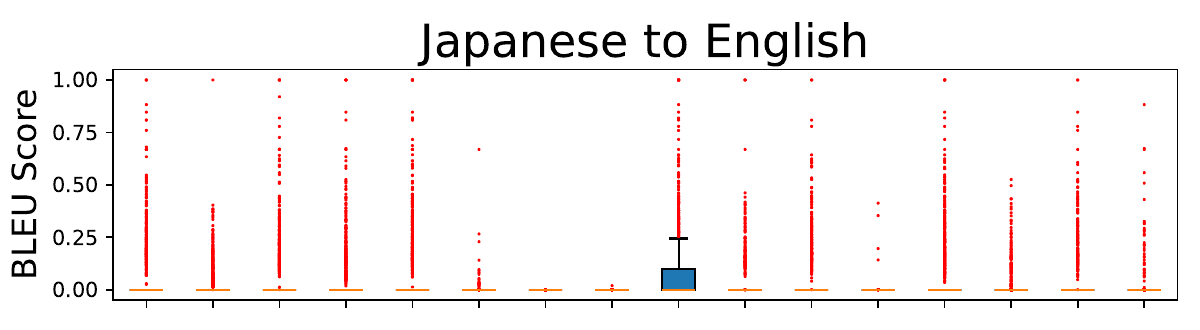}
    \includegraphics[width=0.47\textwidth]{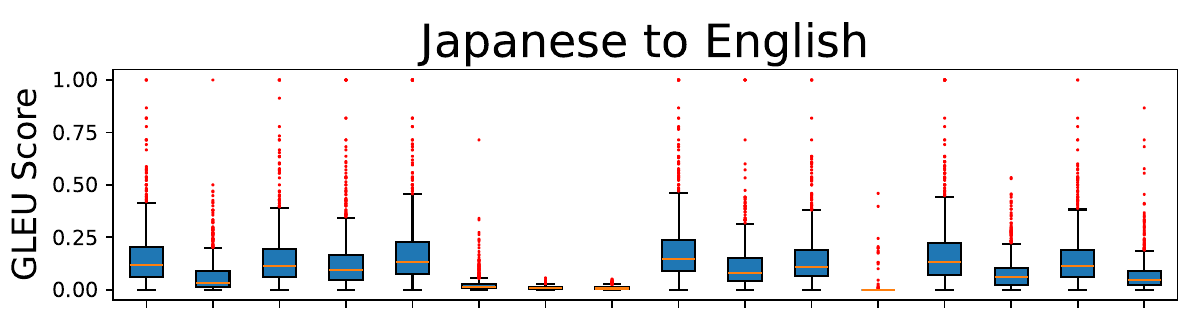}
    \includegraphics[width=0.47\textwidth]{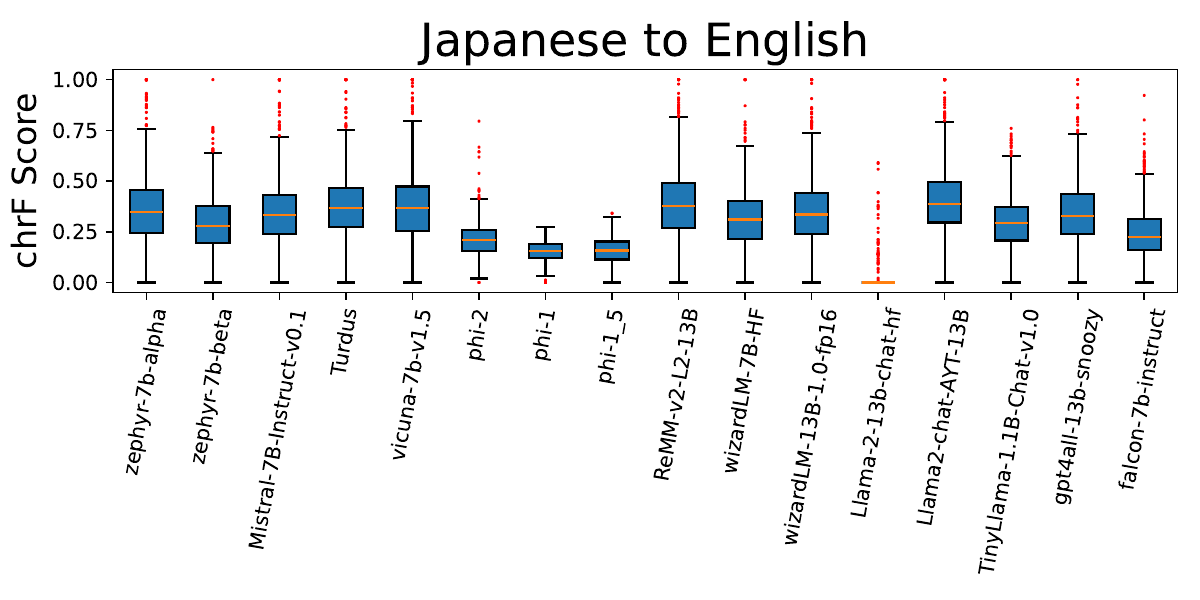}
    \includegraphics[width=0.47\textwidth]{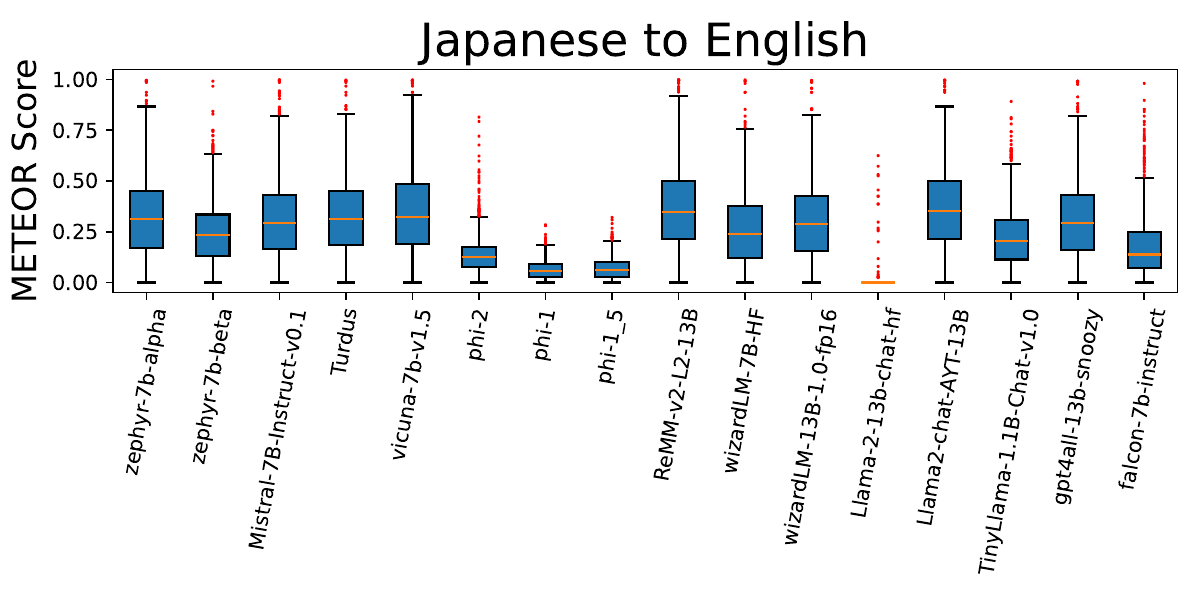}
    \caption{Japanese-to-English dataset per-sentence translation quality and timing statistics  }
    \label{fig:Japanese_translate_stats}
\end{figure}

\begin{figure}[th!]
    \centering
    \includegraphics[width=0.47\textwidth]{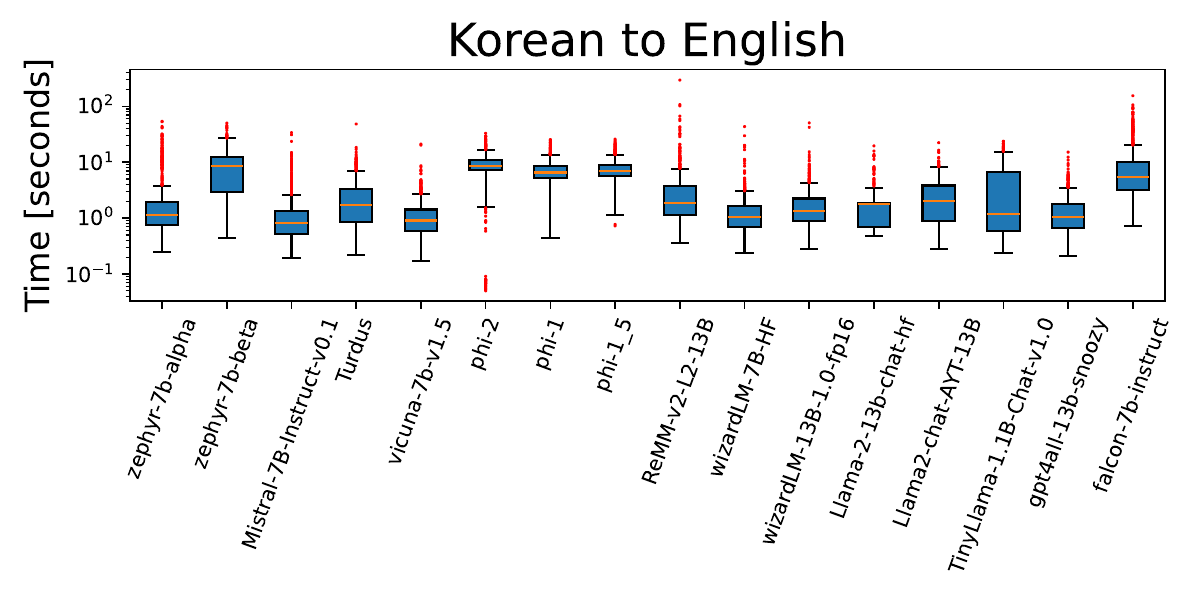}\\
    \includegraphics[width=0.47\textwidth]{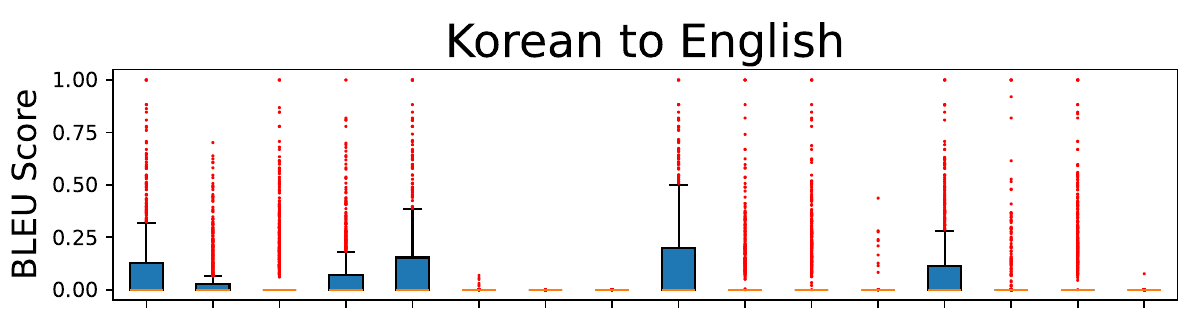}
    \includegraphics[width=0.47\textwidth]{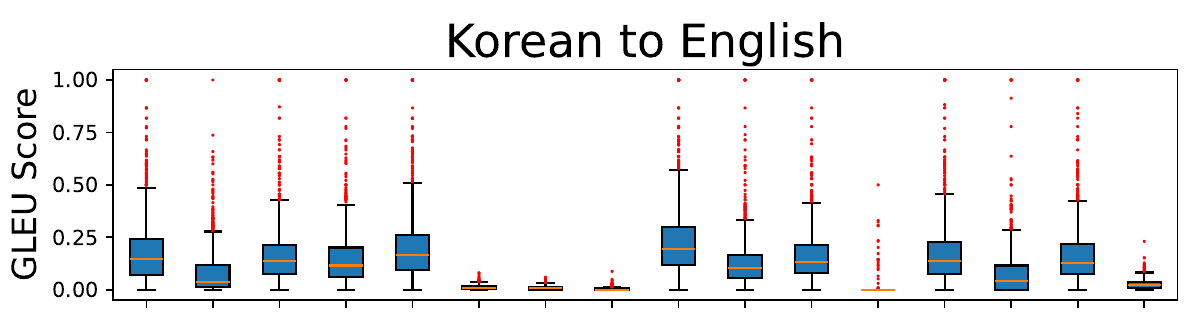}
    \includegraphics[width=0.47\textwidth]{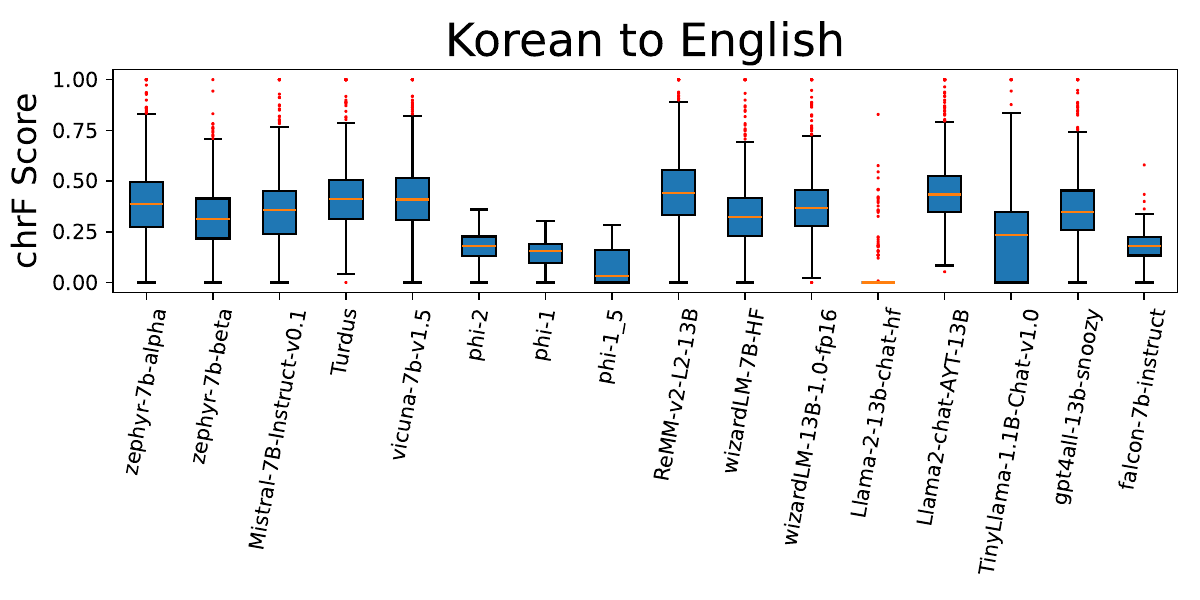}
    \includegraphics[width=0.47\textwidth]{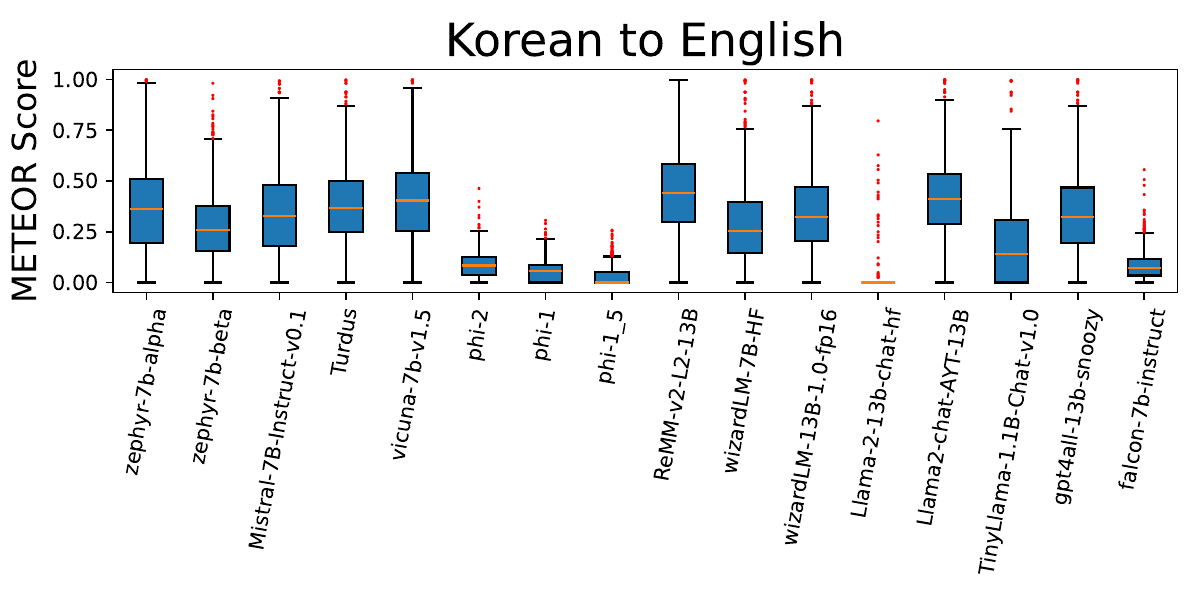}
    \caption{Korean-to-English dataset per-sentence translation quality and timing statistics  }
    \label{fig:Korean_translate_stats}
\end{figure}

\begin{figure}[th!]
    \centering
    \includegraphics[width=0.47\textwidth]{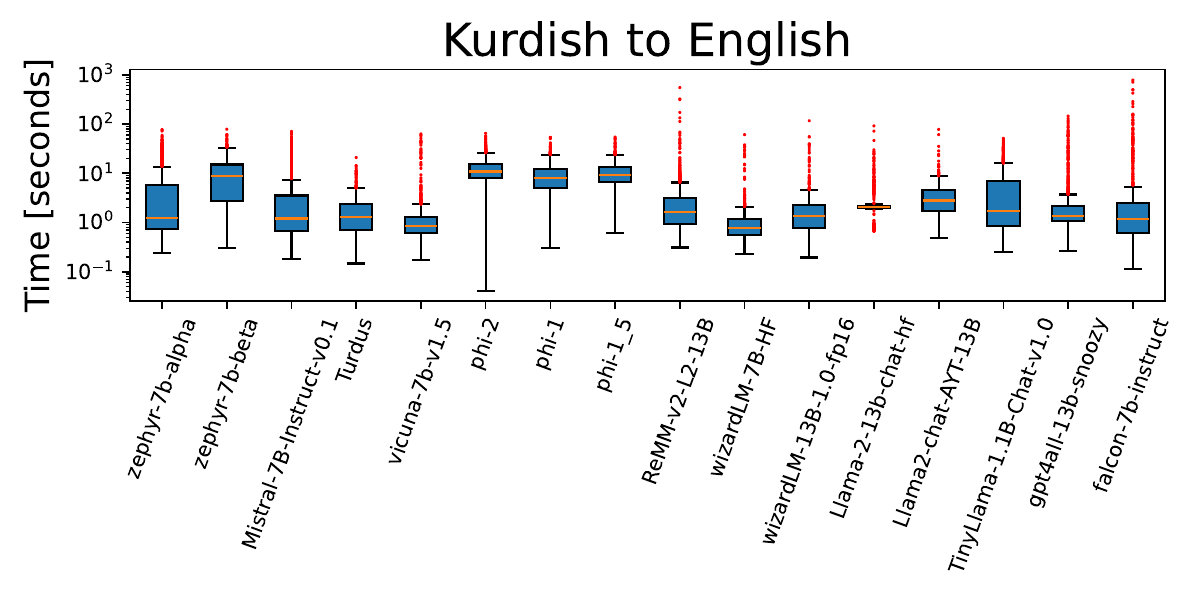}\\
    \includegraphics[width=0.47\textwidth]{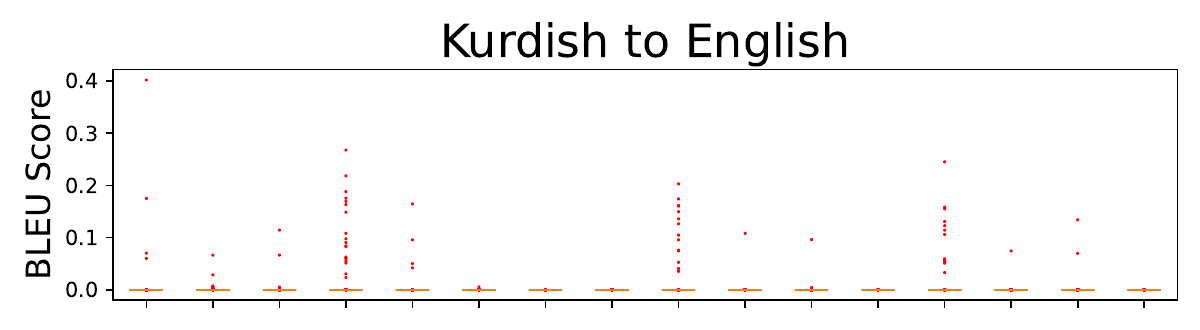}
    \includegraphics[width=0.47\textwidth]{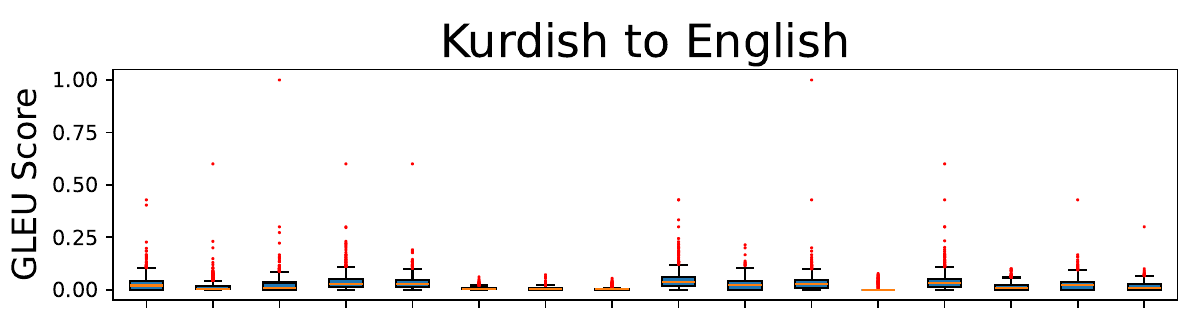}
    \includegraphics[width=0.47\textwidth]{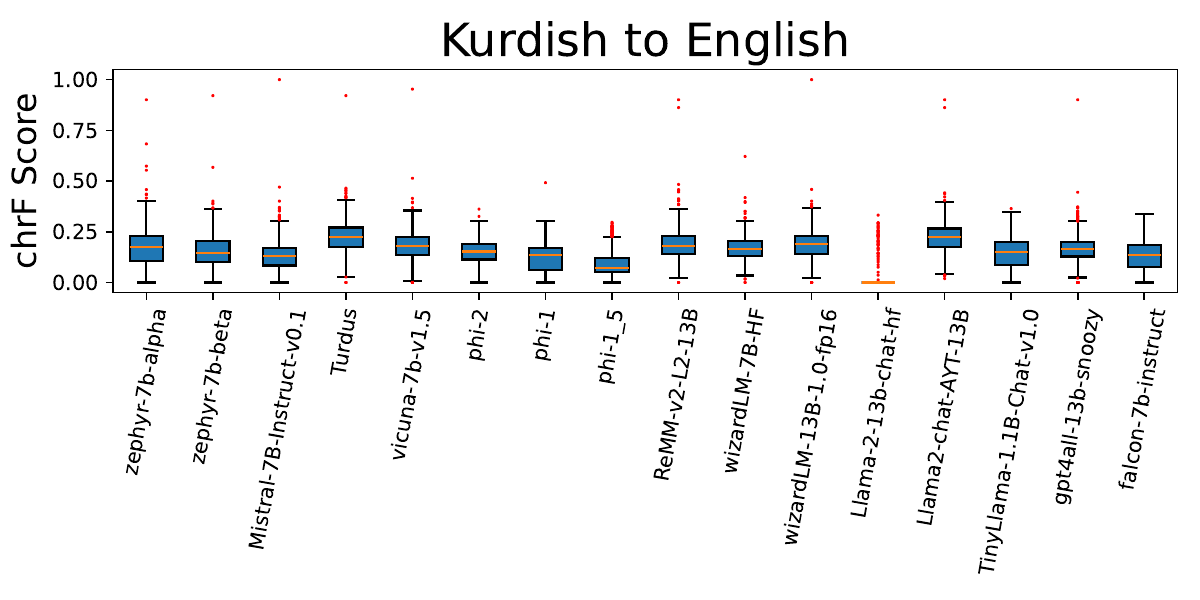}
    \includegraphics[width=0.47\textwidth]{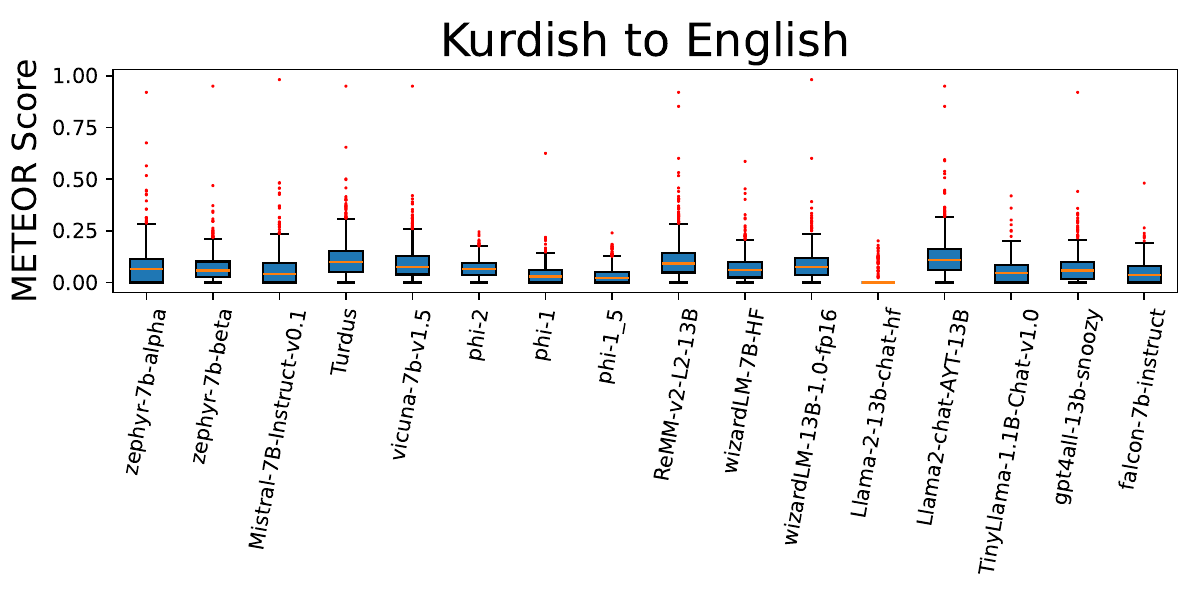}
    \caption{Kurdish-to-English dataset per-sentence translation quality and timing statistics  }
    \label{fig:Kurdish_translate_stats}
\end{figure}

\begin{figure}[th!]
    \centering
    \includegraphics[width=0.47\textwidth]{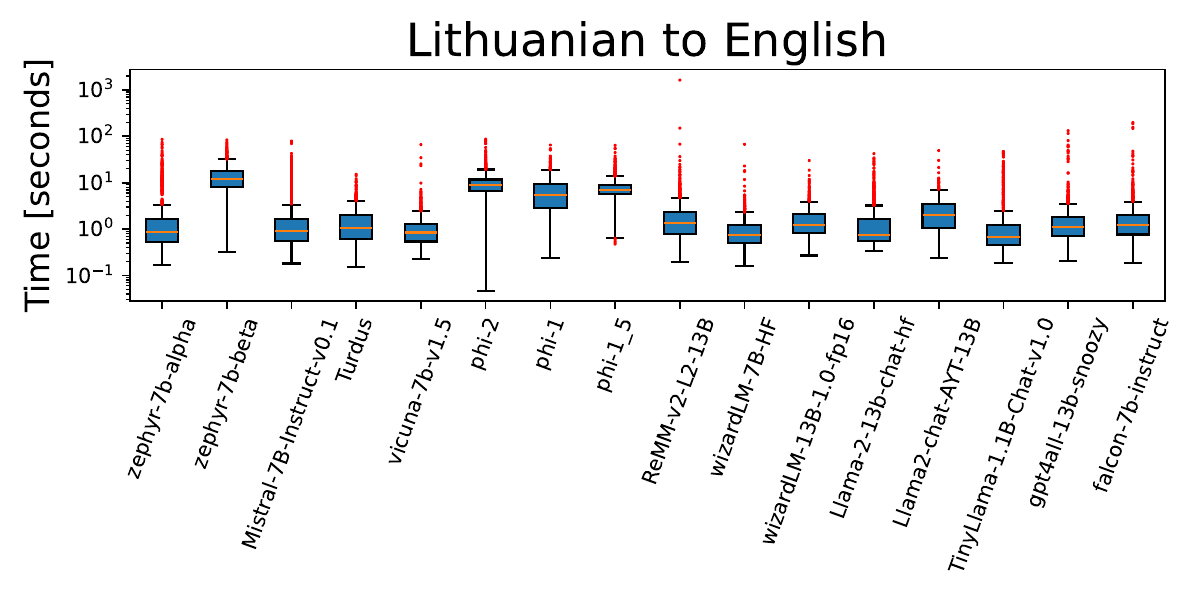}\\
    \includegraphics[width=0.47\textwidth]{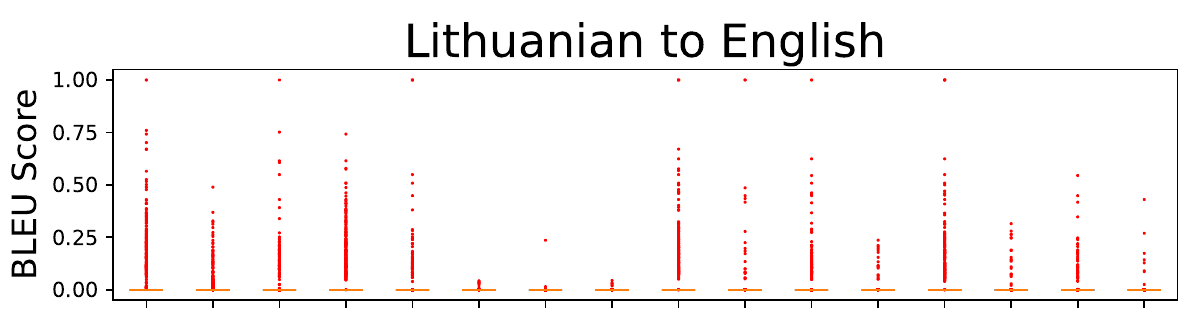}
    \includegraphics[width=0.47\textwidth]{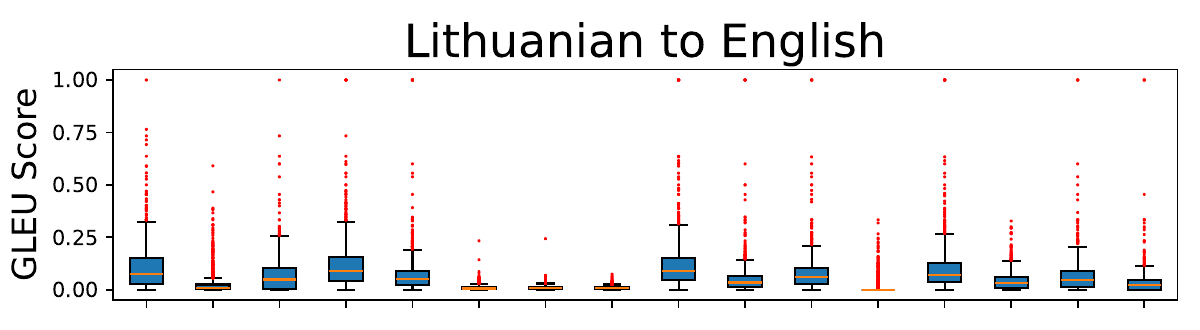}
    \includegraphics[width=0.47\textwidth]{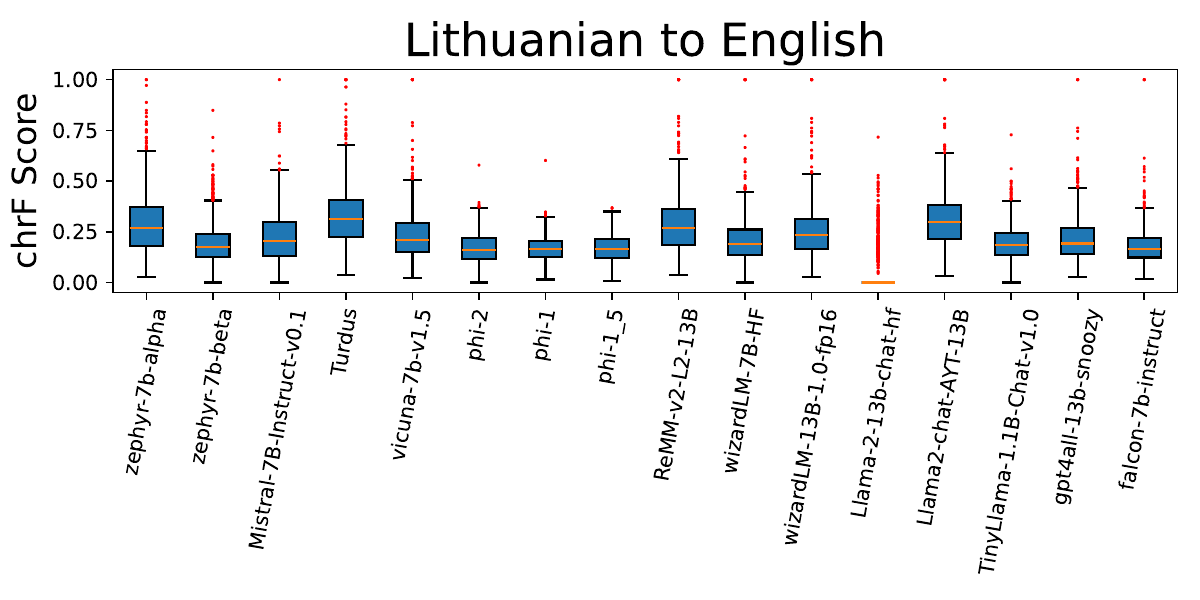}
    \includegraphics[width=0.47\textwidth]{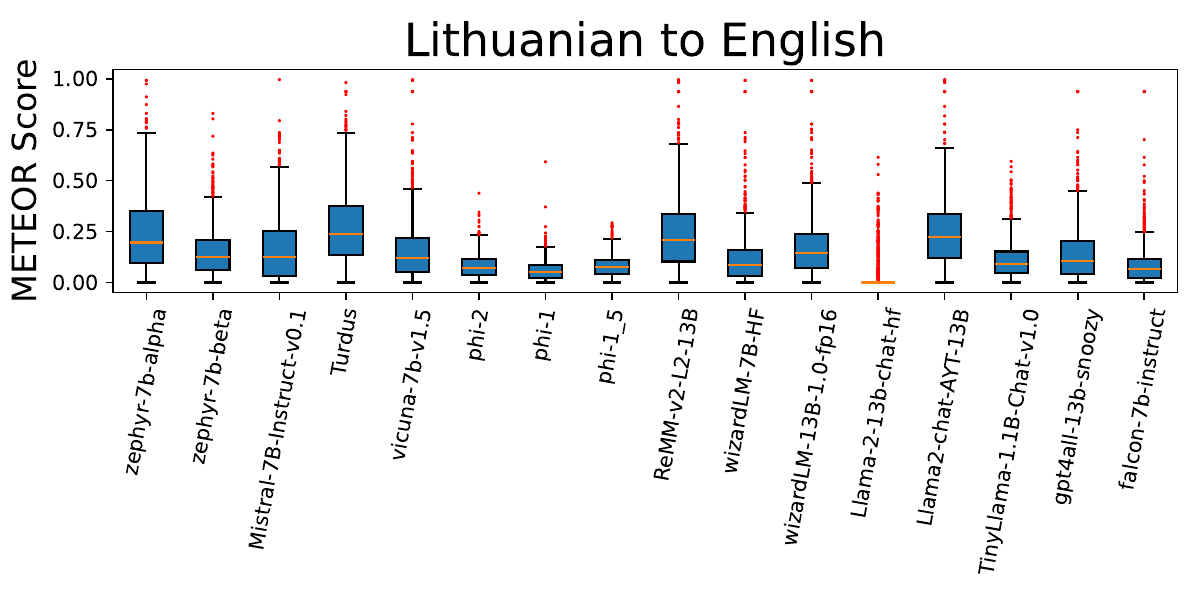}
    \caption{Lithuanian-to-English dataset per-sentence translation quality and timing statistics  }
    \label{fig:Lithuanian_translate_stats}
\end{figure}

\begin{figure}[th!]
    \centering
    \includegraphics[width=0.47\textwidth]{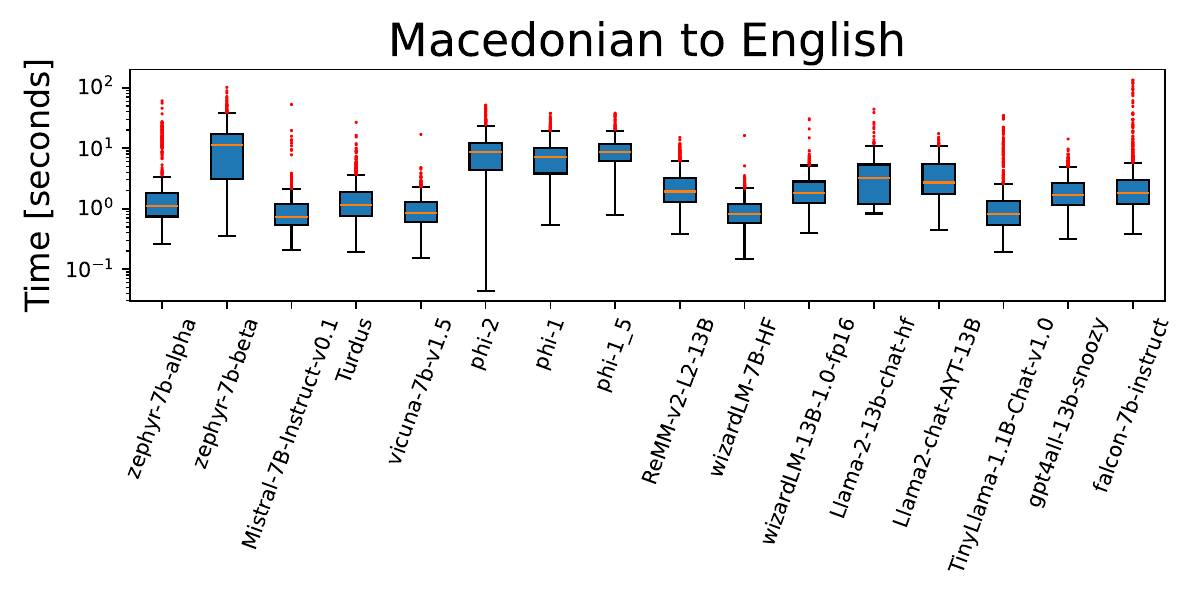}\\
    \includegraphics[width=0.47\textwidth]{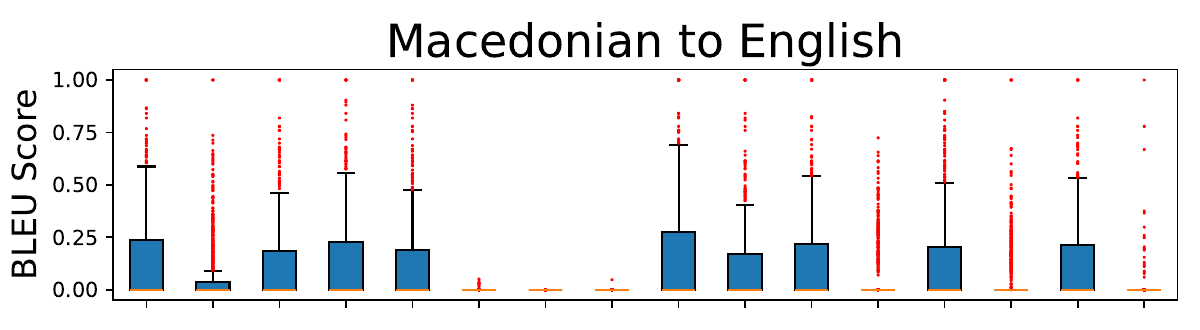}
    \includegraphics[width=0.47\textwidth]{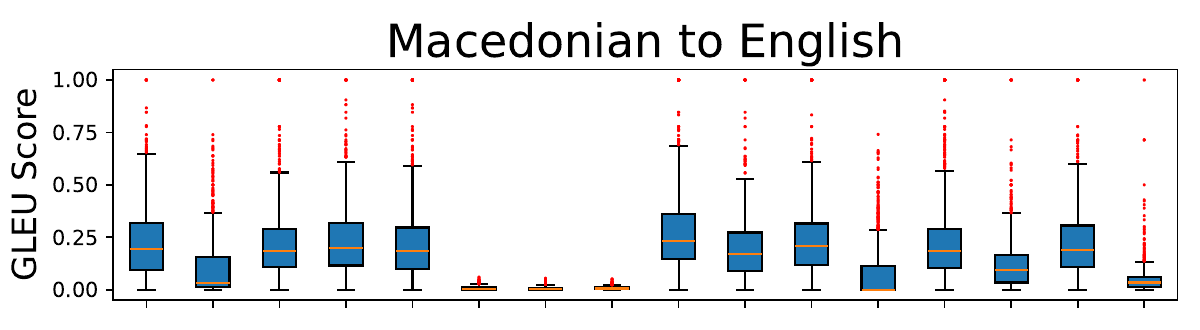}
    \includegraphics[width=0.47\textwidth]{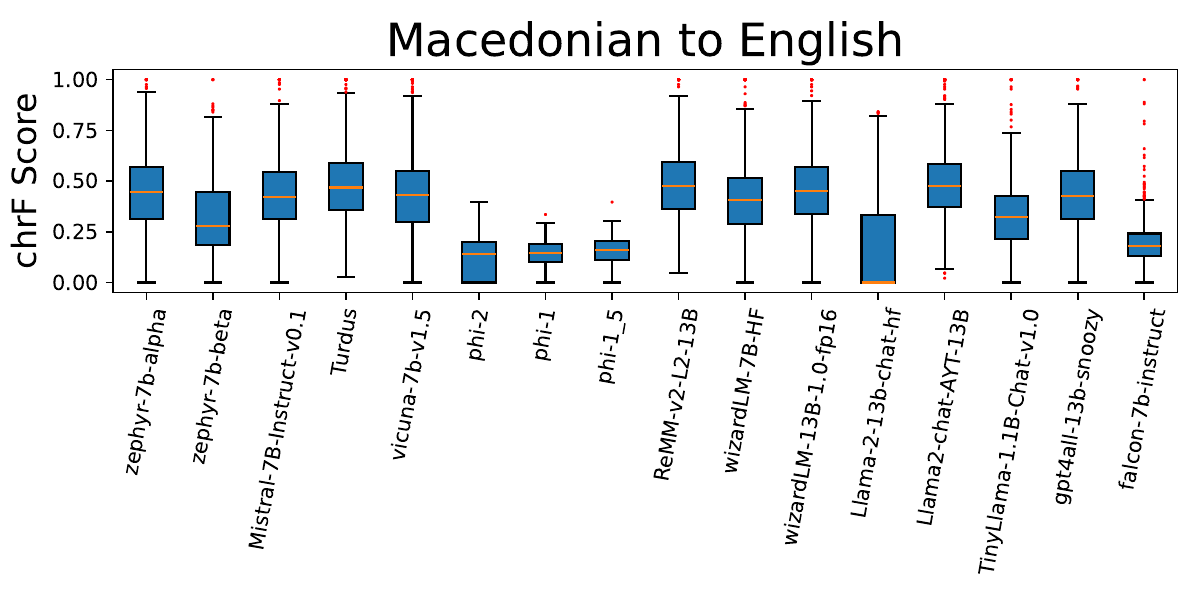}
    \includegraphics[width=0.47\textwidth]{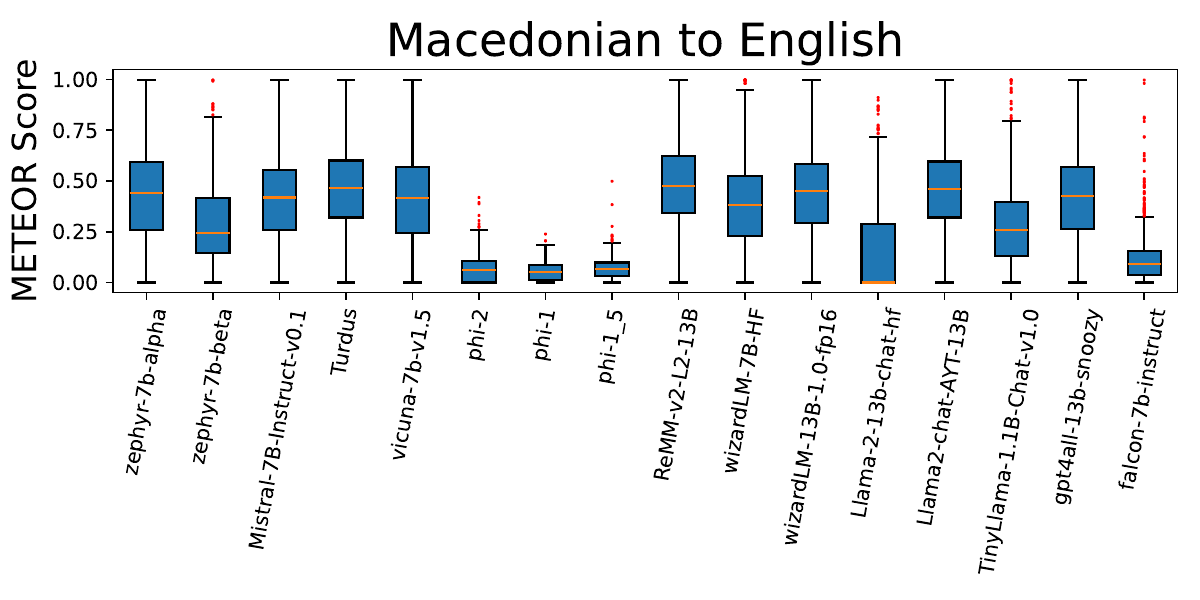}
    \caption{Macedonian-to-English dataset per-sentence translation quality and timing statistics  }
    \label{fig:Macedonian_translate_stats}
\end{figure}

\begin{figure}[th!]
    \centering
    \includegraphics[width=0.47\textwidth]{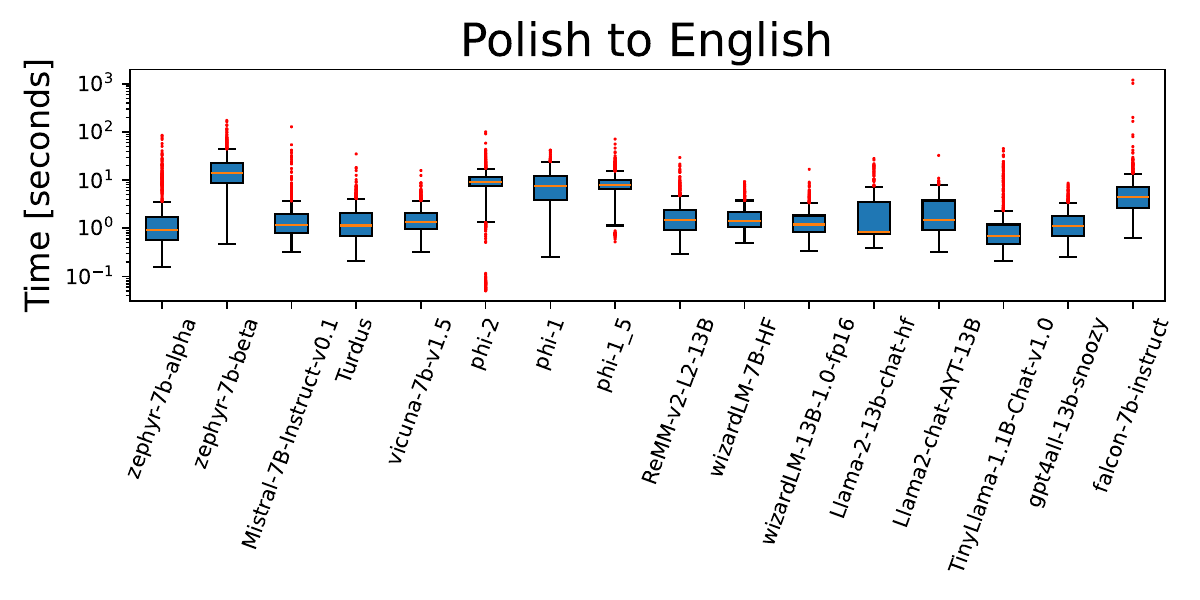}\\
    \includegraphics[width=0.47\textwidth]{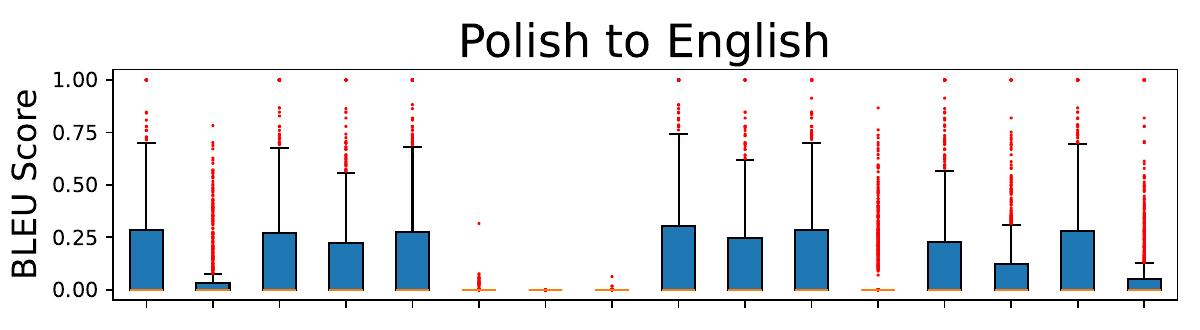}
    \includegraphics[width=0.47\textwidth]{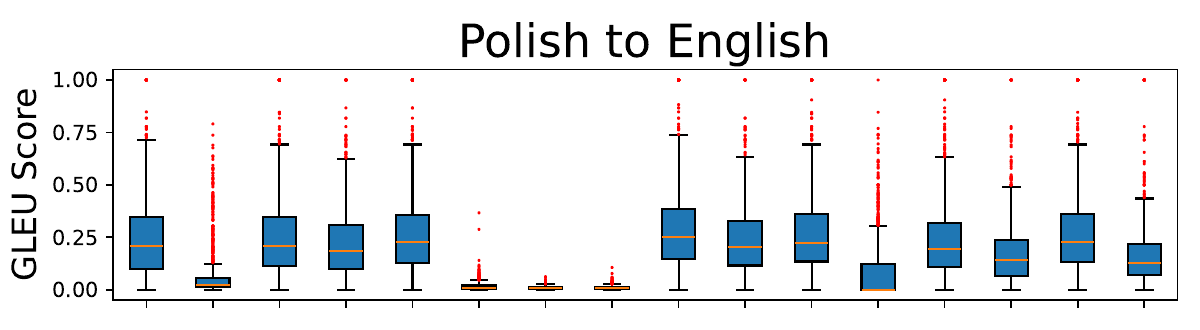}
    \includegraphics[width=0.47\textwidth]{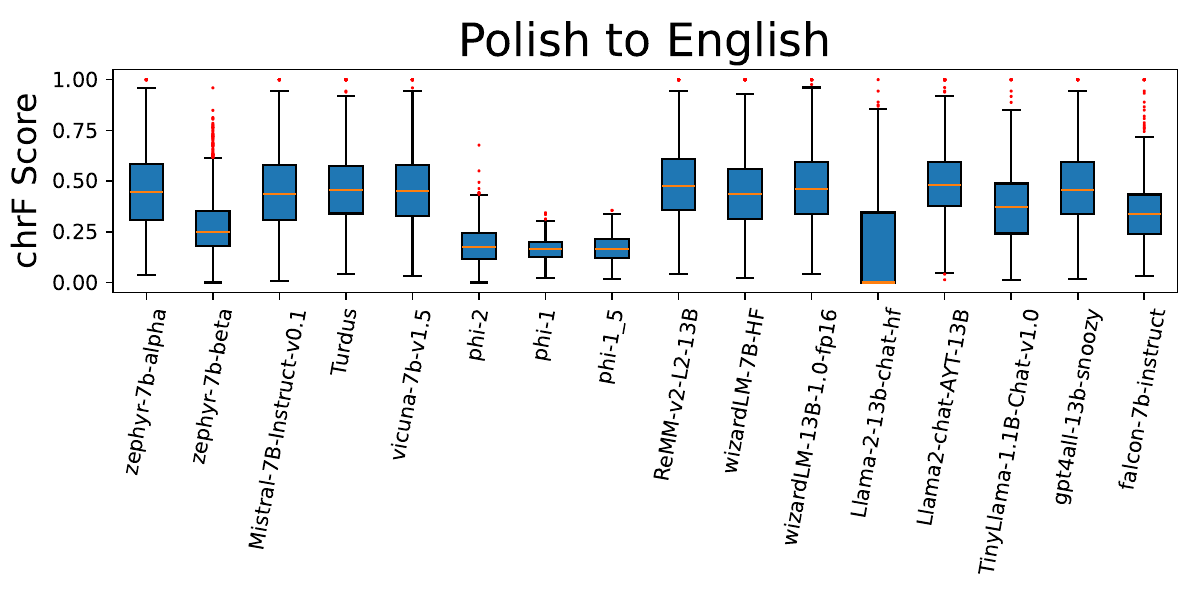}
    \includegraphics[width=0.47\textwidth]{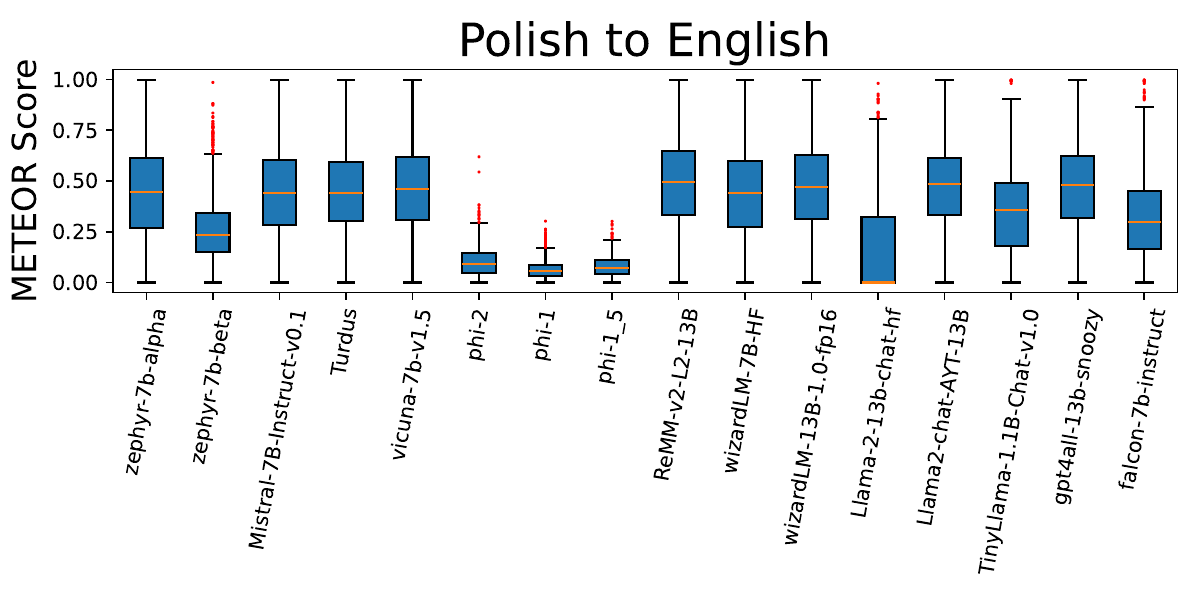}
    \caption{Polish-to-English dataset per-sentence translation quality and timing statistics  }
    \label{fig:Polish_translate_stats}
\end{figure}

\begin{figure}[th!]
    \centering
    \includegraphics[width=0.47\textwidth]{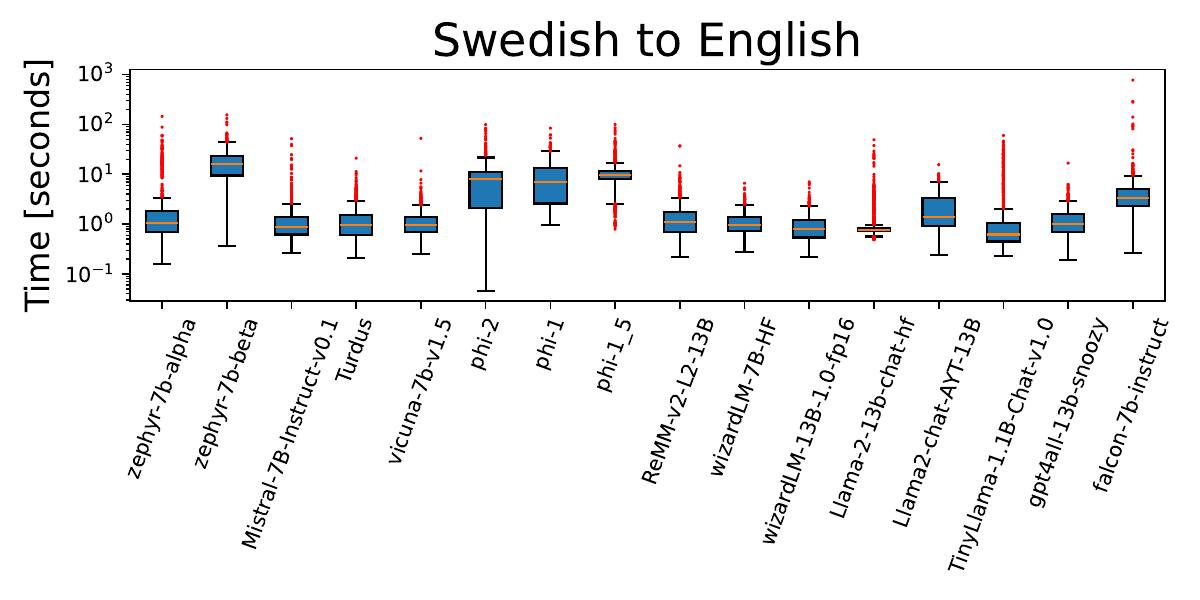}\\
    \includegraphics[width=0.47\textwidth]{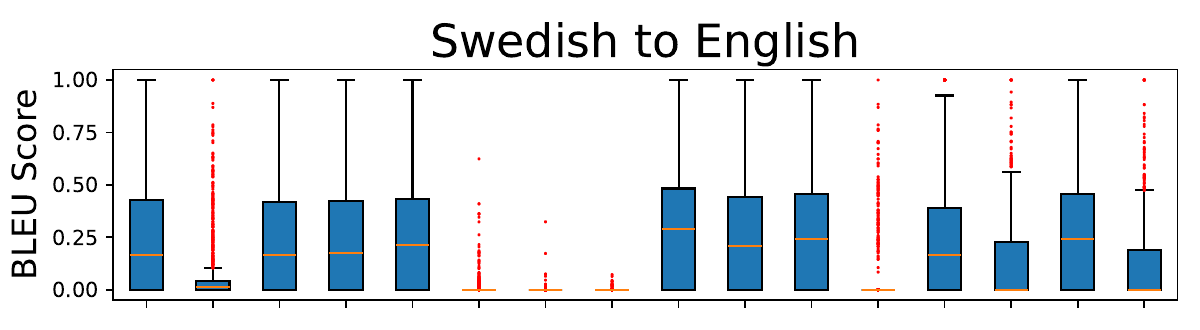}
    \includegraphics[width=0.47\textwidth]{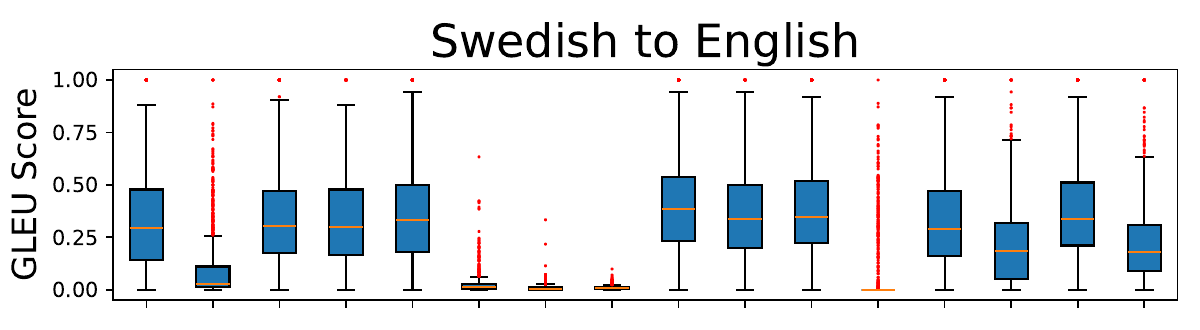}
    \includegraphics[width=0.47\textwidth]{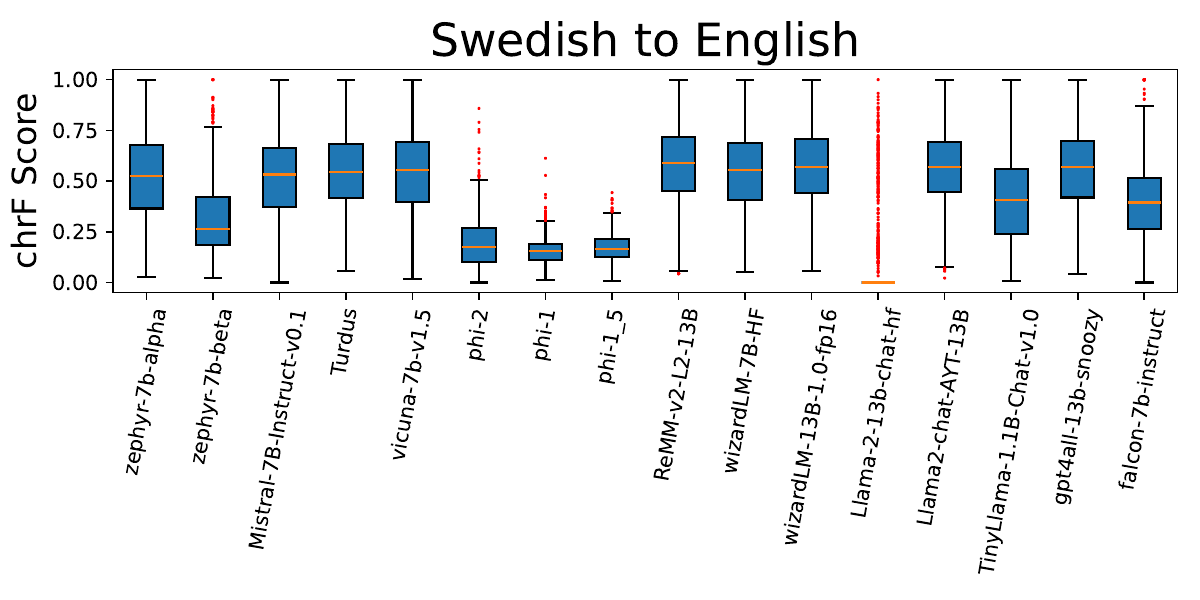}
    \includegraphics[width=0.47\textwidth]{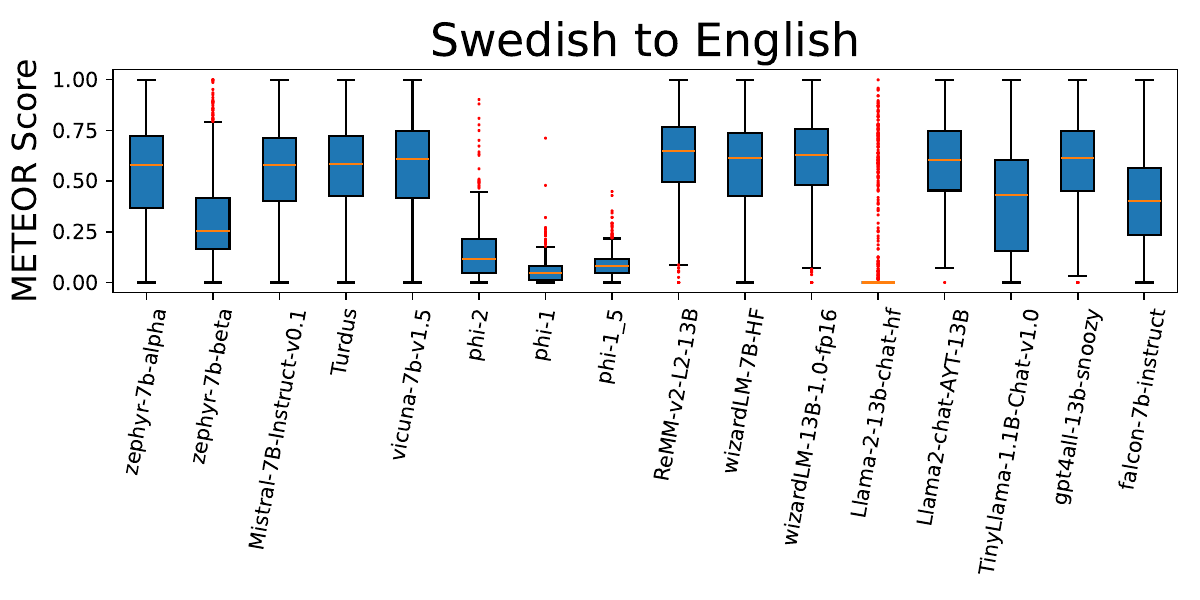}
    \caption{Swedish-to-English dataset per-sentence translation quality and timing statistics }
    \label{fig:Swedish_translate_stats}
\end{figure}

\begin{figure}[th!]
    \centering
    \includegraphics[width=0.47\textwidth]{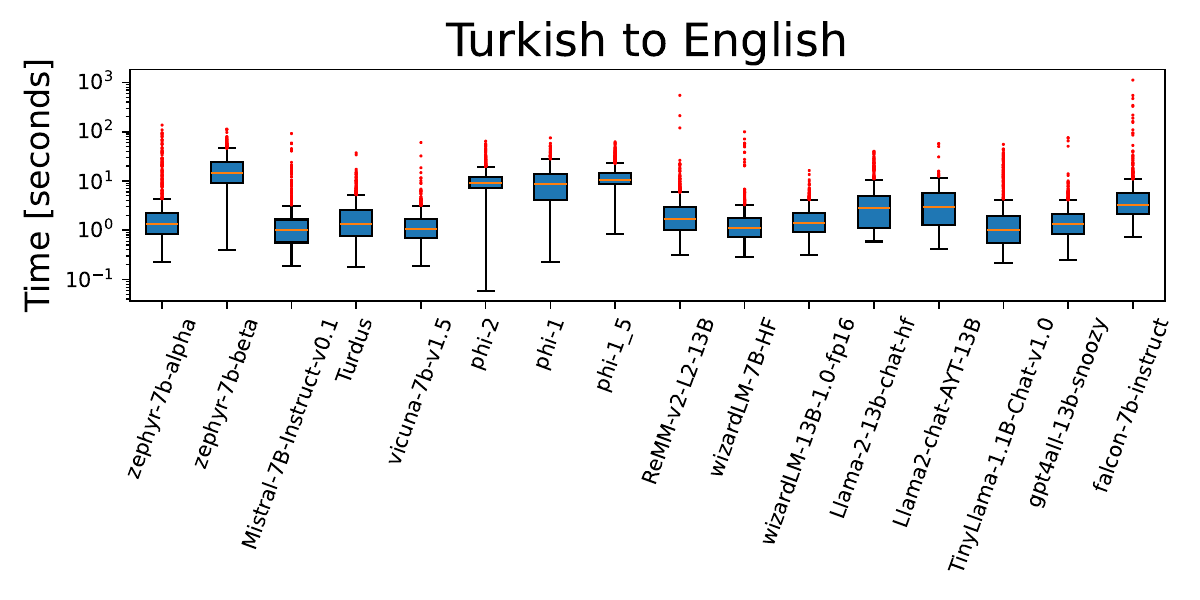}\\
    \includegraphics[width=0.47\textwidth]{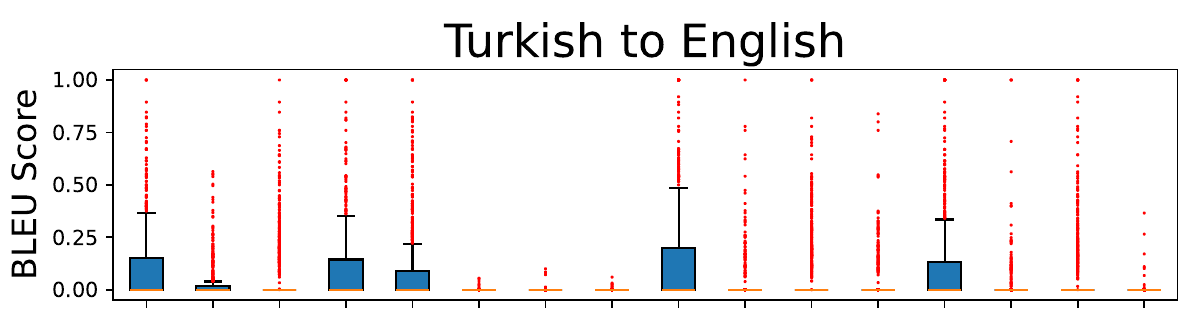}
    \includegraphics[width=0.47\textwidth]{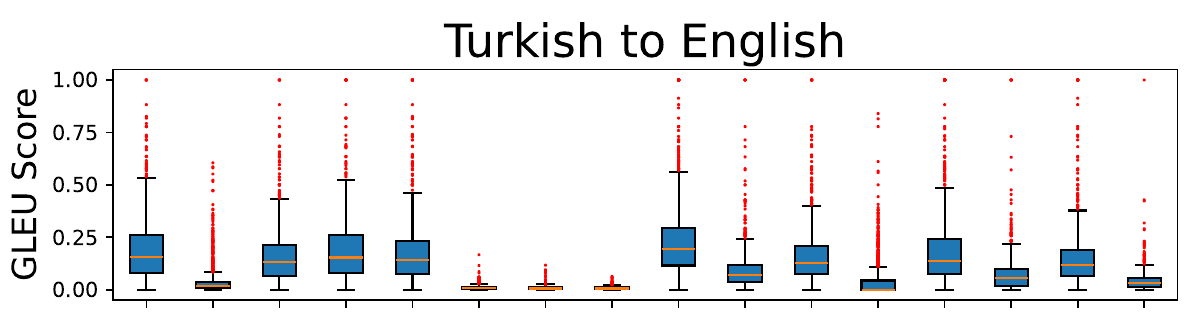}
    \includegraphics[width=0.47\textwidth]{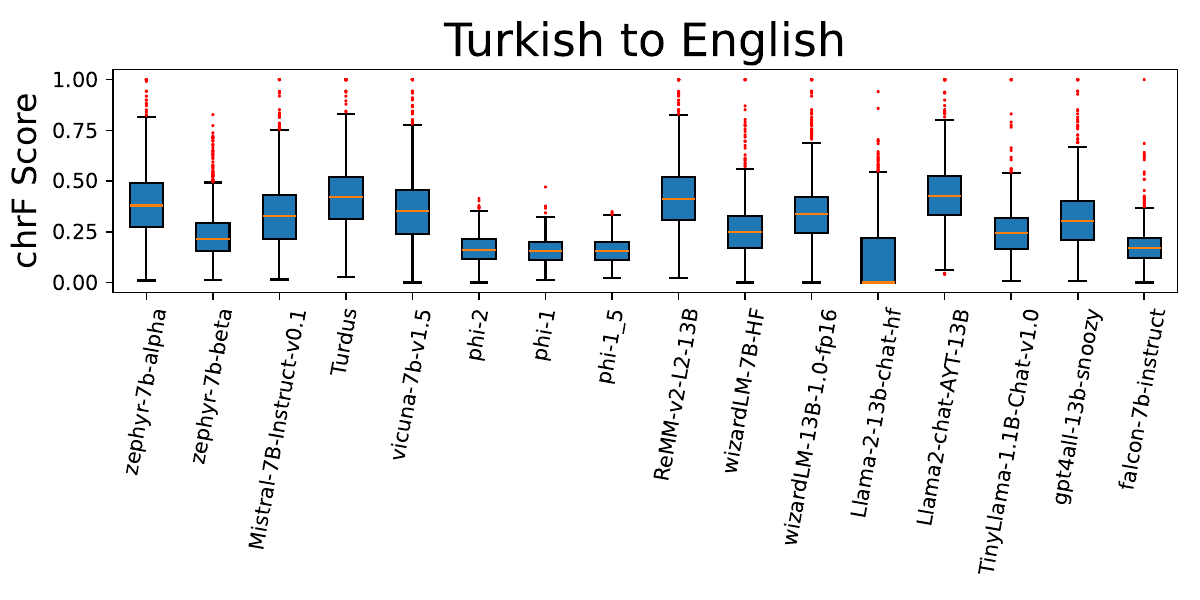}
    \includegraphics[width=0.47\textwidth]{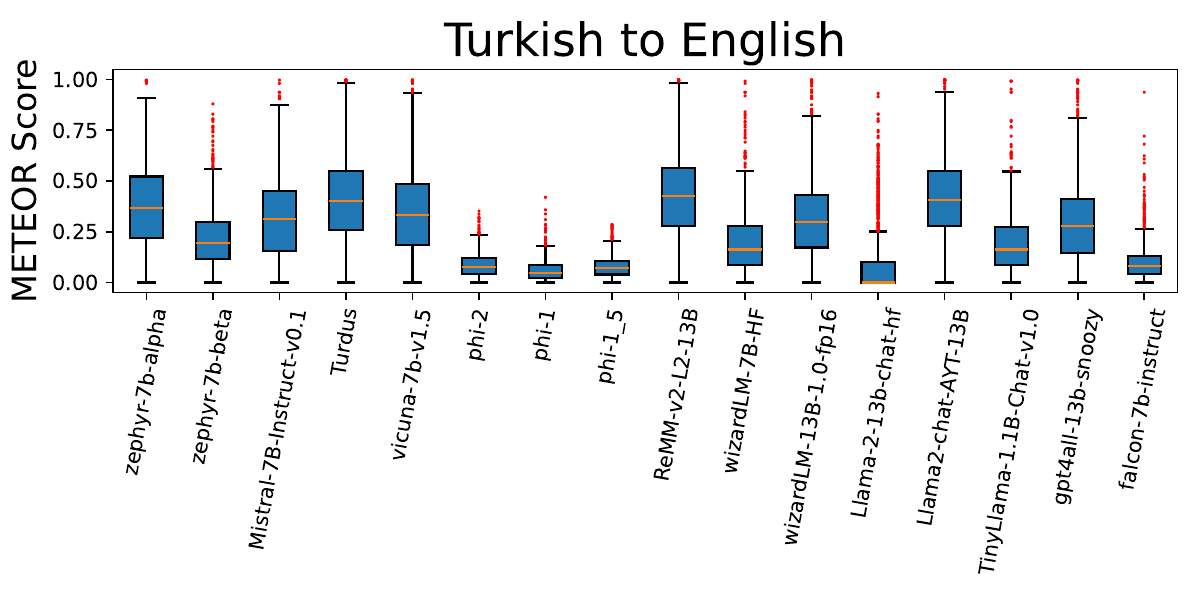}
    \caption{Turkish-to-English dataset per-sentence translation quality and timing statistics  }
    \label{fig:Turkish_translate_stats}
\end{figure}

\begin{figure}[th!]
    \centering
    \includegraphics[width=0.47\textwidth]{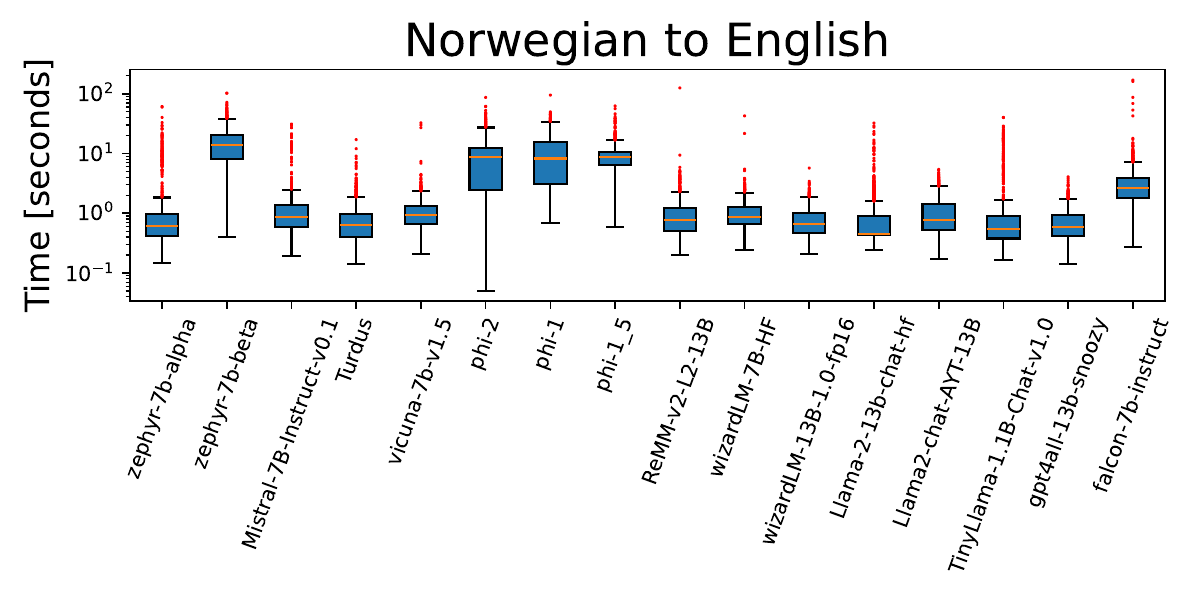}\\
    \includegraphics[width=0.47\textwidth]{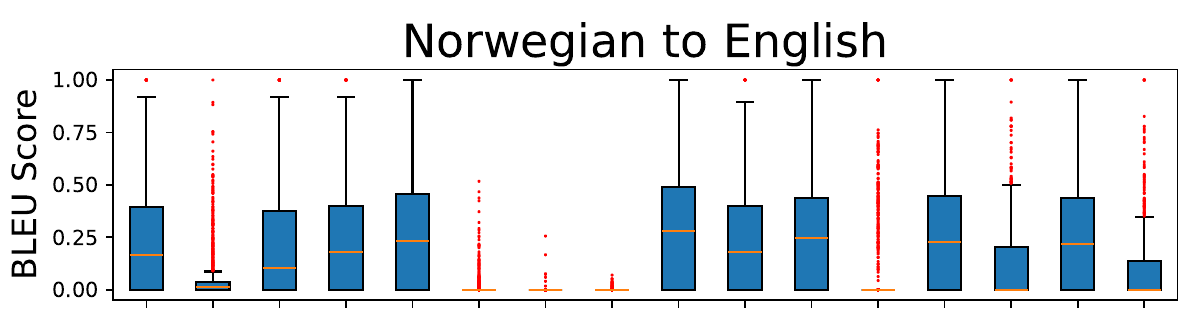}
    \includegraphics[width=0.47\textwidth]{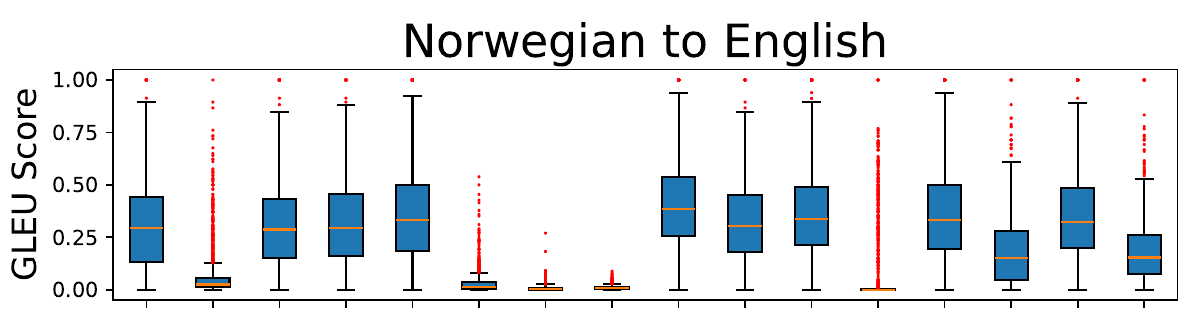}
    \includegraphics[width=0.47\textwidth]{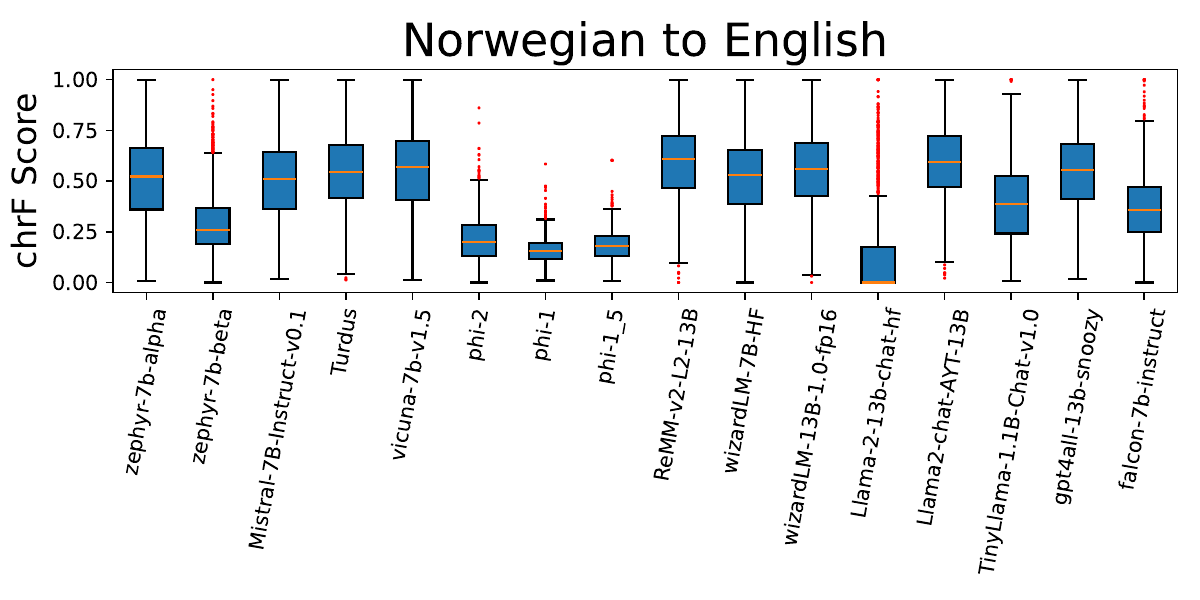}
    \includegraphics[width=0.47\textwidth]{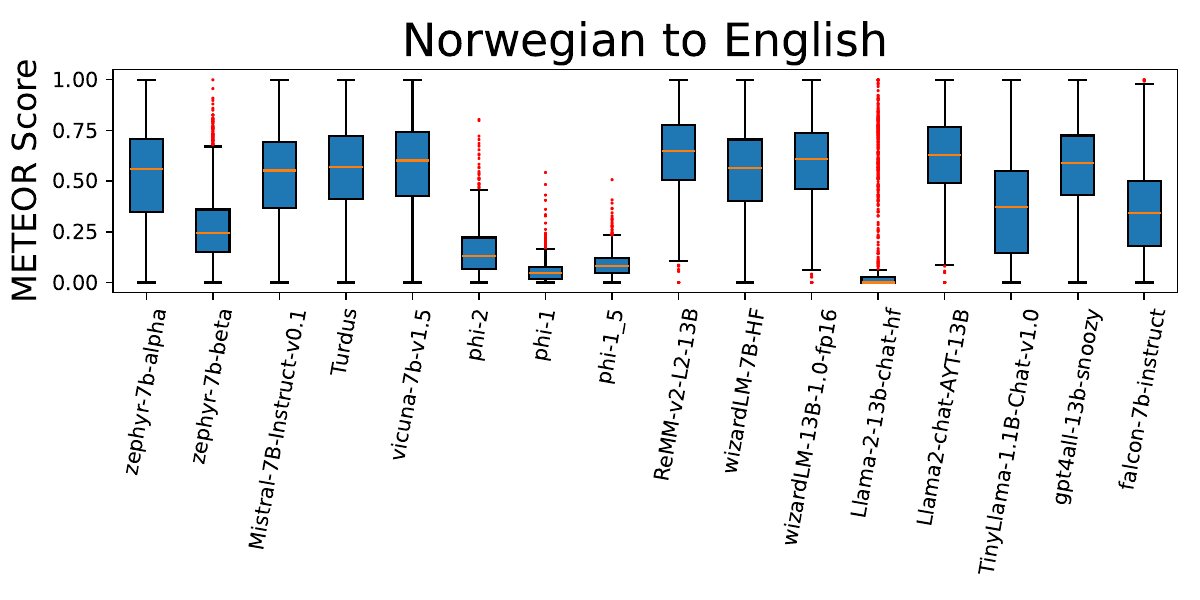}
    \caption{Norwegian-to-English dataset per-sentence translation quality and timing statistics  }
    \label{fig:Norwegian_translate_stats}
\end{figure}

\begin{figure}[th!]
    \centering
    \includegraphics[width=0.47\textwidth]{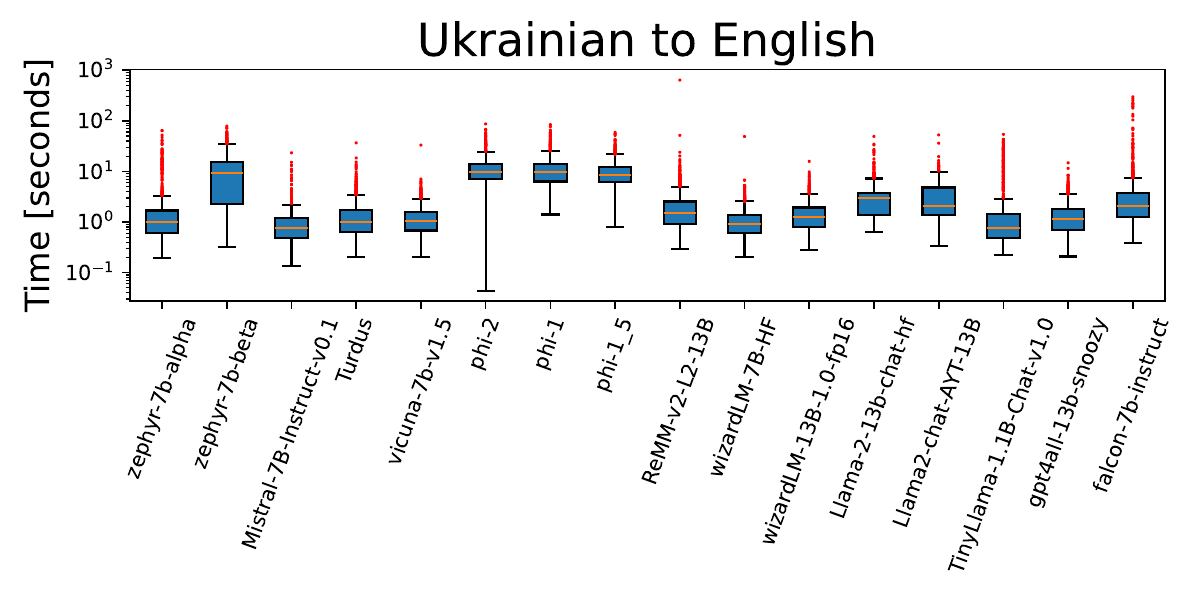}\\
    \includegraphics[width=0.47\textwidth]{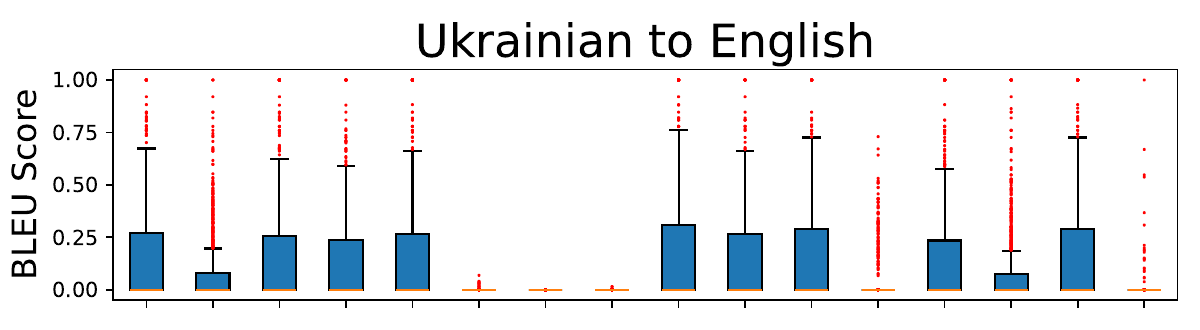}
    \includegraphics[width=0.47\textwidth]{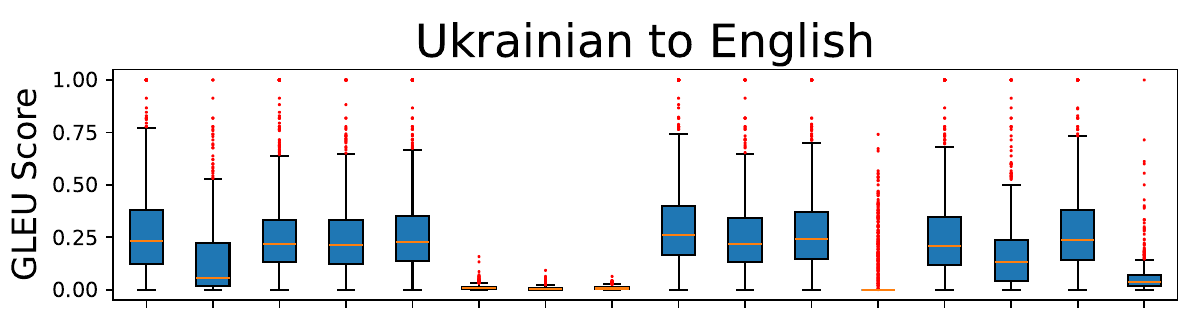}
    \includegraphics[width=0.47\textwidth]{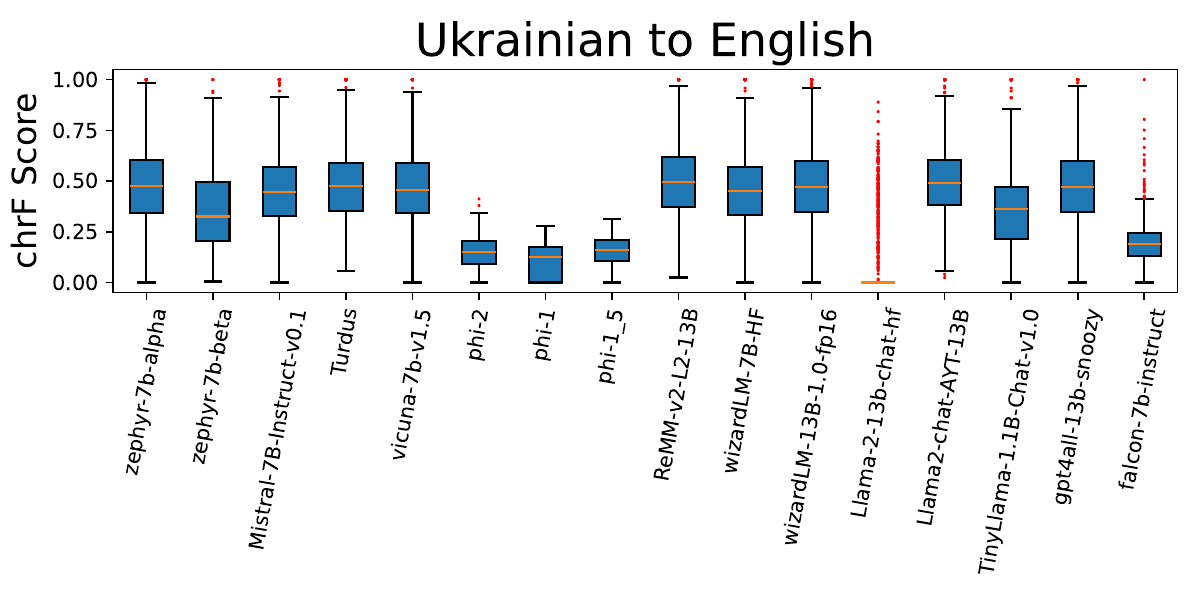}
    \includegraphics[width=0.47\textwidth]{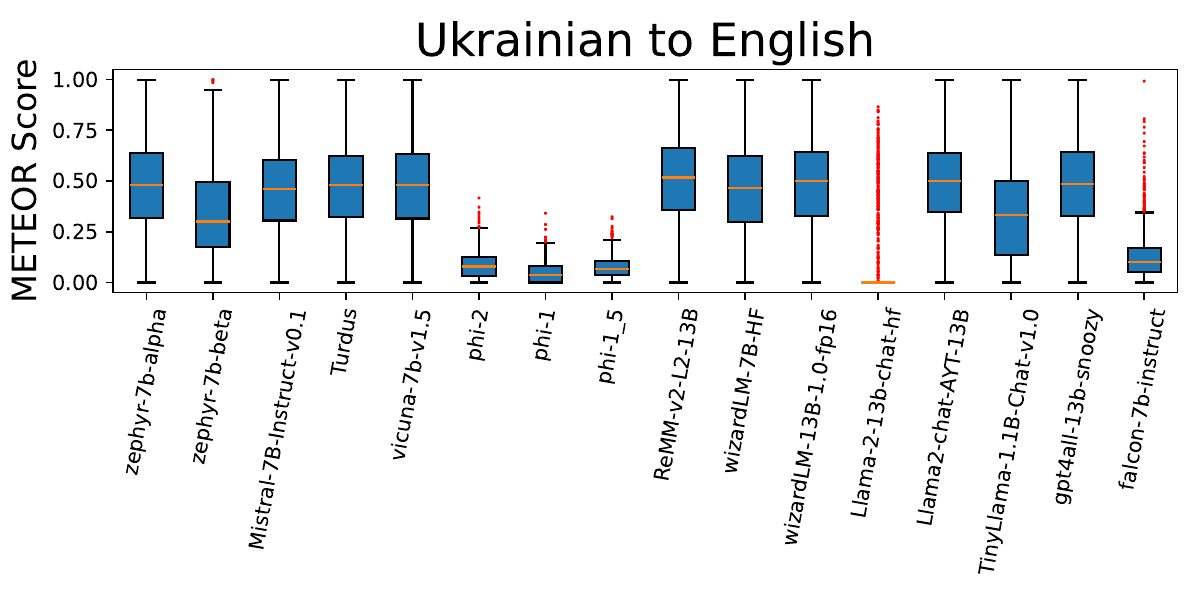}
    \caption{Ukranian-to-English dataset per-sentence translation quality and timing statistics  }
    \label{fig:Ukranian_translate_stats}
\end{figure}

\begin{figure}[th!]
    \centering
    \includegraphics[width=0.47\textwidth]{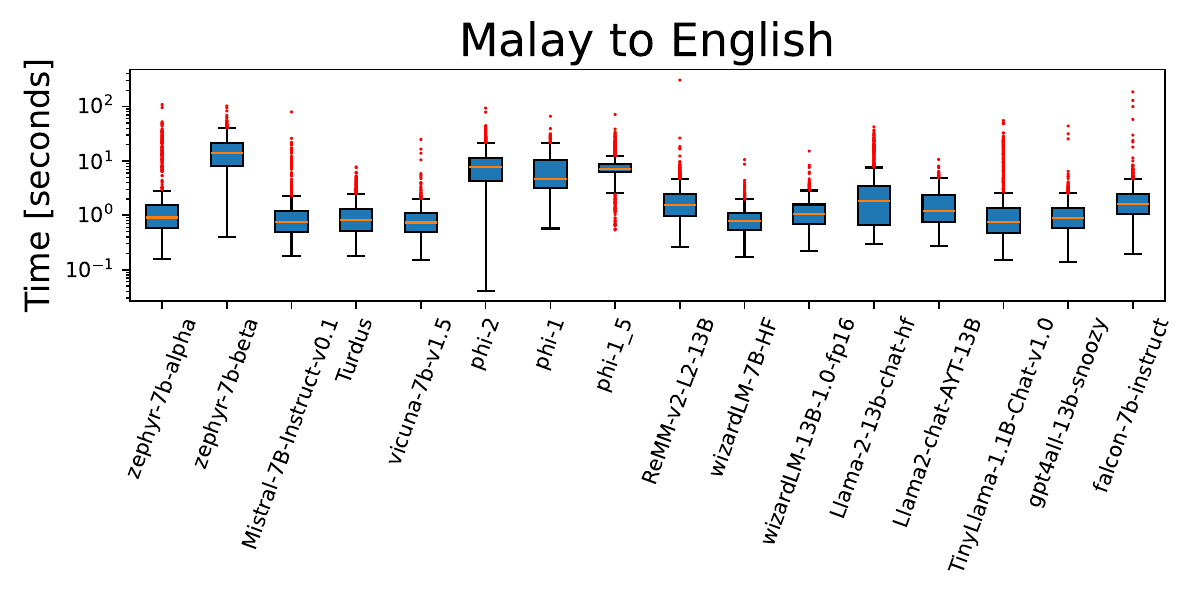}\\
    \includegraphics[width=0.47\textwidth]{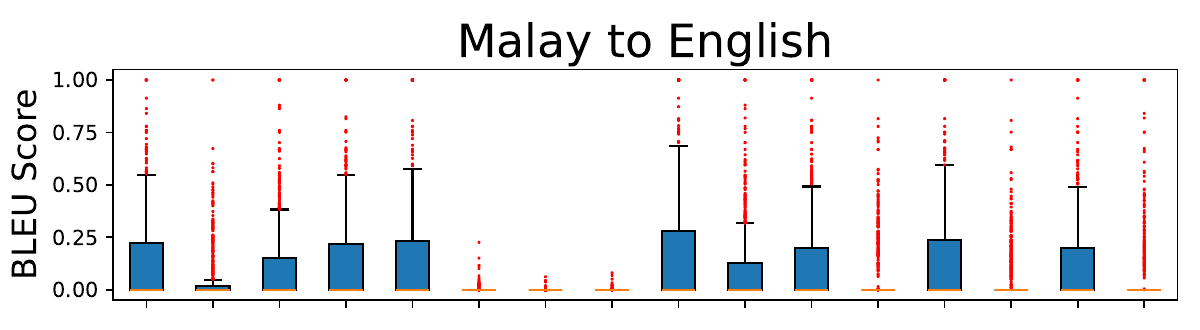}
    \includegraphics[width=0.47\textwidth]{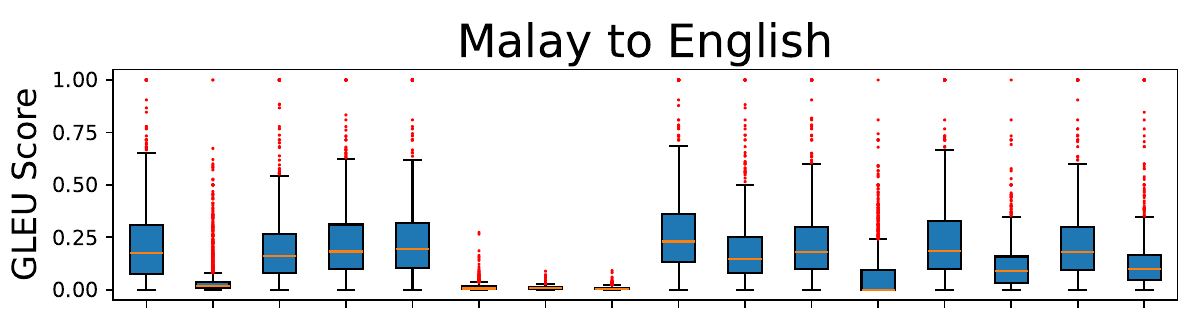}
    \includegraphics[width=0.47\textwidth]{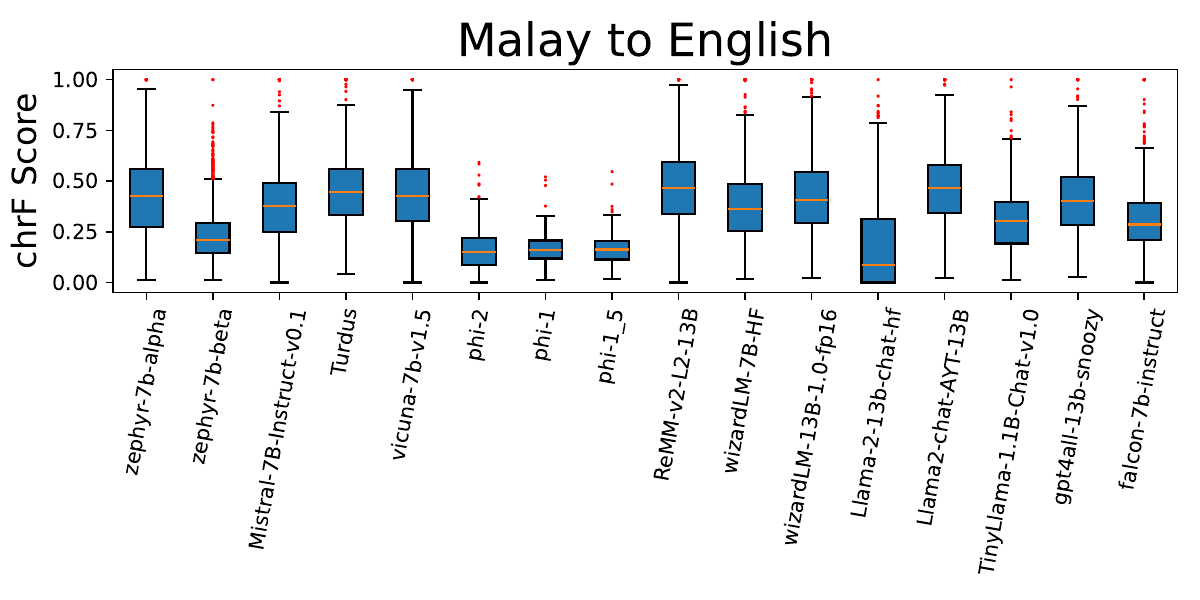}
    \includegraphics[width=0.47\textwidth]{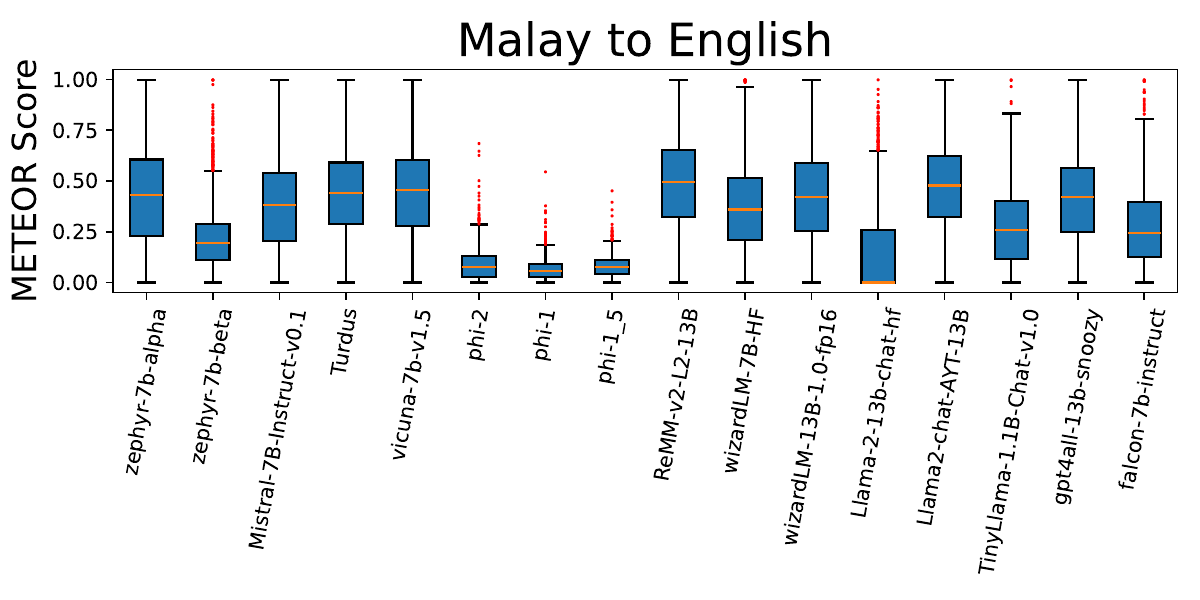}
    \caption{Malay-to-English dataset per-sentence translation quality and timing statistics  }
    \label{fig:Malay_translate_stats}
\end{figure}

\begin{figure}[th!]
    \centering
    \includegraphics[width=0.47\textwidth]{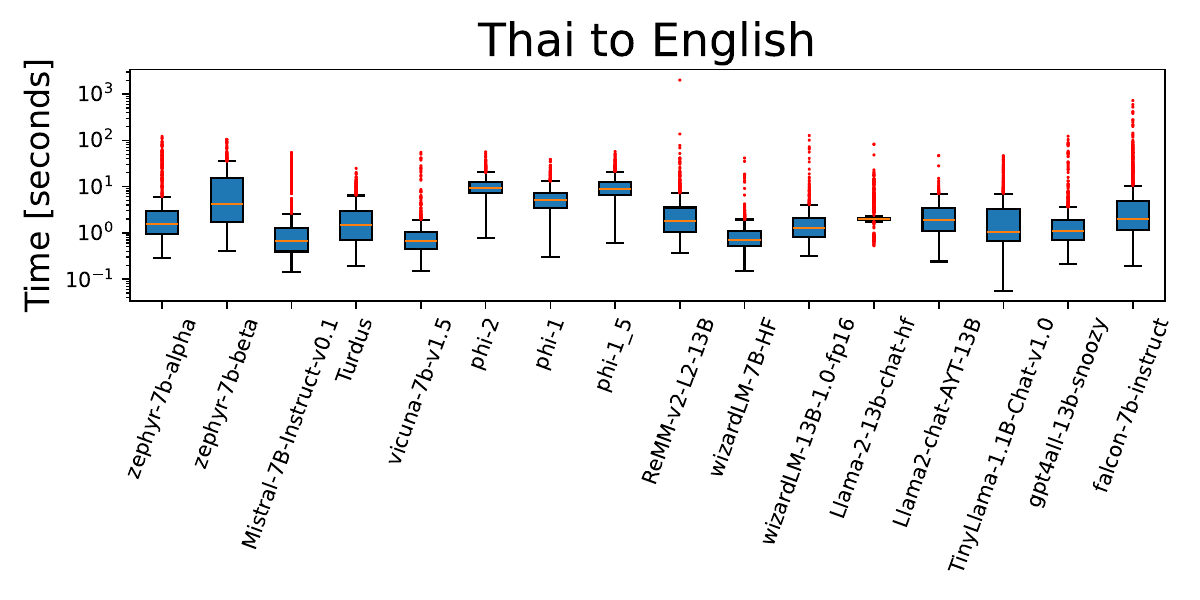}\\
    \includegraphics[width=0.47\textwidth]{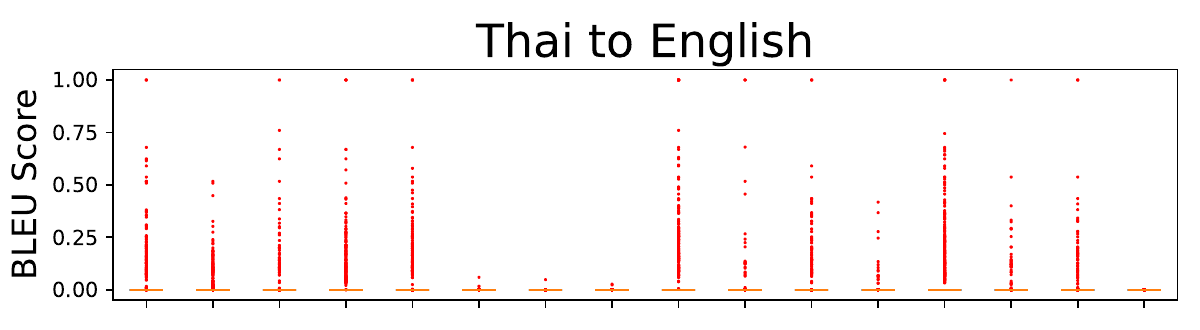}
    \includegraphics[width=0.47\textwidth]{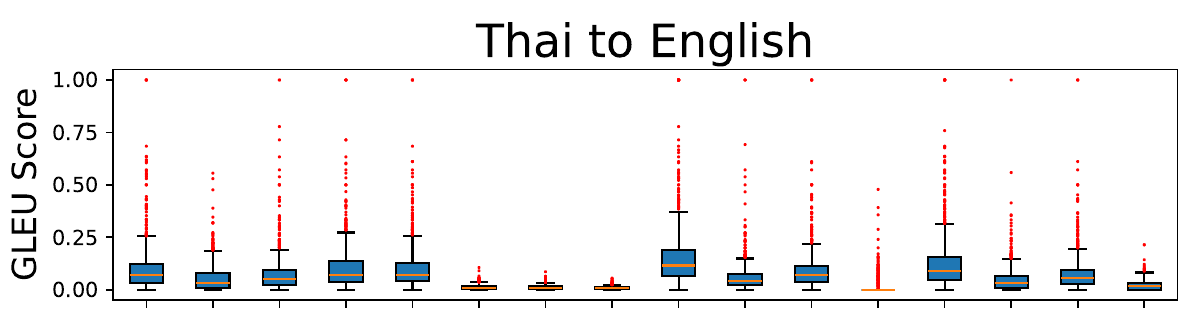}
    \includegraphics[width=0.47\textwidth]{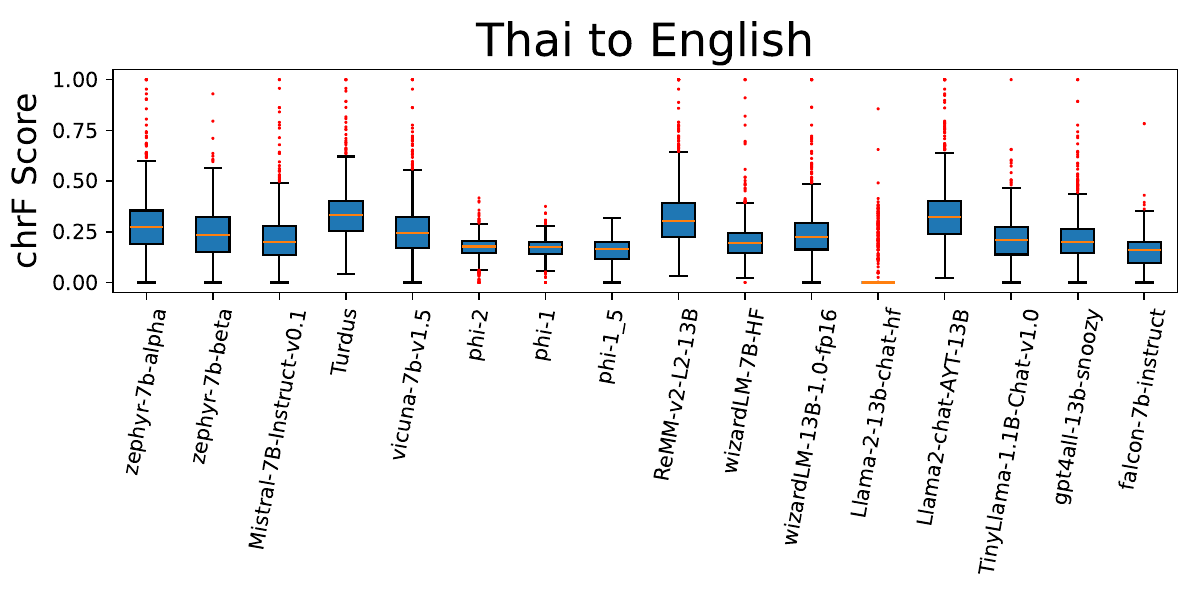}
    \includegraphics[width=0.47\textwidth]{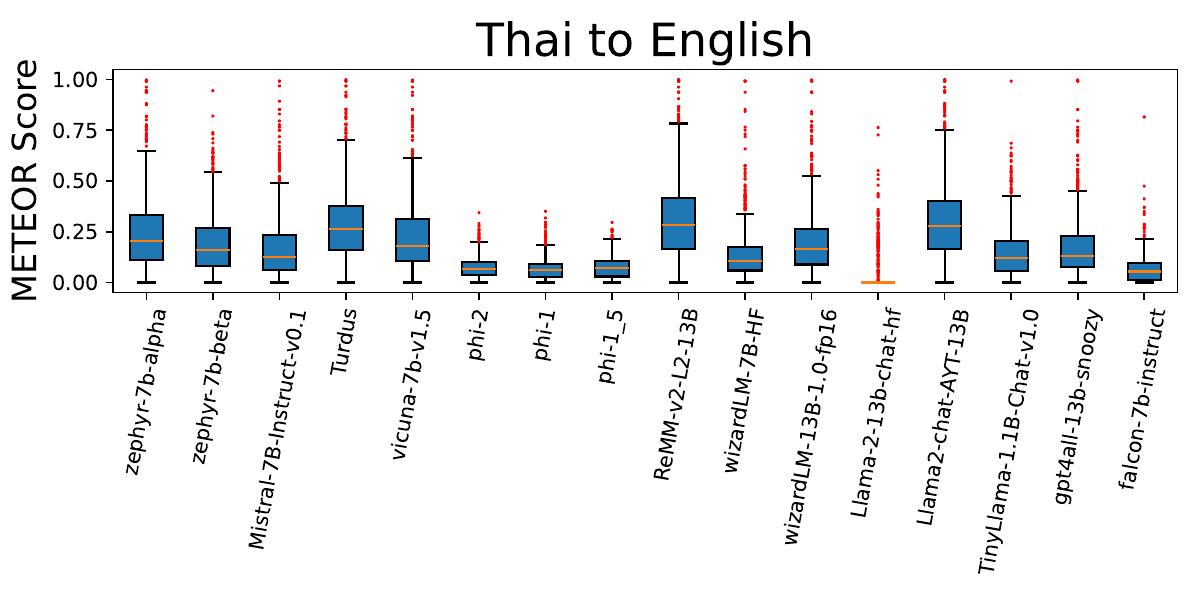}
    \caption{Thai-to-English dataset per-sentence translation quality and timing statistics }
    \label{fig:Thai_translate_stats}
\end{figure}

\begin{figure}[th!]
    \centering
    \includegraphics[width=0.47\textwidth]{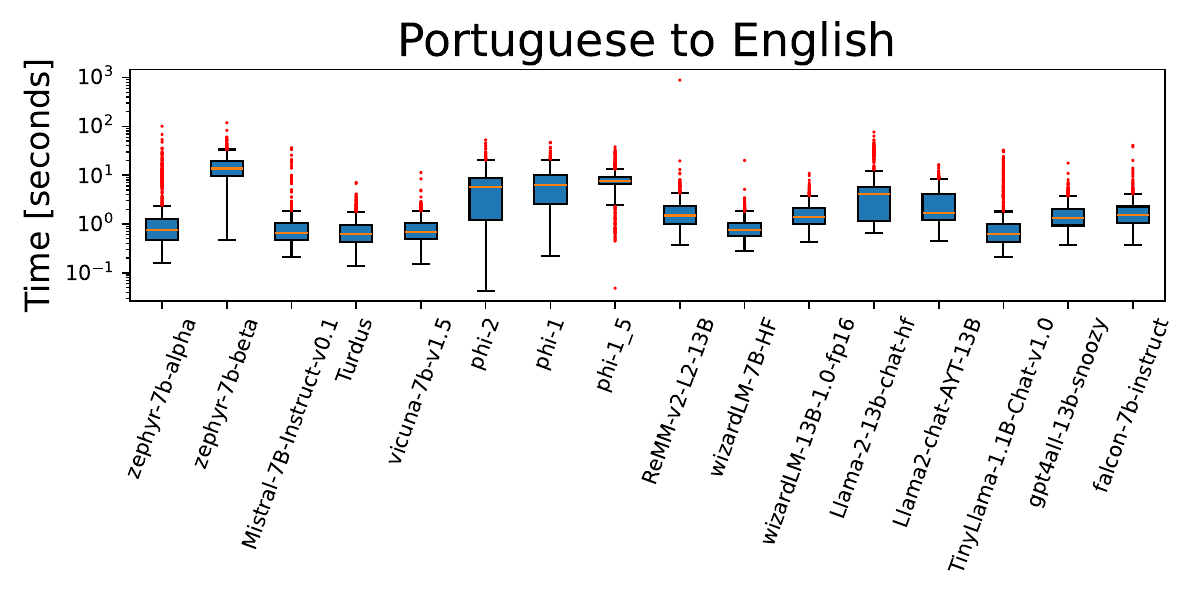}\\
    \includegraphics[width=0.47\textwidth]{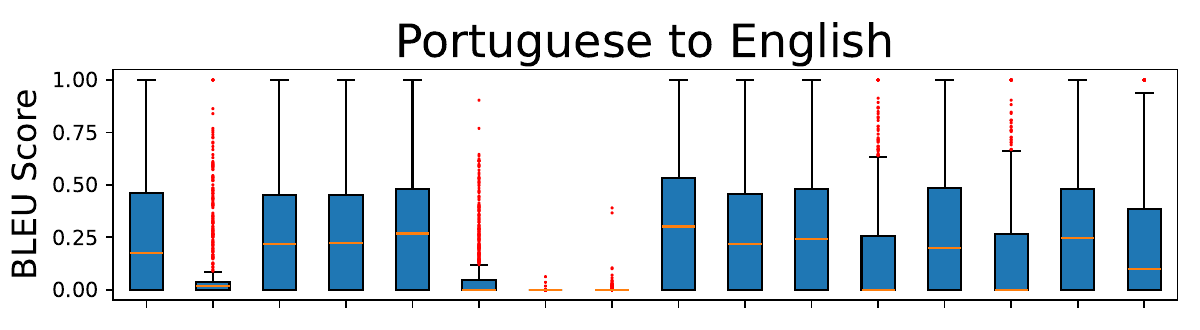}
    \includegraphics[width=0.47\textwidth]{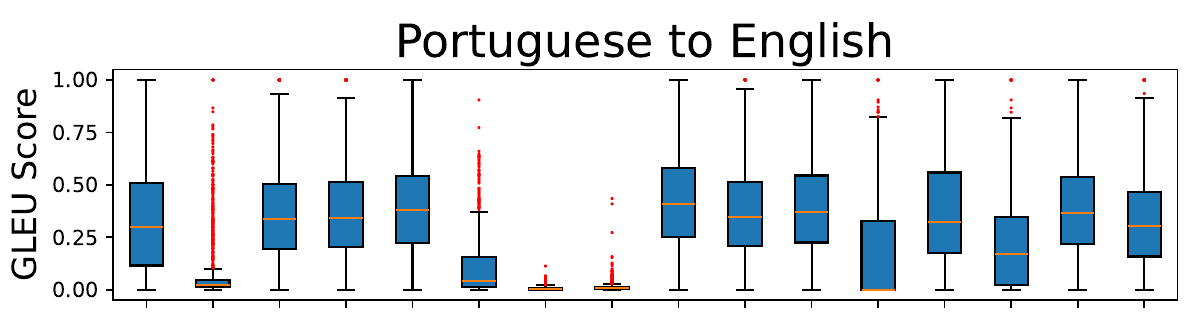}
    \includegraphics[width=0.47\textwidth]{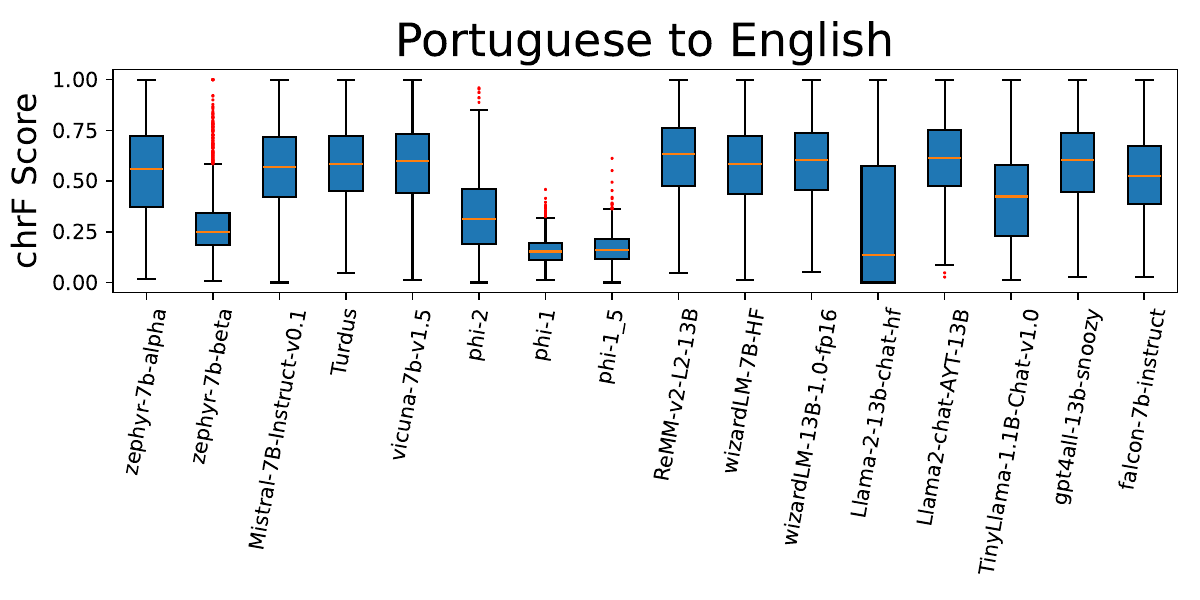}
    \includegraphics[width=0.47\textwidth]{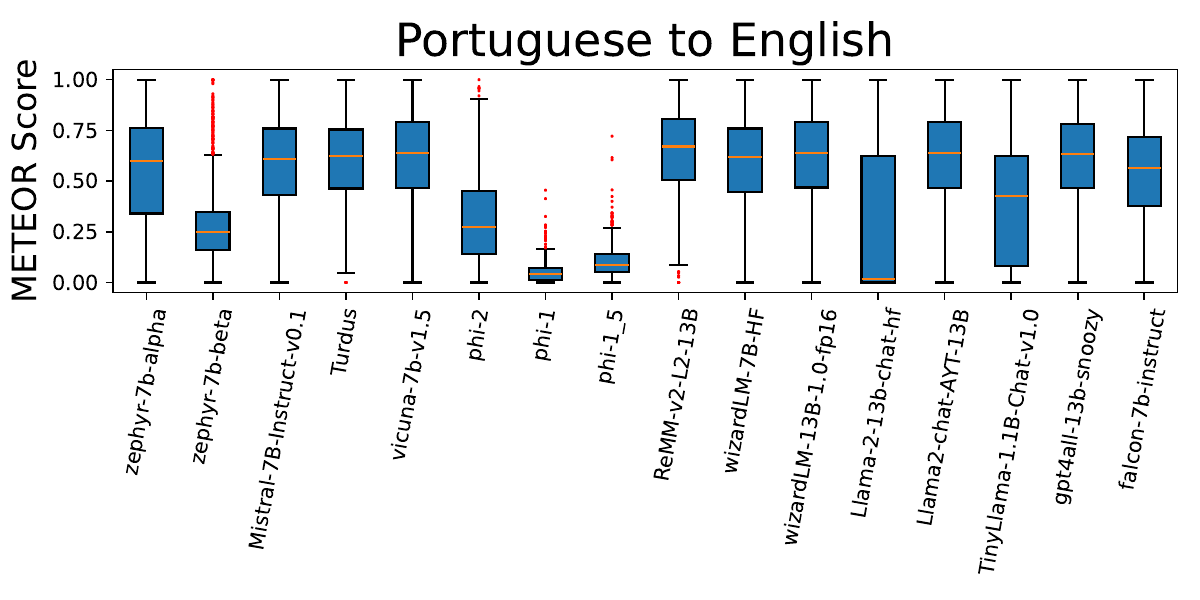}
    \caption{Portuguese-to-English dataset per-sentence translation quality and timing statistics  }
    \label{fig:Portuguese_translate_stats}
\end{figure}

\begin{figure}[th!]
    \centering
    \includegraphics[width=0.47\textwidth]{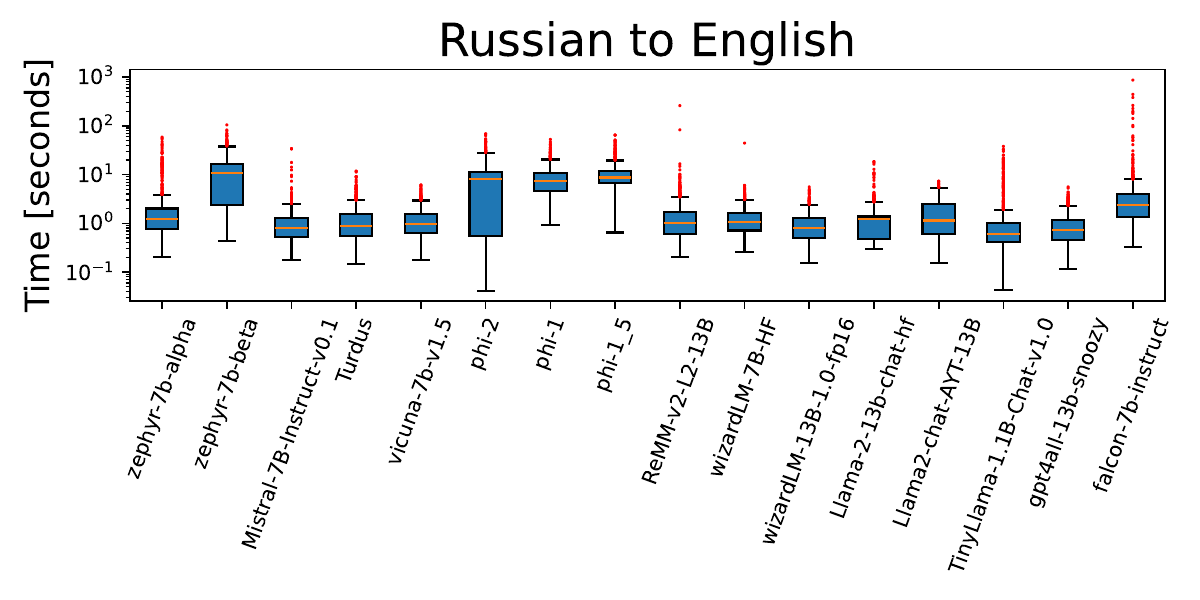}\\
    \includegraphics[width=0.47\textwidth]{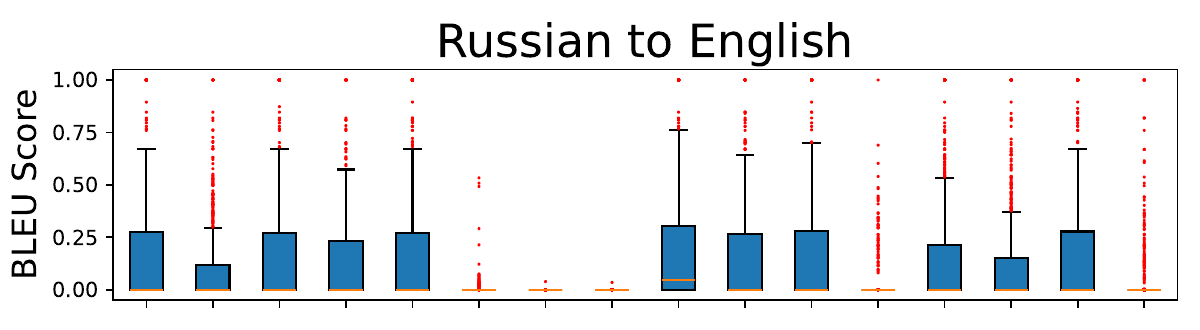}
    \includegraphics[width=0.47\textwidth]{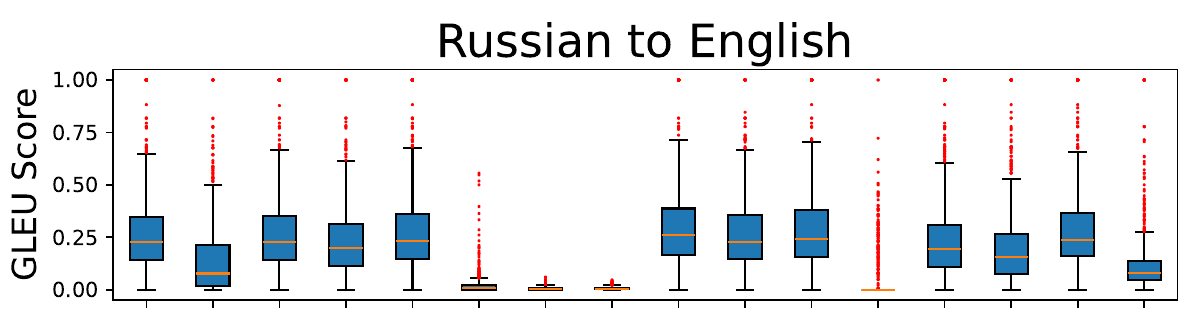}
    \includegraphics[width=0.47\textwidth]{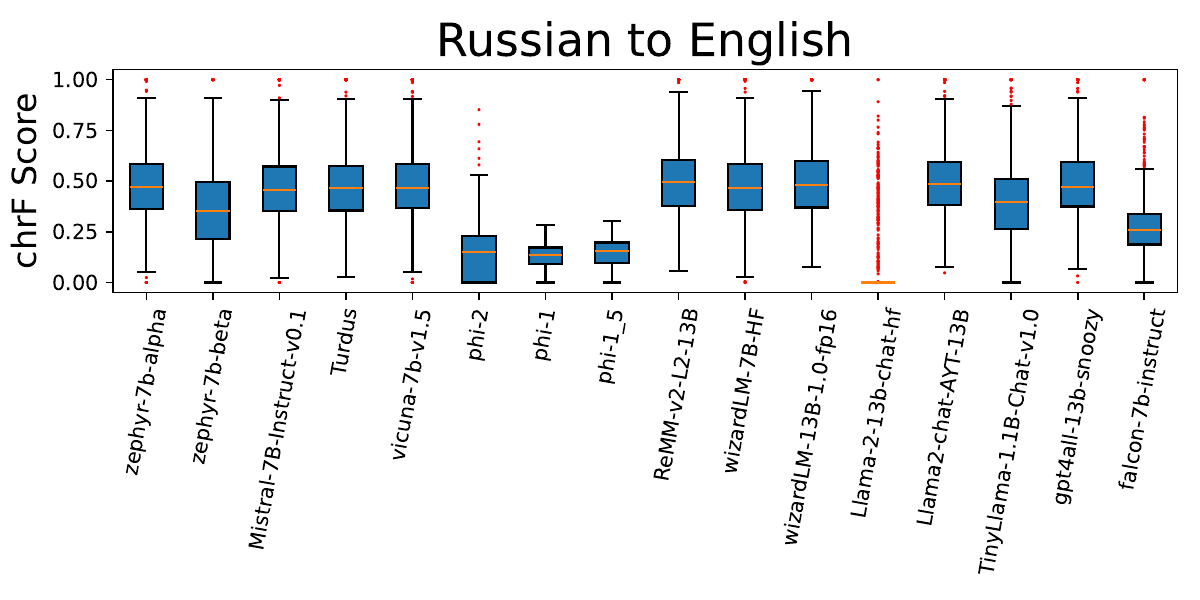}
    \includegraphics[width=0.47\textwidth]{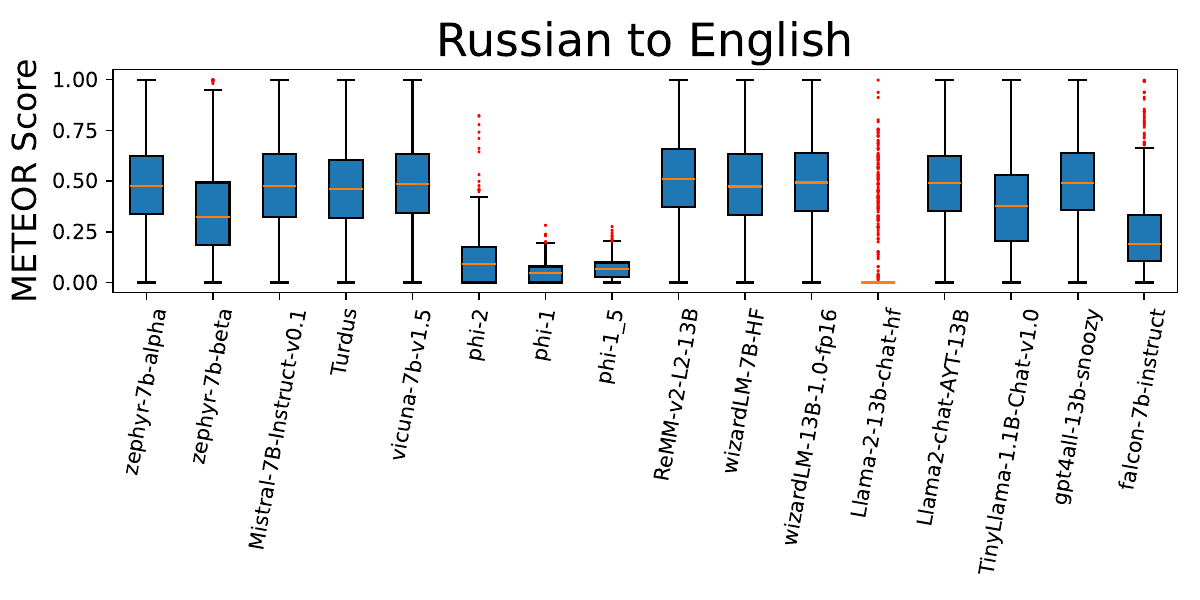}
    \caption{Russian-to-English dataset per-sentence translation quality and timing statistics  }
    \label{fig:Russian_translate_stats}
\end{figure}

\begin{figure}[th!]
    \centering
    \includegraphics[width=0.47\textwidth]{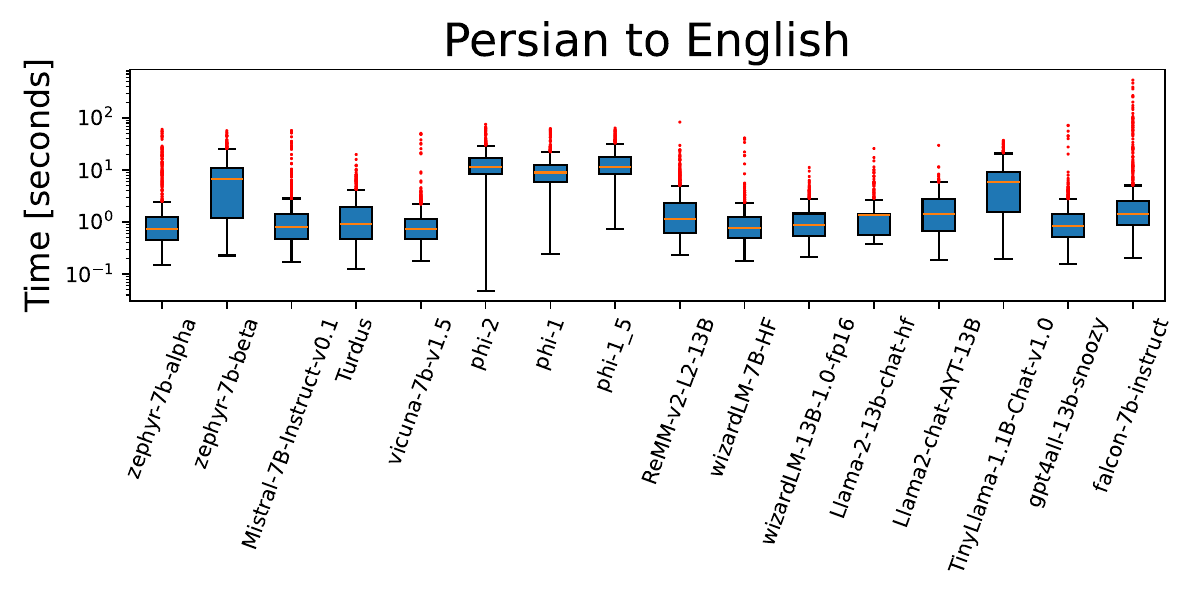}\\
    \includegraphics[width=0.47\textwidth]{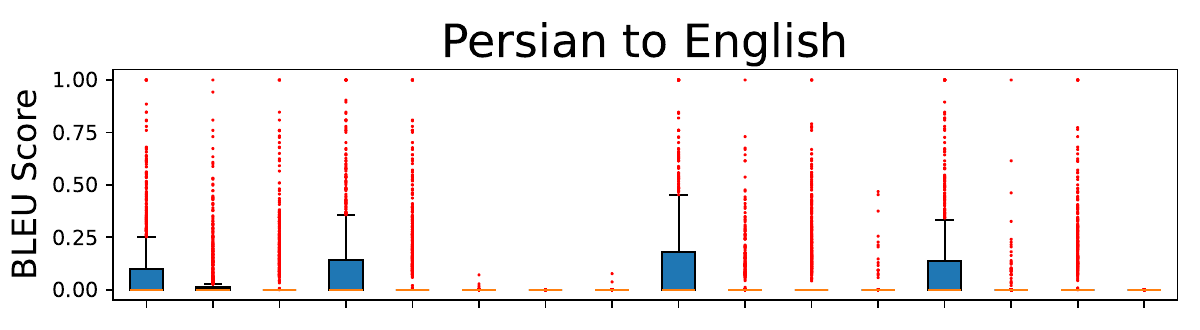}
    \includegraphics[width=0.47\textwidth]{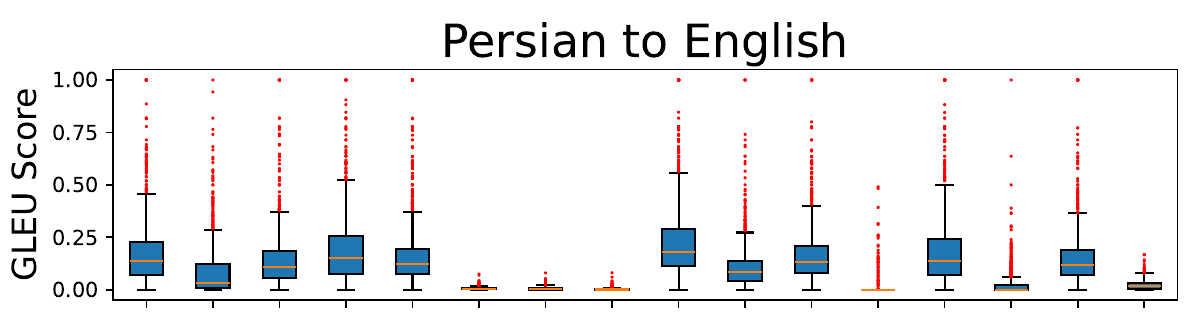}
    \includegraphics[width=0.47\textwidth]{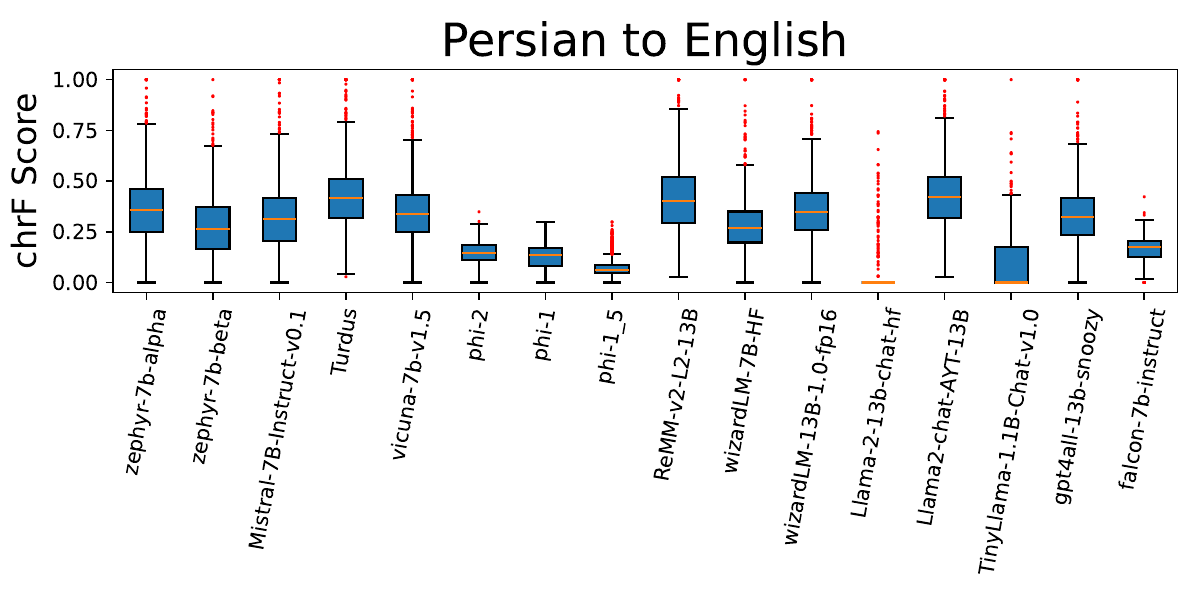}
    \includegraphics[width=0.47\textwidth]{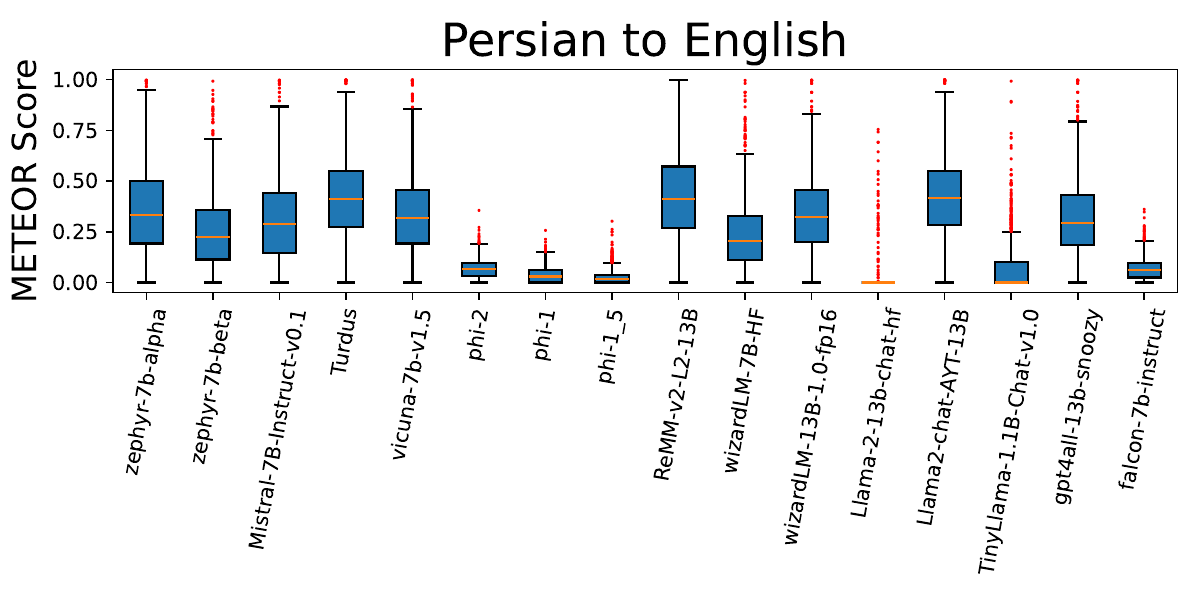}
    \caption{Persian-to-English dataset per-sentence translation quality and timing statistics  }
    \label{fig:Persian_translate_stats}
\end{figure}

\begin{figure}[th!]
    \centering
    \includegraphics[width=0.47\textwidth]{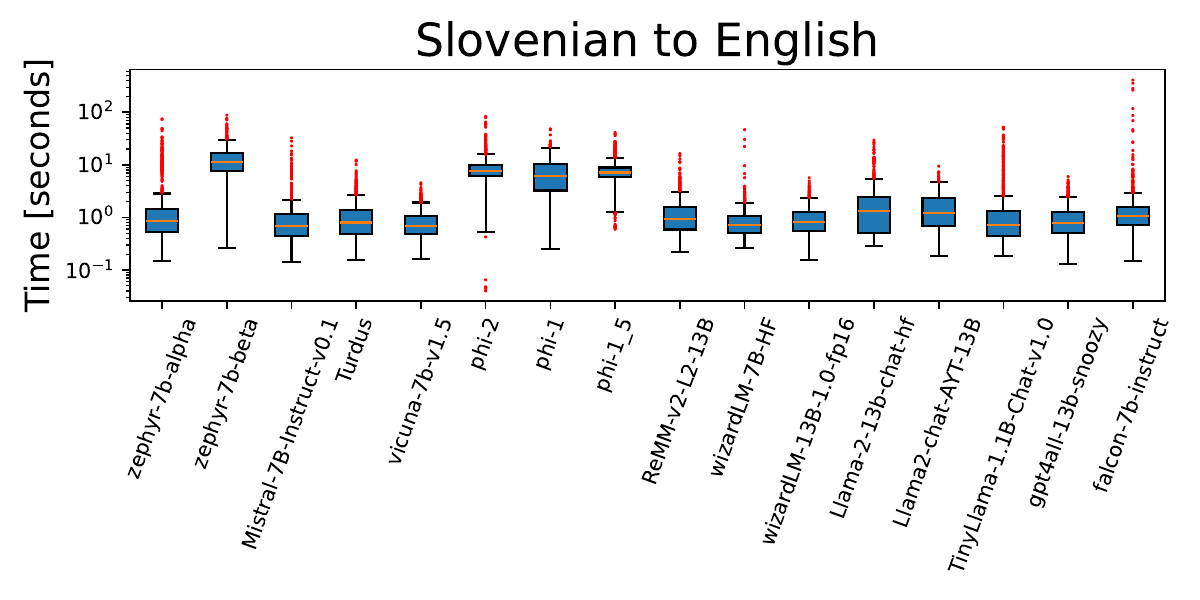}\\
    \includegraphics[width=0.47\textwidth]{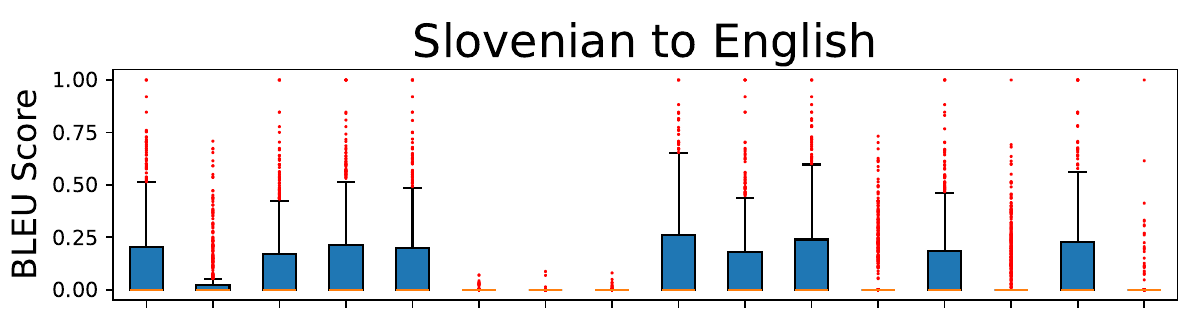}
    \includegraphics[width=0.47\textwidth]{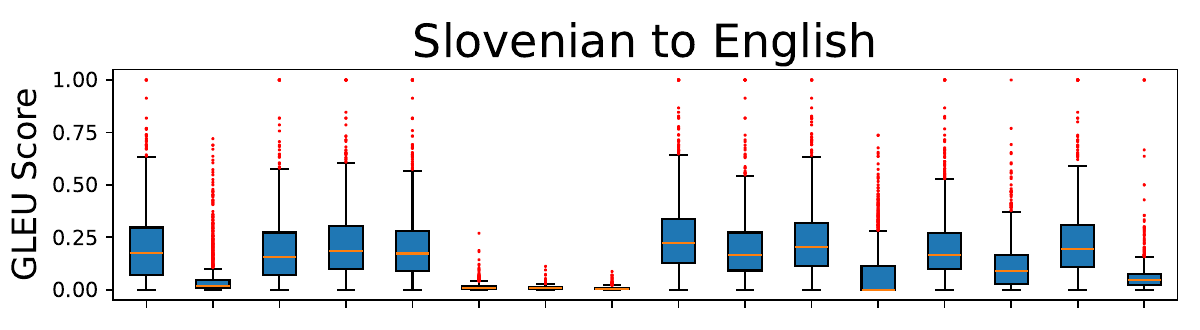}
    \includegraphics[width=0.47\textwidth]{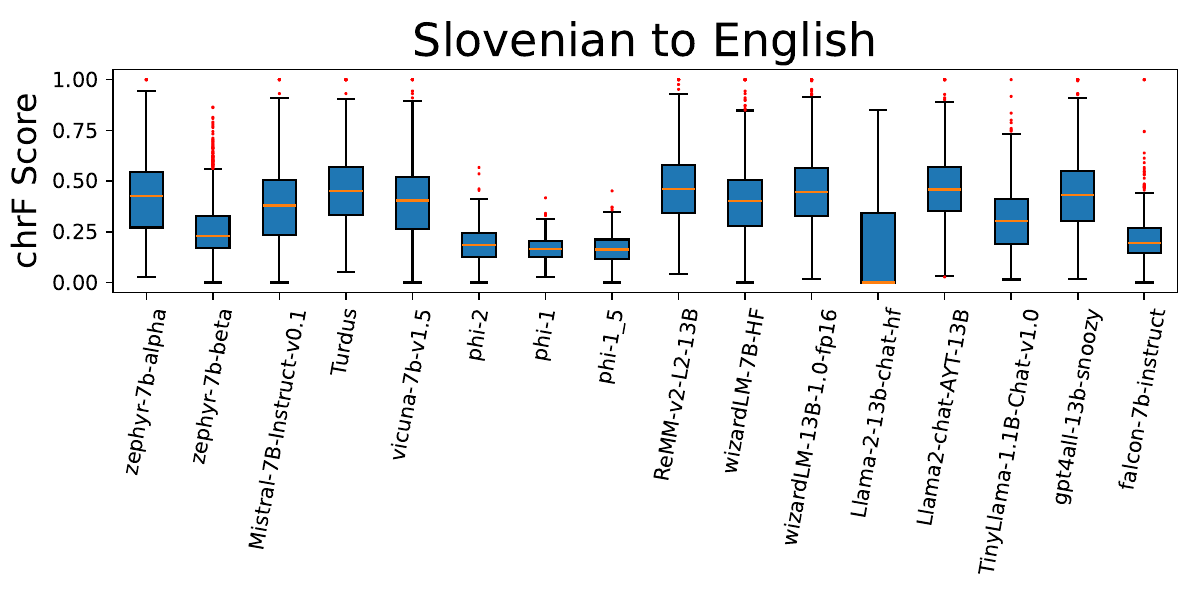}
    \includegraphics[width=0.47\textwidth]{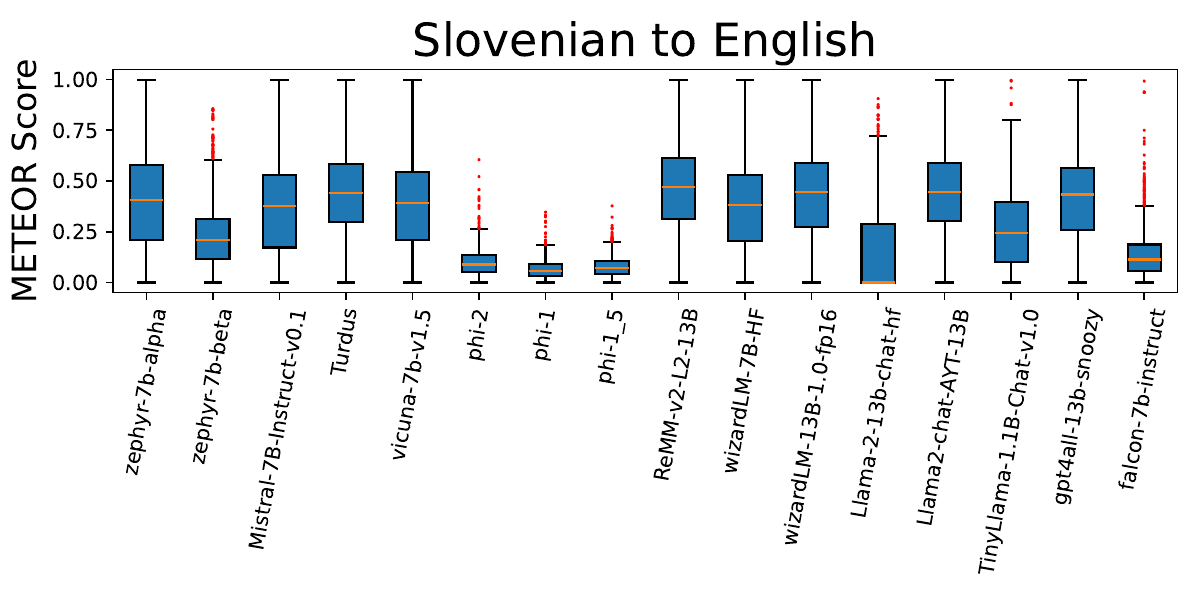}
    \caption{Slovenian-to-English dataset per-sentence translation quality and timing statistics  }
    \label{fig:Slovenian_translate_stats}
\end{figure}

\begin{figure}[th!]
    \centering
    \includegraphics[width=0.47\textwidth]{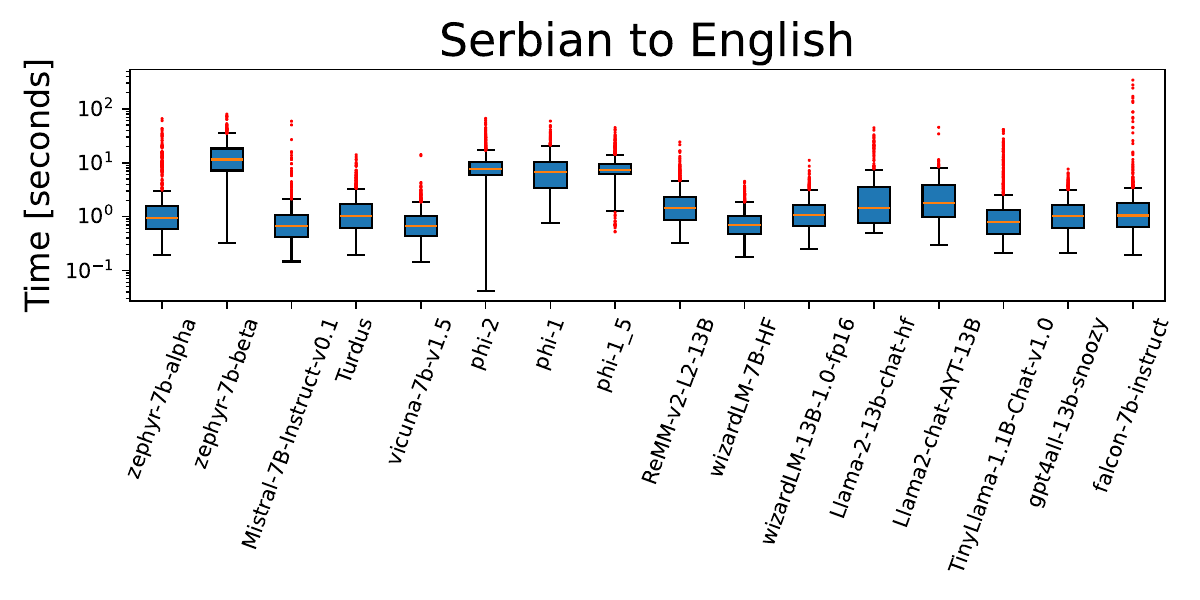}\\
    \includegraphics[width=0.47\textwidth]{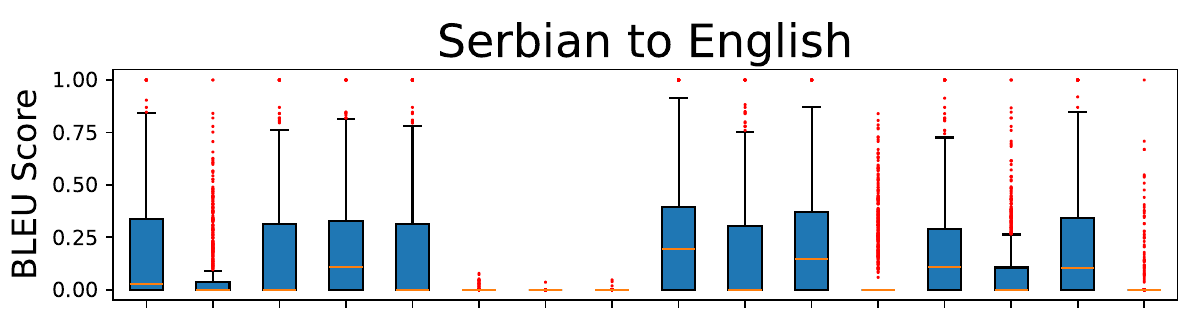}
    \includegraphics[width=0.47\textwidth]{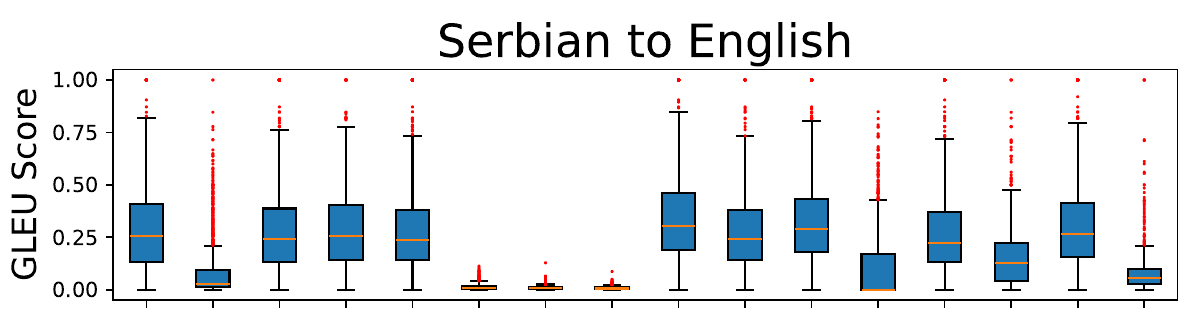}
    \includegraphics[width=0.47\textwidth]{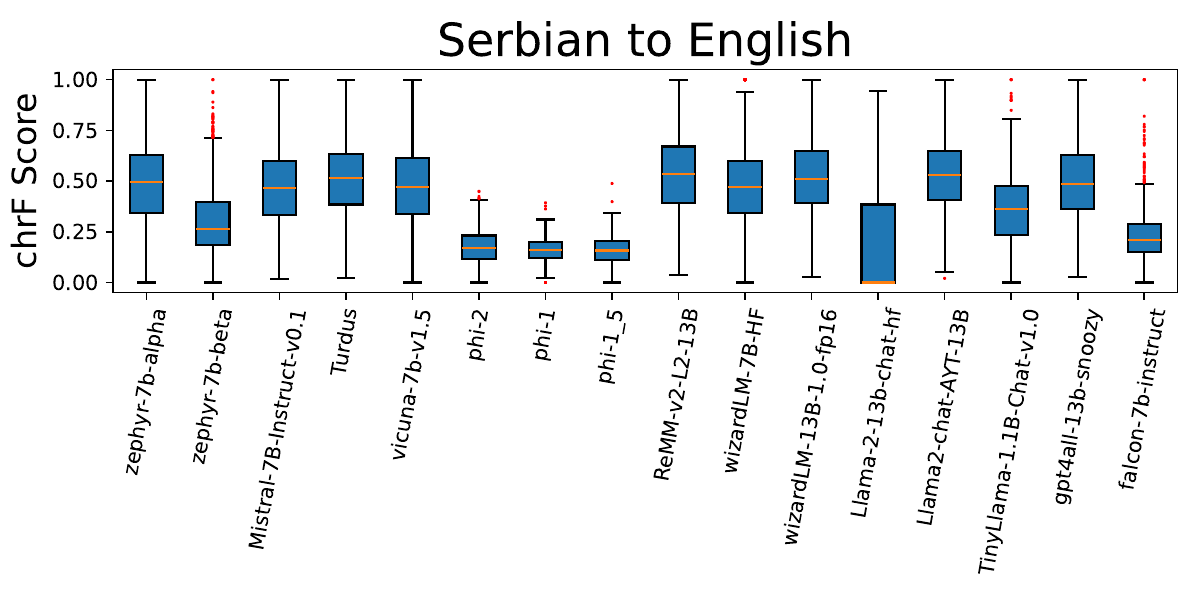}
    \includegraphics[width=0.47\textwidth]{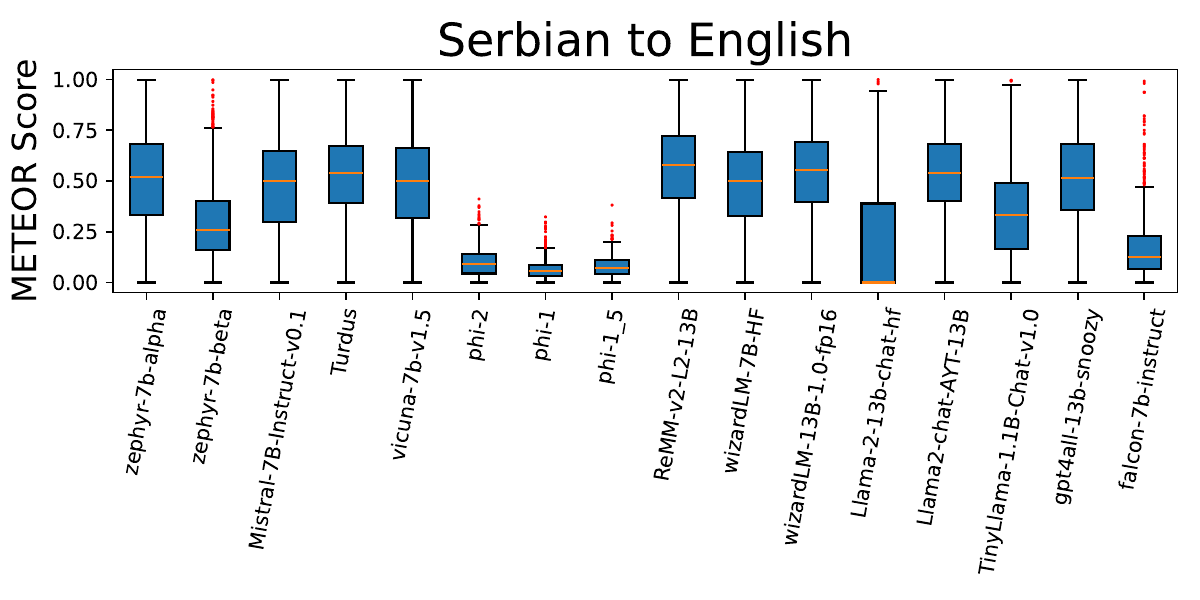}
    \caption{Serbian-to-English dataset per-sentence translation quality and timing statistics  }
    \label{fig:Serbian_translate_stats}
\end{figure}

\begin{figure}[th!]
    \centering
    \includegraphics[width=0.47\textwidth]{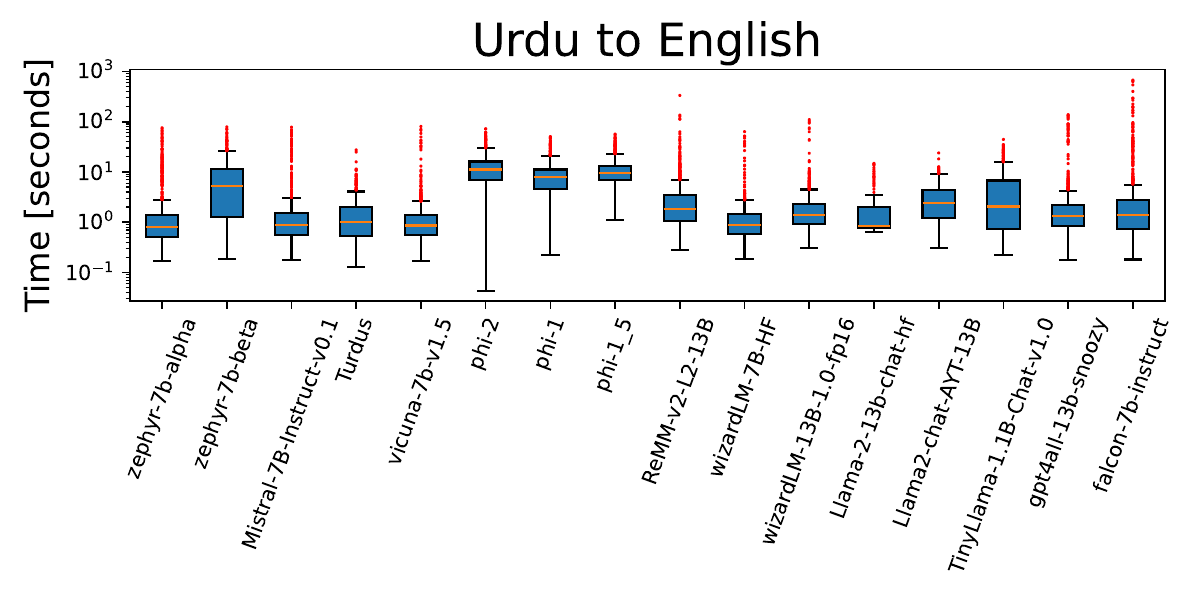}\\
    \includegraphics[width=0.47\textwidth]{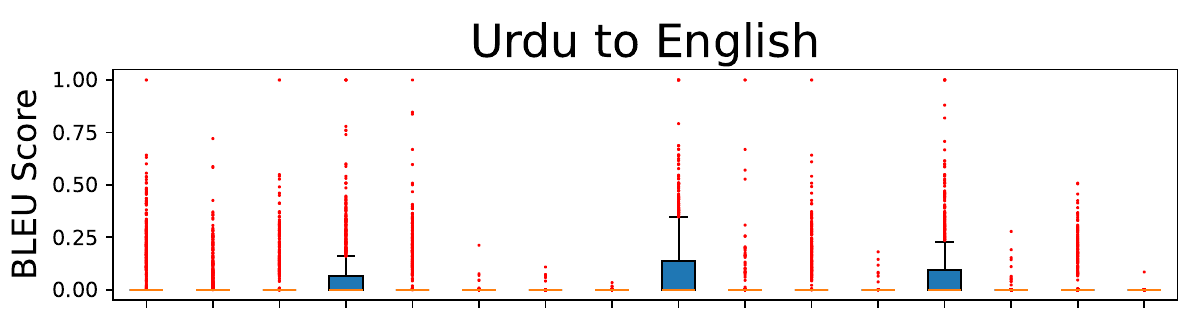}
    \includegraphics[width=0.47\textwidth]{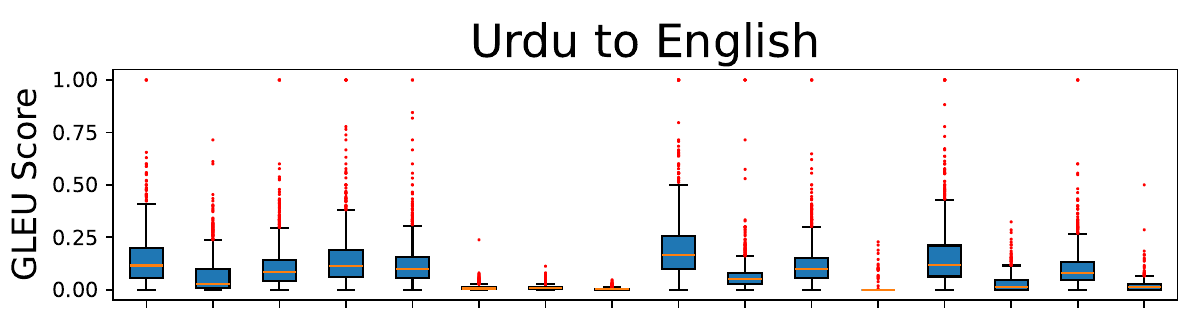}
    \includegraphics[width=0.47\textwidth]{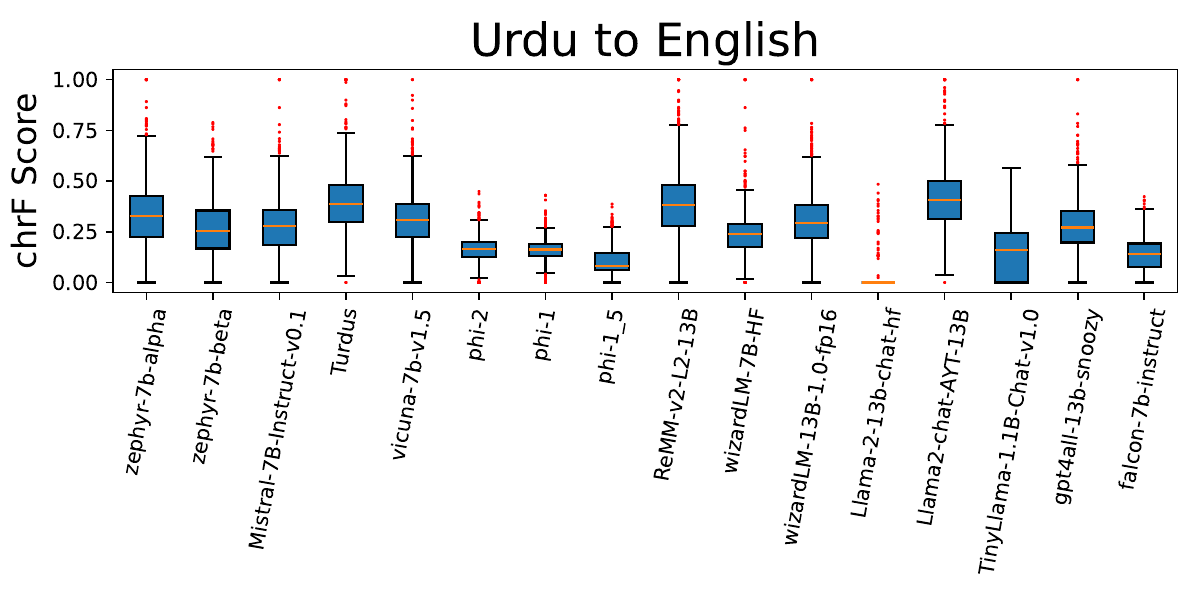}
    \includegraphics[width=0.47\textwidth]{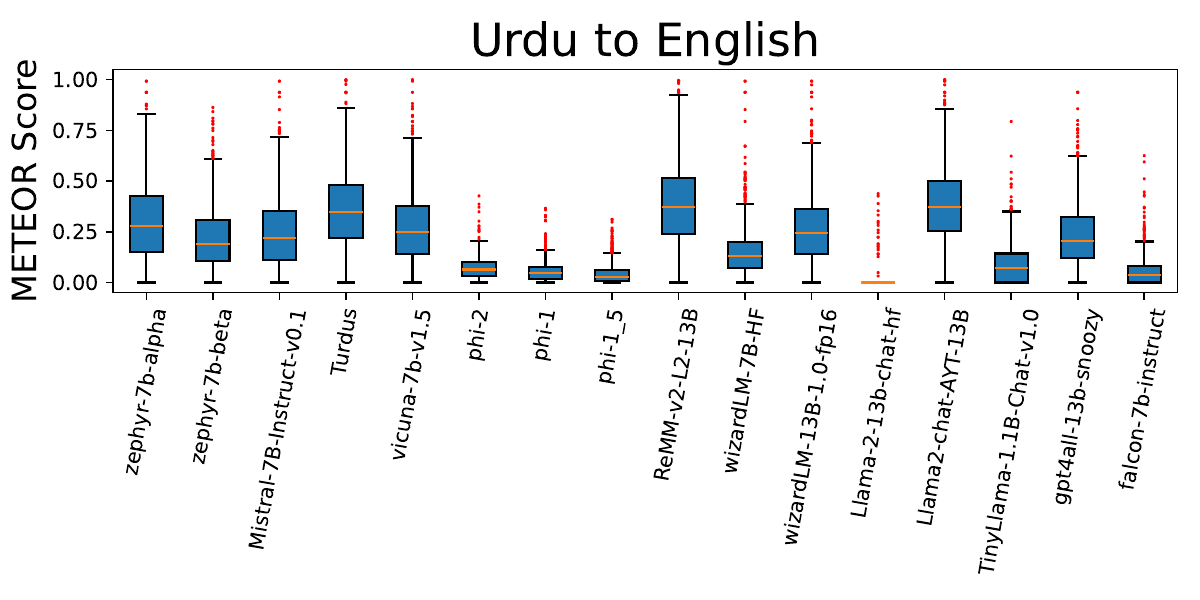}
    \caption{Urdu-to-English dataset per-sentence translation quality and timing statistics  }
    \label{fig:Urdu_translate_stats}
\end{figure}

\begin{figure}[th!]
    \centering
    \includegraphics[width=0.47\textwidth]{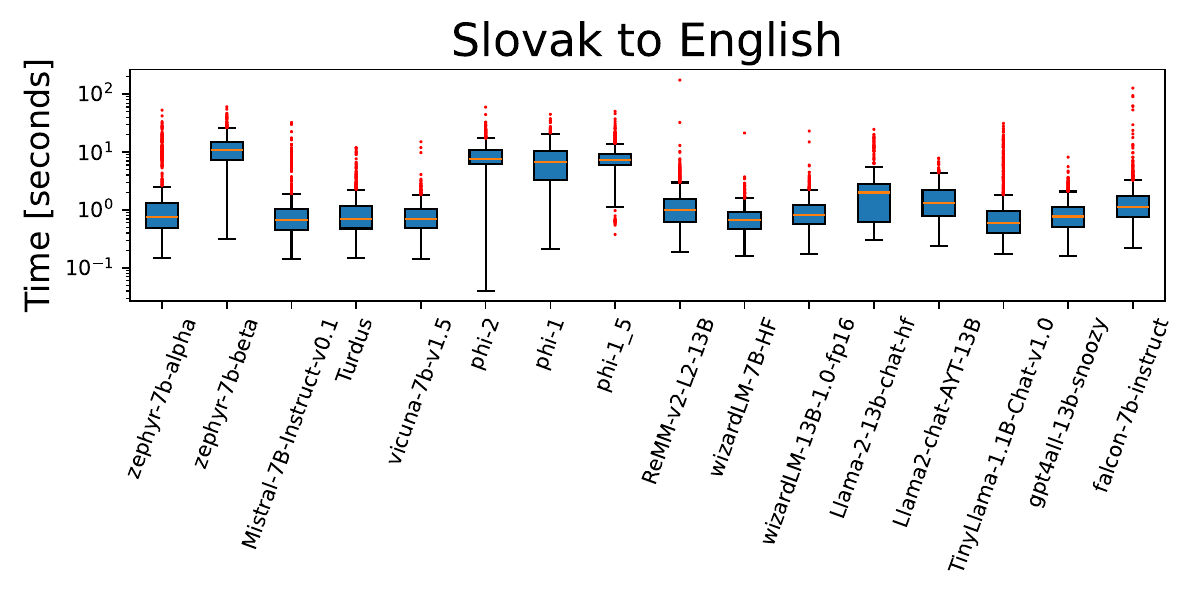}\\
    \includegraphics[width=0.47\textwidth]{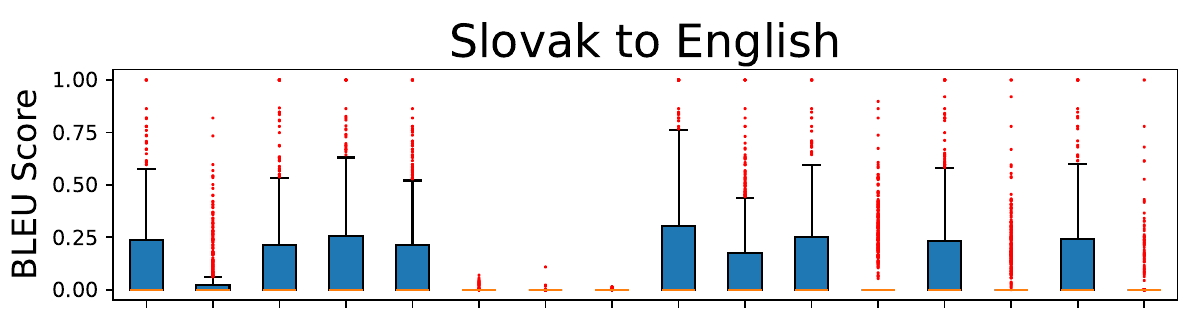}
    \includegraphics[width=0.47\textwidth]{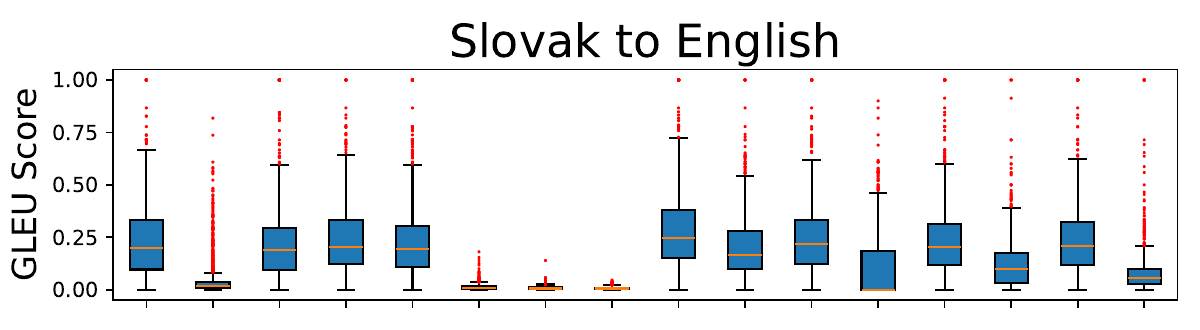}
    \includegraphics[width=0.47\textwidth]{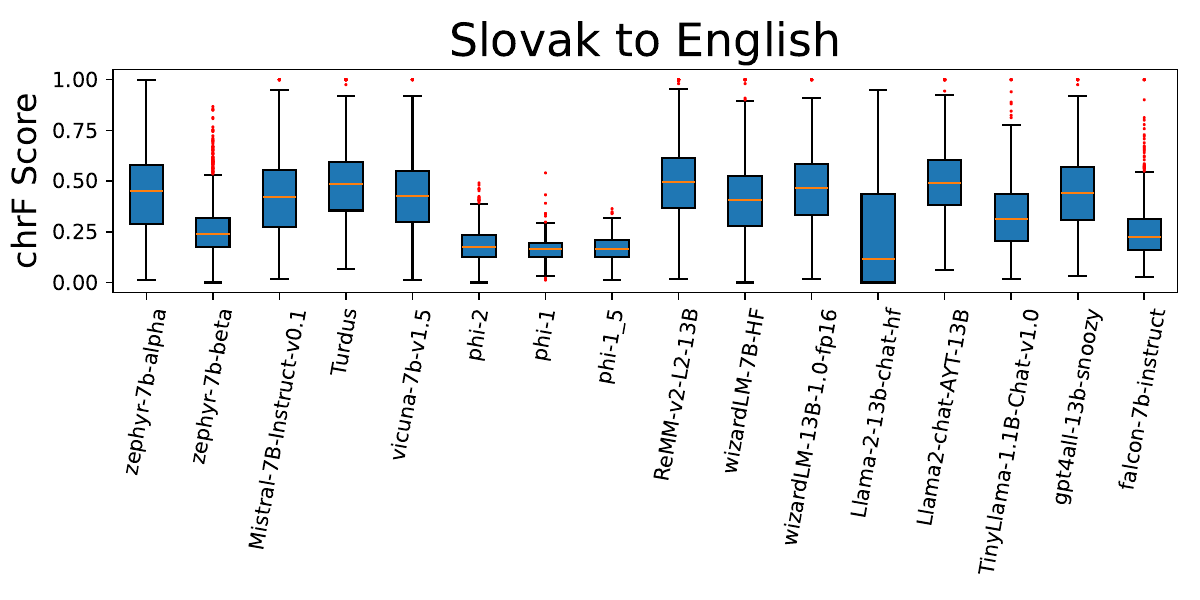}
    \includegraphics[width=0.47\textwidth]{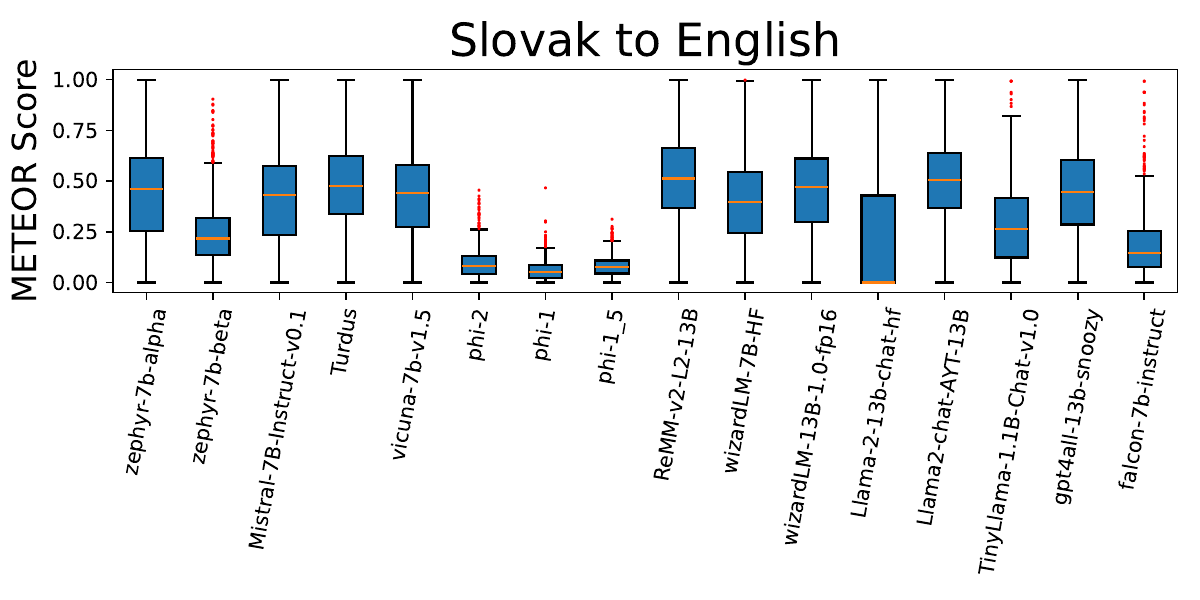}
    \caption{Slovak-to-English dataset per-sentence translation quality and timing statistics  }
    \label{fig:Slovak_translate_stats}
\end{figure}

\begin{figure}[th!]
    \centering
    \includegraphics[width=0.47\textwidth]{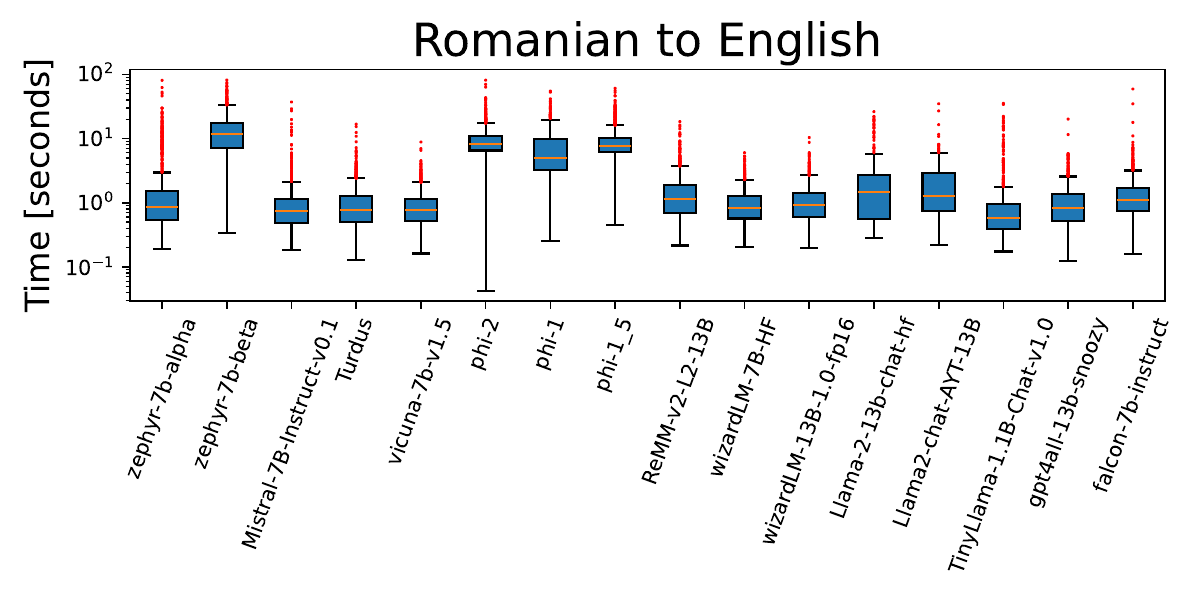}\\
    \includegraphics[width=0.47\textwidth]{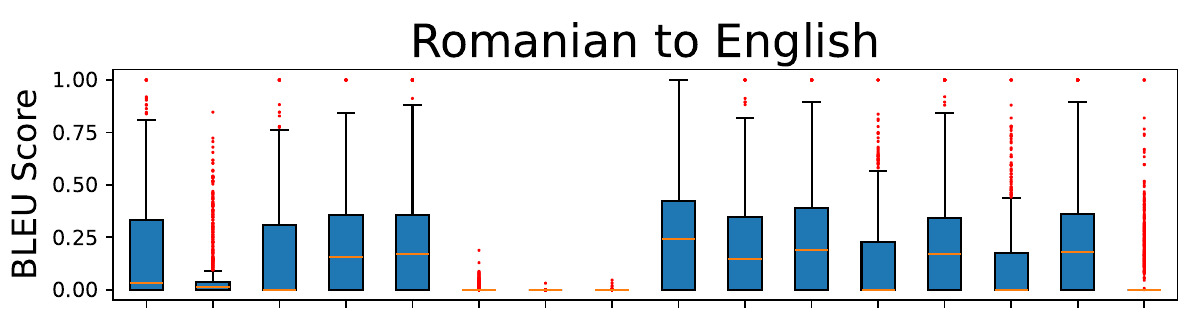}
    \includegraphics[width=0.47\textwidth]{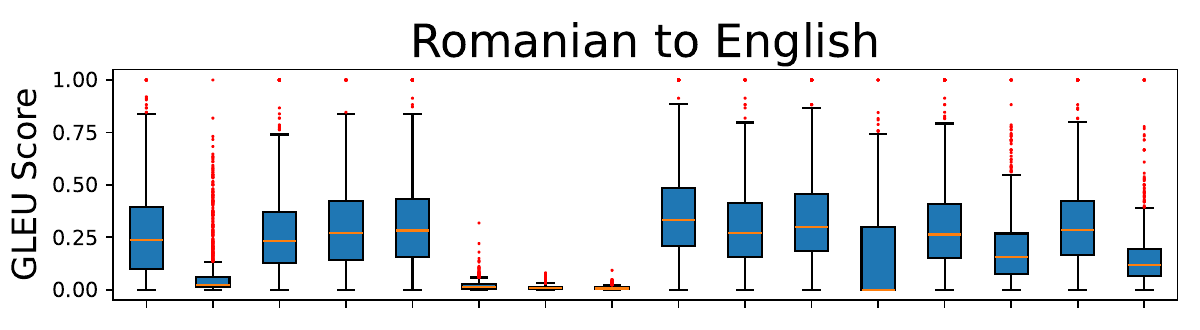}
    \includegraphics[width=0.47\textwidth]{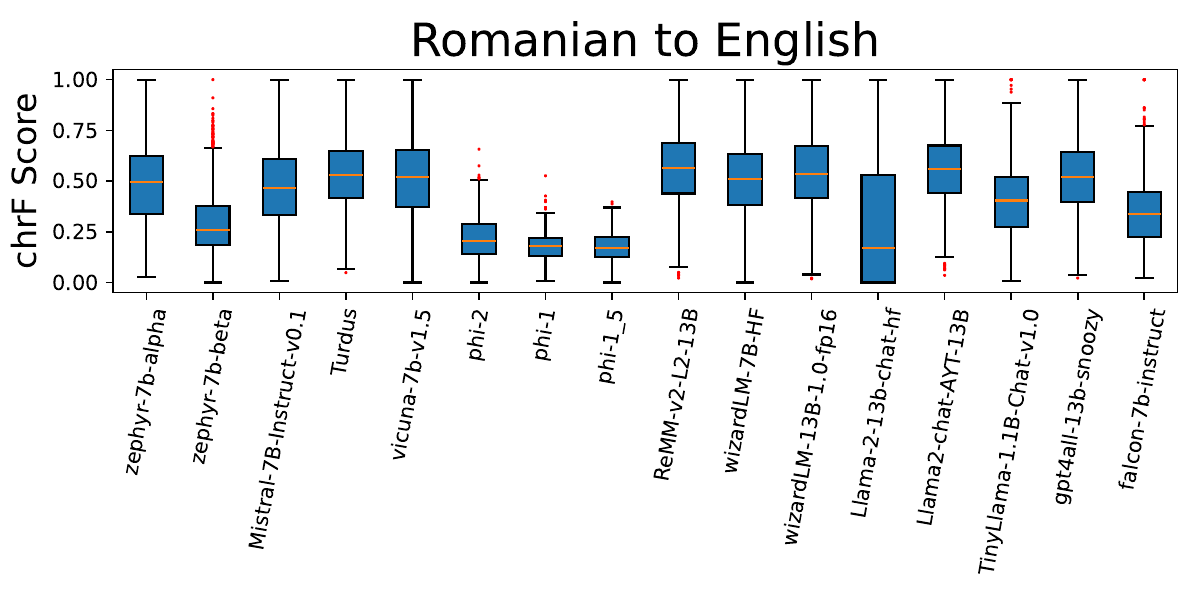}
    \includegraphics[width=0.47\textwidth]{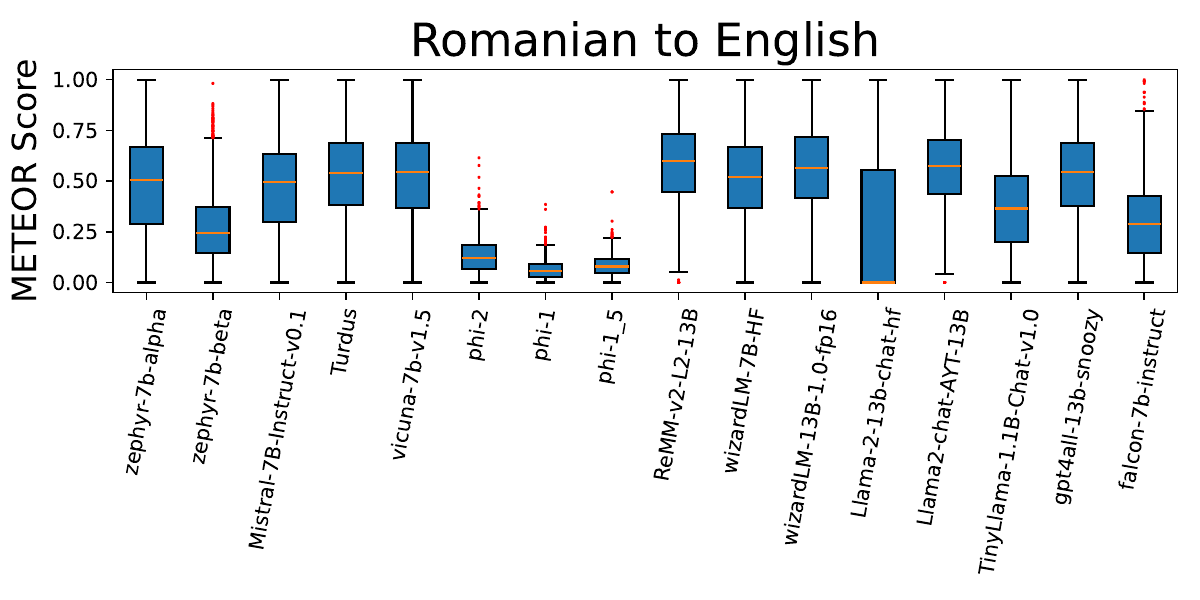}
    \caption{Romanian-to-English dataset per-sentence translation quality and timing statistics  }
    \label{fig:Romanian_translate_stats}
\end{figure}

\begin{figure}[th!]
    \centering
    \includegraphics[width=0.47\textwidth]{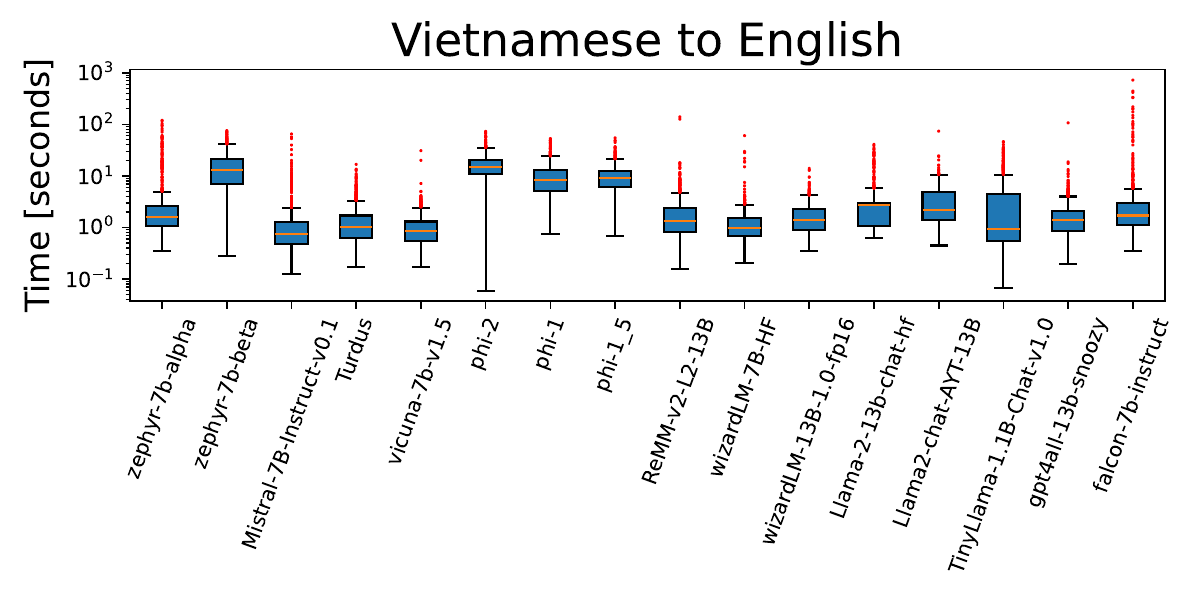}\\
    \includegraphics[width=0.47\textwidth]{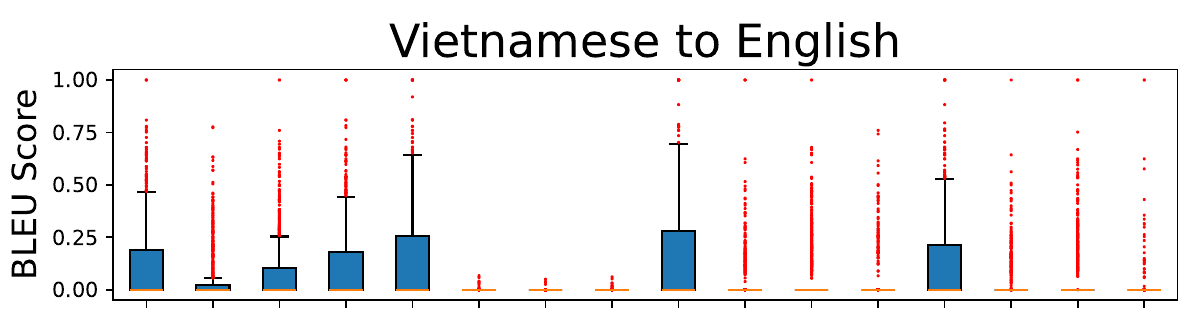}
    \includegraphics[width=0.47\textwidth]{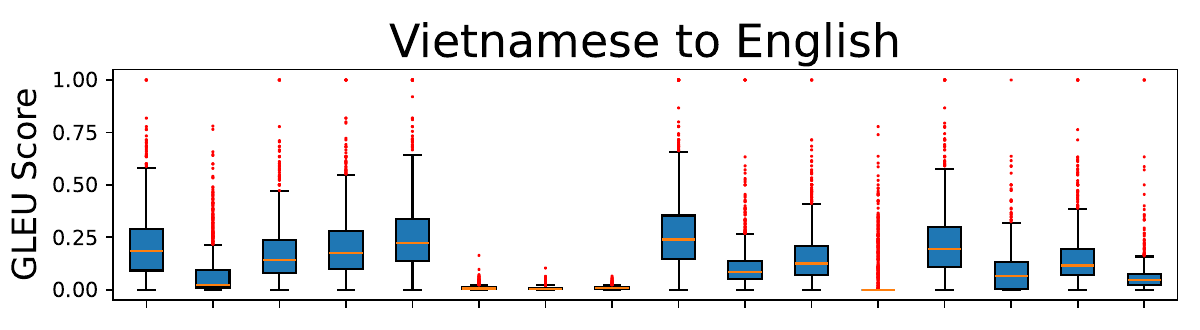}
    \includegraphics[width=0.47\textwidth]{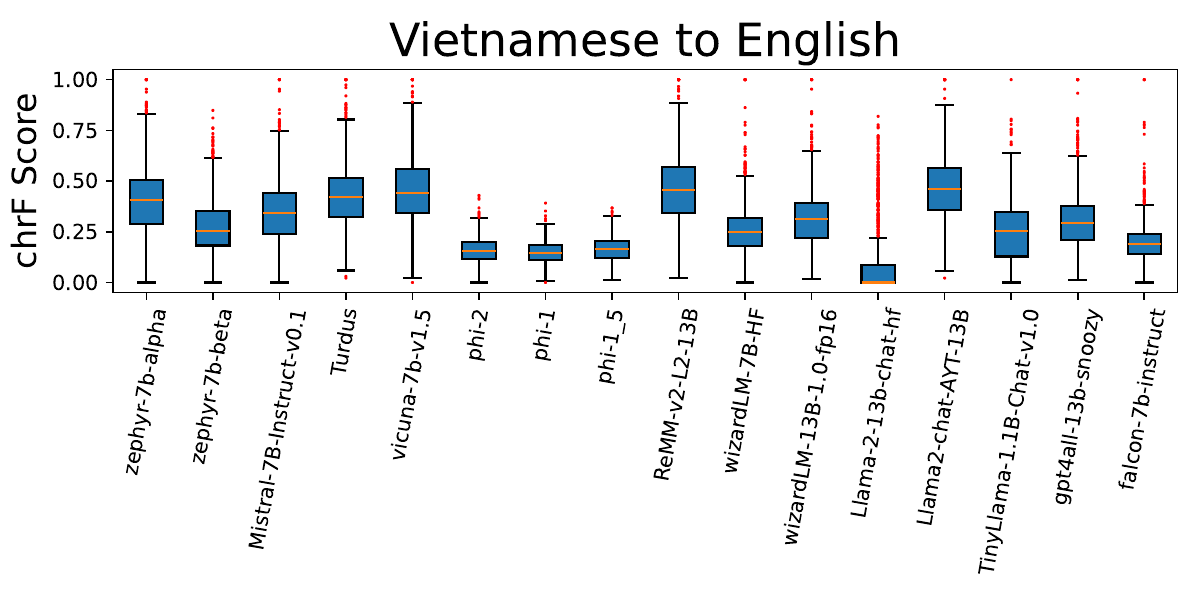}
    \includegraphics[width=0.47\textwidth]{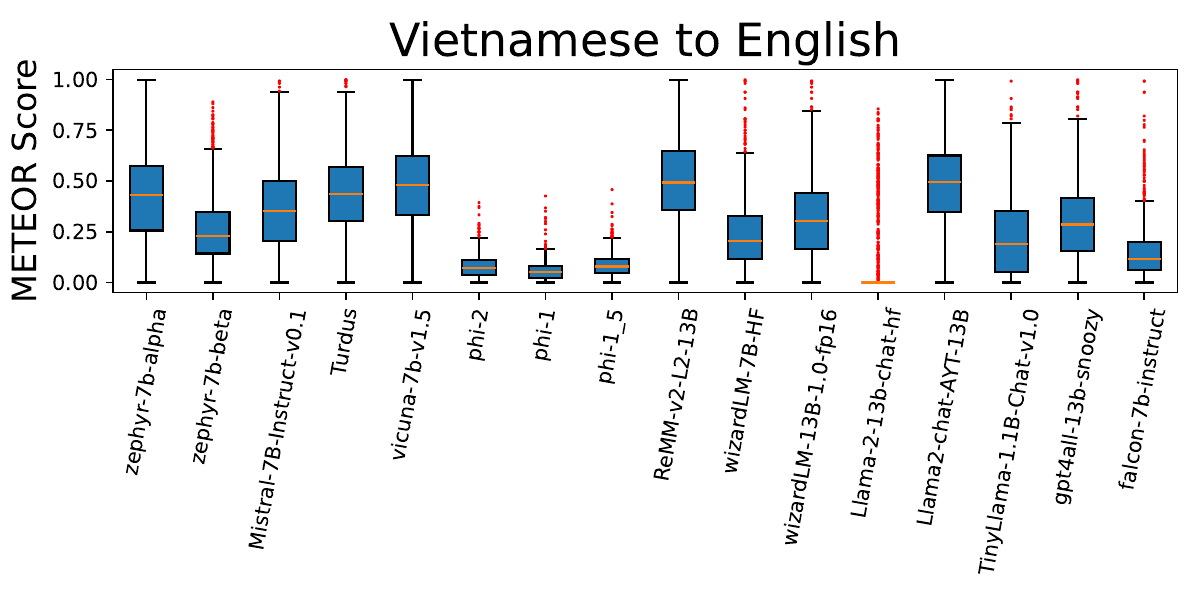}
    \caption{Vietnamese-to-English dataset per-sentence translation quality and timing statistics  }
    \label{fig:Vietnamese_translate_stats}
\end{figure}

\end{document}